\documentclass[journal,twoside]{IEEEtran}

\usepackage{url}
\usepackage{amsmath}
\usepackage{amsfonts}
\newcommand{\texorpdfstring}[2]{#1}
\usepackage{import}
\usepackage{bm}
\usepackage{bbm}
\usepackage[color=red,colorinlistoftodos]{todonotes}
\usepackage{mathtools}
\usepackage{empheq}
\usepackage{multibib}
\newcites{app}{References for the appendices}

\DeclareMathOperator*{\argmin}{arg\,min}

\usepackage{algorithm}
\usepackage{algpseudocode}
\algnewcommand\algorithmicswitch{\textbf{switch}}
\algnewcommand\algorithmiccase{\textbf{case}}
\algdef{SE}[SWITCH]{Switch}{EndSwitch}[1]{\algorithmicswitch\ #1\ \algorithmicdo}{\algorithmicend\ \algorithmicswitch}%
\algdef{SE}[CASE]{Case}{EndCase}[1]{\algorithmiccase\ #1}{\algorithmicend\ \algorithmiccase}%
\algtext*{EndCase}
\algrenewcommand\Return{\State \algorithmicreturn{}}
\algrenewcommand\textproc{}

\usepackage{cleveref}
\crefname{figure}{fig.}{figs.}
\Crefname{Figure}{Fig.}{Figs.}

\providecommand{\sppapathtmp}{}

\begin{document}

\title{Analysis of the
          \texorpdfstring{$(\mu/\mu_I,\lambda)$-$\sigma$-Self-Adaptation}
                         {\$(\textbackslash mu/\textbackslash mu\_I,\textbackslash lambda)\$-\$\textbackslash sigma\$-Self-Adaptation} Evolution Strategy
                         with Repair by Projection Applied to a
                         Conically Constrained Problem}

\author{Patrick~Spettel~and~Hans-Georg~Beyer
  \thanks{Manuscript received Month xx, xxxx; revised Month xx, xxxx
    and Month xx, xxxx; accepted Month xx, xxxx.
    Date of publication Month xx, xxxx; date
    of current version Month xx, xxxx. This work was supported by the
    Austrian Science Fund FWF under grant P29651-N32.
    \textit{(Corresponding author: Patrick Spettel.)}}
  \thanks{The authors are with the
    Research Center Process and Product Engineering
    at the Vorarlberg University of Applied Sciences,
    Dornbirn, Austria
    (e-mail: patrick.spettel@fhv.at; hans-georg.beyer@fhv.at).}
  \thanks{This paper has supplementary downloadable material available at
    http://ieeexplore.ieee.org, provided by the author.
    This consists of an appendix
    containing mathematical derivations and additional figures.
    This material is x.xx MB in size.}
  \thanks{Color versions of one or more of
    the figures in this paper are available
    online at http://ieeexplore.ieee.org.}
  \thanks{Digital Object Identifier xx.xxxx/TEVC.xxxx.xxxxxxx}}

\markboth{IEEE Transactions on Evolutionary Computation,~Vol.~xx,
  No.~x, Month~xxxx}%
{Spettel and Beyer:
  Analysis of the $(\mu/\mu_I,\lambda)$-ES Applied to
  a Conically Constrained Problem}


\IEEEpubid{\begin{minipage}{\textwidth}\centering
    \vspace{0.8cm}
    \copyright~20xx IEEE. Personal use of this material is permitted.
    Permission from IEEE must be obtained for all other uses, in any
    current or future media, including reprinting/republishing this
    material for advertising or promotional purposes, creating new
    collective works, for resale or redistribution to servers or lists,
    or reuse of any copyrighted component of this work in other works.
\end{minipage}}

{\maketitle}
\IEEEpeerreviewmaketitle

\begin{abstract}
  A theoretical performance analysis of the
  $(\mu/\mu_I,\lambda)$-$\sigma$-Self-Adaptation Evolution
  Strategy ($\sigma$SA-ES) is presented considering a conically constrained
  problem. Infeasible offspring are repaired using projection onto the
  boundary of the feasibility region.
  Closed-form approximations are used for the one-generation progress
  of the evolution strategy. Approximate deterministic evolution
  equations are formulated for analyzing the strategy's dynamics.
  By iterating the evolution equations with
  the approximate one-generation expressions, the evolution strategy's
  dynamics can be predicted. The derived theoretical results are
  compared to experiments for assessing the approximation quality.
  It is shown that in the steady state the
  $(\mu/\mu_I,\lambda)$-$\sigma$SA-ES exhibits a performance
  as if the ES were optimizing a sphere model. Unlike the
  non-recombinative $(1,\lambda)$-ES, the parental steady state
  behavior does not evolve on the cone boundary but stays
  away from the boundary to a certain extent.
\end{abstract}


\section{Introduction}
\label{sec:intro}
\IEEEPARstart{C}{urrent} research in evolution strategies includes
the design and analysis of evolution strategies applied to constrained
optimization problems.
It is of particular interest to gain a deep understanding of
evolution strategies on such problems. The insights gained from theory can
help to provide guidance in applying evolution strategies to real
world constrained problems. First, theory can show what
kind of problems are suitably solved by evolution strategies.
Moreover, theoretical investigations can guide the design of
ES algorithms. And furthermore, theoretically
derived suggestions for (optimal) parameter settings can be provided.

Arnold has analyzed a $(1, \lambda)$-ES with constraint handling
by resampling for a single linear constraint~\cite{Arnold2011Behaviour}.
This has been extended in~\cite{Arnold2011Analysis} with the analysis of
repair by projection for a single linear constraint.
This repair approach has been compared with
an approach that reflects infeasible points into the feasible region
and an approach that truncates infeasible points
in~\cite{Hellwig2016Comparison}.
Another idea for constraint handling based on
augmented Lagrangian constraint handling has been presented
in~\cite{Arnold2015Lagrangian} for a $(1+1)$-ES. There, a single
linear inequality constraint with the sphere model is considered.
For this, the one-generation behavior was analyzed.

A multi-recombinative variant of this algorithm has been presented
in~\cite{Atamna2016LagrangianES} for a single linear constraint and multiple
linear constraints in~\cite{Atamna2017LagrangianES}. Markov chains
have been used in both cases for a theoretical investigation.

A conically constrained problem is considered in~\cite{Arnold2013Behaviour}.
There, a $(1, \lambda)$-ES is applied to the problem using death penalty, i.e.,
infeasible offspring are discarded until feasible ones are obtained.
Theoretical investigations have been performed for this constellation.
The same problem has been investigated in~\cite{SpettelBeyer2018SigmaSaEsCone}.
There, a $(1,\lambda)$-$\sigma$-Self-Adaptation ES has been applied.
Repair by projection instead of discarding infeasible offspring has been
used. This paper extends that work to the multi-recombinative variant, the
$(\mu/\mu_I,\lambda)$-$\sigma$-Self-Adaptation ES.

The rest of the paper is organized as follows.
The optimization problem is described in \Cref{sec:problem}.
This is followed by a presentation of the algorithm under
consideration in \Cref{sec:algo}. Next, the
theoretical results are presented. First, the dynamical systems
analysis approach is briefly recapped
in \Cref{sec:theoreticalanalysis:subsec:dynamicalsystem}.
Then, the algorithm's behavior from one generation to the next
(microscopic behavior) is investigated in
\Cref{sec:theoreticalanalysis:subsec:microscopic}.
This is followed by the multi-generation behavior
(macroscopic behavior)
in \Cref{sec:theoreticalanalysis:subsec:evolutiondynamics}.
Closed-form approximations under asymptotic assumptions are derived
for the microscopic behavior. They are then used in deterministic evolution
equations. These evolution equations are iterated in order to predict
the mean value dynamics of the ES. The approximations are compared to
simulations. Finally, the insights gained are discussed in
\Cref{sec:resultsconclusion}. In particular, the differences and similarities
of the $(1,\lambda)$-ES and the multi-recombinative variant are discussed.
\IEEEpubidadjcol

\section{Optimization Problem}
\label{sec:problem}
The optimization problem under consideration is
\begin{equation}
    \label{sec:problem:eq:optgoal}
    f(\mathbf{x}) = x_1 \rightarrow \text{min}!
\end{equation}
subject to constraints
\begin{align}
  x_1^2 - \xi \sum_{k=2}^N x_k^2 &\ge 0
  \label{sec:problem:eq:coneconstraint}\\
  x_1 &\ge 0
\end{align}
where $\mathbf{x} = (x_1, \ldots, x_N)^T \in \mathbb{R}^N$ and $\xi > 0$.

\begin{figure}
  \centering
  \includegraphics{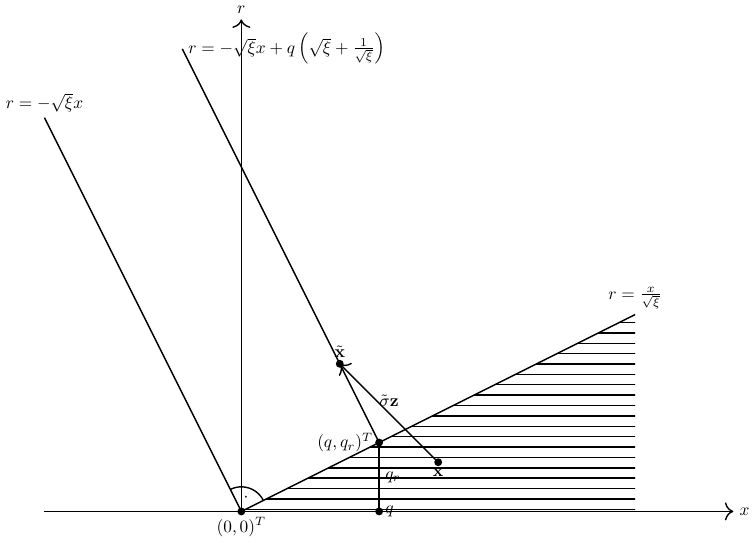}
    \caption[The conically constrained optimization problem in $N$ dimensions
             shown in the $(x,r)^T$-space.]
            {The conically constrained optimization problem in $N$ dimensions
             shown in the $(x,r)^T$-space.
            }
    \label{sec:problem:fig:conicconstraintnd}
\end{figure}

The distance $x$ from $0$ in $x_1$-direction (cone axis) and the distance $r$
from the cone's axis suffices to describe the state of the ES
in the search space. This is denoted as the $(x,r)^T$-space in the following.
Moreover, note that due to the isotropy of the mutations used in the ES,
the coordinate system
can w.l.o.g. be rotated. Thus, an objective parameter vector's distance
from the cone axis coincides with the second component of this rotated
coordinate system, i.e., $(\tilde{x},\tilde{r})^T$ corresponds to
$(\tilde{x},\tilde{r}, 0, \ldots, 0)^T$,
see \Cref{sec:problem:fig:conicconstraintnd}.
The cone boundary is described by the equation
$r=\frac{x_1}{\sqrt{\xi}}$,
which follows from
\Cref{sec:problem:eq:coneconstraint}.
One arrives at the equation of the projection line
by the cone direction $\left(1, \frac{1}{\sqrt{\xi}}\right)^T$ and
$\left(-\frac{1}{\sqrt{\xi}}, 1\right)^T$ (counterclockwise rotation
by 90 degrees).
With this,
$r = -\sqrt{\xi}x_1 + q\left(\sqrt{\xi} + \frac{1}{\sqrt{\xi}}\right)$
follows as the equation for the projection line.
A parent $\mathbf{x}$ and an offspring $\tilde{\mathbf{x}}$
with the mutation $\tilde{\sigma}\mathbf{z}$ are visualized as well.
The values $q$ and $q_r$ denote the $x_1$ and $r$ values
after the projection
step.

\section{Algorithm}
\label{sec:algo}
A $(\mu/\mu_I,\lambda)$-$\sigma$-Self-Adaptation ES is applied
to the optimization problem described in \Cref{sec:problem}.
\Cref{sec:algorithm:alg:es} shows the pseudo code\footnote{$(\mathbf{x})_k$
  denotes the $k$-th element of a vector $\mathbf{x}$.
  $\mathbf{x}_{m;\lambda}$ is the order statistic notation. It denotes
  the $m$-th best (according to fitness) of $\lambda$ elements.}.
First, parameters are initialized
(\crefrange{sec:algorithm:alg:es:init}{sec:algorithm:alg:es:ginit}).
After that, the generation loop is entered.
$\lambda$ offspring are generated
in
\crefrange{sec:algorithm:alg:es:forloopbegin}
{sec:algorithm:alg:es:forloopend}.
A log-normal distribution is used
for every offspring to compute its mutation strength $\tilde{\sigma}_l$
by mutating the parental mutation strength $\sigma^{(g)}$
(\cref{sec:algorithm:alg:es:offspringsigma}).
Using this determined mutation strength, the offspring's parameter vector
is sampled from a multivariate normal distribution
with mean $\mathbf{x}^{(g)}$ and standard deviation $\tilde{\sigma}_l$
(\cref{sec:algorithm:alg:es:generateoffspring}).
Then, the offspring is repaired by projection, if necessary
(\crefrange{sec:algorithm:alg:es:feasibilitycheck}
{sec:algorithm:alg:es:feasibilitycheckend}).
This means that for infeasible offspring, the optimization problem
\begin{align}
\label{sec:algorithm:alg:coneprojection}
\begin{split}
  \mathbf{\hat{x}} = \argmin_{\mathbf{x'}}
    \lVert\mathbf{x'} - \mathbf{x}\rVert^2\\
  \text{s.t. }{x'_1}^2 - \xi \sum_{k=2}^N {x'_k}^2 \ge 0\\
  x'_1 \ge 0
\end{split}
\end{align}
must be solved where
$\mathbf{x}$ is the individual to be projected.
For this, a helper function
\begin{equation}
  \label{sec:algorithm:alg:coneprojectionhelper}
  \mathbf{\hat{x}} = \text{projectOntoCone}(\mathbf{x})
\end{equation}
is introduced that
returns $\hat{\mathbf{x}}$ of~\eqref{sec:algorithm:alg:coneprojection}.
A derivation for a closed-form solution to this projection
optimization problem~\eqref{sec:algorithm:alg:coneprojection}
is presented in the supplementary material in
\Cref{sec:algorithm:subsec:projection}.
$x^{(g)}$, $r^{(g)}$, $q_l$, $\mathbf{q'}_l$,
$\langle q \rangle$, and $\langle q_r \rangle$
(\cref{sec:algorithm:alg:es:x,%
sec:algorithm:alg:es:r,%
sec:algorithm:alg:es:q,%
sec:algorithm:alg:es:qr,%
sec:algorithm:alg:es:bestq,%
sec:algorithm:alg:es:bestqr})
are values used in the theoretical analysis. They can be removed in practical
implementations of the ES.
The offspring's fitness is computed in \cref{sec:algorithm:alg:es:offspringf}.
The next generation's parental individual $\mathbf{x}^{(g + 1)}$
(\cref{sec:algorithm:alg:es:replaceparent})
and the
next generation's mutation strength $\sigma^{(g + 1)}$
(\cref{sec:algorithm:alg:es:replacesigma})
are updated next. The parental parameter vector for the next generation
is set to the mean of the parameter vectors of the $\mu$ best offspring.
Similarly, the parental mutation strength for the next generation
is set to the mean of the mutation strengths of the $\mu$ best offspring.
If the parental parameter vector for the next generation is not feasible,
it is projected onto the cone
(\crefrange{sec:algorithm:alg:es:feasibilitycheckcentroid}
{sec:algorithm:alg:es:feasibilitycheckendcentroid}).
Note that this repair step is not needed
in the real implementation of the algorithm.
Since the conical problem is convex, the intermediate recombination
of the $\mu$ best feasible individuals is feasible
as well. However, in the iteration of the evolution equations in
\Cref{sec:theoreticalanalysis:subsec:evolutiondynamics},
the $(x,r)^T$ can get infeasible due to the approximations used.
Finally, the generation counter is updated, which completes one iteration
of the generational loop.

\begin{algorithm}
  \caption{Pseudo-code of the
    $(\mu/\mu_I,\lambda)$-$\sigma$-Self-Adaptation ES with
    repair by projection applied to the conically constrained problem.}
  \label{sec:algorithm:alg:es}
  \begin{algorithmic}[1]
    \State{Initialize $\mathbf{x}^{(0)}$, $\sigma^{(0)}$, $\tau$,
      $\lambda$, $\mu$}
    \label{sec:algorithm:alg:es:init}
    \State{$g \gets 0$}
    \label{sec:algorithm:alg:es:ginit}
    \Repeat
    \label{sec:algorithm:alg:es:repeatbegin}
    \State{$x^{(g)} = (\mathbf{x}^{(g)})_1$}
    \label{sec:algorithm:alg:es:x}
    \State{$r^{(g)} = \sqrt{\sum_{k=2}^N(\mathbf{x}^{(g)})_k^2}$}
    \label{sec:algorithm:alg:es:r}
    \For{$l \gets 1 \textbf{ to } \lambda$}
    \label{sec:algorithm:alg:es:forloopbegin}
    \State{$\tilde{\sigma}_l
      \gets \sigma^{(g)} e^{\tau\mathcal{N}(0, 1)}$}
    \label{sec:algorithm:alg:es:offspringsigma}
    \State{$\tilde{\mathbf{x}}_l \gets \mathbf{x}^{(g)} +
      \tilde{\sigma}_l\mathcal{N}(\mathbf{0}, \mathbf{I})$}
    \label{sec:algorithm:alg:es:generateoffspring}
    \If{\textbf{not} \Call{isFeasible}{$\tilde{\mathbf{x}}_l$}}
    \Comment{see \Cref{sec:algorithm:alg:feasibilitycheck}}
    \label{sec:algorithm:alg:es:feasibilitycheck}
    \State{$\tilde{\mathbf{x}}_l \gets$
      \Call{projectOntoCone}{$\tilde{\mathbf{x}}_l$}}
    \Comment{see~\eqref{sec:algorithm:alg:coneprojection},~\eqref{sec:algorithm:alg:coneprojectionhelper}}
    \label{sec:algorithm:alg:es:projectionw}
    \EndIf
    \label{sec:algorithm:alg:es:feasibilitycheckend}
    \State{$\tilde{f}_l \gets f(\tilde{\mathbf{x}}_l)=(\tilde{\mathbf{x}}_l)_1$}
    \label{sec:algorithm:alg:es:offspringf}
    \State{$q_l = (\tilde{\mathbf{x}}_l)_1$}
    \label{sec:algorithm:alg:es:q}
    \State{$\mathbf{q'}_l = \tilde{\mathbf{x}}_l$}
    \label{sec:algorithm:alg:es:qr}
    \EndFor
    \label{sec:algorithm:alg:es:forloopend}
    \State{Sort offspring according to $\tilde{f}_l$ in ascending order}
    \label{sec:algorithm:alg:es:sortoffspring}
    \State{$\mathbf{x}^{(g + 1)} \gets \frac{1}{\mu}\sum_{m=1}^\mu\tilde{\mathbf{x}}_{m;\lambda}$}
    \label{sec:algorithm:alg:es:replaceparent}
    \State{$\sigma^{(g + 1)} \gets \frac{1}{\mu}\sum_{m=1}^\mu\tilde{\sigma}_{m;\lambda}$}
    \label{sec:algorithm:alg:es:replacesigma}
    \State{$\langle q \rangle = (\mathbf{x}^{(g+1)})_1$}
    \label{sec:algorithm:alg:es:bestq}
    \State{$\langle q_r \rangle = \sqrt{\sum_{k=2}^N(\mathbf{x}^{(g+1)})_k^2}$}
    \label{sec:algorithm:alg:es:bestqr}
    \If{\textbf{not} \Call{isFeasible}{$\mathbf{x}^{(g+1)}$}}
    \Comment{see \Cref{sec:algorithm:alg:feasibilitycheck}}
    \label{sec:algorithm:alg:es:feasibilitycheckcentroid}
    \State{$\mathbf{x}^{(g+1)} \gets$
      \Call{projectOntoCone}{$\mathbf{x}^{(g+1)}$}}
    \Comment{see~\eqref{sec:algorithm:alg:coneprojection},~\eqref{sec:algorithm:alg:coneprojectionhelper}}
    \label{sec:algorithm:alg:es:projectionwcentroid}
    \EndIf
    \label{sec:algorithm:alg:es:feasibilitycheckendcentroid}
    \State{$g \gets g + 1$}
    \label{sec:algorithm:alg:es:gupdate}
    \Until{termination criteria are met}
    \label{sec:algorithm:alg:es:repeatend}
  \end{algorithmic}
\end{algorithm}
\begin{algorithm}
  \caption{Feasibility check}
  \label{sec:algorithm:alg:feasibilitycheck}
  \begin{algorithmic}[1]
    \Function{isFeasible}{$\mathbf{x}$}
    \Return{($x_1 \ge 0 \land
      x_1^2 - \xi \sum_{k=2}^N x_k^2 \ge 0$)}
    \EndFunction
  \end{algorithmic}
\end{algorithm}

\begin{figure}
  \centering
  \begin{tabular}{@{\hspace{-0.0\textwidth}}c}
    \includegraphics[width=0.43\textwidth]{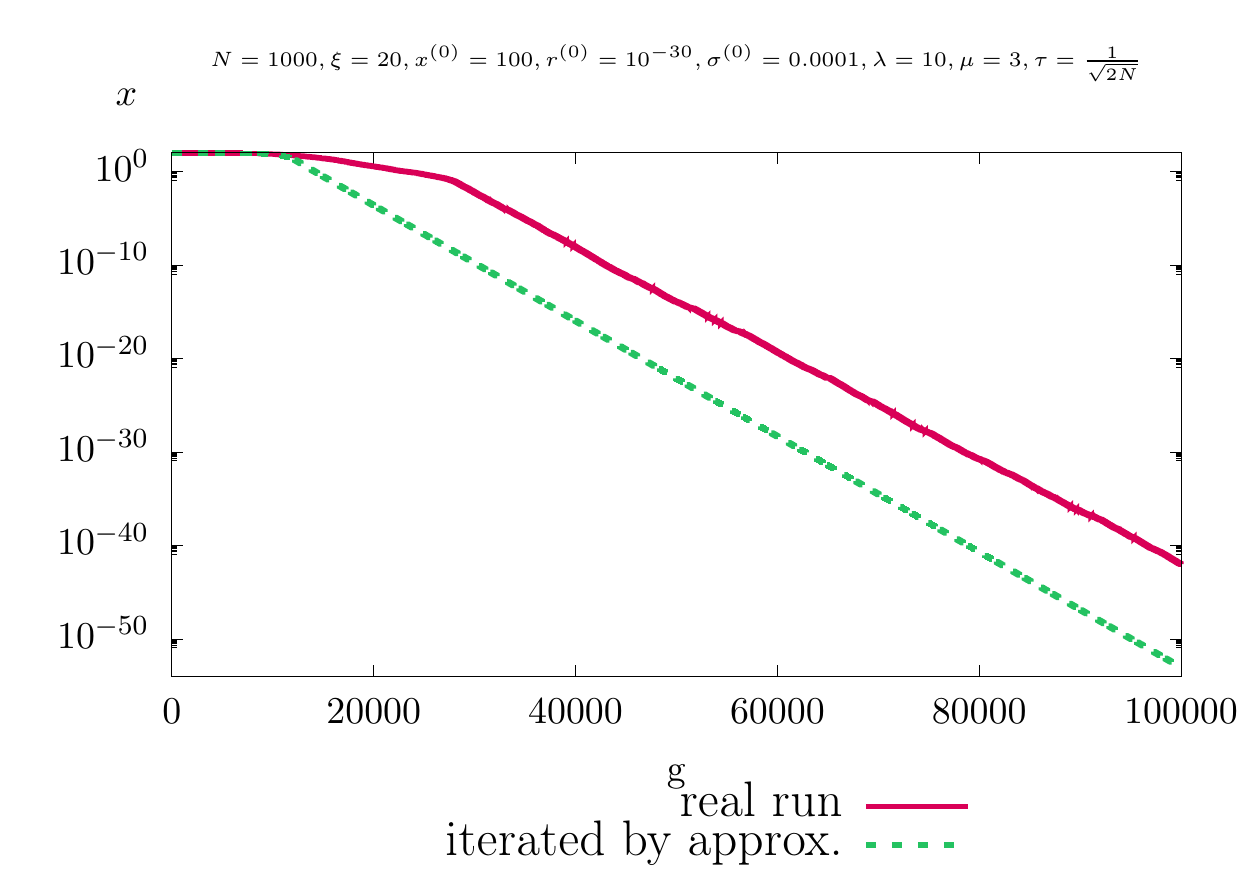}\\
    \includegraphics[width=0.43\textwidth]{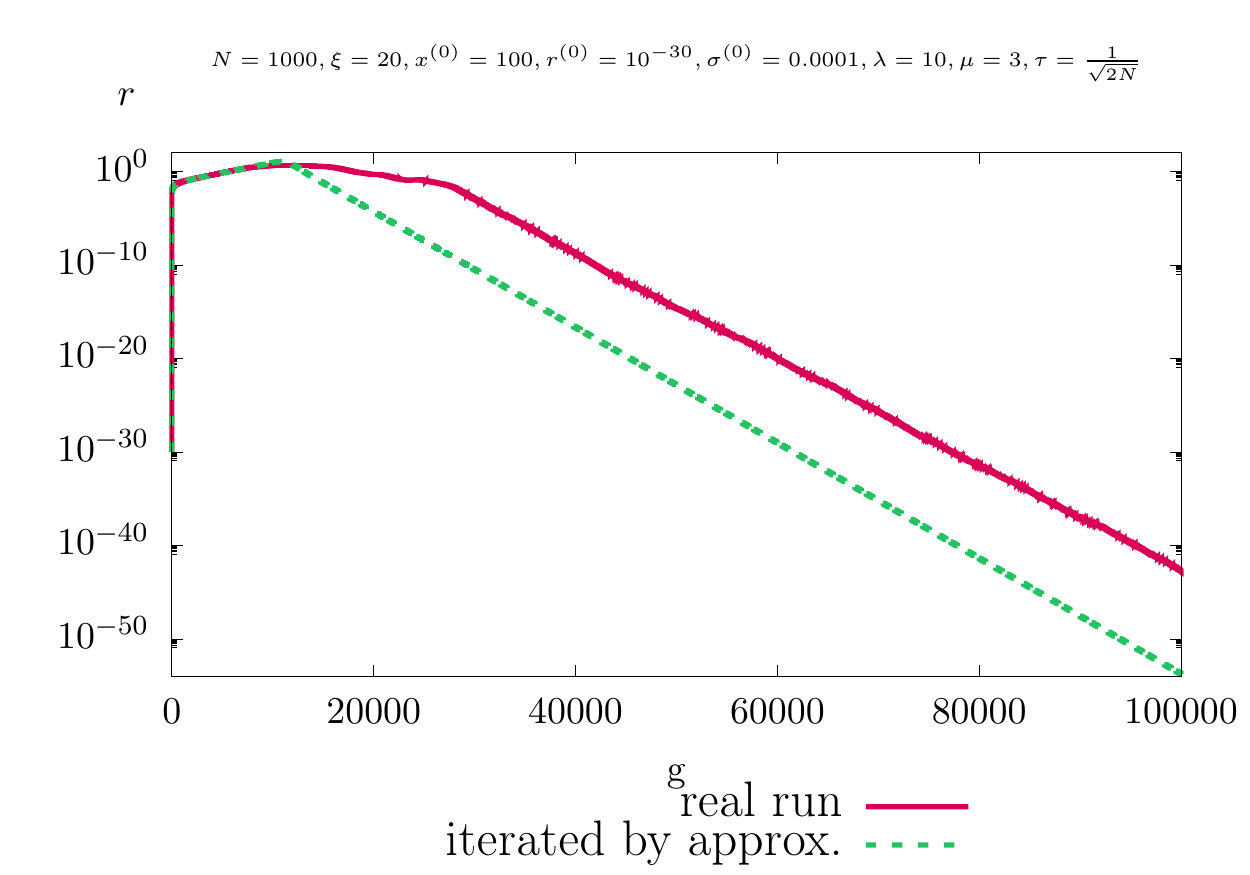}
  \end{tabular}
  \caption{Comparison of the $x$- and $r$-dynamics
    of a real $(3/3_I, 10)$-ES run (solid line) with the iteration of the
    closed-form approximation of the iterative system (dotted line).}
  \label{sec:theoreticalanalysis:fig:dynamicsexample}
\end{figure}

\Cref{sec:theoreticalanalysis:fig:dynamicsexample}
shows an example of the $x$- and $r$-dynamics that are a result of running
\Cref{sec:algorithm:alg:es} (solid line).
The closed-form approximation iterative system (dotted line)
(to be derived in the following sections) is shown in comparison.
The prediction does not coincide completely with the real run.
The ES reaches the stationary state later than predicted.
The approximations that are derived in this work result
in deviations in the transient phase of the ES.
However, our main focus is in the steady state.
There, the predicted slope of
the $x$ and $r$ dynamics is very similar to the one of the real run.

\renewcommand{\sppapathtmp}{./figures/figure02/}

\section{Theoretical Analysis}
\label{sec:theoreticalanalysis}

\subsection{The Dynamical Systems Approach}
\label{sec:theoreticalanalysis:subsec:dynamicalsystem}
The $(x,r)^T$-representation described in \Cref{sec:problem} is used
for the analysis of the $(\mu/\mu_I,\lambda)$-ES (\Cref{sec:algorithm:alg:es}).
The aim is to predict the evolution dynamics of the $(\mu/\mu_I,\lambda)$-ES.
To this end, the dynamical systems method described in \cite{Beyer2001} is used.
State variables of the strategy are modeled over time. For the case
here, the three random variables for $x$, $r$, and $\sigma$ describe the
state of the system. A Markov process can be used for modeling the
transitions between states from one time step to the next. It is often
not possible to derive closed-form expressions for the transition equations.
Therefore, approximate equations are aimed for. Evolution equations
can be stated by making use of the local expected change functions
($\varphi_x, \varphi_r, \psi$)
\begin{align}
  x^{(g + 1)} &= x^{(g)} -
  \varphi_{x}(x^{(g)}, r^{(g)}, \sigma^{(g)}) +
  \epsilon_{x}(x^{(g)}, r^{(g)}, \sigma^{(g)})
  \label{sec:theoreticalanalysis:eq:evolutionequationx}\\
  r^{(g + 1)} &= r^{(g)} -
  \varphi_r(x^{(g)}, r^{(g)}, \sigma^{(g)}) +
  \epsilon_r(x^{(g)}, r^{(g)}, \sigma^{(g)})
  \label{sec:theoreticalanalysis:eq:evolutionequationr}\\
  \sigma^{(g + 1)} &= \sigma^{(g)} +
  \sigma^{(g)}\psi(x^{(g)}, r^{(g)}, \sigma^{(g)}) +
  \epsilon_\sigma(x^{(g)}, r^{(g)}, \sigma^{(g)}).
  \label{sec:theoreticalanalysis:eq:evolutionequationsigma}
\end{align}
The progress rates are defined as
\begin{equation}
  \label{sec:theoreticalanalysis:eq:varphix}
  \varphi_{x}(x^{(g)}, r^{(g)}, \sigma^{(g)}) :=
  \mathrm{E}[x^{(g)} - x^{(g + 1)}\,|\,x^{(g)}, r^{(g)}, \sigma^{(g)}]
\end{equation}
\begin{equation}
  \label{sec:theoreticalanalysis:eq:varphir}
  \varphi_{r}(x^{(g)}, r^{(g)}, \sigma^{(g)}) :=
  \mathrm{E}[r^{(g)} - r^{(g + 1)}\,|\,x^{(g)}, r^{(g)}, \sigma^{(g)}].
\end{equation}
They model the expected change in
the parameter space from one generation to the next.
The random variables
$\epsilon_{x}$, $\epsilon_r$, and $\epsilon_\sigma$
represent the stochastic part. $\mathrm{E}[\epsilon_{x}] = 0$,
$\mathrm{E}[\epsilon_{r}] = 0$, and $\mathrm{E}[\epsilon_{\sigma}] = 0$
necessarily holds.

In what follows, the normalizations
\begin{equation}
  \label{sec:theoreticalanalysis:eq:varphixnormalized}
  \varphi_{x}^*(\cdot) := \frac{N\varphi_{x}(\cdot)}{x^{(g)}},
\end{equation}
\begin{equation}
  \label{sec:theoreticalanalysis:eq:varphirnormalized}
  \varphi_{r}^*(\cdot) := \frac{N\varphi_{r}(\cdot)}{r^{(g)}},
\end{equation}
and
\begin{equation}
  \label{sec:theoreticalanalysis:eq:sigmanormalized}
  \sigma^* := \frac{N\sigma}{r^{(g)}}
\end{equation}
will be used.
For the change of the mutation strength from one generation to the next,
a slightly different progress measure will be applied. It is the so-called
self-adaptation response (SAR). Its definition reads
\begin{equation}
  \label{sec:theoreticalanalysis:eq:psi}
  \psi(x^{(g)}, r^{(g)}, \sigma^{(g)}) :=
  \mathrm{E}\left[\frac{\sigma^{(g + 1)} - \sigma^{(g)}}{\sigma^{(g)}}
    \,\bigg|\,x^{(g)}, r^{(g)}, \sigma^{(g)}\right].
\end{equation}
$\epsilon_{x} = 0$, $\epsilon_r = 0$, and $\epsilon_\sigma = 0$ is assumed
in
\Crefrange{sec:theoreticalanalysis:eq:evolutionequationx}
          {sec:theoreticalanalysis:eq:evolutionequationsigma}
for the analysis in this work. That is, it is assumed that the
fluctuations can be ignored for deriving approximate expressions
for the evolution dynamics of the strategy for the case $N \rightarrow \infty$.
Evolution equations without fluctuation terms are also known as deterministic
evolution equations or mean value evolution equations.
With the goal of arriving at approximate deterministic evolution equations,
approximate expressions for the functions $\varphi_x$, $\varphi_r$, and $\psi$
need to be derived first.

\subsection{The Microscopic Aspects}
\label{sec:theoreticalanalysis:subsec:microscopic}

The microscopic aspects describe the
behavior of the evolution strategy from one generation to the next.
They are expressed by the functions introduced in
\Cref{sec:theoreticalanalysis:subsec:dynamicalsystem}
(\Cref{sec:theoreticalanalysis:eq:varphix,%
  sec:theoreticalanalysis:eq:varphir,%
  sec:theoreticalanalysis:eq:psi}).

In the following analyses, two cases are treated separately.
It turns out that if the parental individual is in the
vicinity of the cone boundary,
the offspring feasibility probability tends to $0$ for
asymptotic considerations ($N \rightarrow \infty$). Hence,
the local progress measures and the SAR are derived separately
for the case of being at the cone boundary and for the opposite case.
In order to have one approximate closed-form expression,
they are combined by weighting with the feasibility probability.
It is referred
to~\cite[Sec. 3.1.2.1.2.8, pp. 44-49]{SpettelBeyer2018SigmaSaEsCone}
for further details. There, the feasibility probability has been
derived as
\begin{align}
  P_{\text{feas}}(x^{(g)}, r^{(g)}, \sigma^{(g)})
  \simeq\Phi\left[
    \frac{1}{\sigma^{(g)}}
    \left(
    \frac{x^{(g)}}{\sqrt{\xi}}-\bar{r}
    \right)
    \right]
  \label{sec:theoreticalanalysis:eq:Pfeasapprox1}
\end{align}
where $\bar{r}$ is the expected value of the $r$ normal approximation
and $\Phi$ is the cumulative distribution function of the standard
normal variate.
In order to simplify the theoretical analysis, the distribution of
the $r$ component is approximated by a normal distribution. The supplementary
material (\Cref{sec:theoreticalanalysis:appendix:rnormalapproximation})
presents a detailed derivation of this approximation.

\subsubsection{Derivation of the \texorpdfstring{$x$}{\$x\$} Progress Rate}
\label{sec:theoreticalanalysis:subsec:microscopic:x}
From the definition of the progress rate
(\Cref{sec:theoreticalanalysis:eq:varphix})
and the pseudo-code of the
ES (\Cref{sec:algorithm:alg:es},
\cref{sec:algorithm:alg:es:bestq,%
sec:algorithm:alg:es:x}),
\begin{align}
  \varphi_{x}(x^{(g)}, r^{(g)}, \sigma^{(g)})
  &=x^{(g)} - \mathrm{E}[x^{(g + 1)}\,|\,x^{(g)}, r^{(g)}, \sigma^{(g)}]
  \label{sec:theoreticalanalysis:eq:xprogressratedef1}\\
  &= x^{(g)} - \mathrm{E}[\langle q \rangle\,|\,x^{(g)}, r^{(g)}, \sigma^{(g)}]
  \label{sec:theoreticalanalysis:eq:xprogressratedef2}
\end{align}
follows.
Therefore, $\mathrm{E}[\langle q \rangle\,|\,x^{(g)}, r^{(g)}, \sigma^{(g)}]
:=\mathrm{E}[\langle q \rangle]$ needs to be derived to proceed
further. Its derivation is presented in detail in the
supplementary material in \Cref{appendix:subsec:expqcentroid}.

Using the result from \Cref{sec:theoreticalanalysis:eq:qcentroidfeasinserted},
the normalized $x$ progress rate for the feasible case can be formulated as
\begin{align}
  {\varphi_x}_{\text{feas}}^*
  &= \frac{N\left(x^{(g)}-\mathrm{E}[{\langle q \rangle}_{\text{feas}}]\right)}{x^{(g)}}
  \label{sec:theoreticalanalysis:eq:varphixnormalizedfeasible1}\\
  &= \frac{N\left(x^{(g)}-x^{(g)}+\sigma^{(g)}c_{\mu/\mu,\lambda}\right)}{x^{(g)}}
  \label{sec:theoreticalanalysis:eq:varphixnormalizedfeasible2}\\
  &= \frac{r^{(g)}}{x^{(g)}}{\sigma^{(g)}}^*c_{\mu/\mu,\lambda}.
  \label{sec:theoreticalanalysis:eq:varphixnormalizedfeasible5}
\end{align}
In \Cref{sec:theoreticalanalysis:eq:varphixnormalizedfeasible5},
one of the so-called generalized progress coefficients
$c_{\mu/\mu,\lambda}=e_{\mu,\lambda}^{1,0}$ defined
in~\cite[Eq. (5.112), p. 172]{Beyer2001} appears.
Their definition reads
\begin{equation}
  \label{sec:theoreticalanalysis:eq:eabml}
  \begin{multlined}
  e_{\mu,\lambda}^{\alpha,\beta} :=
  \frac{\lambda-\mu}{{(\sqrt{2\pi})}^{\alpha+1}}\binom{\lambda}{\mu}\\
  \times\int_{t=-\infty}^{t=\infty}
  t^\beta
  e^{-\frac{\alpha+1}{2}t^2}
  [\Phi(t)]^{\lambda-\mu-1}[1-\Phi(t)]^{\mu-\alpha}
  \,\mathrm{d}t.
  \end{multlined}
\end{equation}

Similarly, use of
\Cref{sec:theoreticalanalysis:eq:qcentroidinfeasinsertedfinal}
results in
\begin{align}
  &\begin{multlined}
  {\varphi_x}_{\text{infeas}}^*
  = \frac{N\left(x^{(g)}-
  \mathrm{E}[{\langle q \rangle}_{\text{infeas}}]\right)}{x^{(g)}}
  \end{multlined}
  \label{sec:theoreticalanalysis:eq:varphixnormalizedinfeasibleminus1}\\
  &\begin{multlined}
  \phantom{{\varphi_x}_{\text{infeas}}^*}=
  \frac{N}{x^{(g)}}\left(\frac{1+\xi}{1+\xi}x^{(g)}
  -\mathrm{E}[{\langle q \rangle}_{\text{infeas}}]\right)
  \end{multlined}
  \label{sec:theoreticalanalysis:eq:varphixnormalizedinfeasible0}\\
  &\begin{multlined}
  \phantom{{\varphi_x}_{\text{infeas}}^*}
  \approx \frac{N}{x^{(g)}}
  \left[
    \frac{x^{(g)}+\xi x^{(g)}-\xi x^{(g)}}{1+\xi}
    -\frac{\sqrt{\xi}\bar{r}}{1+\xi}\right.\\\left.
    +\frac{\xi}{1+\xi}\left(\sqrt{{\sigma^{(g)}}^2+\sigma_r^2/\xi}\right)
      c_{\mu/\mu,\lambda}
  \right].
  \end{multlined}
  \label{sec:theoreticalanalysis:eq:varphixnormalizedinfeasible1}
\end{align}
for the infeasible case.
This can further be simplified using
$\sigma_r$ of the normal approximation
(\Cref{sec:theoreticalanalysis:eq:approximatedr})
together with $\sigma$-normalization yielding
\begin{align}
  &\begin{multlined}
  {\varphi_x}_{\text{infeas}}^*  = \frac{N}{1+\xi}
  \biggl[
    1-\sqrt{\xi}\frac{r^{(g)}}{x^{(g)}}
    \sqrt{1+\frac{{{\sigma^{(g)}}^*}^2}{N}\left(1-\frac{1}{N}\right)}\\
    +\xi\frac{r^{(g)}}{x^{(g)}}
    \sqrt{\frac{{{\sigma^{(g)}}^*}^2}{N^2}
      +\frac{1}{\xi}\frac{{{\sigma^{(g)}}^*}^2}{N^2}
      \frac{1+\frac{{{\sigma^{(g)}}^*}^2}{2N}(1-\frac{1}{N})}
           {1+\frac{{{\sigma^{(g)}}^*}^2}{N}(1-\frac{1}{N})}} c_{\mu/\mu,\lambda}
    \biggr]
  \end{multlined}
  \label{sec:theoreticalanalysis:eq:varphixnormalizedinfeasible2}\\
  &\begin{multlined}
  \phantom{{\varphi_x}_{\text{infeas}}^* }\simeq \frac{N}{1+\xi}
  \left[
    1-\sqrt{\xi}\frac{r^{(g)}}{x^{(g)}}
    \sqrt{1+\frac{{{\sigma^{(g)}}^*}^2}{N}}\right.\\\left.
    \hspace{2cm}+\xi\frac{r^{(g)}}{x^{(g)}}
    \frac{{{\sigma^{(g)}}^*}}{N}\sqrt{
      1+
      \frac{1}{\xi}
      \frac{1+\frac{{{\sigma^{(g)}}^*}^2}{2N}}
           {1+\frac{{{\sigma^{(g)}}^*}^2}{N}}} c_{\mu/\mu,\lambda}
  \right].
  \end{multlined}
  \label{sec:theoreticalanalysis:eq:varphixnormalizedinfeasible3}
\end{align}
From
\Cref{sec:theoreticalanalysis:eq:varphixnormalizedinfeasible2}
to
\Cref{sec:theoreticalanalysis:eq:varphixnormalizedinfeasible3},
$\frac{1}{N}$ has been neglected compared to $1$ as
$N \rightarrow \infty$.
It can be rewritten further yielding
\begin{align}
  &\begin{multlined}
  {\varphi_x}_{\text{infeas}}^* = \frac{N}{1+\xi}
  \left(1-\frac{\sqrt{\xi}r^{(g)}}{x^{(g)}}
    \sqrt{1+\frac{{{\sigma^{(g)}}^*}^2}{N}}\right)\\
    +\frac{\sqrt{\xi}}{1+\xi}
    \frac{\sqrt{\xi}r^{(g)}}{x^{(g)}}
    {\sigma^{(g)}}^* c_{\mu/\mu,\lambda}\sqrt{
      1+
      \frac{1}{\xi}
      \frac{1+\frac{{{\sigma^{(g)}}^*}^2}{2N}}
           {1+\frac{{{\sigma^{(g)}}^*}^2}{N}}}.
  \end{multlined}
  \label{sec:theoreticalanalysis:eq:varphixnormalizedinfeasible5}
\end{align}

Similarly to~\cite[Sec. 3.1.2.1.2.8, pp. 44-49]{SpettelBeyer2018SigmaSaEsCone},
the feasible and infeasible cases can be combined using
the single offspring feasibility probability
\begin{equation}
  \label{sec:theoreticalanalysis:eq:varphixnormalizedcombined}
  \begin{multlined}
  \varphi_x^* \approx
  P_{\text{feas}}(x^{(g)}, r^{(g)}, \sigma^{(g)})
  {\varphi_x}_{\text{feas}}^*\\
  + [1 - P_{\text{feas}}(x^{(g)}, r^{(g)}, \sigma^{(g)})]
  {\varphi_x}_{\text{infeas}}^*.
  \end{multlined}
\end{equation}

\renewcommand{\sppapathtmp}{./figures/figure03/}
\begin{figure}
  \centering
  \begin{tabular}{@{\hspace{-0.0\textwidth}}c@{\hspace{-0.0\textwidth}}c}
    \includegraphics[width=0.43\textwidth]{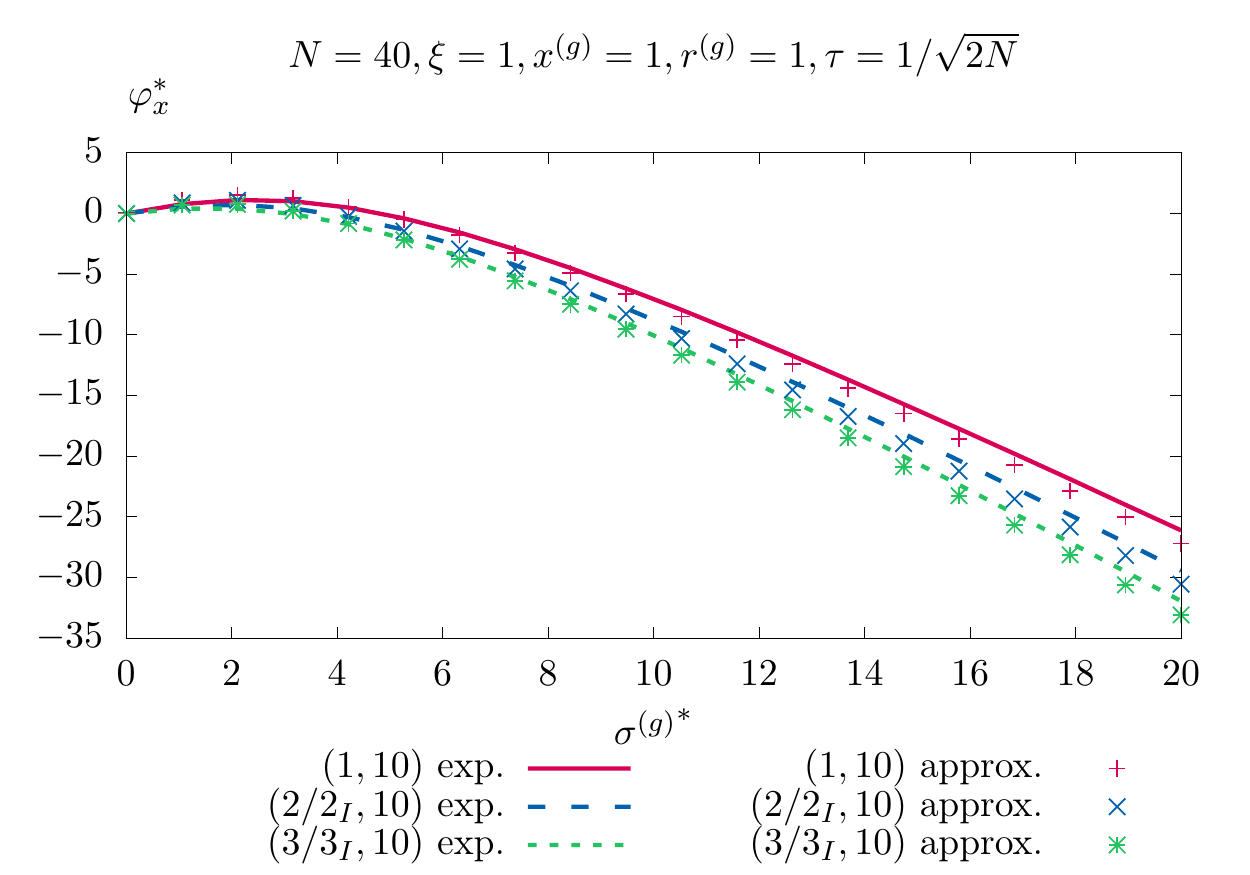}\\
    \includegraphics[width=0.43\textwidth]{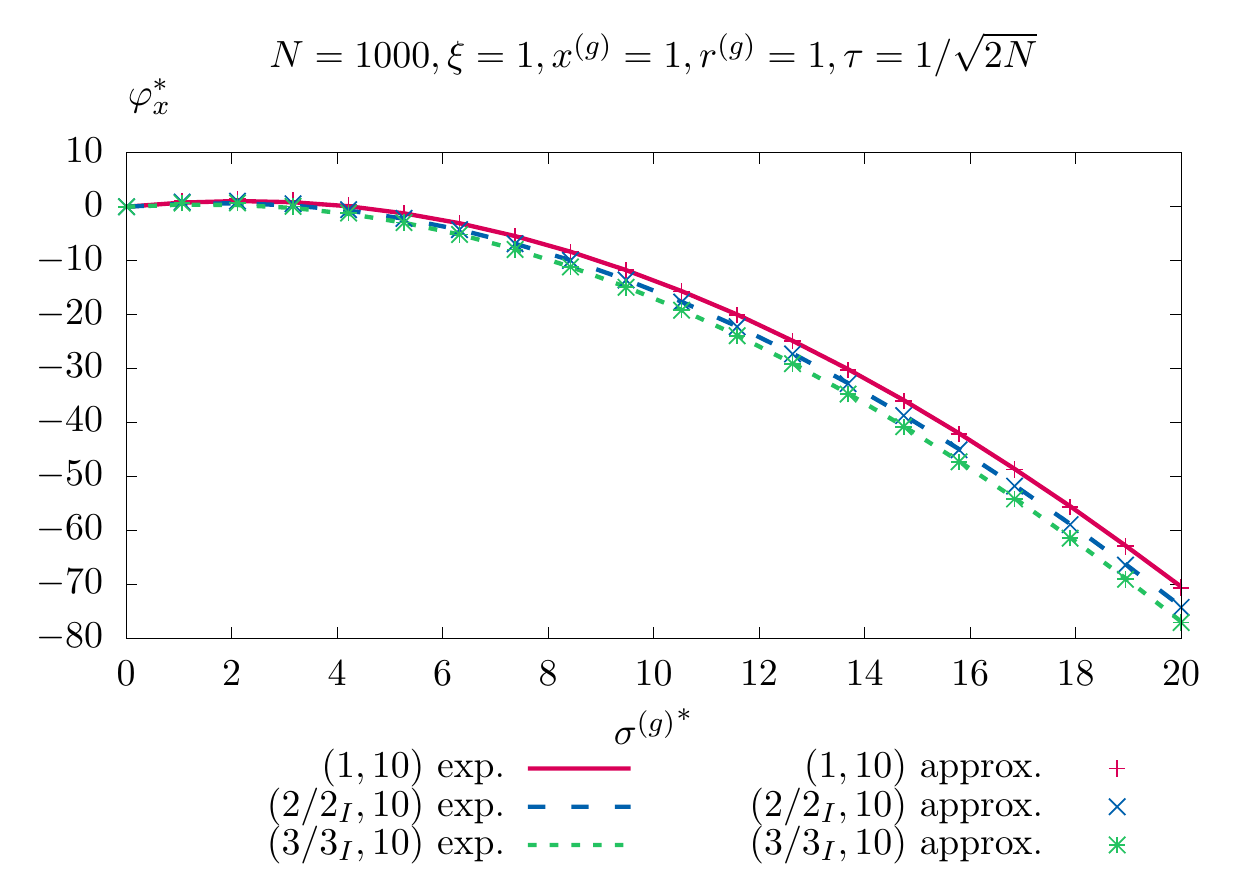}
  \end{tabular}
  \caption{Comparison of the $x$ progress rate approximation with simulations.}
  \label{sec:theoreticalanalysis:fig:xprogresscomparisons}
\end{figure}

\subsubsection{Derivation of the \texorpdfstring{$r$}{\$r\$} Progress Rate}
\label{sec:theoreticalanalysis:subsec:microscopic:r}
From the definition of the progress rate
(\Cref{sec:theoreticalanalysis:eq:varphir})
and the pseudo-code of the
ES (\Cref{sec:algorithm:alg:es},
\cref{sec:algorithm:alg:es:bestqr,sec:algorithm:alg:es:r}),
it follows that
\begin{align}
  \varphi_{r}(x^{(g)}, r^{(g)}, \sigma^{(g)})
  &=r^{(g)} - \mathrm{E}[r^{(g + 1)}\,|\,x^{(g)}, r^{(g)}, \sigma^{(g)}]\\
  &= r^{(g)} - \mathrm{E}[\langle q_r \rangle\,|\,x^{(g)}, r^{(g)}, \sigma^{(g)}].
\end{align}
The detailed derivation is provided in the supplementary material
(\Cref{appendix:subsec:rprogress}) leading to
\begin{align}
  &\begin{multlined}
  \varphi_r^*
  \approx
  P_{\text{feas}}(x^{(g)}, r^{(g)}, \sigma^{(g)})
  N\left(1-\sqrt{1 + \frac{{{\sigma^{(g)}}^*}^2}{\mu N}}\right)\\
  + [1 - P_{\text{feas}}(x^{(g)}, r^{(g)}, \sigma^{(g)})]
  N\left(
  1 -
  \frac
  {x^{(g)}}
  {\sqrt{\xi}r^{(g)}}\right.\\\left.\times
  \left(1-\frac{{\varphi_x^*}_{\text{infeas}}}{N}\right)
  \sqrt{\frac{1+\frac{{{\sigma^{(g)}}^*}^2}{\mu N}}
             {1+\frac{{{\sigma^{(g)}}^*}^2}{N}}}
  \right)
  \end{multlined}
  \label{sec:theoreticalanalysis:eq:varphirnormalizedcombinedmaintext}
\end{align}

\renewcommand{\sppapathtmp}{./figures/figure04/}
\begin{figure}
  \centering
  \begin{tabular}{@{\hspace{-0.0\textwidth}}c@{\hspace{-0.0\textwidth}}c}
    \includegraphics[width=0.43\textwidth]{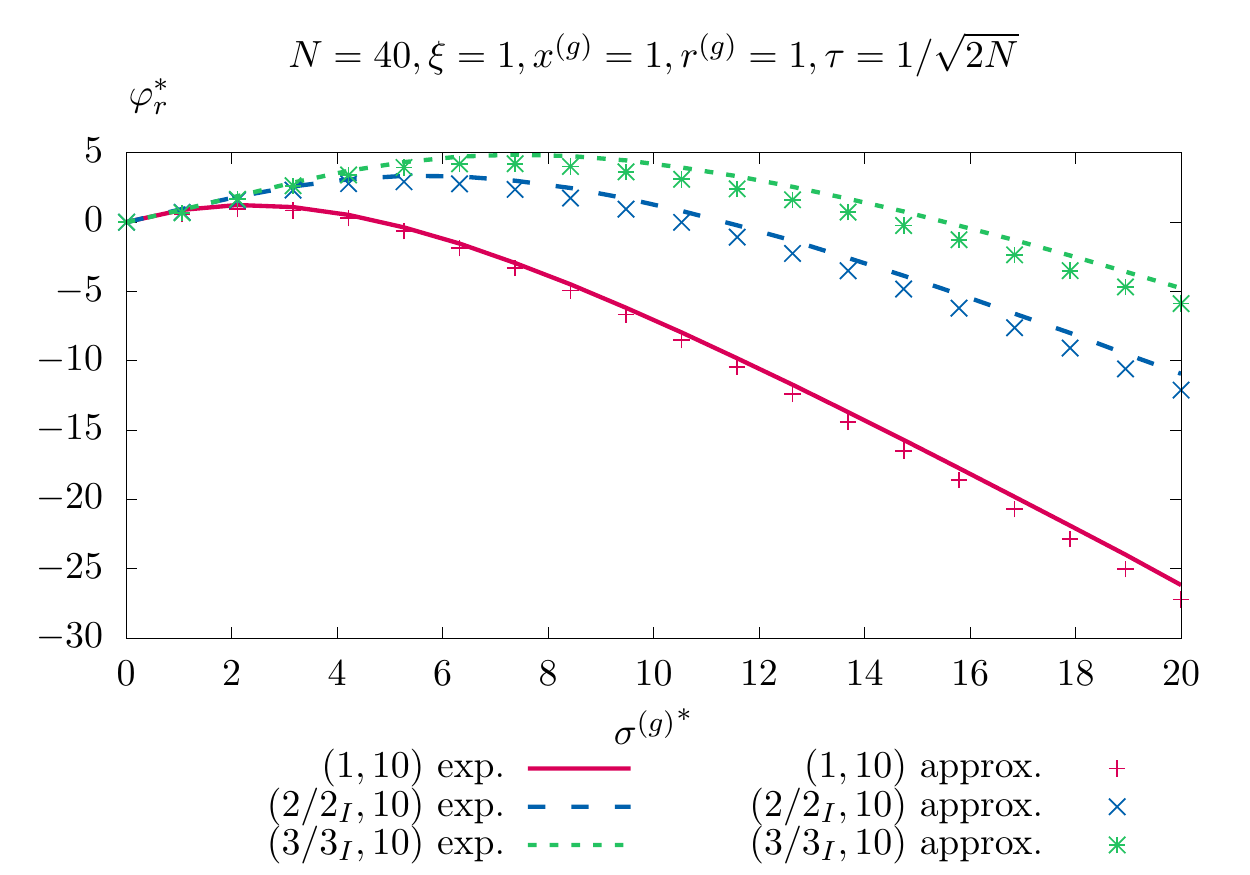}\\
    \includegraphics[width=0.43\textwidth]{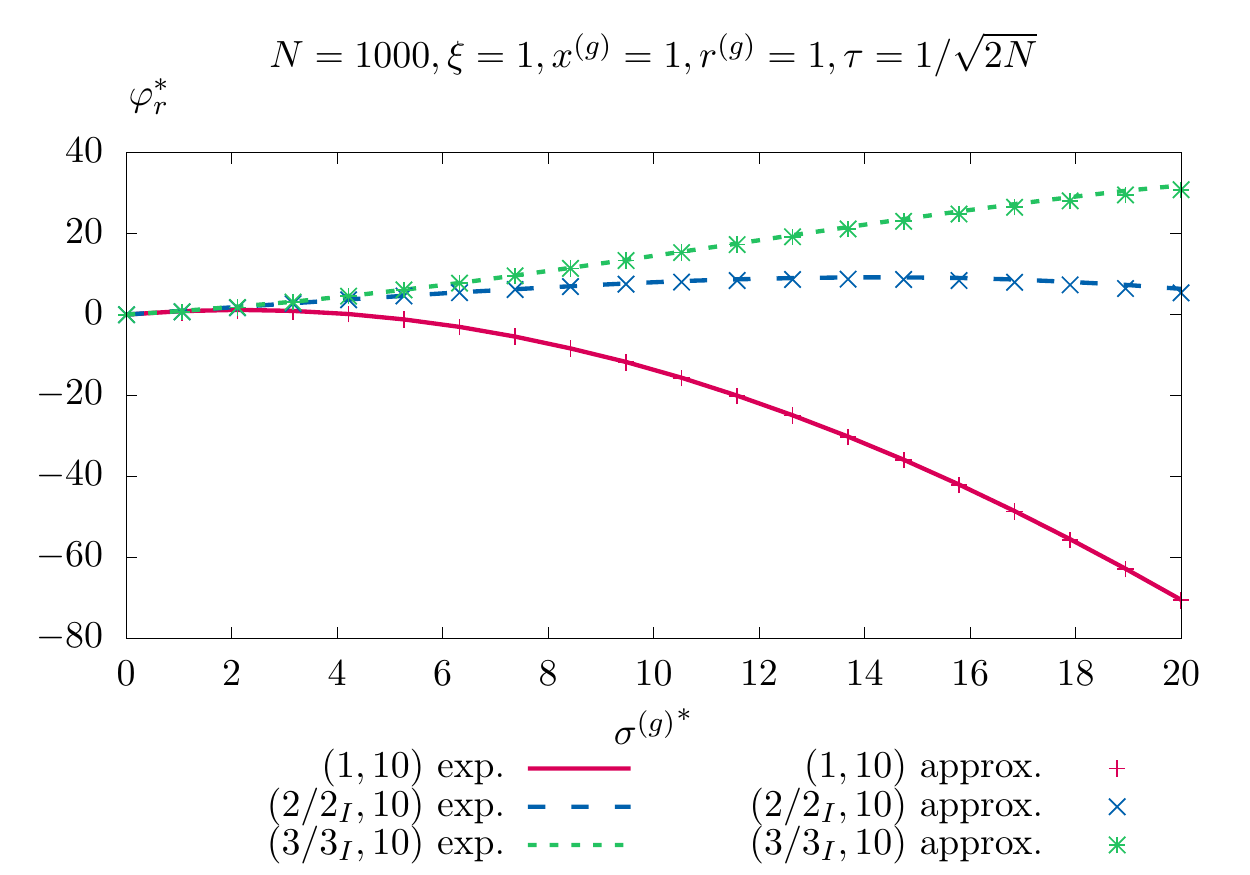}
  \end{tabular}
  \caption{Comparison of the $r$ progress rate approximation with simulations.}
  \label{sec:theoreticalanalysis:fig:rprogresscomparisons}
\end{figure}

\subsubsection{Derivation of the SAR}
\label{sec:theoreticalanalysis:subsec:microscopic:sigma}
With the definition of the SAR
(\Cref{sec:theoreticalanalysis:eq:psi})
and the pseudo-code of the
ES (\Cref{sec:algorithm:alg:es}, \cref{sec:algorithm:alg:es:replacesigma}),
one gets
\begin{align}
  \psi(&x^{(g)}, r^{(g)}, \sigma^{(g)})\\\notag
  &= \mathrm{E}\left[\frac{\sigma^{(g + 1)} - \sigma^{(g)}}{\sigma^{(g)}}
    \,\bigg|\,x^{(g)}, r^{(g)}, \sigma^{(g)}\right]\\
  &= \mathrm{E}\left[\frac{\left(\frac{1}{\mu}\sum_{m=1}^{\mu}
     \tilde{\sigma}_{m;\lambda}\right) -
     \sigma^{(g)}}{\sigma^{(g)}}
    \,\bigg|\,x^{(g)}, r^{(g)}, \sigma^{(g)}\right]\\
  &= \mathrm{E}\left[\frac{1}{\mu}\sum_{m=1}^{\mu}\frac{
     \tilde{\sigma}_{m;\lambda} -
     \sigma^{(g)}}{\sigma^{(g)}}
    \,\bigg|\,x^{(g)}, r^{(g)}, \sigma^{(g)}\right]
\end{align}
An approximation for this expected value is derived in the
supplementary material (\Cref{appendix:subsec:sar})
resulting in
\begin{align}
  &\begin{multlined}
     \psi\approx
     P_{\text{feas}}(x^{(g)}, r^{(g)}, \sigma^{(g)})
     \left[\tau^2\left(\frac{1}{2}+e_{\mu,\lambda}^{1,1}\right)\right]\\
     +[1-P_{\text{feas}}(x^{(g)}, r^{(g)}, \sigma^{(g)})]\\
     \times\left[\tau^2
     \left(\frac{1}{2}
     +e_{\mu,\lambda}^{1,1}
     -\frac{{\sigma^{(g)}}^* c_{\mu/\mu,\lambda}}{\sqrt{1+\xi}}
     \right)\right]
  \end{multlined}\\
  &\begin{multlined}
     \phantom{\psi}=
     \tau^2
     \left[\left(\frac{1}{2}+e_{\mu,\lambda}^{1,1}\right)\right]
     -[1-P_{\text{feas}}(x^{(g)}, r^{(g)}, \sigma^{(g)})]\\
     \times\frac{{\sigma^{(g)}}^* c_{\mu/\mu,\lambda}}{\sqrt{1+\xi}}.
  \end{multlined}
  \label{sec:theoreticalanalysis:eq:psicombinedmaintext}
\end{align}

\Cref{sec:theoreticalanalysis:fig:xprogresscomparisons,%
  sec:theoreticalanalysis:fig:rprogresscomparisons,%
  sec:theoreticalanalysis:fig:sarcomparisons}
show plots comparing the derived closed-form
approximation for $\varphi_x^*$, $\varphi_r^*$, and $\psi$,
respectively, with one-generation experiments.
The pluses, crosses, and stars have been calculated by evaluating
\Cref{sec:theoreticalanalysis:eq:varphixnormalizedcombined}
with
\Cref{sec:theoreticalanalysis:eq:Pfeasapprox1,%
sec:theoreticalanalysis:eq:varphixnormalizedfeasible5,%
sec:theoreticalanalysis:eq:varphixnormalizedinfeasible5},
\Cref{sec:theoreticalanalysis:eq:varphirnormalizedcombinedmaintext}
with
\Cref{sec:theoreticalanalysis:eq:Pfeasapprox1,%
sec:theoreticalanalysis:eq:varphixnormalizedinfeasible5}, and
\Cref{sec:theoreticalanalysis:eq:psicombinedmaintext}
with
\Cref{sec:theoreticalanalysis:eq:Pfeasapprox1},
respectively.
The solid, dashed, and dotted lines have been
generated by one-generation experiments.
For this, the generational loop has been
run $10^5$ times for a fixed parental individual and constant parameters.
The experimentally determined values for
$\varphi_x^*$, $\varphi_r^*$, and $\psi$
from those $10^5$ runs have been averaged.
Further figures showing comparisons for additional configurations
are provided in the supplementary material
(\Crefrange{sec:theoreticalanalysis:fig:xprogresscomparisonsdetail1}
{sec:theoreticalanalysis:fig:sarcomparisonsdetail3}
in \Cref{sec:algorithm:subsec:additionalplots}).

\renewcommand{\sppapathtmp}{./figures/figure05/}
\begin{figure}
  \centering
  \begin{tabular}{@{\hspace{-0.0\textwidth}}c@{\hspace{-0.0\textwidth}}c}
    \includegraphics[width=0.43\textwidth]{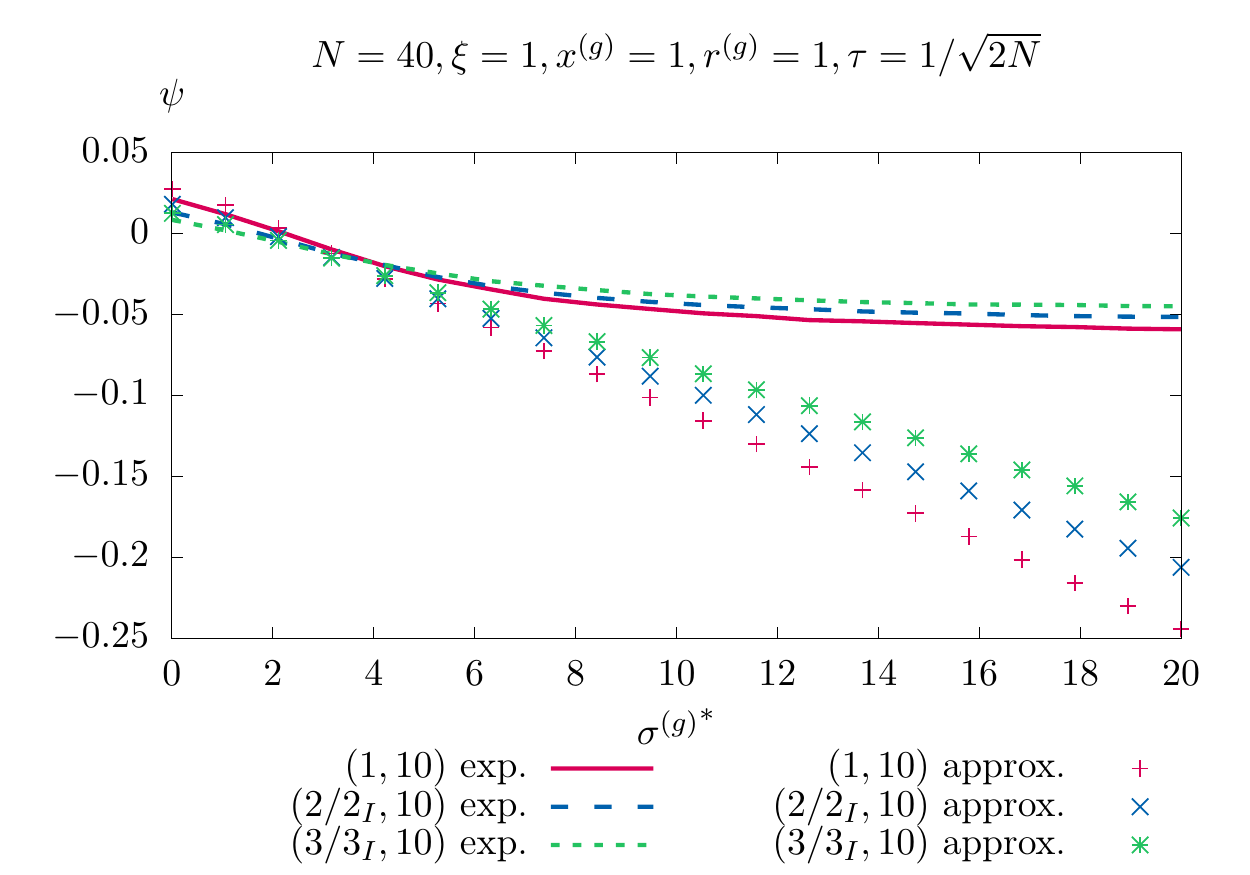}\\
    \includegraphics[width=0.43\textwidth]{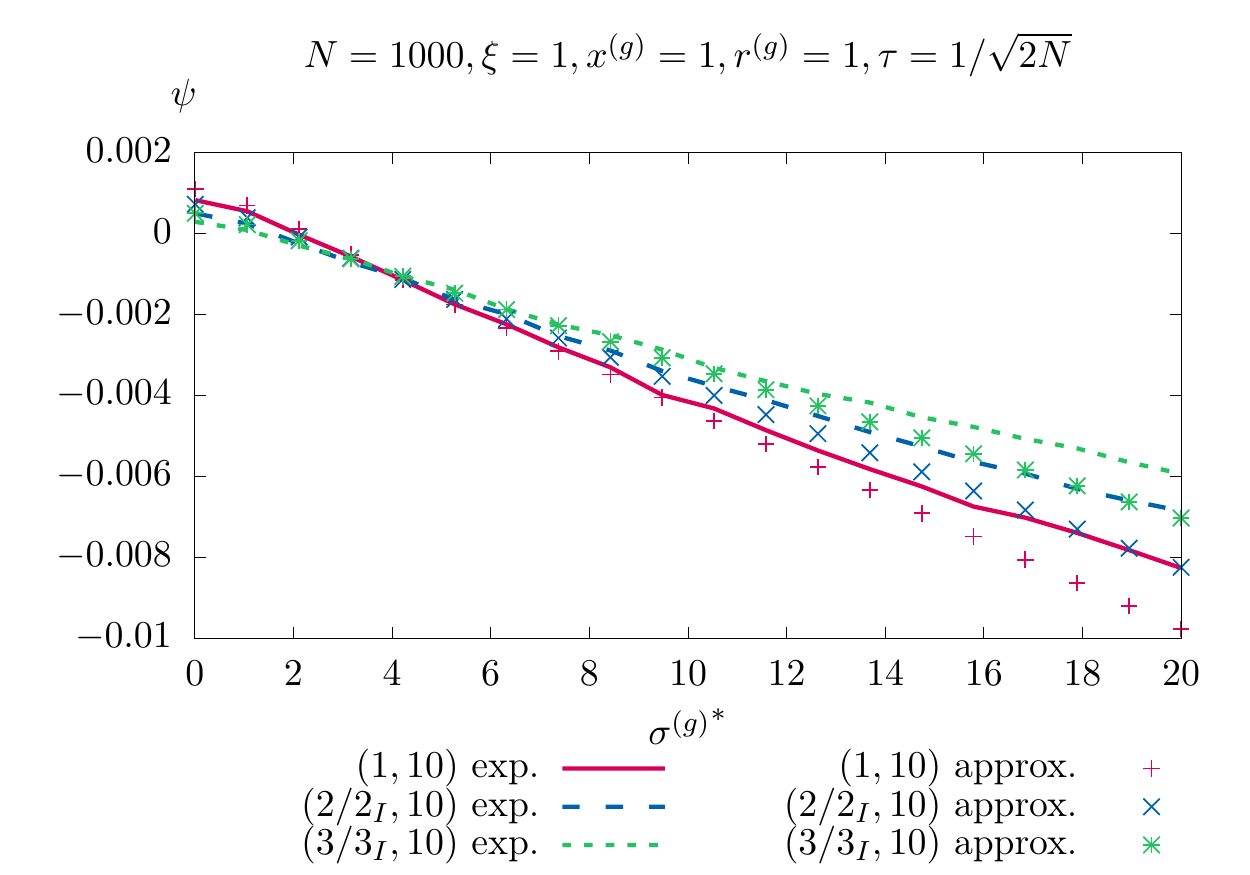}
  \end{tabular}
  \caption{Comparison of the SAR approximation with simulations.}
  \label{sec:theoreticalanalysis:fig:sarcomparisons}
\end{figure}

\subsection{The Evolution Dynamics in the Deterministic Approximation}
\label{sec:theoreticalanalysis:subsec:evolutiondynamics}

\subsubsection{The Evolution Equations}
With the derived progress rates $\varphi_x^*$ and $\varphi_r^*$ and the
SAR $\psi$ from
\Cref{sec:theoreticalanalysis:subsec:microscopic},
the approximated deterministic evolution equations can be iterated.
By doing this, the dynamics can be predicted. The predictions by
the approximations are compared to real runs of the ES.
Using
\Crefrange{sec:theoreticalanalysis:eq:evolutionequationx}
          {sec:theoreticalanalysis:eq:evolutionequationsigma}
together with
\Crefrange{sec:theoreticalanalysis:eq:varphixnormalized}
          {sec:theoreticalanalysis:eq:sigmanormalized},
the evolution equations read
\begin{align}
  x^{(g + 1)} &= x^{(g)} -
  \frac{x^{(g)}{\varphi^{(g)}_{x}}^*}{N}
  =x^{(g)}\left(1-\frac{{\varphi^{(g)}_{x}}^*}{N}\right)
  \label{sec:theoreticalanalysis:eq:meanvalueiterativesystemx}\\
  r^{(g + 1)} &= r^{(g)} -
  \frac{r^{(g)}{\varphi^{(g)}_r}^*}{N}
  =r^{(g)}\left(1-\frac{{\varphi^{(g)}_r}^*}{N}\right)
  \label{sec:theoreticalanalysis:eq:meanvalueiterativesystemr}\\
  \sigma^{(g + 1)} &= \sigma^{(g)} +
  \sigma^{(g)}\psi^{(g)}
  =\sigma^{(g)}\left(1+\psi^{(g)}\right).
  \label{sec:theoreticalanalysis:eq:meanvalueiterativesystemsigma}
\end{align}

\begin{figure}
  \centering
  \begin{tabular}{@{\hspace{-0.0\textwidth}}c@{\hspace{-0.0\textwidth}}c}
    \includegraphics[width=0.43\textwidth]{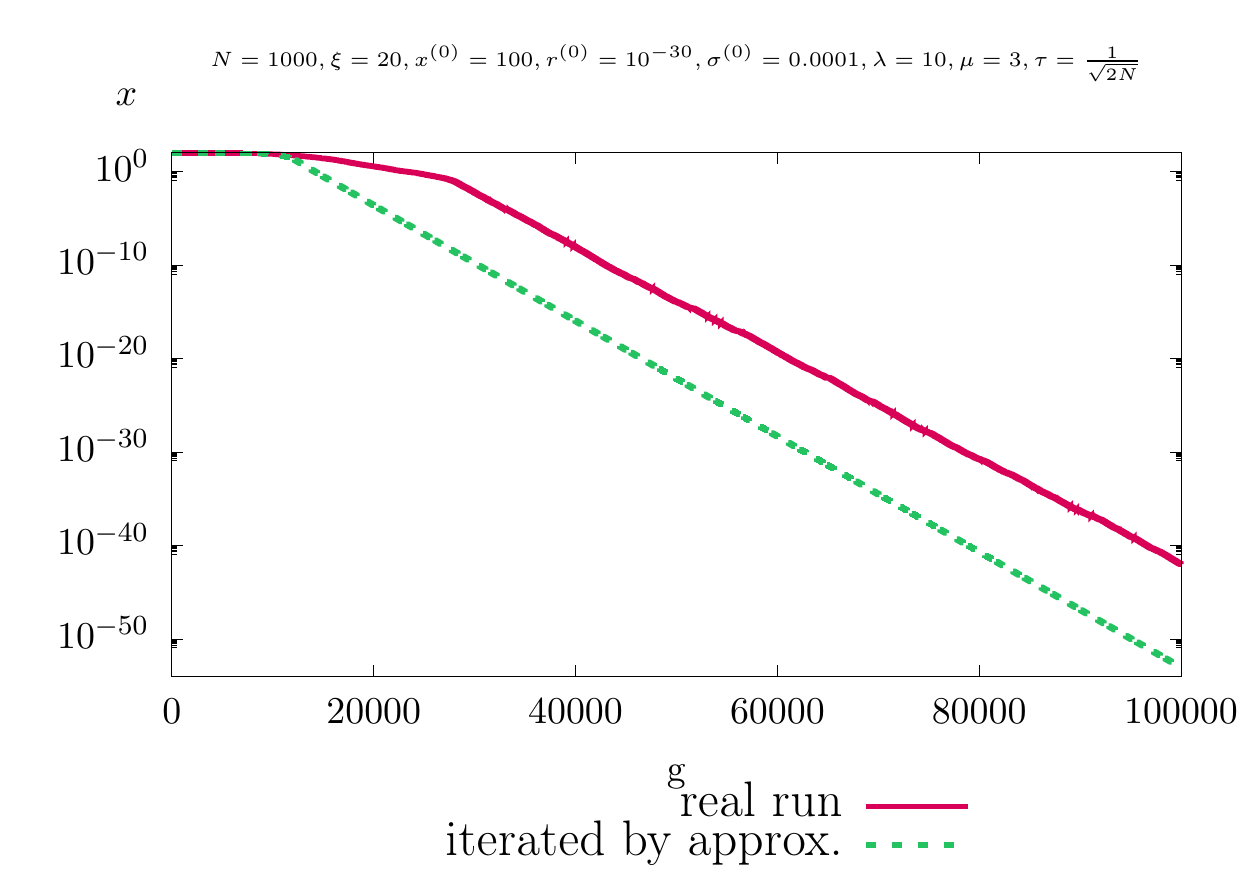}\\
    \includegraphics[width=0.43\textwidth]{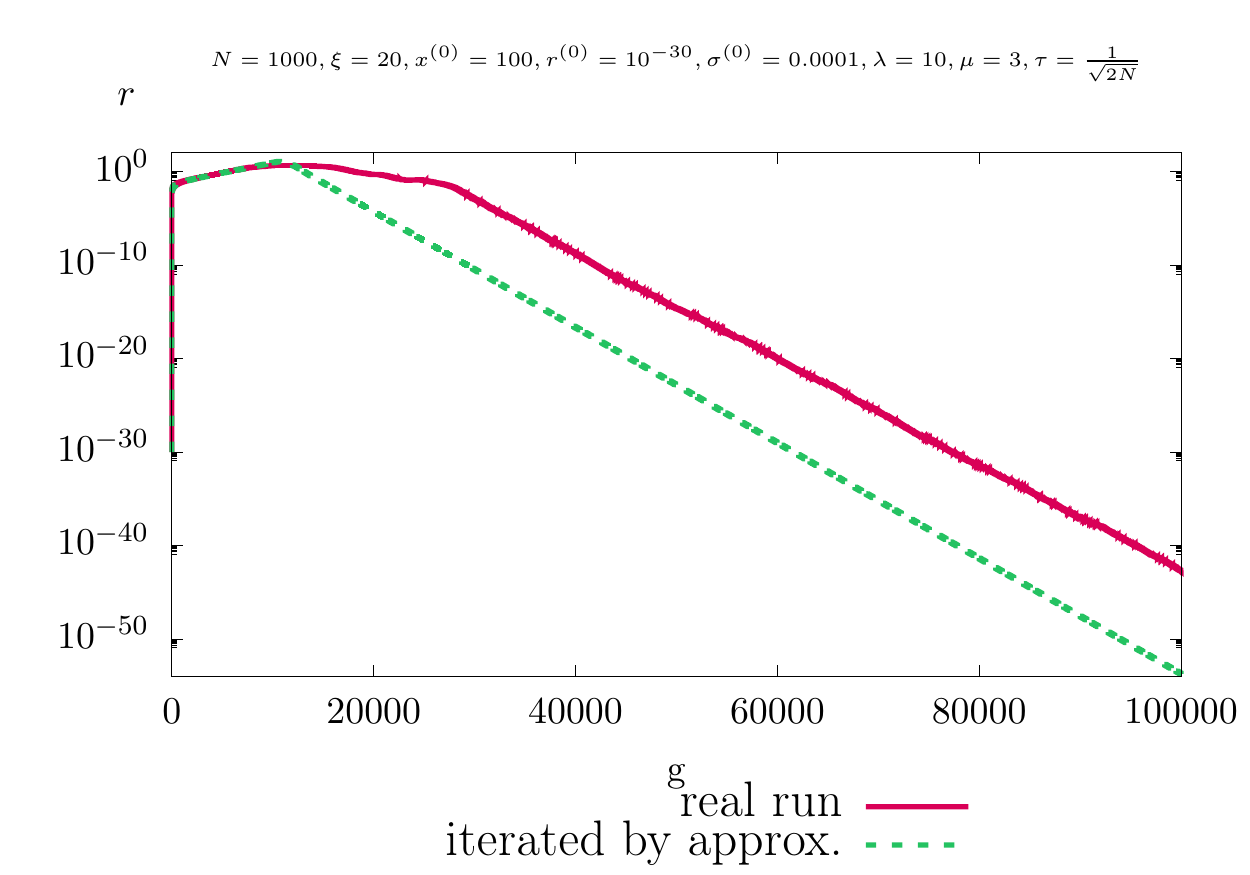}\\
    \includegraphics[width=0.43\textwidth]{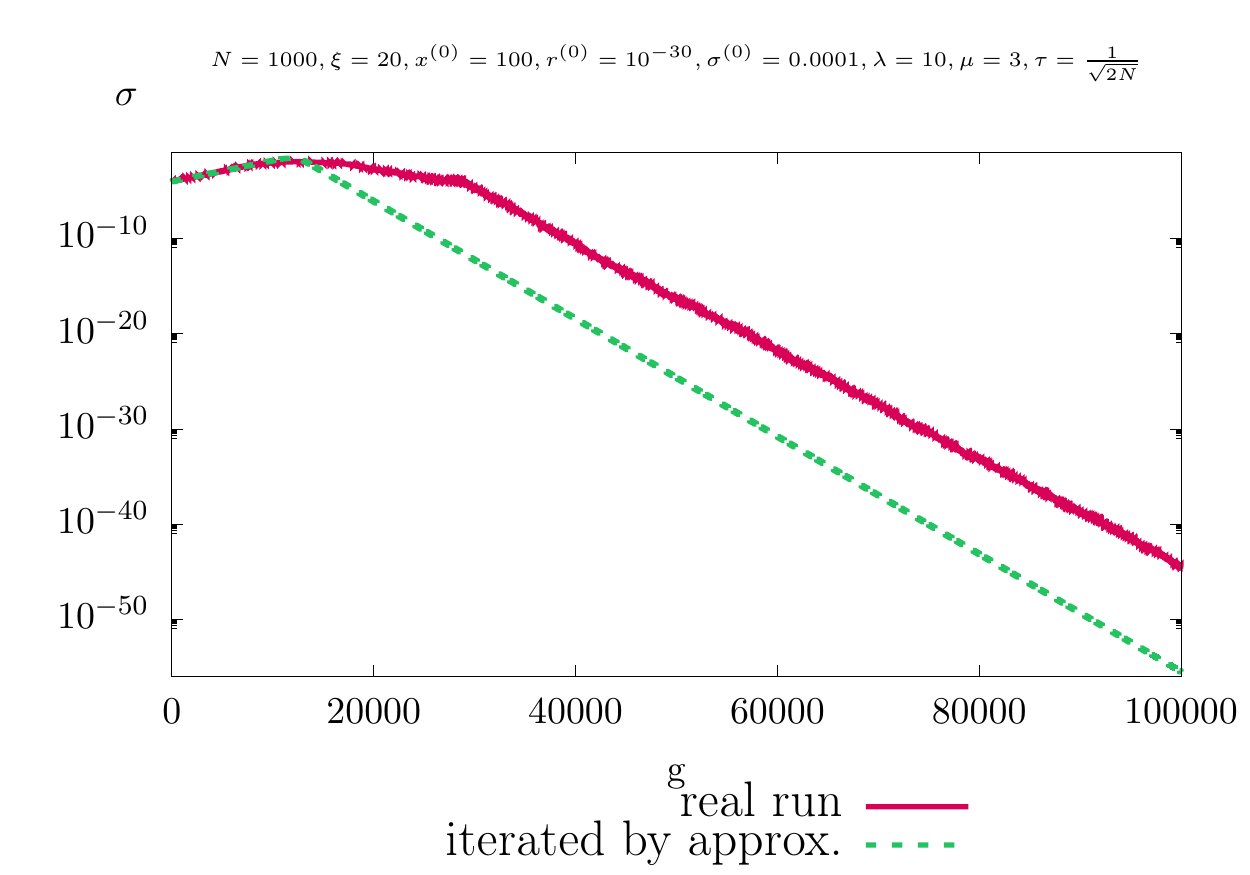}\\
    \includegraphics[width=0.43\textwidth]{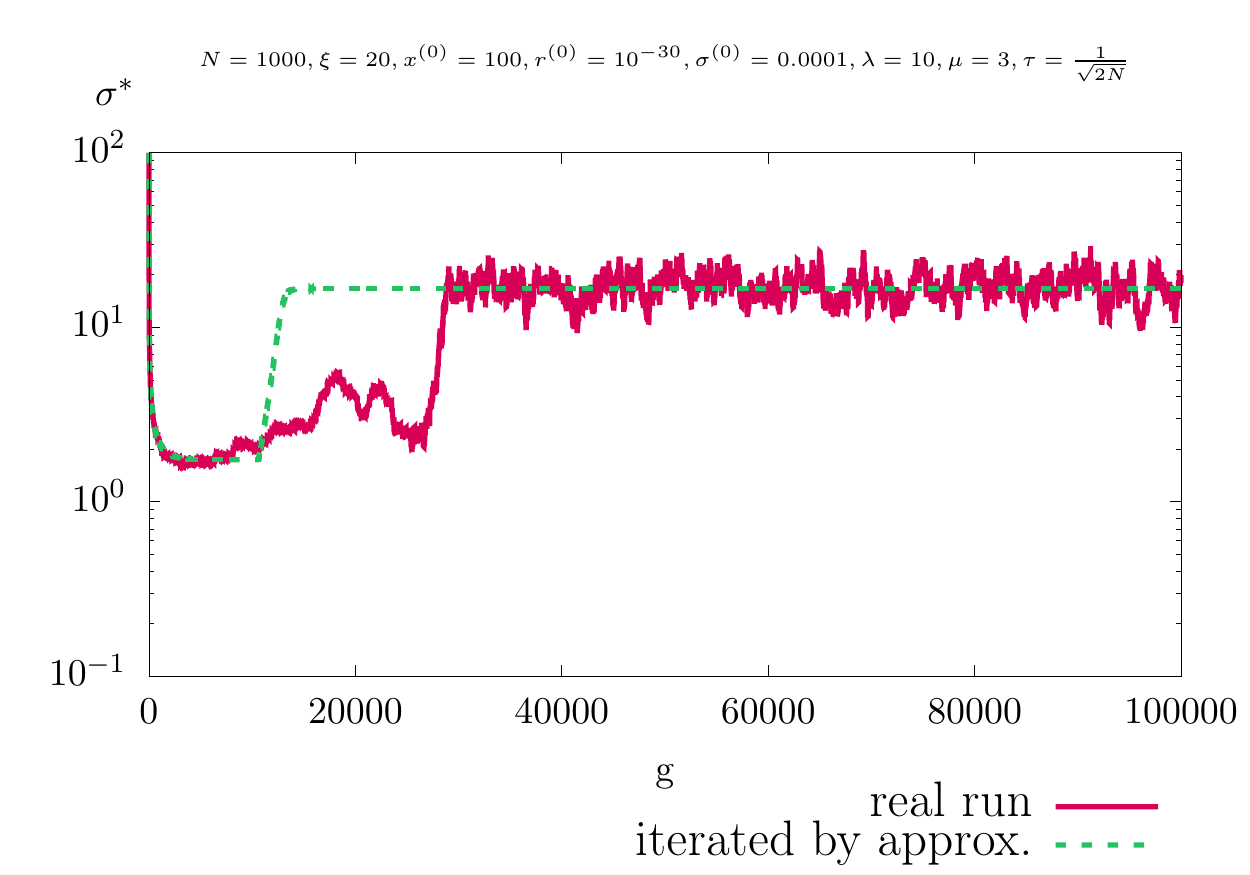}
  \end{tabular}
  \caption{Mean value dynamics closed-form approximation and real-run
    comparison of the
    $(3/3_I,10)$-ES
    with repair by projection
    applied to the conically constrained problem ($N=1000$).}
  \label{sec:theoreticalanalysis:fig:dynamics}
\end{figure}

\Cref{sec:theoreticalanalysis:fig:dynamics}
shows the mean value dynamics of the $(3/3_I,10)$-ES
applied to the conically constrained problem for
$N=1000$, $\xi=20$, and $\tau=\frac{1}{\sqrt{2N}}$.
Additional figures for more configurations are provided
in the supplementary material
(\Cref{sec:theoreticalanalysis:fig:dynamicsdetail1,%
sec:theoreticalanalysis:fig:dynamicsdetail2,%
sec:theoreticalanalysis:fig:dynamicsdetail3,%
sec:theoreticalanalysis:fig:dynamicsdetail4,%
sec:theoreticalanalysis:fig:dynamicsdetail5,%
sec:theoreticalanalysis:fig:dynamicsdetail6}
in \Cref{sec:algorithm:subsec:additionalplots}).
The lines for the real runs have been generated
by averaging $100$ real runs of the ES.
The lines for the iteration by approximation have been
computed by iterating the mean value iterative system
with the derived approximations in
\Cref{sec:theoreticalanalysis:subsec:microscopic}
for ${\varphi^{(g)}_{x}}^*$, ${\varphi^{(g)}_{r}}^*$, and $\psi^{(g)}$.
Note that due to the approximations used
it is possible that the iteration of the mean value
iterative system yields infeasible
$(x^{(g)},r^{(g)})^T$ in some generation $g$.
If this situation has occurred, the corresponding $(x^{(g)},r^{(g)})^T$
have been projected back. Then, in the following iterations,
the projected values have been used. This is what is indicated
by the repair step in
\crefrange{sec:algorithm:alg:es:feasibilitycheckcentroid}
          {sec:algorithm:alg:es:feasibilitycheckendcentroid}
of \Cref{sec:algorithm:alg:es}
that is only needed for the iteration of the mean value
iterative system.

The stationary state is reached by the ES (solid line)
later than predicted (dotted line).
This can be seen in the first three subplots of
\Cref{sec:theoreticalanalysis:fig:dynamics}. The slope of the
lines coincides well between the prediction and the ES. But the
lines are shifted for some number of generations. The same can be
observed in the fourth subplot of \Cref{sec:theoreticalanalysis:fig:dynamics}
where the $\sigma^*$ steady state value is attained later by the ES
(solid line) than predicted (dotted line). Further experiments
to investigate this behavior (not shown here) showed that higher
values of $\tau$ reduce this discrepancy by allowing larger $\sigma$-mutations.

\renewcommand{\sppapathtmp}{./figures/figure06/}

\subsubsection{The ES in the Stationary State}
The state of the $(\mu/\mu_I,\lambda)$-ES on the conically
constrained problem is completely described by
$(x^{(g)},r^{(g)},\sigma^{(g)})^T$
(assuming constant exogenous parameters).
For sufficiently large $g$, the ES transitions into
a stationary state. On average,
the normalized mutation strength should be constant in
this steady state. That is, in the steady state
the expectation of the
normalized mutation strength $\sigma^*_{ss}$ is constant for large
time scales
\begin{equation}
  \sigma^*_{ss} := \lim_{g\rightarrow\infty}\,{\sigma^{(g)}}^*.
\end{equation}
Hence, for sufficiently large $g$,
${\sigma^{(g)}}^*={\sigma^{(g+1)}}^*=\sigma^*_{ss}$.
Requiring this condition, a steady state condition can be derived.
To this end,
\Cref{sec:theoreticalanalysis:eq:meanvalueiterativesystemsigma}
is normalized yielding
\begin{equation}
  \frac{r^{(g + 1)}{\sigma^{(g + 1)}}^*}{N} =
  \frac{r^{(g)}{\sigma^{(g)}}^*}{N}
  \left(1+\psi^{(g)}\right).
\end{equation}
Setting ${\sigma^{(g)}}^*={\sigma^{(g+1)}}^*$ results in
\begin{equation}
  r^{(g + 1)} = r^{(g)}\left(1+\psi^{(g)}\right).
\end{equation}
Inserting
\Cref{sec:theoreticalanalysis:eq:meanvalueiterativesystemr}
yields the steady state condition
\begin{align}
  r^{(g)}\left(1-\frac{{\varphi^{(g)}_r}^*}{N}\right)
  &= r^{(g)}\left(1+\psi^{(g)}\right)
  \label{sec:theoreticalanalysis:eq:steadystateeq1}\\
  1-\frac{{\varphi^{(g)}_r}^*}{N}
  &= 1+\psi^{(g)}
  \label{sec:theoreticalanalysis:eq:steadystateeq2}\\
  \frac{{\varphi^{(g)}_r}^*}{N}
  &= -\psi^{(g)}.
  \label{sec:theoreticalanalysis:eq:steadystateeq3}
\end{align}

In the steady state, the ES moves in the vicinity of the cone
boundary. This can be observed in real ES runs
(see the third rows in \Cref{sec:theoreticalanalysis:fig:dynamicsdetail1,%
sec:theoreticalanalysis:fig:dynamicsdetail2,%
sec:theoreticalanalysis:fig:dynamicsdetail3,%
sec:theoreticalanalysis:fig:dynamicsdetail4,%
sec:theoreticalanalysis:fig:dynamicsdetail5,%
sec:theoreticalanalysis:fig:dynamicsdetail6} in the supplementary
material \Cref{sec:algorithm:subsec:additionalplots}).
Hence, for the derivation of analytical approximations for the steady
state, the condition $P_{\text{feas}} \approx 0$ is assumed similar to
the $(1,\lambda)$ case. However, in contrast to the $(1,\lambda)$ case,
the assumption $\frac{x^{(g)}}{\sqrt{\xi}r^{(g)}} = 1$ is too crude.
There exists a distance from the cone boundary for some parameter
configurations that cannot be neglected
(this can for example be seen in the third row of
\Cref{sec:theoreticalanalysis:fig:dynamicsdetail3} in the supplementary
material \Cref{sec:algorithm:subsec:additionalplots}).
Considering the distance ratio $r^{(g)}/x^{(g)}$, one observes that
this ratio approaches a steady state value.
This results in the condition
\begin{equation}
  \frac{r^{(g)}}{x^{(g)}} = \frac{r^{(g+1)}}{x^{(g+1)}}
\end{equation}
for sufficiently large values of $g$.
Using the progress rates
(\Crefrange{sec:theoreticalanalysis:eq:varphix}
           {sec:theoreticalanalysis:eq:varphirnormalized}),
this can be written as
\begin{equation}
  \frac{r_{ss}}{x_{ss}} =
  \frac{r_{ss}\left(1-\frac{{\varphi_r}_{ss}^*}{N}\right)}
       {x_{ss}\left(1-\frac{{\varphi_x}_{ss}^*}{N}\right)},
\end{equation}
which implies
\begin{equation}
  {\varphi_r}_{ss}^*={\varphi_x}_{ss}^*.
  \label{sec:theoreticalanalysis:eq:steadystatedist1}
\end{equation}
With the assumption $P_{\text{feas}} \approx 0$, use of the
infeasible case approximations
(\Cref{sec:theoreticalanalysis:eq:varphixnormalizedinfeasible5}
and
\Cref{sec:theoreticalanalysis:eq:varphirnormalizedinfeas})
for treating
\Cref{sec:theoreticalanalysis:eq:steadystatedist1}
further, yields
\begin{align}
  &\begin{multlined}
  N\bigg(
  1 -
  \frac
  {x_{ss}}
  {\sqrt{\xi}r_{ss}}
  \left(1-\frac{{{\varphi_x^*}_{ss}}_{\text{infeas}}}{N}\right)
  \sqrt{\frac{1+\frac{{{\sigma_{ss}^*}}^2}{\mu N}}
             {1+\frac{{{\sigma_{ss}^*}}^2}{N}}}
  \bigg)
  ={{\varphi_x^*}_{ss}}_{\text{infeas}}
  \end{multlined}\\
  &\begin{multlined}
  N-\left(
  \frac
  {x_{ss}}
  {\sqrt{\xi}r_{ss}}
  \left(N-{{\varphi_x^*}_{ss}}_{\text{infeas}}\right)
  \sqrt{\frac{1+\frac{{{\sigma_{ss}^*}}^2}{\mu N}}
             {1+\frac{{{\sigma_{ss}^*}}^2}{N}}}
  \right)
  ={{\varphi_x^*}_{ss}}_{\text{infeas}}.
  \end{multlined}
\end{align}
Further,
\begin{align}
  &\begin{multlined}
  \left(
  \frac
  {x_{ss}}
  {\sqrt{\xi}r_{ss}}
  \left(N-{{\varphi_x^*}_{ss}}_{\text{infeas}}\right)
  \sqrt{\frac{1+\frac{{{\sigma_{ss}^*}}^2}{\mu N}}
             {1+\frac{{{\sigma_{ss}^*}}^2}{N}}}
  \right)
  =
  \left(N-{{\varphi_x^*}_{ss}}_{\text{infeas}}\right)
  \end{multlined}\\
  &\begin{multlined}
  \hspace{5cm}\frac
  {x_{ss}}
  {\sqrt{\xi}r_{ss}}
  =
  \frac{1}{\sqrt{\frac{1+\frac{{{\sigma_{ss}^*}}^2}{\mu N}}
      {1+\frac{{{\sigma_{ss}^*}}^2}{N}}}}
  \end{multlined}
  \label{sec:theoreticalanalysis:eq:steadystatedist2}
\end{align}
follows.
Considering
\Cref{sec:theoreticalanalysis:eq:steadystatedist1}
for
\Cref{sec:theoreticalanalysis:eq:steadystateeq3}
results in
\begin{equation}
  \frac{{\varphi_x}^*_{ss}}{N} = -\psi_{ss}.
  \label{sec:theoreticalanalysis:eq:steadystateeq4}
\end{equation}
Assuming $P_{\text{feas}} \approx 0$, the approximations for the infeasible
case can be used. Insertion of
\Cref{sec:theoreticalanalysis:eq:steadystatedist2}
into
\Cref{sec:theoreticalanalysis:eq:varphixnormalizedinfeasible5}
with the assumption
$\frac{1}{\xi}
      \frac{1+\frac{{{\sigma_{ss}^*}}^2}{2N}}
           {1+\frac{{{\sigma_{ss}^*}}^2}{N}} \simeq \frac{1}{\xi}$
for $N \rightarrow \infty$ yields
\begin{align}
  &\begin{multlined}
  {\varphi_x^*}_{ss}
  \approx
  \frac{N}{1+\xi}
  \left(1-\sqrt{\frac{1+\frac{{{\sigma_{ss}^*}}^2}{\mu N}}
                     {1+\frac{{{\sigma_{ss}^*}}^2}{N}}}
    \sqrt{1+\frac{{{\sigma_{ss}^*}}^2}{N}}\right)\\
    +\frac{{\sigma_{ss}^*} c_{\mu/\mu,\lambda}}{\sqrt{1+\xi}}
    \sqrt{\frac{1+\frac{{{\sigma_{ss}^*}}^2}{\mu N}}
               {1+\frac{{{\sigma_{ss}^*}}^2}{N}}}
  \end{multlined}
  \label{sec:theoreticalanalysis:eq:varphixnormalizedss1}\\
  &\begin{multlined}
  \phantom{{\varphi_x^*}_{ss}}
  =
  \frac{N}{1+\xi}
  \left(1-\sqrt{1+\frac{{{\sigma_{ss}^*}}^2}{\mu N}}\right)
    +\frac{{\sigma_{ss}^*} c_{\mu/\mu,\lambda}}{\sqrt{1+\xi}}
    \sqrt{\frac{1+\frac{{{\sigma_{ss}^*}}^2}{\mu N}}
               {1+\frac{{{\sigma_{ss}^*}}^2}{N}}}
  \end{multlined}
  \label{sec:theoreticalanalysis:eq:varphixnormalizedss2}\\
  &\begin{multlined}
  \phantom{{\varphi_x^*}_{ss}}
  \simeq
  \frac{N}{1+\xi}
  \left(1-\left(1+\frac{{{\sigma_{ss}^*}}^2}{2 \mu N}\right)\right)
    +\frac{{\sigma_{ss}^*} c_{\mu/\mu,\lambda}}{\sqrt{1+\xi}}
  \end{multlined}
  \label{sec:theoreticalanalysis:eq:varphixnormalizedss3}\\
  &\begin{multlined}
  \phantom{{\varphi_x^*}_{ss}}
  =
  \frac{N}{1+\xi}
  \left(-\frac{{{\sigma_{ss}^*}}^2}{2 \mu N}\right)
  +\frac{{\sigma_{ss}^*} c_{\mu/\mu,\lambda}}{\sqrt{1+\xi}}.
  \end{multlined}
  \label{sec:theoreticalanalysis:eq:varphixnormalizedss4}
\end{align}
In the step from
\Cref{sec:theoreticalanalysis:eq:varphixnormalizedss2}
to
\Cref{sec:theoreticalanalysis:eq:varphixnormalizedss3},
it has been assumed that
$\sqrt{\frac{1+\frac{{{\sigma_{ss}^*}}^2}{\mu N}}
    {1+\frac{{{\sigma_{ss}^*}}^2}{N}}} \simeq 1$
for $N \gg {\sigma_{ss}^*}^2$ and
a Taylor expansion (neglecting the quadratic and higher order terms)
has been applied to $\sqrt{1+\frac{{{\sigma_{ss}^*}}^2}{\mu N}}$.
With
\Cref{sec:theoreticalanalysis:eq:varphixnormalizedss4},
the steady state equation
\Cref{sec:theoreticalanalysis:eq:steadystateeq3}
reads for the asymptotic case of $N \rightarrow \infty$
\begin{equation}
  \begin{multlined}
  \frac{1}{1+\xi}
  \left(-\frac{{{\sigma_{ss}^*}}^2}{2 \mu N}\right)
    +\frac{{\sigma_{ss}^*} c_{\mu/\mu,\lambda}}{N\sqrt{1+\xi}}\\
  \hspace{2cm}=-\tau^2
  \left(\frac{1}{2}
  +e_{\mu,\lambda}^{1,1}
  -\frac{{\sigma_{ss}^*} c_{\mu/\mu,\lambda}}{\sqrt{1+\xi}}
  \right).
  \end{multlined}
\end{equation}
Solving this quadratic equation yields the steady state normalized mutation
strength (as the mutation strength is positive, the positive root is taken)
\begin{equation}
  \begin{multlined}
  {\sigma_{ss}^*} =
  \sqrt{1+\xi}\mu c_{\mu/\mu,\lambda}
  \left[(1-N\tau^2)+\right.\\\left.\hspace{2cm}
  \sqrt{(1-N\tau^2)^2+
        \frac{2\tau^2N\left(\frac{1}{2}+e_{\mu,\lambda}^{1,1}\right)}
             {\mu c_{\mu/\mu,\lambda}^2}}\right].
  \end{multlined}
  \label{sec:theoreticalanalysis:eq:steadystateroot}
\end{equation}

\begin{figure}
  \centering
  \begin{tabular}{@{\hspace{-0.0\textwidth}}c}
    \includegraphics[width=0.43\textwidth]{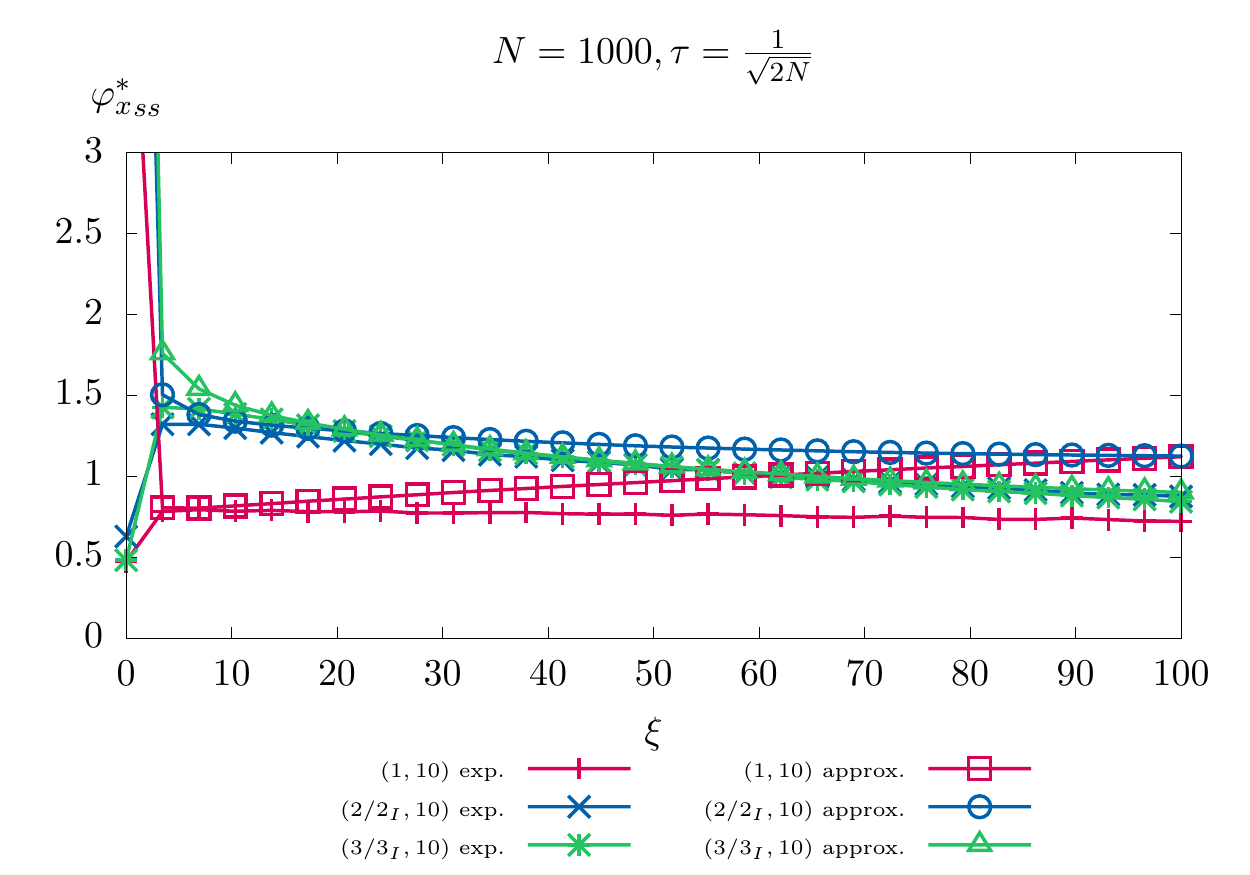}\\
    \includegraphics[width=0.43\textwidth]{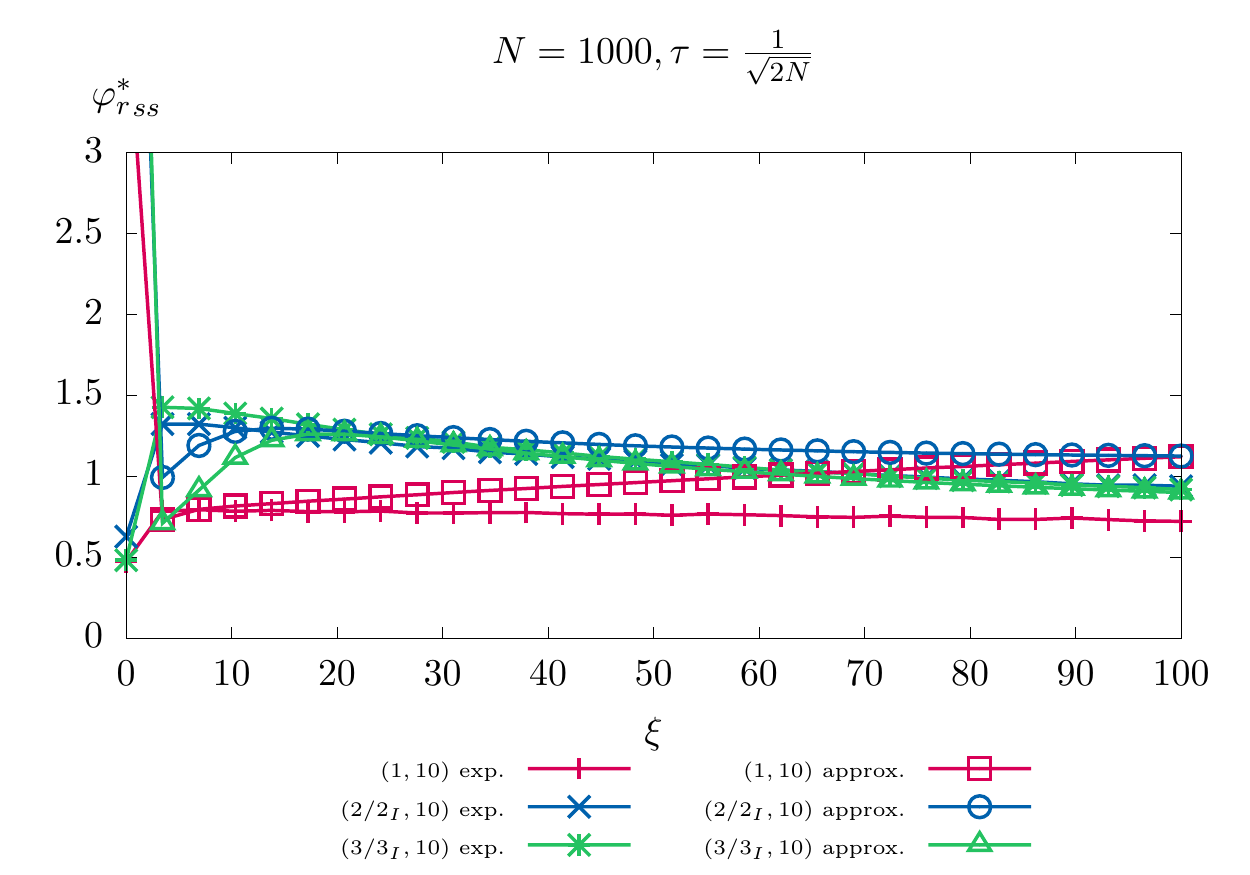}\\
    \includegraphics[width=0.43\textwidth]{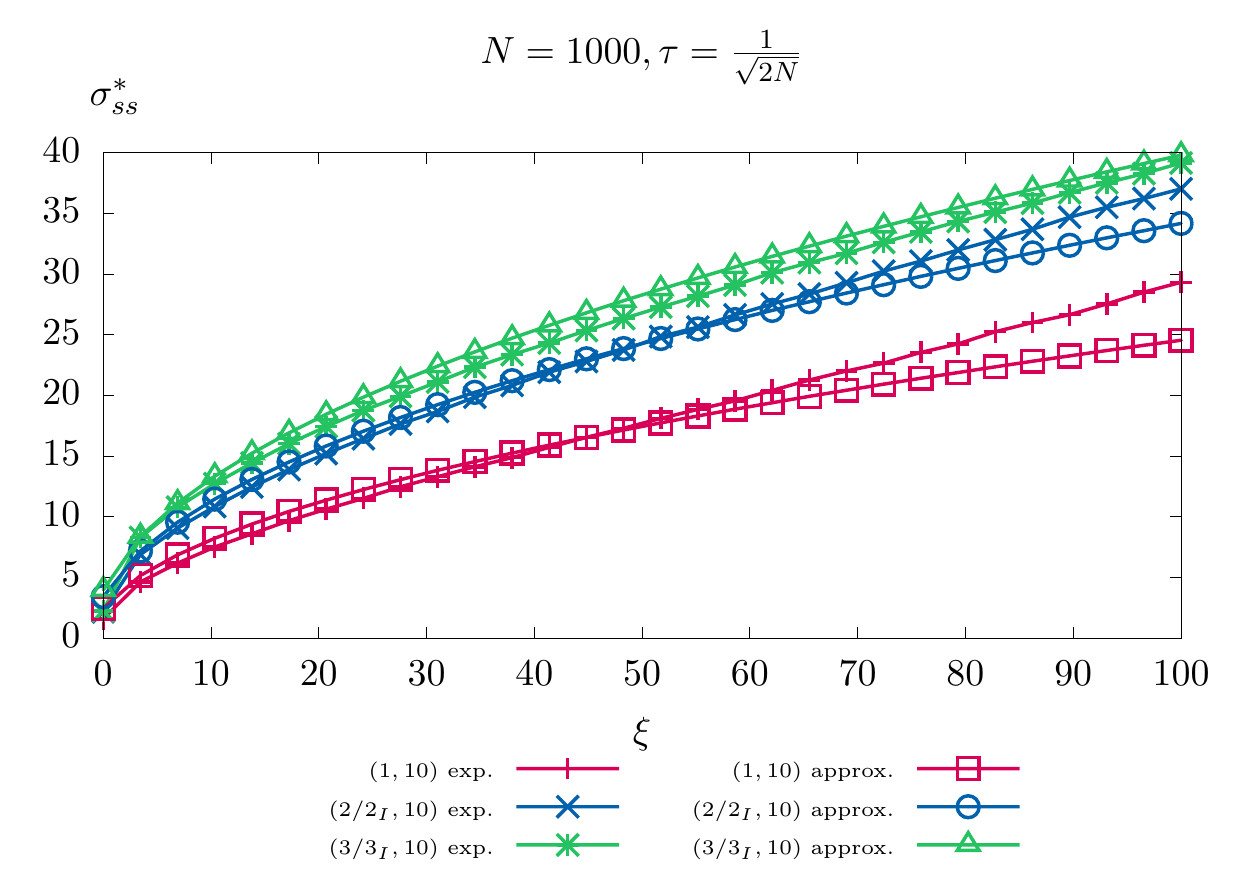}\\
    \includegraphics[width=0.43\textwidth]{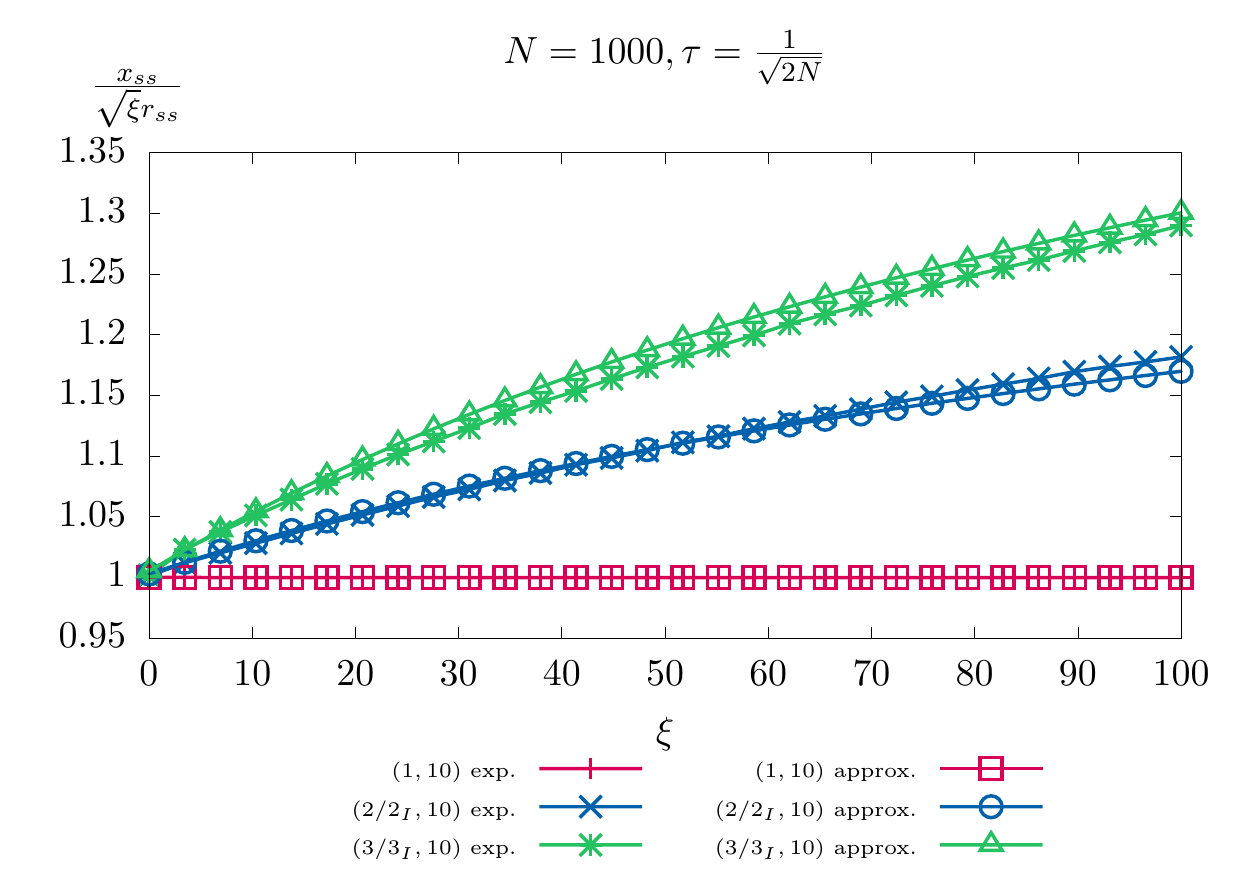}
  \end{tabular}
  \caption{Steady state closed-form approximation and real-run
    comparison of the $(\mu/\mu_I,\lambda)$-ES
    with repair by projection
    applied to the conically constrained problem.}
  \label{sec:theoreticalanalysis:fig:steadystate}
\end{figure}

\Cref{sec:theoreticalanalysis:fig:steadystate}
shows plots of the steady state computations. The derived closed-form
approximation has been compared to real ES runs.
The values for visualizing the approximations
have been calculated as follows. The steady state $\sigma^*_{ss}$ value
has been computed by evaluating
\Cref{sec:theoreticalanalysis:eq:steadystateroot}. This value
was then inserted into the derived approximations for the progress measures
to obtain the values for ${\varphi_x^*}_{ss}$ and ${\varphi_r^*}_{ss}$.
The values from experiments have been determined by
computing the averages of the particular values in real ES runs. For this,
the ES was run for $400N$ generations and the particular values have
been averaged over these generations. This procedure was
repeated $100$ times and those $100$ times have been averaged to compute the
particular steady state values.

\renewcommand{\sppapathtmp}{./figures/figure07/}

\section{Results and Conclusions}
\label{sec:resultsconclusion}
A $(\mu/\mu_I,\lambda)$-ES applied to a conically constrained problem
has been theoretically analyzed.

Approximate expressions for the local progress measures and the SAR have
been derived for asymptotic considerations. Comparison with numerical
one-generation experiments shows the approximation quality. As can
be seen in \Cref{sec:theoreticalanalysis:fig:xprogresscomparisons}
and \Cref{sec:theoreticalanalysis:fig:rprogresscomparisons}, the
derived approximations for the $x$ and $r$ progress rates
coincide well with the experiments, in particular
for higher dimensions ($N=1000$). In the transition from the feasible
to the infeasible case, the weighting with the offspring feasibility
probability does not fit well in all the cases.
This can be seen in the supplementary material
(\Crefrange{sec:theoreticalanalysis:fig:xprogresscomparisonsdetail1}
{sec:theoreticalanalysis:fig:rprogresscomparisonsdetail3} in
\Cref{sec:algorithm:subsec:additionalplots}). The middle row in these
figures always shows a case between the feasible and the infeasible case.
However, the deviation in these cases does not affect the steady state
considerations because the ES moves in the vicinity of the cone boundary
in the stationary state. The approximation for the SAR corresponds well to
the experiments for small values of ${\sigma^{(g)}}^*$. It deviates for
larger values of ${\sigma^{(g)}}^*$. Again, for higher $N$, the approximation
quality fits better as can be seen in
\Cref{sec:theoreticalanalysis:fig:sarcomparisons}.
This behavior is inherent in the SAR approximation.
First, $N \gg {\sigma^{(g)}}^*$ has been assumed in deriving the SAR
approximation. Second, quadratic and higher order terms have been neglected.
Therefore, the SAR approximation is linear and hence necessarily
deviates for large values of ${\sigma^{(g)}}^*$.

The approximate expressions for the local progress measures and the
SAR derived for the $(\mu/\mu_I,\lambda)$-ES generalize the results obtained
for the $(1,\lambda)$-ES. Considering $\mu=1$ for
\Cref{sec:theoreticalanalysis:eq:varphixnormalizedfeasible5}
and
\Cref{sec:theoreticalanalysis:eq:varphixnormalizedinfeasible5}
immediately results
in~\cite[Eq. (3.120), p. 41]{SpettelBeyer2018SigmaSaEsCone}
and~\cite[Eq. (3.149), p. 44]{SpettelBeyer2018SigmaSaEsCone}, respectively,
for the $x$ progress rate. For the $r$ progress rate, it is similar.
Setting $\mu=1$
in \Cref{sec:theoreticalanalysis:eq:varphirnormalizedcombinedmaintext}
results
in~\cite[Eq. (3.221), p. 57]{SpettelBeyer2018SigmaSaEsCone} for the feasible
case and
in~\cite[Eq. (3.209), p. 56]{SpettelBeyer2018SigmaSaEsCone} for the infeasible
case (after applying the definition of $\varphi^*$).
The same holds for $\psi$. It only differs in the progress coefficients.
It can easily be verified that
\Cref{sec:theoreticalanalysis:eq:psicombinedmaintext} with $\mu=1$
corresponds
to~\cite[Eq. (3.281), p. 69]{SpettelBeyer2018SigmaSaEsCone}.

For the macroscopic behavior, the local expected progress functions
have been iterated to predict the mean value dynamics of the ES
and analytic steady state expressions have been derived.

For the evolution dynamics, in \Cref{sec:theoreticalanalysis:fig:dynamics}
the rate of convergence
from the prediction is very similar to the one exhibited by the
real ES run. However, the theory predicts an earlier transitioning
into the stationary state (see the fourth subplot
of \Cref{sec:theoreticalanalysis:fig:dynamics}
showing the $\sigma^*$ dynamics).
Further investigations (not shown here)
showed that the fluctuations of the mean values obtained from real ES
runs are such that the fluctuations cover the predictions.
Larger values of $\tau$ reduce this discrepancy due to bigger
$\sigma$-mutations.

The analytic steady state analysis resulted in a very interesting
conclusion. It turns out that the astonishing result from the $(1,\lambda)$ case
(see the derivations leading
to~\cite[Eq. (3.308), p. 87]{SpettelBeyer2018SigmaSaEsCone})
extends to the multi-recombinative ES. One recognizes the equations for the
sphere model in~\eqref{sec:theoreticalanalysis:eq:steadystateroot}
(see~\cite[Eq. (4.11), p. 35]{MeyerNieberg2007Thesis}), i.e.,
\Cref{sec:theoreticalanalysis:eq:steadystateroot} can be written as
\begin{equation}
  \label{sec:theoreticalanalysis:eq:steadystatespheremulti}
  \sigma^*_{ss} = \sqrt{1+\xi}{\sigma^*_{ss}}_{\text{sphere}}
\end{equation}
where ${\sigma^*_{ss}}_{\text{sphere}}$ denotes the steady state
mutation strength value for the $(\mu/\mu_I,\lambda)$-$\sigma$SA-ES
applied to the sphere model.
This means that in the steady state the $(\mu/\mu_I,\lambda)$-$\sigma$SA-ES
with repair by projection applied to the conically constrained problem
moves towards the optimal value as if a sphere would be optimized.
Additionally, the rate at which the ES approaches this optimal
value is the same as if a sphere were to be optimized, independently
of the value of $\xi$. The $x^{(g)}$ and $r^{(g)}$ dynamics are proportional to
$\exp\left(-\frac{\varphi^*}{N}g\right)$ in
the steady state. This follows
from the definition of the progress rates $\varphi_x^*$ and
$\varphi_r^*$. Consequently, linear convergence order with a convergence rate
of $\frac{\varphi^*}{N}$ is implied.
This means that the conically constrained
problem has been turned into a sphere model by applying the projection as the
repair method. However, the behavior is achieved by a
$\sqrt{1+\xi}$ times larger mutation
strength~\eqref{sec:theoreticalanalysis:eq:steadystatespheremulti}.
That is, the ES performs larger search steps in the search space.
These larger mutation steps are automatically realized by the self-adaptation
as can be seen in~\eqref{sec:theoreticalanalysis:eq:steadystateroot}.

The figure for the steady state (\Cref{sec:theoreticalanalysis:fig:steadystate})
allows for a comparison of the $(1,10)$-ES, the
$(2/2_I,10)$-ES, and the $(3/3_I,10)$-ES. The mutation strength $\sigma_{ss}^*$
and $\frac{x_{ss}}{\sqrt{\xi}r_{ss}}$ ratio
approximations come close to the ones determined by experiments. For the steady
state progress rates, the approximations are not that good. The deviations
are particularly high for small values of $\xi$.
The reason for this is that in the derivations for $P_Q(q)$, $\xi$
has been assumed to be sufficiently large.
Despite that, the figure gives insights. For the considered ESs,
one can observe that the strategies with
intermediate recombination exhibit faster progress in the steady state.
The difference between the $(2/2_I,10)$-ES and the $(3/3_I,10)$-ES
is marginal, which is also predicted by the theory as derived
in the following. Based on
\Cref{sec:theoreticalanalysis:eq:varphixnormalizedss4}, one can
compute the maximal $x$ progress in the steady state under the
assumption $N \gg \sigma_{ss}^*$. Computing the first derivative
with respect to $\sigma_{ss}^*$
of \Cref{sec:theoreticalanalysis:eq:varphixnormalizedss4} yields
\begin{equation}
  \label{sec:theoreticalanalysis:eq:steadystatederivativexprogress}
  \frac{\mathrm{d}}{\mathrm{d}\sigma_{ss}^*}
  {\varphi_x^*}_{ss}
  =
  \frac{c_{\mu/\mu,\lambda}}{\sqrt{1+\xi}}
  -\frac{\sigma_{ss}^*}{\mu (\sqrt{1+\xi})^2}
\end{equation}
Setting \Cref{sec:theoreticalanalysis:eq:steadystatederivativexprogress}
to zero and solving for $\sigma_{ss}^*$ results in the mutation
strength
\begin{equation}
  \label{sec:theoreticalanalysis:eq:steadystatemaxprogressmaxsigma}
  \hat{\sigma}_{ss}^*=\sqrt{1+\xi}\mu c_{\mu/\mu,\lambda}
\end{equation}
with which the maximal $x$ progress in the steady state is obtained.
Further, by insertion of
\Cref{sec:theoreticalanalysis:eq:steadystatemaxprogressmaxsigma}
into
\Cref{sec:theoreticalanalysis:eq:varphixnormalizedss4},
the maximal $x$ progress follows as
${{}\hat{\varphi}_{x}^*}_{ss}(\mu,\lambda)=\frac{\mu c_{\mu/\mu,\lambda}^2}{2}$.
Assuming it is achieved by the ES,
${{}\hat{\varphi}_{x}^*}_{ss}(2,10) \approx 1.61$ and
${{}\hat{\varphi}_{x}^*}_{ss}(3,10) \approx 1.70$ follow, which
are relatively close to each other. In contrast to that,
${{}\hat{\varphi}_{x}^*}_{ss}(1,10) \approx 1.18$.

Insertion of
\Cref{sec:theoreticalanalysis:eq:steadystatemaxprogressmaxsigma}
into
\Cref{sec:theoreticalanalysis:eq:steadystatedist2} yields
\begin{equation}
  \label{sec:theoreticalanalysis:eq:steadystatemaxprogressdist1}
  \frac{\sqrt{\xi}r_{ss}}
       {x_{ss}}
  =
  \sqrt{\frac{1+\frac{(1+\xi)}{N}\mu c_{\mu/\mu,\lambda}^2}
    {1+\frac{(1+\xi)}{N}\mu c_{\mu/\mu,\lambda}^2\mu}}.
\end{equation}
This gives insights
about the deviation of the parental centroid from the cone boundary.
Assuming $\frac{(1+\xi)}{N}\mu c_{\mu/\mu,\lambda}^2$ in
\Cref{sec:theoreticalanalysis:eq:steadystatemaxprogressdist1}
to be sufficiently small, a Taylor expansion around $0$ for
$\frac{(1+\xi)}{N}\mu c_{\mu/\mu,\lambda}^2$ can be applied.
Together with subsequent cutoff after the linear term, it results in
\begin{equation}
  \label{sec:theoreticalanalysis:eq:steadystatemaxprogressdist2}
  \frac{\sqrt{\xi}r_{ss}}
       {x_{ss}}
  \simeq
  1+\frac{(1+\xi)}{N}\frac{\mu c_{\mu/\mu,\lambda}^2}{2}(1-\mu).
\end{equation}
As can be seen, in the case without recombination ($\mu = 1$), the
parental centroid moves on the cone boundary. However, with recombination
($\mu > 1$), the centroid leaves this boundary and moves into the interior
of the cone. The deviation from the cone boundary increases quadratically
with $\mu$.

The theory developed allows further to compute
the optimal value $\hat{\tau}$ for the learning
parameter $\tau$. Requiring that the steady state mutation strength
$\sigma_{ss}^*$ from~\eqref{sec:theoreticalanalysis:eq:steadystateroot}
attains $\hat{\sigma}_{ss}^*$
from~\eqref{sec:theoreticalanalysis:eq:steadystatemaxprogressmaxsigma}, one gets
\begin{equation}
  \begin{multlined}
  \sqrt{1+\xi}\mu c_{\mu/\mu,\lambda}
  \stackrel{!}{=}
  \sqrt{1+\xi}\mu c_{\mu/\mu,\lambda}
  \left[(1-N\tau^2)+\right.\\\left.\hspace{2cm}
  \sqrt{(1-N\tau^2)^2+
        \frac{2\tau^2N\left(\frac{1}{2}+e_{\mu,\lambda}^{1,1}\right)}
             {\mu c_{\mu/\mu,\lambda}^2}}\right].
  \end{multlined}
  \label{sec:theoreticalanalysis:eq:optimaltaucondition}
\end{equation}
Solving \Cref{sec:theoreticalanalysis:eq:optimaltaucondition} for $\tau$
yields the optimal learning parameter
\begin{equation}
  \hat{\tau}(\mu,\lambda)=\frac{1}{\sqrt{2N}}
  \sqrt{\frac{\mu c_{\mu/\mu,\lambda}^2}
    {\mu c_{\mu/\mu,\lambda}^2 -
      \frac{1}{2} - e_{\mu,\lambda}^{1,1}}},
\end{equation}
which equals the optimal learning parameter of the sphere model
(cf. \cite[Eq. (4.116), p. 71]{MeyerNieberg2007Thesis}).
For $\lambda \rightarrow \infty$ with $\mu/\lambda=\text{const.}$,
one has $\hat{\tau} \simeq \frac{1}{\sqrt{2N}}$, which has been
used in the simulations as an approximation to the optimal
value. For $N=1000$, one has
$\frac{1}{\sqrt{2N}} \approx 0.022$ and
$\hat{\tau}(1,10) \approx 0.017$,
$\hat{\tau}(2,10) \approx 0.021$, and
$\hat{\tau}(3,10) \approx 0.022$. Hence, already for $\lambda=10$,
$\frac{1}{\sqrt{2N}}$ approximates the optimal learning parameter
relatively well.

Let us finally outline research goals to be addressed in the future.
The analysis and comparison of different $\sigma$ control methods
is a topic for future work. It is of interest to analyze the ES with
cumulative step size adaptation (CSA) and Meta-ESs
instead of self-adaptation for adapting $\sigma$.
A comparison between the different approaches may give insights for practical
recommendations of their application.
As another research direction, it is of interest to investigate and
compare different repair methods. Truncation and reflection are two
examples. In the truncation approach, an infeasible offspring is repaired by
the intersection of its mutation vector and the cone. For the reflection,
infeasible offspring are mirrored into the feasible region about
the cone boundary.

\bibliographystyle{IEEEtran}
\bibliography{ms}

\newpage

\onecolumn

\author{Patrick~Spettel~and~Hans-Georg~Beyer\\
  \vspace{1.0cm}
  \textsc{\Large Supplementary material}}
\IEEEpubid{}

{\maketitle}

\appendix
\vspace{0.2cm}
\begin{appendices}

\renewcommand{\theequation}{\thesection.\arabic{equation}}

\newcommand{\gls}[1]{#1}
\newcommand{\glsfmttext}[1]{#1}
\newcommand{\glspl}[1]{#1s}
\newcommand{\chapter}[1]{\section{#1}}
\newcommand{\chaptermark}[1]{}

\section{Derivation of Closed-Form Expressions for the Projection}
\label{sec:algorithm:subsec:projection}
\begin{figure}[t]
  \centering
  \includegraphics{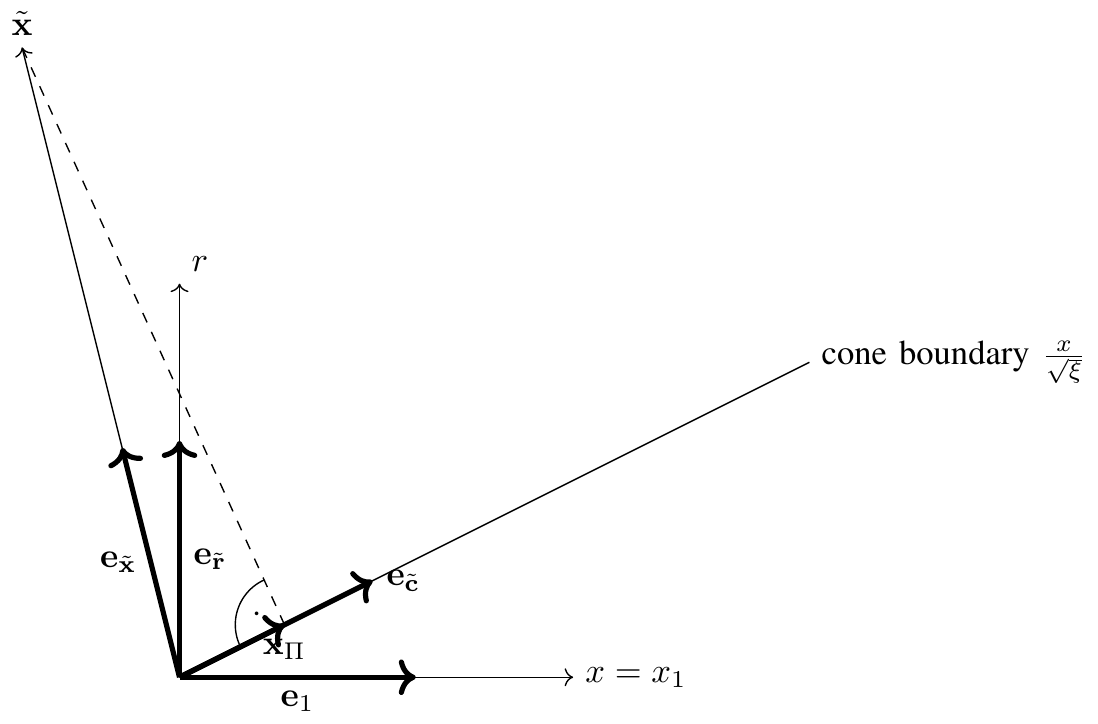}
    \caption[Visualization of the projection onto the cone.]
            {2D-plane
    spanned by an offspring $\tilde{\mathbf{x}}$ and the $x_1$ coordinate axis.
    Vectors introduced for expressing the projection onto the feasible
    region of \Cref{sec:problem:fig:conicconstraintnd} are visualized.
    The vectors $\mathbf{e}_1$, $\mathbf{e}_{\tilde{\mathbf{x}}}$,
    $\mathbf{e}_{\tilde{\mathbf{r}}}$, and
    $\mathbf{e}_{\tilde{\mathbf{c}}}$ are unit vectors of the
    corresponding vectors. The dashed line indicates the orthogonal
    projection of $\tilde{\mathbf{x}}$ onto the cone boundary.}
    \label{sec:algorithm:subsec:projection:fig:projectionvectors}
\end{figure}

In this section, a geometrical approach for the projection of points
that are outside of the feasible region onto the cone boundary
is described.
\Cref{sec:algorithm:subsec:projection:fig:projectionvectors}
shows a visualization of the 2D-plane spanned
by an offspring $\tilde{\mathbf{x}}$ and the $x_1$ coordinate axis.
The equation for the cone boundary is a direct consequence
of the problem definition (\Cref{sec:problem:eq:coneconstraint}).
The projected vector is indicated by $\mathbf{x}_{\Pi}$. The unit vectors
$\mathbf{e}_1$, $\mathbf{e}_{\tilde{\mathbf{x}}}$,
$\mathbf{e}_{\tilde{\mathbf{r}}}$, and
$\mathbf{e}_{\tilde{\mathbf{c}}}$ are introduced. They are unit vectors
in the direction of the $x_1$ axis, $\tilde{\mathbf{x}}$ vector,
$\tilde{\mathbf{r}} = (0, \tilde{x}_2, \ldots, \tilde{x}_N)^T$ vector, and
the cone boundary, respectively. The projection line
is indicated by the dashed line. It indicates the shortest
Euclidean distance
from $\tilde{\mathbf{x}}$ to the cone boundary.
The goal is to compute the parameter vector after projection $\mathbf{x}_{\Pi}$.
By inspecting
\Cref{sec:algorithm:subsec:projection:fig:projectionvectors}
one can see that the projected vector $\mathbf{x}_{\Pi}$ can
be expressed as
\begin{equation}
    \label{sec:algorithm:subsec:projection:eq:xpi}
    \mathbf{x}_{\Pi} =
    \begin{cases}
    (\mathbf{e}_{\tilde{\mathbf{c}}}^T\tilde{\mathbf{x}})
    \mathbf{e}_{\tilde{\mathbf{c}}}
    &\text{ if } \mathbf{e}_{\tilde{\mathbf{c}}}^T\tilde{\mathbf{x}} > 0\\
    \mathbf{0} & \text{ otherwise}.
    \end{cases}
\end{equation}
From \Cref{sec:problem:eq:coneconstraint},
the equation of the cone boundary
$\frac{x}{\sqrt{\xi}}$ follows. Using this, the vector
$\mathbf{e}_{\tilde{\mathbf{c}}}$ can be expressed
as a linear combination of the vectors
$\mathbf{e}_1$ and $\mathbf{e}_{\tilde{\mathbf{r}}}$.
After normalization (such that
$||\mathbf{e}_{\tilde{\mathbf{c}}}||=1$) this writes
\begin{equation}
    \label{sec:algorithm:subsec:projection:eq:ec}
    \mathbf{e}_{\tilde{\mathbf{c}}} =
    \frac{\mathbf{e}_1 + \frac{1}{\sqrt{\xi}}\mathbf{e}_{\tilde{\mathbf{r}}}}
         {\sqrt{1 + \frac{1}{\xi}}}.
\end{equation}
The unit vectors
$\mathbf{e}_1 = (1, 0, \ldots, 0)^T$ and $\mathbf{e}_{\tilde{\mathbf{x}}} =
\frac{\tilde{\mathbf{x}}}{||\mathbf{\tilde{\mathbf{x}}}||}$ are known.
The vector $\mathbf{e}_{\tilde{\mathbf{r}}}$ is a unit vector in direction
of $\tilde{\mathbf{x}}$ in the dimensions $2$ to $N$, i.e.,
\begin{align}
    \label{sec:algorithm:subsec:projection:eq:er}
    \begin{split}
        \mathbf{e}_{\tilde{\mathbf{r}}}
        &= (0,
              (\tilde{\mathbf{x}})_2, \ldots, (\tilde{\mathbf{x}})_N)^T
              / ||\tilde{\mathbf{r}}||.
    \end{split}
\end{align}
Inserting \Cref{sec:algorithm:subsec:projection:eq:er} into
\Cref{sec:algorithm:subsec:projection:eq:ec} results in
\begin{equation}
    \label{sec:algorithm:subsec:projection:eq:ec2}
    \mathbf{e}_{\tilde{\mathbf{c}}} = \left(1,
              \frac{(\tilde{\mathbf{x}})_2}{\sqrt{\xi}||\tilde{\mathbf{r}}||},
              \ldots,
              \frac{(\tilde{\mathbf{x}})_N}{\sqrt{\xi}||\tilde{\mathbf{r}}||}
              \right)^T
              \bigg/ \sqrt{1 + \frac{1}{\xi}}.
\end{equation}
Using \Cref{sec:algorithm:subsec:projection:eq:ec2} it follows that
\begin{align}
    \label{sec:algorithm:subsec:projection:eq:projlen}
    \begin{split}
        \mathbf{e}_{\tilde{\mathbf{c}}}^T\tilde{\mathbf{x}}
        &= \left((\tilde{\mathbf{x}})_1 +
        \frac{(\tilde{\mathbf{x}})_2^2 + \cdots + (\tilde{\mathbf{x}})_N^2}
             {\sqrt{\xi}||\tilde{\mathbf{r}}||}\right)
             \bigg/ \sqrt{1 + \frac{1}{\xi}}\\
        &= \left((\tilde{\mathbf{x}})_1 +
        \frac{||\tilde{\mathbf{r}}||^2}
             {\sqrt{\xi}||\tilde{\mathbf{r}}||}\right)
             \bigg/ \sqrt{1 + \frac{1}{\xi}}\\
        &= \left((\tilde{\mathbf{x}})_1 +
        \frac{||\tilde{\mathbf{r}}||}
             {\sqrt{\xi}}\right)
             \bigg/ \sqrt{1 + \frac{1}{\xi}}
        = \left((\tilde{\mathbf{x}})_1\sqrt{\xi} +
        ||\tilde{\mathbf{r}}||\right)
             \big/ \sqrt{\xi + 1}.
    \end{split}
\end{align}
Use of
\Cref{sec:algorithm:subsec:projection:eq:projlen,%
sec:algorithm:subsec:projection:eq:ec2,%
sec:algorithm:subsec:projection:eq:xpi}
yields for the case $\mathbf{e}_{\tilde{\mathbf{c}}}^T\tilde{\mathbf{x}} > 0$
\begin{align}
    \begin{split}
        \mathbf{x}_{\Pi} &=
        (\mathbf{e}_{\tilde{\mathbf{c}}}^T\tilde{\mathbf{x}})
          \mathbf{e}_{\tilde{\mathbf{c}}}\\
        &= \frac{1}{1 + \frac{1}{\xi}}\left((\tilde{\mathbf{x}})_1 +
        \frac{(\tilde{\mathbf{x}})_2^2 + \cdots + (\tilde{\mathbf{x}})_N^2}
             {\sqrt{\xi}||\tilde{\mathbf{r}}||}\right)
             \left(1,
              \frac{(\tilde{\mathbf{x}})_2}{\sqrt{\xi}||\tilde{\mathbf{r}}||},
              \ldots,
              \frac{(\tilde{\mathbf{x}})_N}{\sqrt{\xi}||\tilde{\mathbf{r}}||}
              \right)^T.
    \end{split}
\end{align}
The first vector component of $\mathbf{x}_{\Pi}$ writes
\begin{align}
    \label{sec:algorithm:subsec:projection:eq:xproj1}
    \begin{split}
        q = (\mathbf{x}_{\Pi})_1 &=
            \frac{1}{1 + \frac{1}{\xi}}\left((\tilde{\mathbf{x}})_1 +
            \frac{(\tilde{\mathbf{x}})_2^2 + \cdots + (\tilde{\mathbf{x}})_N^2}
                 {\sqrt{\xi}||\tilde{\mathbf{r}}||}\right)
            =\frac{\xi}{\xi + 1}\left((\tilde{\mathbf{x}})_1 +
            \frac{||\tilde{\mathbf{r}}||}
                 {\sqrt{\xi}}\right).
    \end{split}
\end{align}
The $k$-th vector component of $\mathbf{x}_{\Pi}$ for $k \in \{2, \ldots, N\}$
writes
\begin{align}
    \begin{split}
        (\mathbf{x}_{\Pi})_k
            &=\frac{\xi}{\xi + 1}\left((\tilde{\mathbf{x}})_1 +
            \frac{||\tilde{\mathbf{r}}||}
                 {\sqrt{\xi}}\right)
            \frac{(\tilde{\mathbf{x}})_k}{\sqrt{\xi}||\tilde{\mathbf{r}}||}
            =\frac{\xi}{\xi + 1}
            \left(\frac{(\tilde{\mathbf{x}})_1}
                       {\sqrt{\xi}||\tilde{\mathbf{r}}||} +
            \frac{1}
                 {\xi}\right)
            (\tilde{\mathbf{x}})_k.
    \end{split}
\end{align}
And the projected distance from the cone axis $||\mathbf{r}_{\Pi}||$ writes
\begin{align}
    \label{sec:algorithm:subsec:projection:eq:rproj}
    \begin{split}
        q_r = ||\mathbf{r}_{\Pi}||
        &= \sqrt{\sum_{k = 2}^N (\mathbf{x}_{\Pi})_k^2}
        = \sqrt{\sum_{k = 2}^N \left(\frac{\xi}{\xi + 1}
            \left(\frac{(\tilde{\mathbf{x}})_1}
                       {\sqrt{\xi}||\tilde{\mathbf{r}}||} +
            \frac{1}
                 {\xi}\right)\right)^2
            (\tilde{\mathbf{x}})_k^2}\\
        &= \left(\frac{\xi}{\xi + 1}
            \left(\frac{(\tilde{\mathbf{x}})_1}
                       {\sqrt{\xi}||\tilde{\mathbf{r}}||} +
            \frac{1}
                 {\xi}\right)\right)
            ||\tilde{\mathbf{r}}||.
    \end{split}
\end{align}
Note that in terms of the $(x,r)^T$ representation, the offspring
$\tilde{\mathbf{x}}$ can be expressed as
$(\mathbf{e}_1^T\tilde{\mathbf{x}},
\mathbf{e}_{\tilde{\mathbf{r}}}^T\tilde{\mathbf{x}})^T
=((\tilde{\mathbf{x}})_1,||\tilde{\mathbf{r}}||)^T$. This is exactly
what is expected. The first component is the value in direction
of the cone axis. The second component is the distance from the
cone axis.

\newpage

\section{Derivation of the Normal Approximation for the Offspring
  Density in \texorpdfstring{$r$}{\$r\$} Direction}
\label{sec:theoreticalanalysis:appendix:rnormalapproximation}
From the offspring generation
(\Cref{sec:algorithm:alg:es:offspringsigma,%
sec:algorithm:alg:es:generateoffspring})
it follows that the distance from the cone's axis of the offspring is
\begin{align}
\begin{split}
  \tilde{r} &= \sqrt{{r^{(g)}}^2 + 2\sigma^{(g)} r^{(g)} z_2 +
                  {\sigma^{(g)}}^2z_2^2 +
                  {\sigma^{(g)}}^2\sum_{k = 3}^N z_k ^ 2}.
\end{split}
\end{align}
Now, $2\sigma^{(g)} r^{(g)} z_2 + {\sigma^{(g)}}^2z_2^2$
and ${\sigma^{(g)}}^2\sum_{k = 3}^N z_k ^ 2$ are replaced with
normally distributed expressions with mean values and standard
deviations of the corresponding expressions.
With the moments of the i.i.d. variables
according to a standard normal distribution
$z_k$, $\mathrm{E}[z_k] = 0$, and $\mathrm{E}[z_k^2] = 1$,
the expected values
$\mathrm{E}\left[2\sigma^{(g)} r^{(g)} z_2 + {\sigma^{(g)}}^2z_2^2\right] =
    {\sigma^{(g)}}^2$
and
$\mathrm{E}\left[{\sigma^{(g)}}^2\sum_{k = 3}^N z_k ^ 2\right] =
  {\sigma^{(g)}}^2(N-2)$
follow.
The variances are computed as
\begin{align}
    \begin{split}
        \mathrm{Var}&\left[2\sigma^{(g)} {r^{(g)}} z_2 +
        {\sigma^{(g)}}^2z_2^2\right]\\
        &= \mathrm{E}\left[(2\sigma^{(g)} {r^{(g)}} z_2 +
        {\sigma^{(g)}}^2z_2^2)^2\right]
           - \mathrm{E}\left[2\sigma^{(g)} {r^{(g)}} z_2 +
           {\sigma^{(g)}}^2z_2^2\right]^2\\
        &=
        \mathrm{E}\left[4{\sigma^{(g)}}^2 {r^{(g)}}^2 z_2^2 +
        4{\sigma^{(g)}}^3 {r^{(g)}} z_2^3 + {\sigma^{(g)}}^4z_2^4\right]
        - {\sigma^{(g)}}^4\\
        &= 4 {\sigma^{(g)}}^2 {r^{(g)}}^2 +
        3 {\sigma^{(g)}}^4 - {\sigma^{(g)}}^4
        = 4 {\sigma^{(g)}}^2 {r^{(g)}}^2 + 2 {\sigma^{(g)}}^4
    \end{split}
\end{align}
and
\begin{align}
    \begin{split}
        \mathrm{Var}\left[{\sigma^{(g)}}^2\sum_{i = 3}^N z_i ^ 2\right]
        &= {\sigma^{(g)}}^4 \sum_{i = 3}^N \mathrm{Var}\left[z_i ^ 2\right]
        = {\sigma^{(g)}}^4 \sum_{i = 3}^N
        \mathrm{E}\left[z_i ^ 4\right] - \mathrm{E}\left[z_i ^ 2\right]^2\\
        &= {\sigma^{(g)}}^4 \sum_{i = 3}^N 2
        = 2{\sigma^{(g)}}^4(N - 2)
    \end{split}
\end{align}
where the moments of the i.i.d. variables according to a standard normal
distribution $z_i$
$\mathrm{E}[z_i^2] = 1$,
$\mathrm{E}[z_i^3] = 0$, and $\mathrm{E}[z_i^4] = 3$ were used.

With these results, the normal approximation follows as
\begin{align}
    \begin{split}
        \tilde{r} &\approx \sqrt{{r^{(g)}}^2 +
                         \mathcal{N}({\sigma^{(g)}}^2,
                         4 {\sigma^{(g)}}^2 {r^{(g)}}^2 + 2 {\sigma^{(g)}}^4) +
                         \mathcal{N}({\sigma^{(g)}}^2(N - 2),
                                     2 {\sigma^{(g)}}^4 (N - 2))
                     }\\
            &= \sqrt{{r^{(g)}}^2 +
                     {\sigma^{(g)}}^2(N - 1) +
                     \sqrt{4 {\sigma^{(g)}}^2 {r^{(g)}}^2 +
                     2 {\sigma^{(g)}}^4 (N - 1)}
                         \mathcal{N}(0, 1).
                     }
    \end{split}
\end{align}
Substitution of the normalized quantity
${\sigma^{(g)}}^* = \frac{\sigma^{(g)} N}{{r^{(g)}}}$ yields
after simplification
\begin{align}
  \tilde{r} &\approx
  {r^{(g)}} \sqrt{1 +
    \frac{{{\sigma^{(g)}}^*}^2}{N}\left(1 - \frac{1}{N}\right)}
  \sqrt{1 + \frac{2{\sigma^{(g)}}^*}{N}
    \frac{\sqrt{1 + \frac{{{\sigma^{(g)}}^*}^2}
        {2N}\left(1 - \frac{1}{N}\right)}}
         {1 +
           \frac{{{\sigma^{(g)}}^*}^2}{N}\left(1 - \frac{1}{N}\right)}
         \mathcal{N}(0, 1)}
  \label{sec:theoreticalanalysis:eq:approximatedr3}.
\end{align}
As the expression
$\frac{2{\sigma^{(g)}}^*}{N}
                  \frac{\sqrt{1 + \frac{{{\sigma^{(g)}}^*}^2}
                      {2N}\left(1 - \frac{1}{N}\right)}}
                      {1 +
                      \frac{{{\sigma^{(g)}}^*}^2}{N}
                      \left(1 - \frac{1}{N}\right)}
              \mathcal{N}(0, 1) \rightarrow 0$
with $N \rightarrow \infty$ (${\sigma^{(g)}}^* < \infty$),
a further asymptotically simplified expression can be obtained
by Taylor expansion of the square root at $0$ and cutoff after the linear term
\begin{align}
    \label{sec:theoreticalanalysis:eq:approximatedr}
    \begin{split}
        \tilde{r} &\approx
        \underbrace{
            {r^{(g)}} \sqrt{1 +
              \frac{{{\sigma^{(g)}}^*}^2}{N}\left(1 - \frac{1}{N}\right)}}
              _{\bar{r}} +
              \underbrace{
              {r^{(g)}}\frac{{\sigma^{(g)}}^*}{N}
                  \sqrt{\frac{1 + \frac{{{\sigma^{(g)}}^*}^2}
                      {2N}\left(1 - \frac{1}{N}\right)}
                      {1 +
                      \frac{{{\sigma^{(g)}}^*}^2}{N}
                      \left(1 - \frac{1}{N}\right)}}}
              _{\sigma_r}
              \mathcal{N}(0, 1).
    \end{split}
\end{align}
Consequently, the mean of the asymptotic normal approximation of $\tilde{r}$ is
$\bar{r}$ and its standard deviation is $\sigma_r$
\begin{align}
  \begin{split}
    p_r(r)
    &\approx \frac{1}{\sigma_r}\phi\left(\frac{r - \bar{r}}{\sigma_r}\right)
    = \frac{1}{\sqrt{2\pi}\sigma_r}\exp\left[
      -\frac{1}{2}\left(\frac{r - \bar{r}}{\sigma_r}\right)^2\right].
  \end{split}
\end{align}

\newpage

\section{Derivation of \texorpdfstring{$\mathrm{E}[\langle q \rangle]$}
{\$\textbackslash mathrm\{E\}[\textbackslash langle q \textbackslash rangle]\$}}
\label{appendix:subsec:expqcentroid}
With
\Cref{sec:algorithm:alg:es:bestq,%
sec:algorithm:alg:es:replaceparent,%
sec:algorithm:alg:es:q}
of
\Cref{sec:algorithm:alg:es},
\begin{align}
  \mathrm{E}[\langle q \rangle\,|\,x^{(g)}, &r^{(g)}, \sigma^{(g)}]
  := \mathrm{E}[\langle q \rangle]\\
  &= \mathrm{E}\left[\frac{1}{\mu}\sum_{m=1}^\mu q_{m;\lambda}\,|\,x^{(g)}, r^{(g)}, \sigma^{(g)}\right]
  \label{sec:theoreticalanalysis:eq:qcentroid1}\\
  &=\frac{1}{\mu}\sum_{m=1}^\mu\mathrm{E}\left[q_{m;\lambda}\,|\,x^{(g)}, r^{(g)}, \sigma^{(g)}\right]
  \label{sec:theoreticalanalysis:eq:qcentroid2}
\end{align}
follows.
This means that the expectation of $q_{m;\lambda}$, i.e.,
the $x$ value after (possible) projection of the $m$-th best offspring,
is needed for the derivation of the progress rate in $x$ direction
\begin{align}
  \mathrm{E}[q_{m;\lambda}\,|\,x^{(g)}, r^{(g)}, \sigma^{(g)}]
  &:= \mathrm{E}[q_{m;\lambda}]\\
  &= \int_{q=0}^{q=\infty}q\,p_{q_{m;\lambda}}(q)\,\mathrm{d}q
  \label{sec:theoreticalanalysis:eq:expectedqmlambda1}\\
  &= \int_{q=0}^{q=\infty}q\,\frac{\lambda!}{(\lambda-m)!(m-1)!}
  p_Q(q)[1-P_Q(q)]^{\lambda-m}[P_Q(q)]^{m-1}
  \,\mathrm{d}q
  \label{sec:theoreticalanalysis:eq:expectedqmlambda2}.
\end{align}
\Cref{sec:theoreticalanalysis:eq:expectedqmlambda1} follows
directly from the definition of expectation where
\begin{equation*}
  p_{q_{m;\lambda}}(q) := p_{q_{m;\lambda}}(q \,|\, x^{(g)}, r^{(g)}, \sigma^{(g)})
\end{equation*}
indicates the probability density function of
the $m$-th best (offspring with $m$-th smallest $q$ value)
offspring's $q$ value.
The random variable $Q$ denotes the random $x$ values after projection.
The step to
\Cref{sec:theoreticalanalysis:eq:expectedqmlambda2}
follows from the calculation of $p_{q_{m;\lambda}}(q)$. Because the objective
function (\Cref{sec:problem:eq:optgoal})
is defined to return an individual's $x$ value, $p_{q_{m;\lambda}}(q)$
is the probability density function of the $m$-th smallest $q$ value among
$\lambda$ values. This calculation is
well-known in order statistics
(see, e.g.,~\citeapp{Arnold1992OrderStatisticsAPP}).
A short derivation is presented here for the case under consideration.
A single offspring's $q$ value has a probability density of
$p_Q(q)$. In order for this particular $q$ value to be the
$m$-th smallest, there must be $(\lambda-m)$ offspring greater than that,
and $(m-1)$ offspring smaller. This results in the density
$p_Q(q)[1-P_Q(q)]^{\lambda-m}[P_Q(q)]^{m-1}$ where
$P_Q(q) = \mathrm{Pr}[Q \le q]$. Because the number of such
combinations is
$\lambda\binom{\lambda-1}{m-1} = \frac{\lambda!}{(\lambda-m)!(m-1)!}$
one obtains
\begin{equation}
  \label{sec:theoreticalanalysis:eq:pqmlambda}
  p_{q_{m;\lambda}}(q) = \frac{\lambda!}{(\lambda-m)!(m-1)!}
  p_Q(q)[1-P_Q(q)]^{\lambda-m}[P_Q(q)]^{m-1}.
\end{equation}
Insertion of
\Cref{sec:theoreticalanalysis:eq:expectedqmlambda2}
into
\Cref{sec:theoreticalanalysis:eq:qcentroid2}
results in
\begin{align}
  \mathrm{E}[\langle q \rangle] &=
  \frac{1}{\mu}\sum_{m=1}^\mu
  \int_{q=0}^{q=\infty}q\,\frac{\lambda!}{(\lambda-m)!(m-1)!}
  p_Q(q)[1-P_Q(q)]^{\lambda-m}[P_Q(q)]^{m-1}\\
  &=\frac{\lambda!}{\mu}
  \int_{q=0}^{q=\infty}q\,p_Q(q)
  \sum_{m=1}^\mu
  \frac{[1-P_Q(q)]^{\lambda-m}[P_Q(q)]^{m-1}}
       {(\lambda-m)!(m-1)!}.
  \label{sec:theoreticalanalysis:eq:qcentroid3}
\end{align}
The identity
\begin{equation}
  \sum_{m=1}^\mu\frac{P^{m-1}[1-P]^{\lambda-m}}{(m-1)!(\lambda-m)!}
  =\frac{1}{(\lambda-\mu-1)!(\mu-1)!}
  \int_{0}^{1-P}z^{\lambda-\mu-1}(1-z)^{\mu-1}\,\mathrm{d}z
  \label{sec:theoreticalanalysis:eq:Beyer2001_5.14}
\end{equation}
stated in~\citeapp[Eq. (5.14), p. 147]{Beyer2001APP} allows expressing
the sum in
\Cref{sec:theoreticalanalysis:eq:qcentroid3}
as an integral.
Application of
\Cref{sec:theoreticalanalysis:eq:Beyer2001_5.14}
to
\Cref{sec:theoreticalanalysis:eq:qcentroid3}
and expressing the fraction using the binomial coefficient yields
\begin{equation}
  \mathrm{E}[\langle q \rangle] =
  (\lambda-\mu)\binom{\lambda}{\mu}
  \int_{q=0}^{q=\infty}q\,p_Q(q)
  \int_{z=0}^{z=1-P_Q(q)}
  z^{\lambda-\mu-1}(1-z)^{\mu-1}\,\mathrm{d}z\,\mathrm{d}q.
\end{equation}
To proceed further, $z=1-P_Q(y)$ is substituted in the inner integral.
This implies $y=P_Q^{-1}(1-z)$
and $\mathrm{d}z=-p_Q(y)$. The upper and lower bounds for the substituted
integral follow as $y_u=P_Q^{-1}(1-1+P_Q(q))=q$ and
$y_l=P_Q^{-1}(1-0)=\infty$, respectively.
Therefore, one obtains
\begin{align}
  \mathrm{E}&[\langle q \rangle]\notag\\
  &= (\lambda-\mu)\binom{\lambda}{\mu}
  \int_{q=0}^{q=\infty}q\,p_Q(q)
  \int_{y=\infty}^{y=q}
      [1-P_Q(y)]^{\lambda-\mu-1}
      [P_Q(y)]^{\mu-1}(-p_Q(y))\,\mathrm{d}y\,\mathrm{d}q\\
  &= (\lambda-\mu)\binom{\lambda}{\mu}
  \int_{q=0}^{q=\infty}q\,p_Q(q)
  \int_{y=q}^{y=\infty}
  p_Q(y)
  [1-P_Q(y)]^{\lambda-\mu-1}
  [P_Q(y)]^{\mu-1}\,\mathrm{d}y\,\mathrm{d}q.
\end{align}
Changing the order of integration, one finally gets
\begin{equation}
  \mathrm{E}[\langle q \rangle]
  = (\lambda-\mu)\binom{\lambda}{\mu}
  \int_{y=0}^{y=\infty}
  p_Q(y)
  [1-P_Q(y)]^{\lambda-\mu-1}
  [P_Q(y)]^{\mu-1}
  \int_{q=0}^{q=y}q\,p_Q(q)\,\mathrm{d}q\,\mathrm{d}y.
  \label{sec:theoreticalanalysis:eq:qcentroid4}
\end{equation}
Approximations for $P_Q(q)$ and $p_Q(q)$ have been derived
in~\citeapp[Sec. 3.1.2.1.2, pp. 21-39]{SpettelBeyer2018SigmaSaEsConeAPP}
\begin{empheq}[left={P_Q(q) \approx\empheqbiglbrace~}]{align}
    &{P_Q}_{\text{feas}}(q) :=
    \Phi\left(\frac{q-x^{(g)}}
             {\sigma^{(g)}}\right),\text{ for } q>\bar{r}\sqrt{\xi}
    \label{sec:theoreticalanalysis:eq:approximatedPQforprogressfeasible}\\
    &{P_Q}_{\text{infeas}}(q) :=
      \Phi\left(\frac{(1+1/\xi)q-x^{(g)}-\bar{r}/\sqrt{\xi}}
               {\sqrt{{\sigma^{(g)}}^2 + \sigma_r^2/\xi}}\right),
               \text{ otherwise}.
    \label{sec:theoreticalanalysis:eq:approximatedPQforprogress}
\end{empheq}
Taking the derivative with respect to $q$ yields
\begin{empheq}[left={p_Q(q) \approx\empheqbiglbrace~}]{align}
    &{p_Q}_{\text{feas}}(q) =
    \frac{1}{\sqrt{2\pi}\sigma^{(g)}}e^{-\frac{1}{2}\left(\frac{q-x^{(g)}}
      {\sigma^{(g)}}\right)^2},\text{ for } q>\bar{r}\sqrt{\xi}
    \label{sec:theoreticalanalysis:eq:approximatedpQforprogressfeasible}\\
    &\begin{multlined}
    {p_Q}_{\text{infeas}}(q) =
    \left(\frac{(1+1/\xi)}
      {\sqrt{{\sigma^{(g)}}^2 + \sigma_r^2/\xi}}\right)\\
      \times
      \frac{1}{\sqrt{2\pi}}\exp
      \left[-\frac{1}{2}\left(\frac{(1+1/\xi)q-x^{(g)}-\bar{r}/\sqrt{\xi}}
        {\sqrt{{\sigma^{(g)}}^2 + \sigma_r^2/\xi}}\right)^2\right],\\
      \text{ otherwise}.
    \end{multlined}
    \label{sec:theoreticalanalysis:eq:approximatedpQforprogress}
\end{empheq}
They are used to proceed further.
Two cases (being feasible with overwhelming probability and
being infeasible with overwhelming probability) have been distinguished
in the derivation of the $P_Q$ and $p_Q$.
Therefore, those two cases are treated separately for
\Cref{sec:theoreticalanalysis:eq:qcentroid4}.

\subsection{Derivation of
  \texorpdfstring{${\mathrm{E}[\langle q \rangle]}_{\text{feas}}$}
                 {\$\{\textbackslash mathrm\{E\}[\textbackslash langle q \textbackslash rangle]\}\_\{\textbackslash text\{feas\}\}\$}}
For treating the feasible case, insertion of
\Cref{sec:theoreticalanalysis:eq:approximatedPQforprogressfeasible,%
sec:theoreticalanalysis:eq:approximatedpQforprogressfeasible}
into
\Cref{sec:theoreticalanalysis:eq:qcentroid4}
yields
\begin{align}
  &\begin{multlined}
     \mathrm{E}[{\langle q \rangle}_{\text{feas}}]
     \approx (\lambda-\mu)\binom{\lambda}{\mu}
     \int_{y=0}^{y=\infty}
         {p_Q}_{\text{feas}}(y)
         [1-{P_Q}_{\text{feas}}(y)]^{\lambda-\mu-1}
         [{P_Q}_{\text{feas}}(y)]^{\mu-1}\\
         \times\int_{q=0}^{q=y}
         q\,{p_Q}_{\text{feas}}(q)\,\mathrm{d}q\,\mathrm{d}y
  \end{multlined}
  \label{sec:theoreticalanalysis:eq:qcentroidfeas1}\\
  &\begin{multlined}
     \phantom{\mathrm{E}[{\langle q \rangle}_{\text{feas}}] }
     = (\lambda-\mu)\binom{\lambda}{\mu}
     \int_{y=\bar{r}\sqrt{\xi}}^{y=\infty}
     \frac{1}{\sqrt{2\pi}\sigma^{(g)}}e^{-\frac{1}{2}\left(\frac{y-x^{(g)}}
       {\sigma^{(g)}}\right)^2}
         \left[1-\Phi\left(\frac{y-x^{(g)}}
           {\sigma^{(g)}}\right)\right]^{\lambda-\mu-1}\\
         \times\left[\Phi\left(\frac{y-x^{(g)}}
           {\sigma^{(g)}}\right)\right]^{\mu-1}
         \int_{q=\bar{r}\sqrt{\xi}}^{q=y}
         q\,
         \frac{1}{\sqrt{2\pi}\sigma^{(g)}}e^{-\frac{1}{2}\left(\frac{q-x^{(g)}}
       {\sigma^{(g)}}\right)^2}\,\mathrm{d}q\,\mathrm{d}y.
  \end{multlined}
  \label{sec:theoreticalanalysis:eq:qcentroidfeas2}
\end{align}
For solving the inner integral, $t:=\frac{q-x^{(g)}}{\sigma^{(g)}}$ is
substituted. It implies $q=\sigma^{(g)}t+x^{(g)}$
and $\mathrm{d}q=\sigma^{(g)}\,\mathrm{d}t$.
The lower bound follows with $\sigma$-normalization, assuming
$N \rightarrow \infty$, $N \gg {\sigma^*}^{(g)}$, and
knowing that $x^{(g)} \ge \bar{r}\sqrt{\xi}$ holds for
the feasible case. It reads
\begin{equation*}
  t_l = \frac{\bar{r}\sqrt{\xi}-x^{(g)}}{\sigma^{(g)}}
  =\frac{N}{{\sigma^*}^{(g)}}\frac{\bar{r}\sqrt{\xi}-x^{(g)}}{r^{(g)}}
  \simeq -\infty.
\end{equation*}
The application of the substitution results in
\begin{align}
  \int_{q=\bar{r}\sqrt{\xi}}^{q=y}
  &q\,
  \frac{1}{\sqrt{2\pi}\sigma^{(g)}}e^{-\frac{1}{2}\left(\frac{q-x^{(g)}}
    {\sigma^{(g)}}\right)^2}\,\mathrm{d}q\notag\\
  &\simeq\int_{t=-\infty}^{t=\frac{y-x^{(g)}}{\sigma^{(g)}}}
  (\sigma^{(g)}t+x^{(g)})
  \frac{1}{\sqrt{2\pi}\sigma^{(g)}}e^{-\frac{1}{2}t^2}\sigma^{(g)}\mathrm{d}t\\
  &=\sigma^{(g)}\frac{1}{\sqrt{2\pi}}
  \int_{t=-\infty}^{t=\frac{y-x^{(g)}}{\sigma^{(g)}}}
  t\,e^{-\frac{1}{2}t^2}\mathrm{d}t
  +x^{(g)}\int_{t=-\infty}^{t=\frac{y-x^{(g)}}{\sigma^{(g)}}}
  \frac{1}{\sqrt{2\pi}}e^{-\frac{1}{2}t^2}\mathrm{d}t.
  \label{sec:theoreticalanalysis:eq:qcentroidfeasinner1}
\end{align}
The integral in the second summand in
\Cref{sec:theoreticalanalysis:eq:qcentroidfeasinner1}
can be expressed using the cumulative distribution function of the
normal distribution $\Phi(\cdot)$. The integral in the first summand in
\Cref{sec:theoreticalanalysis:eq:qcentroidfeasinner1}
can be solved using the identity
\begin{equation}
  \int_{-\infty}^{x}t\,e^{-\frac{1}{2}t^2}\,\mathrm{d}t=-e^{-\frac{1}{2}x^2}
  \label{sec:theoreticalanalysis:eq:Beyer2001A16}
\end{equation}
derived in~\citeapp[Eq. (A.16), p. 331]{Beyer2001APP}.
Hence, one obtains
\begin{align}
  \int_{q=\bar{r}\sqrt{\xi}}^{q=y}
  &q\,
  \frac{1}{\sqrt{2\pi}\sigma^{(g)}}e^{-\frac{1}{2}\left(\frac{q-x^{(g)}}
    {\sigma^{(g)}}\right)^2}\,\mathrm{d}q
  \simeq
  x^{(g)}\Phi\left(\frac{y-x^{(g)}}{\sigma^{(g)}}\right)
  -\sigma^{(g)}\frac{1}{\sqrt{2\pi}}
  e^{-\frac{1}{2}\left(\frac{y-x^{(g)}}{\sigma^{(g)}}\right)^2}.
  \label{sec:theoreticalanalysis:eq:qcentroidfeasinner2}
\end{align}
\Cref{sec:theoreticalanalysis:eq:qcentroidfeasinner2}
can now be inserted into
\Cref{sec:theoreticalanalysis:eq:qcentroidfeas2}
yielding
\begin{align}
  &\begin{multlined}
     \mathrm{E}[{\langle q \rangle}_{\text{feas}}]
     \approx (\lambda-\mu)\binom{\lambda}{\mu}
     \int_{y=\bar{r}\sqrt{\xi}}^{y=\infty}
     \frac{1}{\sqrt{2\pi}\sigma^{(g)}}e^{-\frac{1}{2}\left(\frac{y-x^{(g)}}
       {\sigma^{(g)}}\right)^2}
         \left[1-\Phi\left(\frac{y-x^{(g)}}
           {\sigma^{(g)}}\right)\right]^{\lambda-\mu-1}\\
         \times\left[\Phi\left(\frac{y-x^{(g)}}
           {\sigma^{(g)}}\right)\right]^{\mu-1}
         x^{(g)}\Phi\left(\frac{y-x^{(g)}}{\sigma^{(g)}}\right)
         \,\mathrm{d}y\\
     -(\lambda-\mu)\binom{\lambda}{\mu}
     \int_{y=\bar{r}\sqrt{\xi}}^{y=\infty}
     \frac{1}{\sqrt{2\pi}\sigma^{(g)}}e^{-\frac{1}{2}\left(\frac{y-x^{(g)}}
       {\sigma^{(g)}}\right)^2}
         \left[1-\Phi\left(\frac{y-x^{(g)}}
           {\sigma^{(g)}}\right)\right]^{\lambda-\mu-1}\\
         \times\left[\Phi\left(\frac{y-x^{(g)}}
           {\sigma^{(g)}}\right)\right]^{\mu-1}
         \sigma^{(g)}\frac{1}{\sqrt{2\pi}}
         e^{-\frac{1}{2}\left(\frac{y-x^{(g)}}{\sigma^{(g)}}\right)^2}
         \,\mathrm{d}y.
  \end{multlined}
  \label{sec:theoreticalanalysis:eq:qcentroidfeas3}
\end{align}
The first term in
\Cref{sec:theoreticalanalysis:eq:qcentroidfeas3}
can be simplified by using
${P_Q}_{\text{feas}}(y)=\Phi\left(\frac{y-x^{(g)}}{\sigma^{(g)}}\right)$.
It yields
\begin{align}
  &\begin{multlined}
     (\lambda-\mu)\binom{\lambda}{\mu}
     \int_{y=\bar{r}\sqrt{\xi}}^{y=\infty}
     \frac{1}{\sqrt{2\pi}\sigma^{(g)}}e^{-\frac{1}{2}\left(\frac{y-x^{(g)}}
       {\sigma^{(g)}}\right)^2}
         \left[1-\Phi\left(\frac{y-x^{(g)}}
           {\sigma^{(g)}}\right)\right]^{\lambda-\mu-1}\\
         \times\left[\Phi\left(\frac{y-x^{(g)}}
           {\sigma^{(g)}}\right)\right]^{\mu-1}
         x^{(g)}\Phi\left(\frac{y-x^{(g)}}{\sigma^{(g)}}\right)
         \,\mathrm{d}y
  \end{multlined}\notag\\
  &\begin{multlined}
     =x^{(g)}(\lambda-\mu)\binom{\lambda}{\mu}
     \int_{y=\bar{r}\sqrt{\xi}}^{y=\infty}
         {p_Q}_{\text{feas}}(y)
         [1-{P_Q}_{\text{feas}}(y)]^{\lambda-\mu-1}
         [{P_Q}_{\text{feas}}(y)]^{\mu}\,\mathrm{d}y.
  \end{multlined}
\end{align}
This can be rewritten resulting in
\begin{align}
  &\begin{multlined}
     x^{(g)}(\lambda-\mu)\binom{\lambda}{\mu}
     \int_{y=\bar{r}\sqrt{\xi}}^{y=\infty}
         {p_Q}_{\text{feas}}(y)
         [1-{P_Q}_{\text{feas}}(y)]^{\lambda-\mu-1}
         [{P_Q}_{\text{feas}}(y)]^{\mu}\,\mathrm{d}y
  \end{multlined}\notag\\
  &\begin{multlined}
     =x^{(g)}(\lambda-\mu)\frac{\lambda!}{(\lambda-\mu)!\mu!}
     \int_{y=\bar{r}\sqrt{\xi}}^{y=\infty}
         {p_Q}_{\text{feas}}(y)
         [1-{P_Q}_{\text{feas}}(y)]^{\lambda-\mu-1}
         [{P_Q}_{\text{feas}}(y)]^{\mu}\,\mathrm{d}y
  \end{multlined}\\
  &\begin{multlined}
     =x^{(g)}\frac{\lambda!}{(\lambda-(\mu+1))!((\mu+1)-1)!}
     \int_{y=\bar{r}\sqrt{\xi}}^{y=\infty}
         {p_Q}_{\text{feas}}(y)
         [1-{P_Q}_{\text{feas}}(y)]^{\lambda-(\mu+1)}\\
         \times[{P_Q}_{\text{feas}}(y)]^{(\mu+1)-1}\,\mathrm{d}y.
   \end{multlined}
\end{align}
Now, by comparing the expression inside the integral
with
\Cref{sec:theoreticalanalysis:eq:pqmlambda},
one observes that this is exactly the probability density function
of the $(\mu+1)$-th best offspring. Because in this section the case
under consideration is the feasible case, in the limit the whole mass
is in ${p_q}_{\text{feas}}$ and no mass in ${p_q}_{\text{infeas}}$. Hence,
integration over all values of $q$ equals $1$. This reads
\begin{align}
  &\begin{multlined}
     x^{(g)}\frac{\lambda!}{(\lambda-(\mu+1))!((\mu+1)-1)!}
     \int_{y=0}^{y=\infty}
         {p_Q}_{\text{feas}}(y)
         [1-{P_Q}_{\text{feas}}(y)]^{\lambda-(\mu+1)}\\
         \times[{P_Q}_{\text{feas}}(y)]^{(\mu+1)-1}\,\mathrm{d}y
      =x^{(g)}
     \underbrace{\int_{y=0}^{y=\infty}
       {p}_{{q_\text{feas}}_{(\mu+1);\lambda}}(y)\,\mathrm{d}y}_{=1}
     =x^{(g)}.
  \end{multlined}
  \label{sec:theoreticalanalysis:eq:Eqcentroidfeasaddend}
\end{align}
The second term in
\Cref{sec:theoreticalanalysis:eq:qcentroidfeas3}
can be simplified by solving the integral using substitution.
The substitution $-t:=\frac{y-x^{(g)}}{\sigma^{(g)}}$ is performed.
It implies
$y=-\sigma^{(g)}t+x^{(g)}$ and $\mathrm{d}y=-\sigma^{(g)}\,\mathrm{d}t$.
The lower bound follows with $\sigma$-normalization, assuming
$N \rightarrow \infty$, $N \gg {\sigma^*}^{(g)}$, and
knowing that $x^{(g)} \ge \bar{r}\sqrt{\xi}$
holds for the feasible case. It reads
$t_l = -\frac{\bar{r}\sqrt{\xi}-x^{(g)}}{\sigma^{(g)}}
=-\frac{N}{{\sigma^*}^{(g)}}\frac{\bar{r}\sqrt{\xi}-x^{(g)}}{r^{(g)}}
\simeq \infty$.
Similarly, the upper bound follows with $\sigma$-normalization and assuming
$N \rightarrow \infty$ and $N \gg {\sigma^*}^{(g)}$ as
$t_u = \lim_{y \rightarrow \infty}\frac{-y+x^{(g)}}{\sigma^{(g)}}
=\lim_{y \rightarrow \infty}\frac{N}{{\sigma^*}^{(g)}}\frac{-y+x^{(g)}}{r^{(g)}}
=\frac{N}{{\sigma^*}^{(g)}r^{(g)}}\lim_{y \rightarrow \infty}(-y+x^{(g)})
\simeq -\infty$.
The integral after substitution reads
\begin{align}
  &\begin{multlined}
     -(\lambda-\mu)\binom{\lambda}{\mu}
     \int_{y=\bar{r}\sqrt{\xi}}^{y=\infty}
     \frac{1}{\sqrt{2\pi}\sigma^{(g)}}e^{-\frac{1}{2}\left(\frac{y-x^{(g)}}
       {\sigma^{(g)}}\right)^2}
     \left[1-\Phi\left(\frac{y-x^{(g)}}
       {\sigma^{(g)}}\right)\right]^{\lambda-\mu-1}\\
     \times\left[\Phi\left(\frac{y-x^{(g)}}
       {\sigma^{(g)}}\right)\right]^{\mu-1}
     \sigma^{(g)}\frac{1}{\sqrt{2\pi}}
     e^{-\frac{1}{2}\left(\frac{y-x^{(g)}}{\sigma^{(g)}}\right)^2}
     \,\mathrm{d}y
  \end{multlined}\notag\\
  &\begin{multlined}
     =-\sigma^{(g)}\frac{\lambda-\mu}{{(\sqrt{2\pi})}^{1+1}}\binom{\lambda}{\mu}
     (-1)\int_{t=\infty}^{t=-\infty}
     e^{-\frac{1+1}{2}t^2}
     [1-\Phi(-t)]^{\lambda-\mu-1}[\Phi(-t)]^{\mu-1}
     \,\mathrm{d}t
  \end{multlined}
  \label{sec:theoreticalanalysis:eq:Eqcentroidfeassubtrahend1}\\
  &\begin{multlined}
     =-\sigma^{(g)}\frac{\lambda-\mu}{{(\sqrt{2\pi})}^{1+1}}\binom{\lambda}{\mu}
     \int_{t=-\infty}^{t=\infty}
     e^{-\frac{1+1}{2}t^2}
     [\Phi(t)]^{\lambda-\mu-1}[1-\Phi(t)]^{\mu-1}
     \,\mathrm{d}t=-\sigma^{(g)}c_{\mu/\mu,\lambda}.
  \end{multlined}
  \label{sec:theoreticalanalysis:eq:Eqcentroidfeassubtrahend2}
\end{align}
From
\Cref{sec:theoreticalanalysis:eq:Eqcentroidfeassubtrahend1}
to
\Cref{sec:theoreticalanalysis:eq:Eqcentroidfeassubtrahend2}
the fact that taking the negative of an integral can be expressed by
exchanging the lower and upper bounds and the identity
$1-\Phi(-t)=\Phi(t)$ have been used. In
\Cref{sec:theoreticalanalysis:eq:Eqcentroidfeassubtrahend2},
the progress coefficient $c_{\mu/\mu,\lambda}$ defined
in~\citeapp[Eq. (6.102), p. 247]{Beyer2001APP} appears. It is defined as
\begin{equation}
  \label{sec:theoreticalanalysis:eq:cmumulambda}
  c_{\mu/\mu,\lambda} :=
  \frac{\lambda-\mu}{2\pi}\binom{\lambda}{\mu}
  \int_{t=-\infty}^{t=\infty}
  e^{-t^2}
  [\Phi(t)]^{\lambda-\mu-1}[1-\Phi(t)]^{\mu-1}
  \,\mathrm{d}t.
\end{equation}
Insertion of
\Cref{sec:theoreticalanalysis:eq:Eqcentroidfeasaddend}
and
\Cref{sec:theoreticalanalysis:eq:Eqcentroidfeassubtrahend2}
into
\Cref{sec:theoreticalanalysis:eq:qcentroidfeas3}
results in
\begin{equation}
  \mathrm{E}[{\langle q \rangle}_{\text{feas}}]
  \approx x^{(g)} - \sigma^{(g)}c_{\mu/\mu,\lambda}.
  \label{sec:theoreticalanalysis:eq:qcentroidfeasinserted}
\end{equation}

\subsection{Derivation of
  \texorpdfstring{${\mathrm{E}[\langle q \rangle]}_{\text{infeas}}$}
                 {\$\{\textbackslash mathrm\{E\}[\textbackslash langle q \textbackslash rangle]\}\_\{\textbackslash text\{infeas\}\}\$}}
For treating the infeasible case, insertion of
\Cref{sec:theoreticalanalysis:eq:approximatedPQforprogress,%
sec:theoreticalanalysis:eq:approximatedpQforprogress}
into
\Cref{sec:theoreticalanalysis:eq:qcentroid4}
yields
\begin{align}
  &\begin{multlined}
     \mathrm{E}[{\langle q \rangle}_{\text{infeas}}]
     \approx (\lambda-\mu)\binom{\lambda}{\mu}
     \int_{y=0}^{y=\bar{r}\sqrt{\xi}}
         {p_Q}_{\text{infeas}}(y)
         [1-{P_Q}_{\text{infeas}}(y)]^{\lambda-\mu-1}
         [{P_Q}_{\text{infeas}}(y)]^{\mu-1}\\
         \times\int_{q=0}^{q=y}
         q\,{p_Q}_{\text{infeas}}(q)\,\mathrm{d}q\,\mathrm{d}y
  \end{multlined}
  \label{sec:theoreticalanalysis:eq:qcentroidinfeas1}\\
  &\begin{multlined}
     \phantom{\mathrm{E}[{\langle q \rangle}_{\text{infeas}}] }
     = (\lambda-\mu)\binom{\lambda}{\mu}
     \int_{y=0}^{y=\bar{r}\sqrt{\xi}}
         {p_Q}_{\text{infeas}}(y)
         [1-{P_Q}_{\text{infeas}}(y)]^{\lambda-\mu-1}
         [{P_Q}_{\text{infeas}}(y)]^{\mu-1}\\
         \times\int_{q=0}^{q=y}
         q\left(\frac{(1+1/\xi)}
              {\sqrt{{\sigma^{(g)}}^2 + \sigma_r^2/\xi}}\right)
              \frac{1}{\sqrt{2\pi}}\exp
              \left[-\frac{1}{2}\left(\frac{(1+1/\xi)q-x^{(g)}-\bar{r}/\sqrt{\xi}}
                {\sqrt{{\sigma^{(g)}}^2 + \sigma_r^2/\xi}}\right)^2\right]\,\mathrm{d}q\,\mathrm{d}y.
  \end{multlined}
  \label{sec:theoreticalanalysis:eq:qcentroidinfeas2}
\end{align}
For solving the inner integral in
\Cref{sec:theoreticalanalysis:eq:qcentroidinfeas2},
the substitution
\begin{equation}
  \frac{(1+1/\xi)q-x^{(g)}-\bar{r}/\sqrt{\xi}}
       {\sqrt{{\sigma^{(g)}}^2 + \sigma_r^2/\xi}} := t
\end{equation}
is used.
It follows that
\begin{equation}
  q=\frac{1}{(1+1/\xi)}
  \left(\sqrt{{\sigma^{(g)}}^2+\sigma_r^2/\xi}\,t
  +x^{(g)}+\bar{r}/\sqrt{\xi}\right)
\end{equation}
and
\begin{equation}
  \mathrm{d}q=
  \frac{\sqrt{{\sigma^{(g)}}^2+\sigma_r^2/\xi}}{(1+1/\xi)}\,\mathrm{d}t.
\end{equation}
Using the normalized ${\sigma^{(g)}}^*$, $\sigma_r \simeq \sigma^{(g)}$
for $N \rightarrow \infty$, and ${\sigma^{(g)}}^* \ll N$
(derived from
\Cref{sec:theoreticalanalysis:eq:approximatedr}),
$t$ can be expressed as
\begin{align}
  t &= \frac{(1+1/\xi)q-x^{(g)}-\bar{r}/\sqrt{\xi}}
  {\sqrt{\frac{{{\sigma^{(g)}}^*}^2{r^{(g)}}^2}{N^2}
      +\frac{{{\sigma^{(g)}}^*}^2{r^{(g)}}^2}{N^2\xi}}}\\
  &= \frac{(1+1/\xi)q-x^{(g)}-\bar{r}/\sqrt{\xi}}
  {\frac{1}{N\sqrt{\xi}}\sqrt{\xi{{\sigma^{(g)}}^*}^2{r^{(g)}}^2
      +{{\sigma^{(g)}}^*}^2{r^{(g)}}^2}}\\
  &= N\sqrt{\xi}\left[\frac{(1+1/\xi)q-x^{(g)}-\bar{r}/\sqrt{\xi}}
  {\sqrt{\xi{{\sigma^{(g)}}^*}^2{r^{(g)}}^2
      +{{\sigma^{(g)}}^*}^2{r^{(g)}}^2}}\right]\\
  &= N\sqrt{\xi}\left[\frac{(1+1/\xi)q-x^{(g)}-\bar{r}/\sqrt{\xi}}
  {{{\sigma^{(g)}}^*}{r^{(g)}}\sqrt{\xi+1}}\right].
\end{align}
The lower bound in the transformed integral therefore follows
assuming $\xi \gg 1$, $N \rightarrow \infty$, using
$\infty > \bar{r} \simeq r^{(g)} \ge 0$,
and knowing that $0 \le x^{(g)} < \infty$ as
\begin{align}
  t_l &= N\sqrt{\xi}\left[\frac{(1+1/\xi)0-x^{(g)}-\bar{r}/\sqrt{\xi}}
  {{{\sigma^{(g)}}^*}{r^{(g)}}\sqrt{\xi+1}}\right]\\
  &= N\sqrt{\xi}\left[\frac{-x^{(g)}-\bar{r}/\sqrt{\xi}}
  {{{\sigma^{(g)}}^*}{r^{(g)}}\sqrt{\xi+1}}\right]\\
  &\simeq N\underbrace{\left[\frac{-x^{(g)} - \bar{r}/\sqrt{\xi}}
  {{{\sigma^{(g)}}^*}{r^{(g)}}}\right]}_{\le 0}\\
  &\simeq -\infty.
\end{align}
The transformed integral writes
\begin{align}
  &\begin{multlined}
  \int_{q=0}^{q=y}
  q\left(\frac{(1+1/\xi)}
  {\sqrt{{\sigma^{(g)}}^2 + \sigma_r^2/\xi}}\right)
  \frac{1}{\sqrt{2\pi}}\exp
  \left[-\frac{1}{2}\left(\frac{(1+1/\xi)q-x^{(g)}-\bar{r}/\sqrt{\xi}}
    {\sqrt{{\sigma^{(g)}}^2 + \sigma_r^2/\xi}}\right)^2\right]\,\mathrm{d}q
  \end{multlined}\notag\\
  &\begin{multlined}
     =\int_{t=-\infty}^{t=\frac{(1+1/\xi)y-x^{(g)}-\bar{r}/\sqrt{\xi}}
       {\sqrt{{\sigma^{(g)}}^2 + \sigma_r^2/\xi}}}
     \frac{1}{(1+1/\xi)}
     \left(\sqrt{{\sigma^{(g)}}^2+\sigma_r^2/\xi}\,t
     +x^{(g)}+\bar{r}/\sqrt{\xi}\right)\frac{1}{\sqrt{2\pi}}e^{-\frac{1}{2}t^2}
     \,\mathrm{d}t.
  \end{multlined}
\end{align}
Use of
\Cref{sec:theoreticalanalysis:eq:Beyer2001A16}
for the first summand and the cumulative distribution function of the standard
normal distribution $\Phi(\cdot)$ for the second and third summands,
this can further be rewritten resulting in
\begin{align}
  &\begin{multlined}
     \int_{t=-\infty}^{t=\frac{(1+1/\xi)y-x^{(g)}-\bar{r}/\sqrt{\xi}}
       {\sqrt{{\sigma^{(g)}}^2 + \sigma_r^2/\xi}}}
     \frac{1}{(1+1/\xi)}
     \left(\sqrt{{\sigma^{(g)}}^2+\sigma_r^2/\xi}\,t
     +x^{(g)}+\bar{r}/\sqrt{\xi}\right)\frac{1}{\sqrt{2\pi}}e^{-\frac{1}{2}t^2}
     \,\mathrm{d}t
  \end{multlined}\notag\\
  &\begin{multlined}
     =-\frac{\sqrt{{\sigma^{(g)}}^2 + \sigma_r^2/\xi}}{(1+1/\xi)}
     \frac{1}{\sqrt{2\pi}}\exp\left[-\frac{1}{2}\left(
       \frac{(1+1/\xi)y-x^{(g)}-\bar{r}/\sqrt{\xi}}
            {\sqrt{{\sigma^{(g)}}^2 + \sigma_r^2/\xi}}\right)^2\right]\\
     +\frac{1}{(1+1/\xi)}x^{(g)}
     \underbrace{\Phi\left(\frac{(1+1/\xi)y-x^{(g)}-\bar{r}/\sqrt{\xi}}
       {\sqrt{{\sigma^{(g)}}^2 + \sigma_r^2/\xi}}\right)}
     _{={P_Q}_{\text{infeas}}}\\
     +\frac{1}{(1+1/\xi)}\bar{r}/\sqrt{\xi}
     \underbrace{\Phi\left(\frac{(1+1/\xi)y-x^{(g)}-\bar{r}/\sqrt{\xi}}
       {\sqrt{{\sigma^{(g)}}^2 +
           \sigma_r^2/\xi}}\right)}_{={P_Q}_{\text{infeas}}}
     \,\mathrm{d}t.
  \end{multlined}
  \label{sec:theoreticalanalysis:eq:qcentroidinfeasinner1}
\end{align}
Insertion of
\Cref{sec:theoreticalanalysis:eq:qcentroidinfeasinner1}
back into
\Cref{sec:theoreticalanalysis:eq:qcentroidinfeas2}
leads to
\begin{align}
  &\begin{multlined}
     \mathrm{E}[{\langle q \rangle}_{\text{infeas}}]
     \approx (\lambda-\mu)\binom{\lambda}{\mu}
     \int_{y=0}^{y=\bar{r}\sqrt{\xi}}
         {p_Q}_{\text{infeas}}(y)
         [1-{P_Q}_{\text{infeas}}(y)]^{\lambda-\mu-1}
         [{P_Q}_{\text{infeas}}(y)]^{\mu-1}\\
         \times(-1)\frac{\sqrt{{\sigma^{(g)}}^2 + \sigma_r^2/\xi}}{(1+1/\xi)}
         \frac{1}{\sqrt{2\pi}}\exp\left[-\frac{1}{2}\left(
           \frac{(1+1/\xi)y-x^{(g)}-\bar{r}/\sqrt{\xi}}
                {\sqrt{{\sigma^{(g)}}^2 + \sigma_r^2/\xi}}\right)^2\right]
         \,\mathrm{d}y\\
         +\frac{1}{(1+1/\xi)}x^{(g)}
         \underbrace{
         \int_{y=0}^{\infty}p_{{q_{\text{infeas}}}_{(\mu+1);\lambda}}
         \,\mathrm{d}y}_{=1}\\
         +\frac{1}{(1+1/\xi)}\bar{r}/\sqrt{\xi}
         \underbrace{
         \int_{y=0}^{\infty}p_{{q_{\text{infeas}}}_{(\mu+1);\lambda}}
         \,\mathrm{d}y}_{=1}.
  \end{multlined}
  \label{sec:theoreticalanalysis:eq:qcentroidinfeasinserted}
\end{align}
The second and third summands are the result of the same argument as the
one leading to
\Cref{sec:theoreticalanalysis:eq:Eqcentroidfeasaddend}.
For solving the integral in the first summand,
\begin{equation}
  \frac{(1+1/\xi)y-x^{(g)}-\bar{r}/\sqrt{\xi}}
       {\sqrt{{\sigma^{(g)}}^2 + \sigma_r^2/\xi}} := -t
\end{equation}
is substituted.
It follows that
\begin{equation}
  y=-\frac{1}{(1+1/\xi)}
  \left(\sqrt{{\sigma^{(g)}}^2+\sigma_r^2/\xi}\,t
  +x^{(g)}+\bar{r}/\sqrt{\xi}\right)
\end{equation}
and
\begin{equation}
  \mathrm{d}y=
  -\frac{\sqrt{{\sigma^{(g)}}^2+\sigma_r^2/\xi}}{(1+1/\xi)}\,\mathrm{d}t.
\end{equation}
Using the normalized ${\sigma^{(g)}}^*$, $\sigma_r \simeq \sigma^{(g)}$
for $N \rightarrow \infty$, and ${\sigma^{(g)}}^* \ll N$
(derived from
\Cref{sec:theoreticalanalysis:eq:approximatedr}),
$t$ can be expressed as
\begin{align}
  t &= -\frac{(1+1/\xi)q-x^{(g)}-\bar{r}/\sqrt{\xi}}
  {\sqrt{\frac{{{\sigma^{(g)}}^*}^2{r^{(g)}}^2}{N^2}
      +\frac{{{\sigma^{(g)}}^*}^2{r^{(g)}}^2}{N^2\xi}}}\\
  &= -\frac{(1+1/\xi)q-x^{(g)}-\bar{r}/\sqrt{\xi}}
  {\frac{1}{N\sqrt{\xi}}\sqrt{\xi{{\sigma^{(g)}}^*}^2{r^{(g)}}^2
      +{{\sigma^{(g)}}^*}^2{r^{(g)}}^2}}\\
  &= -N\sqrt{\xi}\left[\frac{(1+1/\xi)q-x^{(g)}-\bar{r}/\sqrt{\xi}}
  {\sqrt{\xi{{\sigma^{(g)}}^*}^2{r^{(g)}}^2
      +{{\sigma^{(g)}}^*}^2{r^{(g)}}^2}}\right]\\
  &= -N\sqrt{\xi}\left[\frac{(1+1/\xi)q-x^{(g)}-\bar{r}/\sqrt{\xi}}
  {{{\sigma^{(g)}}^*}{r^{(g)}}\sqrt{\xi+1}}\right].
\end{align}
The lower bound in the transformed integral therefore follows
assuming $\xi \gg 1$, $N \rightarrow \infty$, using
$\infty > \bar{r} \simeq r^{(g)} \ge 0$,
and knowing that $0 \le x^{(g)} < \infty$ as
\begin{align}
  t_l &= -N\sqrt{\xi}\left[\frac{(1+1/\xi)0-x^{(g)}-\bar{r}/\sqrt{\xi}}
  {{{\sigma^{(g)}}^*}{r^{(g)}}\sqrt{\xi+1}}\right]\\
  &= -N\sqrt{\xi}\left[\frac{-x^{(g)}-\bar{r}/\sqrt{\xi}}
  {{{\sigma^{(g)}}^*}{r^{(g)}}\sqrt{\xi+1}}\right]\\
  &\simeq -N\underbrace{\left[\frac{-x^{(g)} - \bar{r}/\sqrt{\xi}}
  {{{\sigma^{(g)}}^*}{r^{(g)}}}\right]}_{\le 0}\\
  &\simeq \infty.
\end{align}
Similarly, the upper bound follows with the same assumptions
and using the fact that the case under consideration is
the infeasible case, i.e., $x^{(g)}\le\bar{r}\sqrt{\xi}$
\begin{align}
  t_u &= -N\sqrt{\xi}\left[\frac{(1+1/\xi)\bar{r}\sqrt{\xi}
  -x^{(g)}-\bar{r}/\sqrt{\xi}}
  {{{\sigma^{(g)}}^*}{r^{(g)}}\sqrt{\xi+1}}\right]\\
  &=-N\underbrace{\left[\frac{\bar{r}\sqrt{\xi}-x^{(g)}}
  {{{\sigma^{(g)}}^*}{r^{(g)}}}\right]}_{\ge 0}\\
  &\simeq -\infty.
\end{align}
The transformed integral writes
\begin{align}
  &\begin{multlined}
    (\lambda-\mu)\binom{\lambda}{\mu}
     \int_{y=0}^{y=\bar{r}\sqrt{\xi}}
         {p_Q}_{\text{infeas}}(y)
         [1-{P_Q}_{\text{infeas}}(y)]^{\lambda-\mu-1}
         [{P_Q}_{\text{infeas}}(y)]^{\mu-1}\\
         \times(-1)\frac{\sqrt{{\sigma^{(g)}}^2 + \sigma_r^2/\xi}}{(1+1/\xi)}
         \frac{1}{\sqrt{2\pi}}\exp\left[-\frac{1}{2}\left(
           \frac{(1+1/\xi)y-x^{(g)}-\bar{r}/\sqrt{\xi}}
                {\sqrt{{\sigma^{(g)}}^2 + \sigma_r^2/\xi}}\right)^2\right]
         \,\mathrm{d}y
  \end{multlined}\notag\\
  &\begin{multlined}
    =-\int_{t=\infty}^{t=-\infty}
     (\lambda-\mu)\binom{\lambda}{\mu}
     (-1)\frac{1}{\sqrt{2\pi}}e^{-\frac{1}{2}t^2}
     [1-\Phi(-t)]^{\lambda-\mu-1}[\Phi(-t)]^{\mu-1}
     \frac{\sqrt{{\sigma^{(g)}}^2+\sigma_r^2/\xi}}{(1+1/\xi)}
     \,\mathrm{d}t
  \end{multlined}\\
  &\begin{multlined}
    =-\frac{\sqrt{{\sigma^{(g)}}^2+\sigma_r^2/\xi}}{(1+1/\xi)}
    \underbrace{\frac{\lambda-\mu}{{(\sqrt{2\pi})}^{1+1}}\binom{\lambda}{\mu}
    \int_{t=-\infty}^{t=\infty}
     e^{-\frac{1+1}{2}t^2}
     [\Phi(t)]^{\lambda-\mu-1}[1-\Phi(t)]^{\mu-1}
     \,\mathrm{d}t}_{=c_{\mu/\mu,\lambda}}.
  \end{multlined}
  \label{sec:theoreticalanalysis:eq:qcentroidinfeasinsertedintegral}
\end{align}
Insertion of
\Cref{sec:theoreticalanalysis:eq:qcentroidinfeasinsertedintegral}
into
\Cref{sec:theoreticalanalysis:eq:qcentroidinfeasinserted}
finally yields
\begin{align}
  \mathrm{E}[{\langle q \rangle}_{\text{infeas}}]
  &\approx
  \frac{x^{(g)}}{(1+1/\xi)}
  + \frac{\bar{r}}{\sqrt{\xi}(1+1/\xi)}
  - \frac{\sqrt{{\sigma^{(g)}}^2+\sigma_r^2/\xi}}{(1+1/\xi)}
  c_{\mu/\mu,\lambda}\\
  &= \frac{\xi}{1+\xi}\left(x^{(g)} + \bar{r}/\sqrt{\xi}\right)
  - \frac{\xi}{1+\xi}\sqrt{{\sigma^{(g)}}^2+\sigma_r^2/\xi}
  c_{\mu/\mu,\lambda}.
  \label{sec:theoreticalanalysis:eq:qcentroidinfeasinsertedfinal}
\end{align}

\newpage

\section{Derivation of the \texorpdfstring{$r$}{\$r\$} Progress Rate}
\label{appendix:subsec:rprogress}
From the definition of the progress rate
(\Cref{sec:theoreticalanalysis:eq:varphir})
and the pseudo-code of the
ES (\Cref{sec:algorithm:alg:es},
\Cref{sec:algorithm:alg:es:bestqr,%
sec:algorithm:alg:es:r})
it follows that
\begin{align}
  \varphi_{r}(x^{(g)}, r^{(g)}, \sigma^{(g)})
  &=r^{(g)} - \mathrm{E}[r^{(g + 1)}\,|\,x^{(g)}, r^{(g)}, \sigma^{(g)}]
  \label{sec:theoreticalanalysis:eq:rprogressratedef1}\\
  &= r^{(g)} - \mathrm{E}[\langle q_r \rangle\,|\,x^{(g)}, r^{(g)}, \sigma^{(g)}].
  \label{sec:theoreticalanalysis:eq:rprogressratedef2}
\end{align}
The computation
$\mathrm{E}[\langle q_r \rangle\,|\,x^{(g)}, r^{(g)}, \sigma^{(g)}]$
requires the evaluation of the square root of a sum of squares.
This is because $\langle q_r \rangle$ is the distance from the cone
axis \emph{after} centroid computation. Therefore, the computation of
$\mathrm{E}[\langle q_r \rangle\,|\,x^{(g)}, r^{(g)}, \sigma^{(g)}]$
is not analogous to the computation of the $x$ progress rate.
For the computation of the $r$ progress rate, two cases are distinguished.
First, the case of ``being far'' from the cone boundary is considered
(feasible offspring with overwhelming probability). Second,
the case of ``being on'' the cone boundary is considered
(infeasible offspring with overwhelming probability). Both cases
are then combined with the single offspring feasibility probability.
The feasible case is similar to the sphere model because the projection
can be ignored. For the infeasible case, a geometric approach is pursued.

\subsection{The Approximate \texorpdfstring{$r$}{\$r\$}
  Progress Rate in the case that the
  probability of feasible offspring tends to \texorpdfstring{$1$}{\$1\$}}
For the case that the probability for infeasible offspring is negligible,
generated offspring do not need to be repaired. The mutation vectors
are therefore not altered.
Hence, from
\Crefrange{sec:algorithm:alg:es:offspringsigma}
{sec:algorithm:alg:es:generateoffspring}
of
\Cref{sec:algorithm:alg:es},
\begin{equation}
    \tilde{\mathbf{x}}_{m;\lambda} =
    \mathbf{x}^{(g)} + \tilde{\sigma}_{m;\lambda}\mathbf{z}_{m;\lambda}
\end{equation}
follows directly.
Now, $\tau$ is assumed to be small such that
$\tilde{\sigma}_{l} \simeq \sigma^{(g)}$ for $l \in \{1,\ldots,\lambda\}$.
Hence,
\begin{equation}
    \tilde{\mathbf{x}}_{m;\lambda} \simeq
    \mathbf{x}^{(g)} + \sigma^{(g)}\mathbf{z}_{m;\lambda}
\end{equation}
and subsequently
\begin{equation}
    \langle\tilde{\mathbf{x}}\rangle \simeq
    \mathbf{x}^{(g)} +
    \sigma^{(g)}\frac{1}{\mu}\sum_{m=1}^{\mu}\mathbf{z}_{m;\lambda}
\end{equation}
can be written. With this,
\begin{align}
    \begin{split}
        \langle{q_r}\rangle_{\text{feas}} = \langle\tilde{r}\rangle
        &\simeq
        \sqrt{\sum_{k=2}^{N}\left(
        (\mathbf{x}^{(g)})_k +
        \sigma^{(g)}\frac{1}{\mu}\sum_{m=1}^{\mu}(\mathbf{z}_{m;\lambda})_k
        \right)^2}\\
        &=
        \sqrt{\underbrace{\sum_{k=2}^{N}{(\mathbf{x}^{(g)})}_k^2
              + \sum_{k=2}^{N}\left(2\sigma^{(g)}{(\mathbf{x}^{(g)})}_k
                \frac{1}{\mu}\sum_{m=1}^{\mu}(\mathbf{z}_{m;\lambda})_k\right)
              + \sum_{k=2}^{N}\left(\sigma^{(g)}\frac{1}{\mu}\sum_{m=1}^{\mu}
                (\mathbf{z}_{m;\lambda})_k\right)^2}_{=\langle\tilde{r}\rangle^2}}
    \end{split}
    \label{sec:theoreticalanalysis:eq:rcentroid}
\end{align}
follows for the distance from the cone axis $\langle\tilde{r}\rangle$ of
$\langle\tilde{\mathbf{x}}\rangle$.
To proceed with
\Cref{sec:theoreticalanalysis:eq:rcentroid},
it is assumed that the deviation of the expression under the square root
from its mean is small. This allows the use of a Taylor expansion that yields
the linear approximation
\begin{align}
  \sqrt{\langle\tilde{r}\rangle^2}
  &=\sqrt{\mathrm{E}\left[\langle\tilde{r}\rangle^2\right]}
  +\frac{1}{2}
   \frac{\langle\tilde{r}\rangle^2
           -\mathrm{E}\left[\langle\tilde{r}\rangle^2\right]}
        {\sqrt{\mathrm{E}\left[\langle\tilde{r}\rangle^2\right]}}
  + \cdots.
  \label{sec:theoreticalanalysis:eq:rcentroid3}
\end{align}
Taking the expected value, the second summand in
\Cref{sec:theoreticalanalysis:eq:rcentroid3}
vanishes. With the further assumption that the higher order terms
are negligible, the approximation writes
\begin{equation}
  \mathrm{E}\left[\sqrt{\langle\tilde{r}\rangle^2}\right]
  \approx\sqrt{\mathrm{E}\left[\langle\tilde{r}\rangle^2\right]}.
  \label{sec:theoreticalanalysis:eq:rcentroid4}
\end{equation}
Reinsertion of
\Cref{sec:theoreticalanalysis:eq:rcentroid}
into the approximation
(\Cref{sec:theoreticalanalysis:eq:rcentroid4})
results in
\begin{align}
    &\mathrm{E}[\langle\tilde{r}\rangle]\notag\\
    &\approx
    \sqrt{\underbrace{\sum_{k=2}^{N}{(\mathbf{x}^{(g)})}_k^2}_{={r^{(g)}}^2}
        + \sum_{k=2}^{N}\left(2\sigma^{(g)}{(\mathbf{x}^{(g)})}_k
            \frac{1}{\mu}\sum_{m=1}^{\mu}\mathrm{E}[(\mathbf{z}_{m;\lambda})_k]\right)
        + \sum_{k=2}^{N}\frac{{\sigma^{(g)}}^2}{\mu^2}\mathrm{E}\left[\left(\sum_{m=1}^{\mu}
            (\mathbf{z}_{m;\lambda})_k\right)^2\right]}
  \label{sec:theoreticalanalysis:eq:rcentroid5}
\end{align}
As selection is done in direction of the cone axis,
the $(\mathbf{z}_{m;\lambda})_k$ for $k \in \{2,\ldots,N\}$
are selectively neutral (because they are perpendicular to
the cone axis). Hence, $(\mathbf{z}_{m;\lambda})_k \sim \mathcal{N}(0,1)$
and consequently
\begin{equation}
    \mathrm{E}[(\mathbf{z}_{m;\lambda})_k] = 0
    \label{sec:theoreticalanalysis:eq:expectation}
\end{equation}
holds. The expression
$\sum_{k=2}^{N}\frac{{\sigma^{(g)}}^2}{\mu^2}
\mathrm{E}\left[\left(\sum_{m=1}^{\mu}
(\mathbf{z}_{m;\lambda})_k\right)^2\right]$ can be rewritten by
expanding the squared sum term. It writes

\begin{align}
\sum_{k=2}^{N}\frac{{\sigma^{(g)}}^2}{\mu^2}
\mathrm{E}\left[\left(\sum_{m=1}^{\mu}
(\mathbf{z}_{m;\lambda})_k\right)^2\right]
&=\frac{{\sigma^{(g)}}^2}{\mu^2}\sum_{k=2}^{N}\sum_{m=1}^{\mu}\sum_{p=1}^{\mu}
  \mathrm{E}\left[(\mathbf{z}_{m;\lambda})_k
                  (\mathbf{z}_{p;\lambda})_k\right]
  \label{sec:theoreticalanalysis:eq:rcentroidnext2}\\
&\begin{multlined}=\frac{{\sigma^{(g)}}^2}{\mu^2}\sum_{k=2}^{N}\sum_{m=1}^{\mu}
                   \mathrm{E}\left[(\mathbf{z}_{m;\lambda})_k^2\right]\\
                   \hspace{0.5cm}+\frac{{\sigma^{(g)}}^2}{\mu^2}\sum_{k=2}^{N}\sum_{m=1}^{\mu}
                                   \sum_{\substack{p=1\\p \neq m}}^{\mu}
                   \mathrm{E}\left[(\mathbf{z}_{m;\lambda})_k
                                   (\mathbf{z}_{p;\lambda})_k\right]
                   \end{multlined}
  \label{sec:theoreticalanalysis:eq:rcentroidnext3}
\end{align}
It is important to note that the second summand in
\Cref{sec:theoreticalanalysis:eq:rcentroidnext3}
vanishes. The reason for that is twofold.
First, as the selection only depends on the direction of the cone
axis ($x_1$ direction), $(\mathbf{z}_{m;\lambda})_{2..N}$ and
$(\mathbf{z}_{p;\lambda})_{2..N}$ are statistically
independent because they are perpendicular to the $x_1$ direction.
Second, all directions of $(\mathbf{z}_{m;\lambda})_{2..N}$ and
$(\mathbf{z}_{p;\lambda})_{2..N}$ are equally probable
because of the isotropy of the mutations.
Hence, $(\mathbf{z}_{m;\lambda})_{2..N}$ and
$(\mathbf{z}_{p;\lambda})_{2..N}$ average out in expectation. Consequently, for
fixed $m$ and $p$ with $m \neq p$,
\begin{equation}
  \begin{multlined}
    \sum_{k=2}^{N}
    \mathrm{E}\left[(\mathbf{z}_{m;\lambda})_k
                   (\mathbf{z}_{p;\lambda})_k\right]
    =\mathrm{E}\left[(\mathbf{z}_{m;\lambda})_{2..N}^T
                    (\mathbf{z}_{p;\lambda})_{2..N}\right]\\
    =\mathrm{E}\left[(\mathbf{z}_{m;\lambda})_{2..N}^T\right]
     \mathrm{E}\left[(\mathbf{z}_{p;\lambda})_{2..N}\right] = 0
  \end{multlined}
  \label{sec:theoreticalanalysis:eq:rcentroidnext4}
\end{equation}
holds.
With
\Cref{sec:theoreticalanalysis:eq:expectation}
and
\Cref{sec:theoreticalanalysis:eq:rcentroidnext4},
\Cref{sec:theoreticalanalysis:eq:rcentroid5}
simplifies to
\begin{align}
    \mathrm{E}[\langle\tilde{r}\rangle]
    &\approx\sqrt{{r^{(g)}}^2 + \frac{{\sigma^{(g)}}^2}{\mu^2}
    \sum_{m=1}^{\mu}\mathrm{E}
    \left[\sum_{k=2}^N(\mathbf{z}_{m;\lambda})_k^2\right]}
\end{align}
Again, because selection is done in the $x_1$ direction,
the expression
$\mathrm{E}\left[\sum_{k=2}^N(\mathbf{z}_{m;\lambda})_k^2\right]$
is the expected value of the sum of $N-1$ standard normally
distributed random variables. It it therefore the expected value
of a $\chi$-distributed random variable with $N-1$ degrees of
freedom. Hence,
\begin{equation}
    \mathrm{E}\left[\sum_{k=2}^N(\mathbf{z}_{m;\lambda})_k^2\right] = N - 1
\end{equation}
holds which implies
\begin{align}
    \mathrm{E}[{\langle{q_r}\rangle}_{\text{feas}}]
    =\mathrm{E}[\langle\tilde{r}\rangle]
    &\approx
    \sqrt{{r^{(g)}}^2 + \frac{{\sigma^{(g)}}^2}{\mu}(N-1)}\\
    &=
    r^{(g)}\sqrt{1 + \frac{{{\sigma^{(g)}}^*}^2}{N^2\mu}(N-1)}\\
    &\simeq
    r^{(g)}\sqrt{1 + \frac{{{\sigma^{(g)}}^*}^2}{\mu N}}
\end{align}
With this result, the normalized $r$ progress rate
for the feasible case can be formulated as
\begin{align}
    {\varphi_r^*}_{\text{feas}}
    &=
    \frac{N(r^{(g)}-\mathrm{E}[{\langle{q_r}\rangle}_{\text{feas}}])}{r^{(g)}}\\
    &\approx
    N\left(1-\sqrt{1 + \frac{{{\sigma^{(g)}}^*}^2}{\mu N}}\right)
    \label{sec:theoreticalanalysis:eq:varphirnormalizedfeas}
\end{align}

\subsection{The Approximate \texorpdfstring{$r$}{\$r\$}
  Progress Rate in the case that the
  probability of feasible offspring tends to \texorpdfstring{$0$}{\$0\$}}
\begin{figure}
  \centering
  \includegraphics{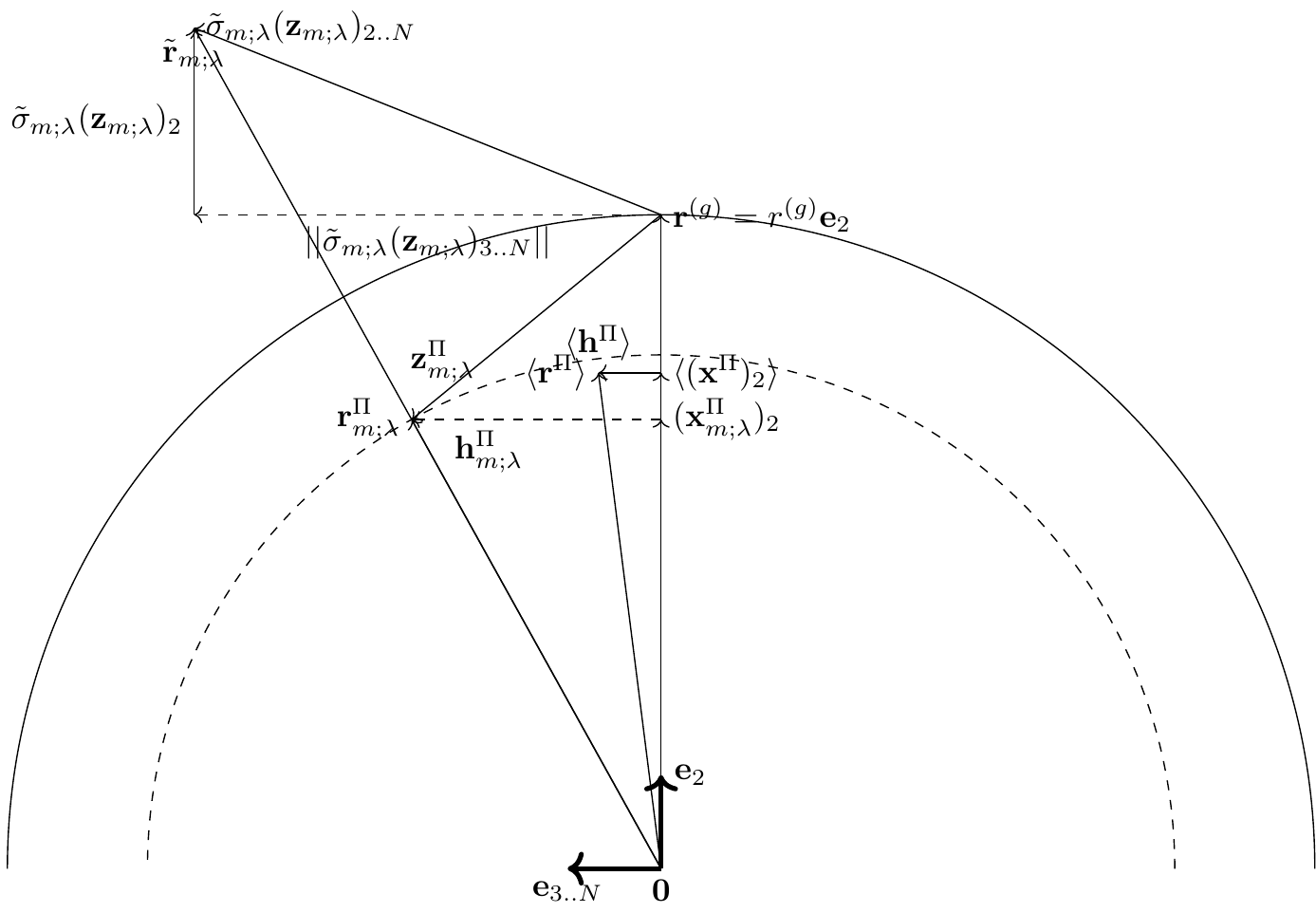}
    \caption[Visualization of the geometrical approach for deriving
             ${\varphi_r}_{\text{infeas}}$ for the multi-recombinative
             ES applied to the conically constrained problem.]
            {Visualization of the geometrical approach for deriving
             ${\varphi_r}_{\text{infeas}}$ for the multi-recombinative
             ES applied to the conically constrained problem.}
    \label{sec:theoreticalanalysis:fig:rprogressinfeasgeometry}
\end{figure}

For the case that the probability of feasible offspring tends to $0$,
the aim is to relate the $r$ progress rate to the $x$ progress rate.
It has also been done in this way for the $(1,\lambda)$ case in
\Cref{sec:theoreticalanalysis:subsec:microscopic:r}.
Here, a geometrical approach is pursued that is visualized in
\Cref{sec:theoreticalanalysis:fig:rprogressinfeasgeometry}.
The visualization shows the $2..N$ space of the conically constrained problem.
That is, the view in direction of the cone axis ($x_1$ direction) is shown.
The $x_1$ direction can be imagined ``coming out'' of the paper
perpendicular to $\mathbf{e}_2$ and $\mathbf{e}_{3..N}$. For this constellation,
the optimal value can be imagined being directly ``on the paper''.
The origin is indicated as $\mathbf{0}$ and the coordinate axes in the
$2..N$ space are indicated as $\mathbf{e}_2$ and $\mathbf{e}_{3..N}$.
The coordinate
system is rotated such that the distance from the cone axis of the
parental individual corresponds to the parental parameter vector's
value in the second dimension, i.e.,
$\mathbf{x}^{(g)} = (x^{(g)}, r^{(g)}, 0, \ldots, 0)^T$. The parental
individual is therefore located at the position indicated by
$\mathbf{r}^{(g)}$. The solid half circle indicates the cone boundary
at $x_1 = x^{(g)}$. As the case under consideration is the one
in which the probability of feasible offspring tends to $0$,
the parental individual is assumed to be on the cone
boundary\footnote{An asymptotic expression for the feasibility probability
has been derived as
$
  P_{\text{feas}}(x^{(g)}, r^{(g)}, \sigma^{(g)})
  \simeq
    \Phi\left[
      \frac{1}{\sigma^{(g)}}
      \left(\frac{x^{(g)}}{\sqrt{\xi}}-\bar{r}\right)
      \right]
  =
    \Phi\left[
      \frac{N}{{\sigma^{(g)}}^*}
      \left(\frac{x^{(g)}}{r^{(g)}\sqrt{\xi}}-\frac{\bar{r}}{r^{(g)}}\right)
      \right]
$
(see~\citeapp[Sec. 3.1.2.1.2.8, pp. 44-49]{SpettelBeyer2018SigmaSaEsConeAPP}).
Note that for the case that the parental individual is on the cone
boundary, $r^{(g)}=\frac{x^{(g)}}{\sqrt{\xi}}$ holds. Further,
$\bar{r} > r^{(g)}$ follows from
\Cref{sec:theoreticalanalysis:eq:approximatedr}.
Hence, the feasibility probability tends to $0$ for the case of
the parental individual being on the cone boundary for $N \rightarrow \infty$
with $N \gg {\sigma^{(g)}}^*$.}.
The mutation vector $\tilde{\sigma}_{m;\lambda}(\mathbf{z}_{m;\lambda})_{2..N}$
indicates the mutation vector in the $2..N$ space leading to the
$m$-th best offspring before projection $\tilde{\mathbf{r}}_{m;\lambda}$.
This mutation vector can be decomposed
into vectors in direction of $\mathbf{e}_2$ and $\mathbf{e}_{3..N}$.
The lengths of those vectors
are indicated as $\tilde{\sigma}_{m;\lambda}(\mathbf{z}_{m;\lambda})_{2}$
and $||\tilde{\sigma}_{m;\lambda}(\mathbf{z}_{m;\lambda})_{3..N}||$,
respectively. The value in direction
of the cone boundary of the $m$-th best offspring after projection
$(\mathbf{x}^{\Pi}_{m;\lambda})_{1}$
is assumed such that the dashed half circle indicates the cone boundary
corresponding to $(\mathbf{x}^{\Pi}_{m;\lambda})_{1}$.
As the dashed half
circle is drawn smaller than the solid half circle, it is assumed that
the $m$-th best offspring after projection $\mathbf{r}^{\Pi}_{m;\lambda}$
has a smaller value in direction of $x_1$ than the parental individual.
The $m$-th best offspring under consideration is considered infeasible. Hence,
it is projected onto the cone boundary.
Its decomposition into vectors along the $\mathbf{e}_2$ and $\mathbf{e}_{3..N}$
axes are indicated as $(\mathbf{x}^{\Pi}_{m;\lambda})_2$ and
$\mathbf{h}^{\Pi}_{m;\lambda}$, respectively. The centroid of the best
$\mu$ offspring after projection has to be computed for the calculation
of the $r$ progress rate. This centroid is
indicated as $\langle \mathbf{r}^{\Pi} \rangle$ with its decompositions
in $\mathbf{e}_2$ and $\mathbf{e}_{3..N}$ directions
$\langle (\mathbf{x}^{\Pi})_2 \rangle$ and
$\langle \mathbf{h}^{\Pi} \rangle$, respectively.

${\varphi_r}_{\text{infeas}}$
can be computed by geometric considerations with the help of
\Cref{sec:theoreticalanalysis:fig:rprogressinfeasgeometry}.
Note that
${\langle{q_r}\rangle}_{\text{infeas}}=||\langle\mathbf{r}^{\Pi}\rangle||$
holds.
Hence,
\begin{equation}
    {\varphi_r}_{\text{infeas}} =
    r^{(g)} - \mathrm{E}[||\langle\mathbf{r}^{\Pi}\rangle||]
    \label{sec:theoreticalanalysis:eq:progressinfeasdef}
\end{equation}
follows and $\mathrm{E}[||\langle\mathbf{r}^{\Pi}\rangle||]$ needs
to be derived. By Pythagoras' theorem,
\begin{equation}
    \mathrm{E}[||\langle\mathbf{r}^{\Pi}\rangle||] =
    \mathrm{E}\left[\sqrt{{\langle (\mathbf{x}^{\Pi})_2 \rangle}^2+
                          {||\langle \mathbf{h}^{\Pi} \rangle||}^2}\right]
\end{equation}
follows. With the assumption that the expression under the square root
deviates only slightly from its mean, a Taylor expansion can be used.
The idea is the same as for the derivation leading to
\Cref{sec:theoreticalanalysis:eq:rcentroid4}
and results in
\begin{align}
    \mathrm{E}[||\langle\mathbf{r}^{\Pi}\rangle||] =
    \mathrm{E}\left[\sqrt{{\langle (\mathbf{x}^{\Pi})_2 \rangle}^2+
                          {||\langle \mathbf{h}^{\Pi} \rangle||}^2}\right]
    &=
    \sqrt{\mathrm{E}\left[{\langle (\mathbf{x}^{\Pi})_2 \rangle}^2+
                         {||\langle \mathbf{h}^{\Pi} \rangle||}^2\right]} +
                         \cdots\\
    &\approx
    \sqrt{\mathrm{E}\left[{\langle (\mathbf{x}^{\Pi})_2 \rangle}^2+
                          {||\langle \mathbf{h}^{\Pi} \rangle||}^2\right]}\\
    &=
    \sqrt{\mathrm{E}\left[{\langle (\mathbf{x}^{\Pi})_2 \rangle}^2\right]+
          \mathrm{E}\left[{||\langle \mathbf{h}^{\Pi} \rangle||}^2\right]}
    \label{sec:theoreticalanalysis:eq:exptaylorapprox}
\end{align}
where the last step follows from linearity of expectation. To proceed
further, $\mathrm{E}\left[{\langle (\mathbf{x}^{\Pi})_2 \rangle}^2\right]$
and $\mathrm{E}\left[{||\langle \mathbf{h}^{\Pi} \rangle||}^2\right]$ need to
be derived.

For $\mathrm{E}\left[{\langle (\mathbf{x}^{\Pi})_2 \rangle}^2\right]$,
\begin{equation}
    \label{sec:theoreticalanalysis:eq:xproj2centroid}
    \mathrm{E}\left[{\langle (\mathbf{x}^{\Pi})_2 \rangle}^2\right]
    =
    \mathrm{E}\left[
    \left(
    \frac{1}{\mu}\sum_{m=1}^{\mu}(\mathbf{x}^{\Pi}_{m;\lambda})_2
    \right)^2
    \right]
\end{equation}
can be written by expanding the notation for the centroid computation.
By the intercept theorem,
\begin{equation}
    \frac{||\mathbf{r}^{\Pi}_{m;\lambda}||}{(\mathbf{x}^{\Pi}_{m;\lambda})_2}
    =
    \frac{||\tilde{\mathbf{r}}_{m;\lambda}||}
         {r^{(g)}+\tilde{\sigma}_{m;\lambda}(\mathbf{z}_{m;\lambda})_2}
\end{equation}
follows
(see \Cref{sec:theoreticalanalysis:fig:rprogressinfeasgeometry}).
This implies
\begin{equation}
    \label{sec:theoreticalanalysis:eq:xproj2}
    (\mathbf{x}^{\Pi}_{m;\lambda})_2
    =
    \frac
    {||\mathbf{r}^{\Pi}_{m;\lambda}||
    (r^{(g)}+\tilde{\sigma}_{m;\lambda}(\mathbf{z}_{m;\lambda})_2)}
    {||\tilde{\mathbf{r}}_{m;\lambda}||}.
\end{equation}

$||\mathbf{r}^{\Pi}_{m;\lambda}||$ can be expressed in a different way
by using the projection equation
(see \Cref{sec:algorithm:subsec:projection:eq:rproj})
from
\Cref{sec:algorithm:subsec:projection}
yielding
\begin{equation}
    \label{sec:theoreticalanalysis:eq:rlenproj}
    ||\mathbf{r}^{\Pi}_{m;\lambda}||
    =
    \frac{(\mathbf{x}^{\Pi}_{m;\lambda})_1}{\sqrt{\xi}}.
\end{equation}

$||\tilde{\mathbf{r}}_{m;\lambda}||$ can be written in terms of the parental
individual and the mutation as
\begin{equation}
    \label{sec:theoreticalanalysis:eq:rlen}
    ||\tilde{\mathbf{r}}_{m;\lambda}||
    =
    ||r^{(g)}\mathbf{e}_2+\tilde{\sigma}_{m;\lambda}(\mathbf{z}_{m;\lambda})_{2..N}||.
\end{equation}

Insertion of
\Cref{sec:theoreticalanalysis:eq:rlenproj}
and
\Cref{sec:theoreticalanalysis:eq:rlen}
into
\Cref{sec:theoreticalanalysis:eq:xproj2}
results in
\begin{equation}
    \label{sec:theoreticalanalysis:eq:xproj2inserted}
    (\mathbf{x}^{\Pi}_{m;\lambda})_2
    =
    \frac
    {(\mathbf{x}^{\Pi}_{m;\lambda})_1
     (r^{(g)}+\tilde{\sigma}_{m;\lambda}(\mathbf{z}_{m;\lambda})_2)}
    {\sqrt{\xi}
     ||r^{(g)}\mathbf{e}_2+\tilde{\sigma}_{m;\lambda}(\mathbf{z}_{m;\lambda})_{2..N}||}.
\end{equation}

Inserting
\Cref{sec:theoreticalanalysis:eq:xproj2inserted}
into
\Cref{sec:theoreticalanalysis:eq:xproj2centroid}
yields
\begin{equation}
    \label{sec:theoreticalanalysis:eq:xproj2centroid2}
    \mathrm{E}\left[{\langle (\mathbf{x}^{\Pi})_2 \rangle}^2\right]
    =
    \mathrm{E}\left[
    \left(
    \frac{1}{\mu}\sum_{m=1}^{\mu}
    \frac
    {(\mathbf{x}^{\Pi}_{m;\lambda})_1
     (r^{(g)}+\tilde{\sigma}_{m;\lambda}(\mathbf{z}_{m;\lambda})_2)}
    {\sqrt{\xi}
     ||r^{(g)}\mathbf{e}_2+\tilde{\sigma}_{m;\lambda}(\mathbf{z}_{m;\lambda})_{2..N}||}
    \right)^2
    \right].
\end{equation}
In order to proceed further, sufficiently small $\tau$ is assumed
such that $\tilde{\sigma}_{m;\lambda} \simeq \sigma^{(g)}$ holds. Additionally,
$\sigma^{(g)}$ is assumed to be sufficiently small as well. These assumptions
allow writing
\begin{equation}
    \label{sec:theoreticalanalysis:eq:xproj2centroidhelper1}
    r^{(g)}+\tilde{\sigma}_{m;\lambda}(\mathbf{z}_{m;\lambda})_2
    \simeq r^{(g)}+\sigma^{(g)}(\mathbf{z}_{m;\lambda})_2
    \simeq r^{(g)}.
\end{equation}
For simplifying the expression
$||r^{(g)}\mathbf{e}_2+\tilde{\sigma}_{m;\lambda}(\mathbf{z}_{m;\lambda})_{2..N}||$,
note that $(\mathbf{z}_{m;\lambda})_{2..N}$ corresponds to
$(\mathbf{z})_{2..N}$ because selection is done in the $x_1$ direction.
This allows writing
\begin{equation}
    \label{sec:theoreticalanalysis:eq:xproj2centroidhelper2}
    ||r^{(g)}\mathbf{e}_2+\tilde{\sigma}_{m;\lambda}(\mathbf{z}_{m;\lambda})_{2..N}||
    \simeq
    ||r^{(g)}\mathbf{e}_2+\sigma^{(g)}(\mathbf{z})_{2..N}||
    =\tilde{r}
    \simeq r^{(g)}\sqrt{1+\frac{{{\sigma^{(g)}}^*}^2}{N}}
\end{equation}
where the last step follows from the normal approximation of
the distribution of a single offspring's distance from the cone
axis $\tilde{r}$
(see \Cref{sec:theoreticalanalysis:eq:approximatedr}).

Consideration of
\Cref{sec:theoreticalanalysis:eq:xproj2centroidhelper1}
and
\Cref{sec:theoreticalanalysis:eq:xproj2centroidhelper2}
for
\Cref{sec:theoreticalanalysis:eq:xproj2centroid2}
results in
\begin{align}
    \mathrm{E}\left[{\langle (\mathbf{x}^{\Pi})_2 \rangle}^2\right]
    &\simeq
    \mathrm{E}\left[
    \left(
    \frac{1}{\mu}\sum_{m=1}^{\mu}
    \frac
    {(\mathbf{x}^{\Pi}_{m;\lambda})_1 r^{(g)}}
    {\sqrt{\xi}r^{(g)}\sqrt{1+\frac{{{\sigma^{(g)}}^*}^2}{N}}}
    \right)^2
    \right]\\
    &=
    \mathrm{E}\left[
    \left(
    \frac
    {1}
    {\sqrt{\xi}\sqrt{1+\frac{{{\sigma^{(g)}}^*}^2}{N}}}
    \frac{1}{\mu}\sum_{m=1}^{\mu}(\mathbf{x}^{\Pi}_{m;\lambda})_1
    \right)^2
    \right]\\
    &=
    \mathrm{E}\left[
    \left(
    \frac
    {\langle (\mathbf{x}^{\Pi})_1 \rangle}
    {\sqrt{\xi}\sqrt{1+\frac{{{\sigma^{(g)}}^*}^2}{N}}}
    \right)^2
    \right].
    \label{sec:theoreticalanalysis:eq:xproj2centroid3}
\end{align}

For $\mathrm{E}\left[{||\langle \mathbf{h}^{\Pi} \rangle||}^2\right]$,
\begin{align}
    \mathrm{E}\left[{||\langle \mathbf{h}^{\Pi} \rangle||}^2\right]
    &=
    \mathrm{E}\left[
    \left(
    \frac{1}{\mu}\sum_{m=1}^{\mu}\mathbf{h}^{\Pi}_{m;\lambda}
    \right)^2
    \right]\\
    &=
    \frac{1}{\mu^2}
    \sum_{m=1}^{\mu}\sum_{p=1}^{\mu}
    \mathrm{E}\left[{\mathbf{h}_{m;\lambda}^{\Pi}}^T
                    \mathbf{h}_{p;\lambda}^{\Pi}\right]\\
    &=
    \frac{1}{\mu^2}
    \sum_{m=1}^{\mu}
    \mathrm{E}\left[{\mathbf{h}_{m;\lambda}^{\Pi}}^T
                    \mathbf{h}_{m;\lambda}^{\Pi}\right]
    +
    \frac{1}{\mu^2}
    \sum_{m=1}^{\mu}\sum_{\substack{p=1\\p \neq m}}^{\mu}
    \mathrm{E}\left[{\mathbf{h}_{m;\lambda}^{\Pi}}^T
                    \mathbf{h}_{p;\lambda}^{\Pi}\right]
    \label{sec:theoreticalanalysis:eq:hprojcentroid}
\end{align}
can be written by expanding the notation for the centroid computation.
The second summand in
\Cref{sec:theoreticalanalysis:eq:hprojcentroid}
vanishes. Note that the $\mathbf{h}_{m;\lambda}$ and
$\mathbf{h}_{p;\lambda}$ vectors for particular $m$ and $p$
are in the $3..N$ space.
Because selection is done in $x_1$ direction, they are statistically
independent. Additionally, every direction for those vectors
is equally probable because the $3..N$ is spherical
(assuming the distortion through the projection
is negligible). As a result, for fixed $m$ and $p$,
\begin{equation}
    \mathrm{E}\left[{\mathbf{h}_{m;\lambda}^{\Pi}}^T
                     \mathbf{h}_{p;\lambda}^{\Pi}\right]
    =
    \mathrm{E}\left[{\mathbf{h}_{m;\lambda}^{\Pi}}^T\right]
    \mathrm{E}\left[\mathbf{h}_{p;\lambda}^{\Pi}\right]
    =
    0
\end{equation}
holds. Consequently, these considerations result in
\begin{equation}
    \label{sec:theoreticalanalysis:eq:hprojcentroid1}
    \mathrm{E}\left[{||\langle \mathbf{h}^{\Pi} \rangle||}^2\right]
    =
    \frac{1}{\mu^2}
    \sum_{m=1}^{\mu}
    \mathrm{E}\left[{\mathbf{h}_{m;\lambda}^{\Pi}}^T
                     \mathbf{h}_{m;\lambda}^{\Pi}\right]
    =
    \frac{1}{\mu^2}
    \sum_{m=1}^{\mu}
    \mathrm{E}\left[||\mathbf{h}_{m;\lambda}^{\Pi}||^2\right].
\end{equation}
Thus, $\mathrm{E}\left[||\mathbf{h}_{m;\lambda}^{\Pi}||^2\right]$ needs to be
derived to proceed.
By the intercept theorem,
\begin{equation}
    \frac{||\mathbf{h}^{\Pi}_{m;\lambda}||}{||\mathbf{r}^{\Pi}_{m;\lambda}||}
    =
    \frac{||\tilde{\sigma}_{m;\lambda}(\mathbf{z}_{m;\lambda})_{3..N}||}
         {||\tilde{\mathbf{r}}_{m;\lambda}||}
\end{equation}
follows
(see \Cref{sec:theoreticalanalysis:fig:rprogressinfeasgeometry}).
With this and consideration of
\Cref{sec:theoreticalanalysis:eq:rlenproj},
\begin{equation}
    \label{sec:theoreticalanalysis:eq:hprojlensquared1}
    \mathrm{E}[||\mathbf{h}^{\Pi}_{m;\lambda}||^2]
    =
    \mathrm{E}\left[
    \left(
    \frac{(\mathbf{x}^{\Pi}_{m;\lambda})_1
          ||\tilde{\sigma}_{m;\lambda}(\mathbf{z}_{m;\lambda})_{3..N}||}
         {\sqrt{\xi}
          ||\tilde{\mathbf{r}}_{m;\lambda}||}
    \right)^2
    \right]
\end{equation}
can be written. In order to simplify
\Cref{sec:theoreticalanalysis:eq:hprojlensquared1},
it is assumed that the deviation of
$||\tilde{\sigma}_{m;\lambda}(\mathbf{z}_{m;\lambda})_{3..N}||^2$ and
$||\tilde{\mathbf{r}}_{m;\lambda}||^2$ from their mean values
is negligible. This allows replacing those random variable expressions
with their mean values yielding
\begin{align}
    \mathrm{E}[||\mathbf{h}^{\Pi}_{m;\lambda}||^2]
    &\approx
    \mathrm{E}\left[
    \frac{(\mathbf{x}^{\Pi}_{m;\lambda})_1^2}{\xi}
    \frac{\mathrm{E}[
          ||\tilde{\sigma}_{m;\lambda}(\mathbf{z}_{m;\lambda})_{3..N}||^2]}
         {\mathrm{E}[||\tilde{\mathbf{r}}_{m;\lambda}||^2]}
    \right]\\
    &=
    \mathrm{E}\left[
    \frac{(\mathbf{x}^{\Pi}_{m;\lambda})_1^2}{\xi}
    \frac{\mathrm{E}[
          ||\tilde{\sigma}_{m;\lambda}(\mathbf{z}_{m;\lambda})_{3..N}||^2]}
         {\mathrm{E}[||r^{(g)}\mathbf{e}_2+
               \tilde{\sigma}_{m;\lambda}(\mathbf{z}_{m;\lambda})_{2..N}||^2]}
    \right].
    \label{sec:theoreticalanalysis:eq:hprojlensquared2}
\end{align}
Now $\tau$ is assumed to be sufficiently small such that
$\tilde{\sigma}_{m;\lambda} \simeq \sigma^{(g)}$. Moreover,
because selection is done in the $x_1$ direction, the
$\mathbf{z}_{m;\lambda}$ components from $2$ to $N$
are all standard normally distributed. With these observations,
\begin{align}
    \mathrm{E}[||\mathbf{h}^{\Pi}_{m;\lambda}||^2]
    &\approx
    \mathrm{E}\left[
    \frac{(\mathbf{x}^{\Pi}_{m;\lambda})_1^2}{\xi}
    \frac{{\sigma^{(g)}}^2(N-2)}
         {\mathrm{E}[||r^{(g)}\mathbf{e}_2+
                       {\sigma}^{(g)}(\mathbf{z})_{2..N}||^2]}
    \right]
    \label{sec:theoreticalanalysis:eq:hprojlensquared3}\\
    &=
    \mathrm{E}\left[
    \frac{(\mathbf{x}^{\Pi}_{m;\lambda})_1^2}{\xi}
    \frac{{\sigma^{(g)}}^2(N-2)}
         {\mathrm{E}[\tilde{r}^2]}
    \right]
    \label{sec:theoreticalanalysis:eq:hprojlensquared4}\\
    &\simeq
    \mathrm{E}\left[
    \frac{(\mathbf{x}^{\Pi}_{m;\lambda})_1^2}{\xi}
    \frac{{\sigma^{(g)}}^2(N-2)}
         {{r^{(g)}}^2\left(1+\frac{{{\sigma^{(g)}}^*}^2}{N}\right)}
    \right]
    \label{sec:theoreticalanalysis:eq:hprojlensquared5}
\end{align}
can be derived where the last step follows from the normal approximation of
the distribution of a single offspring's distance from the cone
axis $\tilde{r}$
(see \Cref{sec:theoreticalanalysis:eq:approximatedr}).
Insertion of
\Cref{sec:theoreticalanalysis:eq:hprojlensquared5}
into
\Cref{sec:theoreticalanalysis:eq:hprojcentroid1}
yields
\begin{align}
    &\begin{multlined}
    \mathrm{E}\left[{||\langle \mathbf{h}^{\Pi} \rangle||}^2\right]
    \approx
    \frac{1}{\mu^2}
    \sum_{m=1}^{\mu}
    \mathrm{E}\left[
    \frac{(\mathbf{x}^{\Pi}_{m;\lambda})_1^2}{\xi}
    \frac{{\sigma^{(g)}}^2(N-2)}
         {{r^{(g)}}^2\left(1+\frac{{{\sigma^{(g)}}^*}^2}{N}\right)}
    \right]
    \end{multlined}\\
    &\begin{multlined}
    \phantom{\mathrm{E}\left[{||\langle \mathbf{h}^{\Pi} \rangle||}^2\right]}
    =
    \left(
    \frac{1}{\mu}
    \frac{1}{\xi}
    \frac{{\sigma^{(g)}}^2(N-2)}
         {{r^{(g)}}^2\left(1+\frac{{{\sigma^{(g)}}^*}^2}{N}\right)}
    \right)
    \left(
    \frac{1}{\mu}
    \sum_{m=1}^{\mu}
    \mathrm{E}\left[(\mathbf{x}^{\Pi}_{m;\lambda})_1^2\right]
    \right)
    \end{multlined}\\
    &\begin{multlined}
    \phantom{\mathrm{E}\left[{||\langle \mathbf{h}^{\Pi} \rangle||}^2\right]}
    =
    \left(
    \frac{1}{\mu}
    \frac{1}{\xi}
    \frac{{\sigma^{(g)}}^2(N-2)}
         {{r^{(g)}}^2\left(1+\frac{{{\sigma^{(g)}}^*}^2}{N}\right)}
    \right)
    \mathrm{E}[{\langle (\mathbf{x}^{\Pi})_1^2 \rangle}].
    \end{multlined}
    \label{sec:theoreticalanalysis:eq:hprojcentroidlensquared}
\end{align}

Insertion of
\Cref{sec:theoreticalanalysis:eq:xproj2centroid3}
and
\Cref{sec:theoreticalanalysis:eq:hprojcentroidlensquared}
into
\Cref{sec:theoreticalanalysis:eq:exptaylorapprox}
results in
\begin{align}
    \mathrm{E}[||\langle\mathbf{r}^{\Pi}\rangle||]
    &\approx
    \sqrt{
    \frac
    {\mathrm{E}\left[\langle (\mathbf{x}^{\Pi})_1 \rangle^2\right]}
    {\xi\left(1+\frac{{{\sigma^{(g)}}^*}^2}{N}\right)}
    +
    \left(
    \frac{1}{\mu}
    \frac{1}{\xi}
    \frac{{\sigma^{(g)}}^2(N-2)}
         {{r^{(g)}}^2\left(1+\frac{{{\sigma^{(g)}}^*}^2}{N}\right)}
    \right)
    \mathrm{E}[{\langle (\mathbf{x}^{\Pi})_1^2 \rangle}]
    }\\
    &\approx
    \sqrt{
    \frac
    {\mathrm{E}\left[\langle (\mathbf{x}^{\Pi})_1 \rangle\right]^2}
    {\xi\left(1+\frac{{{\sigma^{(g)}}^*}^2}{N}\right)}
    +
    \left(
    \frac{1}{\mu}
    \frac{1}{\xi}
    \frac{{\sigma^{(g)}}^2(N-2)}
         {{r^{(g)}}^2\left(1+\frac{{{\sigma^{(g)}}^*}^2}{N}\right)}
    \right)
    \mathrm{E}[{\langle (\mathbf{x}^{\Pi})_1^2 \rangle}]
    }
    \label{sec:theoreticalanalysis:eq:rprojcentroidlen}
\end{align}
where in the last step it has been assumed that the deviation
of $\langle (\mathbf{x}^{\Pi})_1 \rangle$ from its mean is negligible.
More formally, $\langle (\mathbf{x}^{\Pi})_1 \rangle$ can be written
as the sum of its expected value and a stochastic term
\begin{equation}
    \langle (\mathbf{x}^{\Pi})_1 \rangle =
    \mathrm{E}[\langle (\mathbf{x}^{\Pi})_1 \rangle] +
    \epsilon_{\langle (\mathbf{x}^{\Pi})_1 \rangle}.
\end{equation}
This stochastic term $\epsilon_{\langle (\mathbf{x}^{\Pi})_1 \rangle}$ is
assumed to be negligible which leads to
\begin{equation}
    \mathrm{E}[{\langle (\mathbf{x}^{\Pi})_1 \rangle}^2]
    \approx \mathrm{E}[{\langle (\mathbf{x}^{\Pi})_1 \rangle}]^2.
    \label{sec:theoreticalanalysis:eq:assumptionexpectedvalues1}
\end{equation}
Note that the $x_1$ value after projection corresponds to $q$. Further, due to
\Cref{sec:theoreticalanalysis:eq:varphix}
and
\Cref{sec:theoreticalanalysis:eq:varphixnormalized},
\begin{align}
    \mathrm{E}\left[\langle (\mathbf{x}^{\Pi})_1 \rangle\right]
    =
    \mathrm{E}\left[{\langle q \rangle}_{\text{infeas}}\right]
    =x^{(g)}\left(1-\frac{{\varphi_x^*}_{\text{infeas}}}{N}\right)
\end{align}
holds and an approximation for ${\varphi_x^*}_{\text{infeas}}$
has already been derived as
\Cref{sec:theoreticalanalysis:eq:varphixnormalizedinfeasible5}.
Because
\begin{equation}
  \mathrm{E}[{\langle (\mathbf{x}^{\Pi})_1^2 \rangle}]
  =
  \mathrm{E}[{\langle q^2 \rangle}_{\text{infeas}}],
  \label{sec:theoreticalanalysis:eq:assumptionexpectedvalues2}
\end{equation}
the approximation derived in
\Cref{sec:appendix:derivation_expectation_qrsquarecentroid}
(\Cref{sec:appendix:derivation_expectation_qrsquarecentroid:eq:squarecentroidinserted}\footnote{Note that for the infeasible case only the second summand of \Cref{sec:appendix:derivation_expectation_qrsquarecentroid:eq:squarecentroidinserted} is relevant.})
can be used. In order to simplify
\Cref{sec:theoreticalanalysis:eq:rprojcentroidlen}
further, an additional assumption is made. It is assumed that
\begin{equation}
    \mathrm{E}[{\langle (\mathbf{x}^{\Pi})_1^2 \rangle}]
    \approx
    \mathrm{E}[{\langle (\mathbf{x}^{\Pi})_1 \rangle}^2].
    \label{sec:theoreticalanalysis:eq:assumptionexpectedvalues3}
\end{equation}

In order to derive a condition for which the assumption in
\Cref{sec:theoreticalanalysis:eq:assumptionexpectedvalues3}
is justified,
\Cref{sec:theoreticalanalysis:eq:assumptionexpectedvalues1,%
sec:theoreticalanalysis:eq:assumptionexpectedvalues2,%
sec:theoreticalanalysis:eq:assumptionexpectedvalues3}
are investigated further. By
\Cref{sec:theoreticalanalysis:eq:assumptionexpectedvalues2,%
sec:appendix:derivation_expectation_qrsquarecentroid:eq:squarecentroidinserted},
one has
\begin{equation}
  \label{sec:theoreticalanalysis:eq:assumptionexpectedvalues4}
  \begin{multlined}
    \mathrm{E}[{\langle (\mathbf{x}^{\Pi})_1^2 \rangle}]
    =
    \mathrm{E}[{\langle q^2 \rangle}_{\text{infeas}}]
    =
    \bigg(\frac{({\sigma^{(g)}}^2+\sigma_r^2/\xi)}{(1+1/\xi)^2}
    \left[1+e_{\mu,\lambda}^{1,1}\right]\\
    -\frac{\sqrt{{\sigma^{(g)}}^2+\sigma_r^2/\xi}}{(1+1/\xi)^2}2(x^{(g)}+\bar{r}/\sqrt{\xi})
    c_{\mu/\mu,\lambda}
    +\frac{(x^{(g)}+\bar{r}/\sqrt{\xi})^2}{(1+1/\xi)^2}\bigg).
  \end{multlined}
\end{equation}
Insertion of
\Cref{sec:theoreticalanalysis:eq:qcentroidinfeasinsertedfinal}
into
\Cref{sec:theoreticalanalysis:eq:assumptionexpectedvalues1}
with subsequent calculation of the square and rewriting results in
\begin{align}
  \mathrm{E}[{\langle (\mathbf{x}^{\Pi})_1 \rangle}]^2
  &=\mathrm{E}[{\langle q \rangle}_{\text{infeas}}]^2\notag\\
  &\begin{multlined}
  \approx
  \left[\frac{\xi}{1+\xi}\left(x^{(g)} + \bar{r}/\sqrt{\xi}\right)
  - \frac{\xi}{1+\xi}\sqrt{{\sigma^{(g)}}^2+\sigma_r^2/\xi}
  c_{\mu/\mu,\lambda}\right]^2
  \end{multlined}\\
  &\begin{multlined}
  =\bigg(\frac{({\sigma^{(g)}}^2+\sigma_r^2/\xi)}{(1+1/\xi)^2}
    c_{\mu/\mu,\lambda}^2\\
    -\frac{\sqrt{{\sigma^{(g)}}^2+\sigma_r^2/\xi}}{(1+1/\xi)^2}2(x^{(g)}+\bar{r}/\sqrt{\xi})
    c_{\mu/\mu,\lambda}
    +\frac{(x^{(g)}+\bar{r}/\sqrt{\xi})^2}{(1+1/\xi)^2}\bigg).
  \end{multlined}
  \label{sec:theoreticalanalysis:eq:assumptionexpectedvalues5}
\end{align}
Comparison of
\Cref{sec:theoreticalanalysis:eq:assumptionexpectedvalues4}
with
\Cref{sec:theoreticalanalysis:eq:assumptionexpectedvalues5}
reveals that only the first term differs.
For $N \rightarrow \infty$, one has
$\sigma_r \simeq {r^{(g)}}\frac{{\sigma^{(g)}}^*}{N}$ by
\Cref{sec:theoreticalanalysis:eq:approximatedr}.
Together with $\sigma^{(g)}={r^{(g)}}\frac{{\sigma^{(g)}}^*}{N}$, which
follows from the definition of
${\sigma^{(g)}}^*=\frac{{\sigma^{(g)}}N}{r^{(g)}}$, this yields
\begin{equation}
  \frac{({\sigma^{(g)}}^2+\sigma_r^2/\xi)}{(1+1/\xi)^2}
  \simeq
  \frac{{\sigma^{(g)}}^2+
  \left({r^{(g)}}\frac{{\sigma^{(g)}}^*}{N}\right)^2/\xi}{(1+1/\xi)^2}
  =\frac{\left({r^{(g)}}\frac{{\sigma^{(g)}}^*}{N}\right)^2}{(1+1/\xi)}
  ={{r^{(g)}}^2}\frac{{{\sigma^{(g)}}^*}^2}{N^2(1+1/\xi)}.
\end{equation}
Hence, for $N^2(1+1/\xi) \gg {{\sigma^{(g)}}^*}^2$,
\Cref{sec:theoreticalanalysis:eq:assumptionexpectedvalues4}
and
\Cref{sec:theoreticalanalysis:eq:assumptionexpectedvalues5}
are approximately equal, which justifies
\Cref{sec:theoreticalanalysis:eq:assumptionexpectedvalues3}.

As it has already been assumed that
$\mathrm{E}[{\langle (\mathbf{x}^{\Pi})_1 \rangle}^2]
\approx
\mathrm{E}[{\langle (\mathbf{x}^{\Pi})_1 \rangle}]^2$,
\begin{align}
    \mathrm{E}[||\langle\mathbf{r}^{\Pi}\rangle||]
    &\approx
    \sqrt{
    \frac
    {\mathrm{E}\left[\langle (\mathbf{x}^{\Pi})_1 \rangle\right]^2}
    {\xi\left(1+\frac{{{\sigma^{(g)}}^*}^2}{N}\right)}
    +
    \left(
    \frac{1}{\mu}
    \frac{1}{\xi}
    \frac{{\sigma^{(g)}}^2(N-2)}
         {{r^{(g)}}^2\left(1+\frac{{{\sigma^{(g)}}^*}^2}{N}\right)}
    \right)
    \mathrm{E}[{\langle (\mathbf{x}^{\Pi})_1 \rangle}]^2
    }\\
    &=
    \sqrt{
    \frac
    {\mathrm{E}\left[\langle (\mathbf{x}^{\Pi})_1 \rangle\right]^2}
    {\xi\left(1+\frac{{{\sigma^{(g)}}^*}^2}{N}\right)}
    \left(
    1+
    \frac{1}{\mu}
    \frac{{\sigma^{(g)}}^2(N-2)}
         {{r^{(g)}}^2}
    \right)}\\
    &=
    \frac
    {\mathrm{E}\left[\langle (\mathbf{x}^{\Pi})_1 \rangle\right]}
    {\sqrt{\xi}\sqrt{1+\frac{{{\sigma^{(g)}}^*}^2}{N}}}
    \sqrt{
    1+
    \frac{1}{\mu}
    \underbrace{\frac{{{\sigma^{(g)}}^*}^2(N-2)}
                 {N^2}}_{\simeq \frac{{{\sigma^{(g)}}^*}^2}
                 {N}\text{ for } N \rightarrow \infty}
    }\\
    &\simeq
    \frac
    {\mathrm{E}\left[\langle (\mathbf{x}^{\Pi})_1 \rangle\right]}
    {\sqrt{\xi}\sqrt{1+\frac{{{\sigma^{(g)}}^*}^2}{N}}}
    \sqrt{
    1+
    \frac{{{\sigma^{(g)}}^*}^2}
         {\mu N}
    }\\
    &=
    \frac
    {\mathrm{E}\left[{\langle q \rangle}_{\text{infeas}}\right]}
    {\sqrt{\xi}}
    \sqrt{\frac{1+\frac{{{\sigma^{(g)}}^*}^2}{\mu N}}
               {1+\frac{{{\sigma^{(g)}}^*}^2}{N}}}
    \label{sec:theoreticalanalysis:eq:progressinfeashelper}
\end{align}
can now be written under the assumptions stated before.
Insertion of
\Cref{sec:theoreticalanalysis:eq:progressinfeashelper}
into
\Cref{sec:theoreticalanalysis:eq:progressinfeasdef}
yields
\begin{align}
    {\varphi_r}_{\text{infeas}}
    &\approx
    r^{(g)} -
    \frac
    {\mathrm{E}\left[{\langle q \rangle}_{\text{infeas}}\right]}
    {\sqrt{\xi}}
    \sqrt{\frac{1+\frac{{{\sigma^{(g)}}^*}^2}{\mu N}}
               {1+\frac{{{\sigma^{(g)}}^*}^2}{N}}}\\
    &=
    r^{(g)} -
    \frac
    {x^{(g)}-{\varphi_x}_{\text{infeas}}}
    {\sqrt{\xi}}
    \sqrt{\frac{1+\frac{{{\sigma^{(g)}}^*}^2}{\mu N}}
               {1+\frac{{{\sigma^{(g)}}^*}^2}{N}}}.
\end{align}
Normalization of $\varphi_r$ and $\varphi_x$ yields
\begin{align}
    {\varphi_r^*}_{\text{infeas}}
    &\approx
    N\left(
    1 -
    \frac
    {x^{(g)}}
    {\sqrt{\xi}r^{(g)}}
    \left(1-\frac{{\varphi_x^*}_{\text{infeas}}}{N}\right)
    \sqrt{\frac{1+\frac{{{\sigma^{(g)}}^*}^2}{\mu N}}
               {1+\frac{{{\sigma^{(g)}}^*}^2}{N}}}
    \right)
    \label{sec:theoreticalanalysis:eq:varphirnormalizedinfeas}
\end{align}
where an approximation for ${\varphi_x^*}_{\text{infeas}}$ has already been
derived as
\Cref{sec:theoreticalanalysis:eq:varphixnormalizedinfeasible5}.

\subsection{The Approximate \texorpdfstring{$r$}{\$r\$}
  Progress Rate - Combination Using the
  Single Offspring Feasibility Probability}
Similarly to the $x$ progress rate,
the feasible and infeasible cases can be combined using
the single offspring feasibility probability yielding
\begin{align}
  &\begin{multlined}
  \varphi_r^*
  \approx
  P_{\text{feas}}(x^{(g)}, r^{(g)}, \sigma^{(g)})
  {\varphi_r}_{\text{feas}}^*
  + [1 - P_{\text{feas}}(x^{(g)}, r^{(g)}, \sigma^{(g)})]
  {\varphi_r}_{\text{infeas}}^*
  \end{multlined}\\
  &\begin{multlined}
  \phantom{\varphi_r^*}
  \approx
  P_{\text{feas}}(x^{(g)}, r^{(g)}, \sigma^{(g)})
  N\left(1-\sqrt{1 + \frac{{{\sigma^{(g)}}^*}^2}{\mu N}}\right)\\
  + [1 - P_{\text{feas}}(x^{(g)}, r^{(g)}, \sigma^{(g)})]
  N\left(
  1 -
  \frac
  {x^{(g)}}
  {\sqrt{\xi}r^{(g)}}
  \left(1-\frac{{\varphi_x^*}_{\text{infeas}}}{N}\right)
  \sqrt{\frac{1+\frac{{{\sigma^{(g)}}^*}^2}{\mu N}}
             {1+\frac{{{\sigma^{(g)}}^*}^2}{N}}}
  \right)
  \end{multlined}
  \label{sec:theoreticalanalysis:eq:varphirnormalizedcombined}
\end{align}

\newpage

\section{Derivation of the SAR}
\label{appendix:subsec:sar}
From the definition of the SAR
(\Cref{sec:theoreticalanalysis:eq:psi})
and the pseudo-code of the
ES (\Cref{sec:algorithm:alg:es}, \Cref{sec:algorithm:alg:es:replacesigma})
it follows that
\begin{align}
  \psi(x^{(g)}, r^{(g)}, \sigma^{(g)})
  &= \mathrm{E}\left[\frac{\sigma^{(g + 1)} - \sigma^{(g)}}{\sigma^{(g)}}
    \,\bigg|\,x^{(g)}, r^{(g)}, \sigma^{(g)}\right]\\
  &= \mathrm{E}\left[\frac{\left(\frac{1}{\mu}\sum_{m=1}^{\mu}
     \tilde{\sigma}_{m;\lambda}\right) -
     \sigma^{(g)}}{\sigma^{(g)}}
    \,\bigg|\,x^{(g)}, r^{(g)}, \sigma^{(g)}\right]\\
  &= \mathrm{E}\left[\frac{1}{\mu}\sum_{m=1}^{\mu}\frac{
     \tilde{\sigma}_{m;\lambda} -
     \sigma^{(g)}}{\sigma^{(g)}}
    \,\bigg|\,x^{(g)}, r^{(g)}, \sigma^{(g)}\right]\\
  &= \frac{1}{\mu}\sum_{m=1}^{\mu}\mathrm{E}\left[\frac{
     \tilde{\sigma}_{m;\lambda} -
     \sigma^{(g)}}{\sigma^{(g)}}
    \,\bigg|\,x^{(g)}, r^{(g)}, \sigma^{(g)}\right]\\
  &= \frac{1}{\mu}\sum_{m=1}^{\mu}\int_{\sigma=0}^{\sigma=\infty}
  \left(\frac{\sigma - \sigma^{(g)}}{\sigma^{(g)}}\right)
  p_{\tilde{\sigma}_{m;\lambda}}(\sigma)
  \,\mathrm{d}\sigma
  \label{chapter:analysis_repairbyprojection_multirecombinative:sec:theoreticalanalysis:eq:psimoredetail1}
\end{align}
where $p_{\tilde{\sigma}_{m;\lambda}}(\sigma) :=
p_{\tilde{\sigma}_{m;\lambda}}(\sigma\,|\,x^{(g)}, r^{(g)}, \sigma^{(g)})$
denotes the probability density function of the $m$-th best offspring's
mutation strength. Note that $\tilde{\sigma}_{m;\lambda}$ is not obtained
by direct selection. It is the $\sigma$ value of the individual with
the $m$-th best objective function value among the $\lambda$ offspring.
Its probability density function can be derived as follows.
A random sample $\sigma$ from the conditional distribution density
\begin{equation}
  \label{chapter:analysis_repairbyprojection_multirecombinative:sec:theoreticalanalysis:eq:lognormaldensityforpsi}
  p_\sigma(\sigma \,|\, \sigma^{(g)}) =
  \frac{1}{\sqrt{2\pi}\tau}\frac{1}{\sigma}
  \exp\left[-\frac{1}{2}\left(
    \frac{\ln(\sigma / \sigma^{(g)})}{\tau}\right)^2\right]
\end{equation}
is considered.
In order for this $\sigma$ to be selected,
the corresponding offspring's $q$ value has to be the
$m$-th best value among all the $\lambda$ offspring.
This is the case if $(\lambda-m)$ offspring have a greater
projected value, and $(m-1)$ offspring have a smaller projected value.
The projected $q$ density for a given mutation strength is given by
$p_Q(q \,|\, x^{(g)}, r^{(g)}, \sigma)$.
As there are
$\lambda\binom{\lambda-1}{m-1} = \frac{\lambda!}{(\lambda-m)!(m-1)!}$
different possibilities for
one offspring being the $m$-th best among $\lambda$ descendants,
it must be multiplied by this factor. Integrating over all possible
$q$ values yields
\begin{equation}
  \label{chapter:analysis_repairbyprojection_multirecombinative:sec:theoreticalanalysis:eq:psigma1lambda2}
  \begin{multlined}
  p_{\tilde{\sigma}_{m;\lambda}}(\sigma) =
  p_\sigma(\sigma \,|\, \sigma^{(g)})
    \frac{\lambda!}{(\lambda-m)!(m-1)!}
    \int_{q=0}^{q=\infty}
    p_Q(q \,|\, x^{(g)}, r^{(g)}, \sigma)\\
    \times[1-P_Q(q)]^{\lambda-m}[P_Q(q)]^{m-1}
    \,\mathrm{d}q.
  \end{multlined}
\end{equation}
This result can now be inserted into
\Cref{chapter:analysis_repairbyprojection_multirecombinative:sec:theoreticalanalysis:eq:psimoredetail1}
resulting in
\begin{align}
  &\begin{multlined}
    \psi(x^{(g)}, r^{(g)}, \sigma^{(g)})
    =\frac{1}{\mu}\sum_{m=1}^{\mu}\int_{\sigma=0}^{\sigma=\infty}
      \left(\frac{\sigma - \sigma^{(g)}}{\sigma^{(g)}}\right)
      p_\sigma(\sigma \,|\, \sigma^{(g)})\\
      \times\frac{\lambda!}{(\lambda-m)!(m-1)!}
      \int_{q=0}^{q=\infty}
      p_Q(q \,|\, x^{(g)}, r^{(g)}, \sigma)
      [1-P_Q(q)]^{\lambda-m}[P_Q(q)]^{m-1}
      \,\mathrm{d}q
      \,\mathrm{d}\sigma
  \end{multlined}
  \label{chapter:analysis_repairbyprojection_multirecombinative:sec:theoreticalanalysis:eq:psimoredetail2}\\
  &\begin{multlined}
    \phantom{\psi(x^{(g)}, r^{(g)}, \sigma^{(g)}) }
    =\int_{\sigma=0}^{\sigma=\infty}
      \left(\frac{\sigma - \sigma^{(g)}}{\sigma^{(g)}}\right)
      p_\sigma(\sigma \,|\, \sigma^{(g)})\\
      \times\frac{\lambda!}{\mu}
      \int_{q=0}^{q=\infty}
      p_Q(q \,|\, x^{(g)}, r^{(g)}, \sigma)
      \sum_{m=1}^{\mu}
      \frac{[1-P_Q(q)]^{\lambda-m}[P_Q(q)]^{m-1}}{(\lambda-m)!(m-1)!}
      \,\mathrm{d}q
      \,\mathrm{d}\sigma.
  \end{multlined}
  \label{chapter:analysis_repairbyprojection_multirecombinative:sec:theoreticalanalysis:eq:psimoredetail3}
\end{align}
Use of
\Cref{sec:theoreticalanalysis:eq:Beyer2001_5.14}
with subsequent expression of the fraction using the binomial coefficient,
one can rewrite
\Cref{chapter:analysis_repairbyprojection_multirecombinative:sec:theoreticalanalysis:eq:psimoredetail3}
to get
\begin{align}
  &\begin{multlined}
    \psi(x^{(g)}, r^{(g)}, \sigma^{(g)})
    =\int_{\sigma=0}^{\sigma=\infty}
      \left(\frac{\sigma - \sigma^{(g)}}{\sigma^{(g)}}\right)
      p_\sigma(\sigma \,|\, \sigma^{(g)})\\
      \times(\lambda-\mu)\binom{\lambda}{\mu}
      \int_{q=0}^{q=\infty}
      p_Q(q \,|\, x^{(g)}, r^{(g)}, \sigma)
      \int_{z=0}^{z=1-P_Q(q)}
      z^{\lambda-\mu-1}(1-z)^{\mu-1}
      \,\mathrm{d}z
      \,\mathrm{d}q
      \,\mathrm{d}\sigma.
  \end{multlined}
  \label{chapter:analysis_repairbyprojection_multirecombinative:sec:theoreticalanalysis:eq:psimoredetail4}
\end{align}
To proceed further, $z=1-P_Q(y)$ is substituted in the inner integral.
This implies $y=P_Q^{-1}(1-z)$
and $\mathrm{d}z=-p_Q(y)$. The upper and lower bounds for the substituted
integral follow as $y_u=P_Q^{-1}(1-1+P_Q(q))=q$ and
$y_l=P_Q^{-1}(1-0)=\infty$, respectively.
Therefore, one obtains
\begin{align}
  &\begin{multlined}
    \psi(x^{(g)}, r^{(g)}, \sigma^{(g)})
    =\int_{\sigma=0}^{\sigma=\infty}
      \left(\frac{\sigma - \sigma^{(g)}}{\sigma^{(g)}}\right)
      p_\sigma(\sigma \,|\, \sigma^{(g)})
      (\lambda-\mu)\binom{\lambda}{\mu}\\
      \times\int_{q=0}^{q=\infty}
      p_Q(q \,|\, x^{(g)}, r^{(g)}, \sigma)\\
      \times\int_{y=\infty}^{y=q}
      [1-P_Q(y)]^{\lambda-\mu-1}[P_Q(y)]^{\mu-1}
      (-p_Q(y))\,\mathrm{d}y
      \,\mathrm{d}q
      \,\mathrm{d}\sigma
  \end{multlined}
  \label{chapter:analysis_repairbyprojection_multirecombinative:sec:theoreticalanalysis:eq:psimoredetail5}\\
  &\begin{multlined}
    \psi(x^{(g)}, r^{(g)}, \sigma^{(g)})
    =\int_{\sigma=0}^{\sigma=\infty}
      \left(\frac{\sigma - \sigma^{(g)}}{\sigma^{(g)}}\right)
      p_\sigma(\sigma \,|\, \sigma^{(g)})
      (\lambda-\mu)\binom{\lambda}{\mu}\\
      \times\int_{q=0}^{q=\infty}
      p_Q(q \,|\, x^{(g)}, r^{(g)}, \sigma)\\
      \times\int_{y=q}^{y=\infty}
      p_Q(y)
      [1-P_Q(y)]^{\lambda-\mu-1}[P_Q(y)]^{\mu-1}
      \,\mathrm{d}y
      \,\mathrm{d}q
      \,\mathrm{d}\sigma.
  \end{multlined}
  \label{chapter:analysis_repairbyprojection_multirecombinative:sec:theoreticalanalysis:eq:psimoredetail6}
\end{align}
Changing the order of the two inner-most integrals one finally gets
\begin{align}
  &\begin{multlined}
    \psi(x^{(g)}, r^{(g)}, \sigma^{(g)})
    =\int_{\sigma=0}^{\sigma=\infty}
      \left(\frac{\sigma - \sigma^{(g)}}{\sigma^{(g)}}\right)
      p_\sigma(\sigma \,|\, \sigma^{(g)})
      (\lambda-\mu)\binom{\lambda}{\mu}\\
      \times\int_{y=0}^{y=\infty}
      p_Q(y)
      [1-P_Q(y)]^{\lambda-\mu-1}[P_Q(y)]^{\mu-1}\\
      \times\int_{q=0}^{q=y}
      p_Q(q \,|\, x^{(g)}, r^{(g)}, \sigma)
      \,\mathrm{d}q
      \,\mathrm{d}y
      \,\mathrm{d}\sigma
  \end{multlined}
  \label{chapter:analysis_repairbyprojection_multirecombinative:sec:theoreticalanalysis:eq:psimoredetail7}\\
  &\begin{multlined}
    \psi(x^{(g)}, r^{(g)}, \sigma^{(g)})
    =\int_{\sigma=0}^{\sigma=\infty}
      \left(\frac{\sigma - \sigma^{(g)}}{\sigma^{(g)}}\right)
      p_\sigma(\sigma \,|\, \sigma^{(g)})\\
      \times(\lambda-\mu)\binom{\lambda}{\mu}
      \int_{y=0}^{y=\infty}
      p_Q(y)
      [1-P_Q(y)]^{\lambda-\mu-1}[P_Q(y)]^{\mu-1}\\
      \times P_Q(y \,|\, x^{(g)}, r^{(g)}, \sigma)
      \,\mathrm{d}y
      \,\mathrm{d}\sigma.
  \end{multlined}
  \label{chapter:analysis_repairbyprojection_multirecombinative:sec:theoreticalanalysis:eq:psimoredetail8}
\end{align}
Due to difficulties in analytically solving
\Cref{chapter:analysis_repairbyprojection_multirecombinative:sec:theoreticalanalysis:eq:psimoredetail8},
an approximate solution is derived.
To this end, the approach used in~\citeapp[Sec. 7.3.2.4]{Beyer2001APP}
is followed.
For this,
\Cref{chapter:analysis_repairbyprojection_multirecombinative:sec:theoreticalanalysis:eq:psimoredetail8}
is written as
\begin{align}
  \psi(x^{(g)}, r^{(g)}, \sigma^{(g)})
  &= \int_{\sigma=0}^{\sigma=\infty}
  f(\sigma)
  p_\sigma(\sigma \,|\, \sigma^{(g)})
  \,\mathrm{d}\sigma\\
  &= \int_{\sigma=0}^{\sigma=\infty}
  f(\sigma)
  \frac{1}{\sqrt{2\pi}\tau}\frac{1}{\sigma}
  \exp\left[-\frac{1}{2}\left(
    \frac{\ln(\sigma / \sigma^{(g)})}{\tau}\right)^2\right]
  \,\mathrm{d}\sigma
  \label{chapter:analysis_repairbyprojection_multirecombinative:sec:theoreticalanalysis:eq:psimoredetail9}
\end{align}
with
\begin{equation}
  \begin{multlined}
    f(\sigma) =
    \left(\frac{\sigma - \sigma^{(g)}}{\sigma^{(g)}}\right)
    (\lambda-\mu)\binom{\lambda}{\mu}
    \int_{y=0}^{y=\infty}
    p_Q(y)
    [1-P_Q(y)]^{\lambda-\mu-1}[P_Q(y)]^{\mu-1}\\
    P_Q(y \,|\, x^{(g)}, r^{(g)}, \sigma)
    \,\mathrm{d}y.
  \end{multlined}
\end{equation}
Now, $\tau$ is assumed to be small. Therefore,
\Cref{chapter:analysis_repairbyprojection_multirecombinative:sec:theoreticalanalysis:eq:psimoredetail9}
can be expanded into a Taylor series at $\tau = 0$. Further, as
$\tau$ is small, the probability mass of the log-normally distributed
values is concentrated around $\sigma^{(g)}$.
Therefore, $f(\sigma)$ can be expanded into
a Taylor series at $\sigma=\sigma^{(g)}$. After further calculation
(it is referred to~\citeapp[Sec. 7.3.2.4]{Beyer2001APP} for all the details)
one obtains
\begin{equation}
  \label{chapter:analysis_repairbyprojection_multirecombinative:sec:theoreticalanalysis:eq:psiapproxwithf}
  \psi(x^{(g)}, r^{(g)}, \sigma^{(g)}) =
  f(\sigma^{(g)})
  + \frac{\tau^2}{2}\sigma^{(g)}
  \frac{\partial f}{\partial \sigma}\Biggr|_{\sigma=\sigma^{(g)}}
  + \frac{\tau^2}{2}{\sigma^{(g)}}^2
  \frac{\partial^2 f}{\partial \sigma^2}\Biggr|_{\sigma=\sigma^{(g)}}
  + O(\tau^4).
\end{equation}
To proceed further, the expressions
$f(\sigma^{(g)})$,
$\frac{\partial f}{\partial \sigma}\Bigr|_{\sigma=\sigma^{(g)}}$,
and
$\frac{\partial^2 f}{\partial \sigma^2}\Bigr|_{\sigma=\sigma^{(g)}}$
need to be evaluated. Because
$f(\sigma^{(g)}) =
\frac{\sigma^{(g)} - \sigma^{(g)}}{\sigma^{(g)}} \times \cdots = 0$,
\begin{equation}
  \label{chapter:analysis_repairbyprojection_multirecombinative:sec:theoreticalanalysis:eq:fsigmap}
  f(\sigma^{(g)}) = 0
\end{equation}
immediately follows. For
$\frac{\partial f}{\partial \sigma}\Bigr|_{\sigma=\sigma^{(g)}}$,
one derives with the product rule
\begin{equation}
  \label{chapter:analysis_repairbyprojection_multirecombinative:sec:theoreticalanalysis:eq:ffirstderivative}
  \begin{multlined}
    \frac{\partial f}{\partial \sigma}
    =\frac{1}{\sigma^{(g)}}
    (\lambda-\mu)\binom{\lambda}{\mu}
    \int_{y=0}^{y=\infty}
    p_Q(y)
    [1-P_Q(y)]^{\lambda-\mu-1}[P_Q(y)]^{\mu-1}
    P_Q(y \,|\, x^{(g)}, r^{(g)}, \sigma)
    \,\mathrm{d}y\\
    +\left(\frac{\sigma-\sigma^{(g)}}{\sigma^{(g)}}\right)
    (\lambda-\mu)\binom{\lambda}{\mu}
    \int_{y=0}^{y=\infty}
    p_Q(y)
    [1-P_Q(y)]^{\lambda-\mu-1}[P_Q(y)]^{\mu-1}\\
    \times
    \frac{\partial}{\partial \sigma}P_Q(y \,|\, x^{(g)}, r^{(g)}, \sigma)
    \,\mathrm{d}y.
  \end{multlined}
\end{equation}
Therefore, it follows that
\begin{align}
  &\begin{multlined}
    \frac{\partial f}{\partial \sigma}\Bigr|_{\sigma=\sigma^{(g)}}
    =\frac{1}{\sigma^{(g)}}
    (\lambda-\mu)\binom{\lambda}{\mu}
    \int_{y=0}^{y=\infty}
    p_Q(y)
    [1-P_Q(y)]^{\lambda-\mu-1}[P_Q(y)]^{\mu-1}\\
    \times \underbrace{P_Q(y \,|\, x^{(g)}, r^{(g)}, \sigma^{(g)})}_{=P_Q(y)}
    \,\mathrm{d}y\\
    \hspace{2cm}+\underbrace{
    \left(\frac{\sigma^{(g)}-\sigma^{(g)}}{\sigma^{(g)}}\right)}_{=0}
    (\lambda-\mu)\binom{\lambda}{\mu}
    \int_{y=0}^{y=\infty}
    p_Q(y)
    [1-P_Q(y)]^{\lambda-\mu-1}[P_Q(y)]^{\mu-1}\\
    \times \frac{\partial}{\partial \sigma}P_Q(y \,|\, x^{(g)}, r^{(g)}, \sigma)
    \,\mathrm{d}y\Bigr|_{\sigma=\sigma^{(g)}}
  \end{multlined}\\
  &\begin{multlined}
    \phantom{\frac{\partial f}{\partial \sigma}\Bigr|_{\sigma=\sigma^{(g)}} }
    =\frac{1}{\sigma^{(g)}}
    \frac{\lambda!}{(\lambda-(\mu+1))!((\mu+1)-1)!}\\
    \hspace{2cm}\times\int_{y=0}^{y=\infty}
    p_Q(y)
    [1-P_Q(y)]^{\lambda-(\mu+1)}[P_Q(y)]^{(\mu+1)-1}
    \,\mathrm{d}y
  \end{multlined}\\
  &\begin{multlined}
    \phantom{\frac{\partial f}{\partial \sigma}\Bigr|_{\sigma=\sigma^{(g)}} }
    =\frac{1}{\sigma^{(g)}}
    \underbrace{\int_{y=0}^{y=\infty}
    p_{q_{(\mu+1);\lambda}}(y)\,\mathrm{d}y}_{=1}
  \end{multlined}\\
  &\begin{multlined}
    \phantom{\frac{\partial f}{\partial \sigma}\Bigr|_{\sigma=\sigma^{(g)}} }
    =\frac{1}{\sigma^{(g)}}.
  \end{multlined}
\end{align}
Computing the derivative with respect to $\sigma$ of
\Cref{chapter:analysis_repairbyprojection_multirecombinative:sec:theoreticalanalysis:eq:ffirstderivative}
using the product rule for the second summand results in
\begin{align}
  &\begin{multlined}
    \frac{\partial^2 f}{\partial \sigma^2}
    =\frac{1}{\sigma^{(g)}}(\lambda-\mu)\binom{\lambda}{\mu}
    \int_{y=0}^{y=\infty}
    p_Q(y)
    [1-P_Q(y)]^{\lambda-\mu-1}[P_Q(y)]^{\mu-1}\\
    \times
    \frac{\partial}{\partial \sigma}P_Q(y \,|\, x^{(g)}, r^{(g)}, \sigma)
    \,\mathrm{d}y\\
    +\frac{1}{\sigma^{(g)}}(\lambda-\mu)\binom{\lambda}{\mu}
    \int_{y=0}^{y=\infty}
    p_Q(y)
    [1-P_Q(y)]^{\lambda-\mu-1}[P_Q(y)]^{\mu-1}\\
    \times
    \frac{\partial}{\partial \sigma}P_Q(y \,|\, x^{(g)}, r^{(g)}, \sigma)
    \,\mathrm{d}y\\
    +\left(\frac{\sigma-\sigma^{(g)}}{\sigma^{(g)}}\right)
    (\lambda-\mu)\binom{\lambda}{\mu}
    \int_{y=0}^{y=\infty}
    p_Q(y)
    [1-P_Q(y)]^{\lambda-\mu-1}[P_Q(y)]^{\mu-1}\\
    \times
    \frac{\partial^2}{\partial \sigma^2}P_Q(y \,|\, x^{(g)}, r^{(g)}, \sigma)
    \,\mathrm{d}y.
  \end{multlined}
\end{align}
This implies
\begin{align}
  &\begin{multlined}
    \frac{\partial^2 f}{\partial \sigma^2}\Bigr|_{\sigma=\sigma^{(g)}}
    =\frac{2}{\sigma^{(g)}}
    (\lambda-\mu)\binom{\lambda}{\mu}
    \int_{y=0}^{y=\infty}
    p_Q(y)
    [1-P_Q(y)]^{\lambda-\mu-1}[P_Q(y)]^{\mu-1}\\
    \times
    \frac{\partial}{\partial \sigma}P_Q(y \,|\, x^{(g)}, r^{(g)}, \sigma)
    \,\mathrm{d}y
    \Bigr|_{\sigma=\sigma^{(g)}}.
  \end{multlined}
  \label{chapter:analysis_repairbyprojection_multirecombinative:sec:theoreticalanalysis:eq:fsecondderivative}
\end{align}
The $P_Q(q)$ and $p_Q(q)$ approximations
\Cref{sec:theoreticalanalysis:eq:approximatedPQforprogressfeasible,%
sec:theoreticalanalysis:eq:approximatedPQforprogress,%
sec:theoreticalanalysis:eq:approximatedpQforprogressfeasible,%
sec:theoreticalanalysis:eq:approximatedpQforprogress}
are used to proceed further.
Two cases (being feasible with overwhelming probability and
being infeasible with overwhelming probability) have been distinguished
in the derivation of the $P_Q$ and $p_Q$ approximations.
Therefore, those two cases are treated separately for
\Cref{chapter:analysis_repairbyprojection_multirecombinative:sec:theoreticalanalysis:eq:fsecondderivative}.

\subsection{The Approximate \glsfmttext{SAR}
  in the case that the
  probability of feasible offspring tends to \texorpdfstring{$1$}{\$1\$}}
In order to treat
\Cref{chapter:analysis_repairbyprojection_multirecombinative:sec:theoreticalanalysis:eq:fsecondderivative}
further for the feasible case,
$\frac{\partial}{\partial \sigma}{P_Q}_{\text{feas}}
(y \,|\, x^{(g)}, r^{(g)}, \sigma)$
needs to be derived.
Insertion of
\Cref{sec:theoreticalanalysis:eq:approximatedPQforprogressfeasible}
into
\Cref{chapter:analysis_repairbyprojection_multirecombinative:sec:theoreticalanalysis:eq:fsecondderivative}
and subsequent application of the chain rule results in
\begin{align}
  \frac{\partial}{\partial \sigma}
  {P_Q}_{\text{feas}}
  (y \,|\, x^{(g)}, r^{(g)}, \sigma)
  &=\frac{\partial}{\partial \sigma}
  \Phi\left(\frac{y-x^{(g)}}
  {\sigma}\right)
  \label{chapter:analysis_repairbyprojection_multirecombinative:sec:theoreticalanalysis:eq:PQfeasderivative1}\\
  &=\phi\left(\frac{y-x^{(g)}}
  {\sigma}\right)
  (-1)\left(\frac{y-x^{(g)}}
  {\sigma^2}\right)
  \label{chapter:analysis_repairbyprojection_multirecombinative:sec:theoreticalanalysis:eq:PQfeasderivative2}\\
  &=-\frac{1}{\sqrt{2\pi}}
  e^{-\frac{1}{2}\left(\frac{y-x^{(g)}}
  {\sigma}\right)^2}
  \left(\frac{y-x^{(g)}}
  {\sigma^2}\right).
  \label{chapter:analysis_repairbyprojection_multirecombinative:sec:theoreticalanalysis:eq:PQfeasderivative3}
\end{align}
Insertion of
\Cref{sec:theoreticalanalysis:eq:approximatedPQforprogressfeasible,%
sec:theoreticalanalysis:eq:approximatedpQforprogressfeasible,%
chapter:analysis_repairbyprojection_multirecombinative:sec:theoreticalanalysis:eq:PQfeasderivative3}
into
\Cref{chapter:analysis_repairbyprojection_multirecombinative:sec:theoreticalanalysis:eq:fsecondderivative}
yields
\begin{align}
  &\begin{multlined}
    \frac{\partial^2 f_{\text{feas}}}{\partial \sigma^2}
    \Bigr|_{\sigma=\sigma^{(g)}}
    =\frac{2}{\sigma^{(g)}}
    (\lambda-\mu)\binom{\lambda}{\mu}
    \int_{y=\bar{r}\sqrt{\xi}}^{y=\infty}
    \frac{1}{\sigma^{(g)}}
    \left(-\frac{y-x^{(g)}}
             {\sigma^{(g)}}\right)
    \frac{1}{{(\sqrt{2\pi})}^2{\sigma^{(g)}}}
    e^{-\frac{2}{2}\left(\frac{y-x^{(g)}}
      {\sigma^{(g)}}\right)^2}\\
    \times\left[1-
    \Phi\left(\frac{y-x^{(g)}}
             {\sigma^{(g)}}\right)
    \right]^{\lambda-\mu-1}\left[
    \Phi\left(\frac{y-x^{(g)}}
             {\sigma^{(g)}}\right)
    \right]^{\mu-1}\,\mathrm{d}y.
  \end{multlined}
\end{align}
For solving this integral, $-t:=\frac{y-x^{(g)}}{\sigma^{(g)}}$ is
substituted. It implies $y=-\sigma^{(g)}t+x^{(g)}$
and $\mathrm{d}y=-\sigma^{(g)}\,\mathrm{d}t$.
The lower bound follows with $\sigma$-normalization, assuming
$N \rightarrow \infty$, $N \gg {\sigma^*}^{(g)}$, and
knowing that $x^{(g)} \ge \bar{r}\sqrt{\xi}$ holds for
the feasible case. It reads
\begin{equation*}
  t_l = -\frac{\bar{r}\sqrt{\xi}-x^{(g)}}{\sigma^{(g)}}
  =-\frac{N}{{\sigma^*}^{(g)}}\frac{\bar{r}\sqrt{\xi}-x^{(g)}}{r^{(g)}}
  \simeq \infty.
\end{equation*}
Similarly, the upper bound
\begin{equation*}
  t_u
  = \lim_{y \rightarrow \infty}
  -\frac{N}{{\sigma^*}^{(g)}r^{(g)}}\left(y-x^{(g)}\right)
  = -\infty
\end{equation*}
follows.
Hence, the transformed integral reads
\begin{align}
  &\begin{multlined}
    \frac{\partial^2 f_{\text{feas}}}{\partial \sigma^2}
    \Bigr|_{\sigma=\sigma^{(g)}}
    =-\frac{2}{\sigma^{(g)}}
    (\lambda-\mu)\binom{\lambda}{\mu}
    \int_{t=\infty}^{t=-\infty}
    t\frac{1}{{(\sqrt{2\pi})}^2\sigma^{(g)}}
    e^{-\frac{2}{2}t^2}\\
    \times[1-
    \Phi\left(-t\right)
    ]^{\lambda-\mu-1}[
    \Phi\left(-t\right)
    ]^{\mu-1}\,\mathrm{d}t.
  \end{multlined}
\end{align}
This integral can further be rewritten by making use of the fact that
taking the negative of an integral is equivalent to exchanging the
lower and upper bounds. Further, the identity $\Phi(t) = 1-\Phi(-t)$
is used. The resulting expression reads
\begin{align}
  &\begin{multlined}
    \frac{\partial^2 f_{\text{feas}}}{\partial \sigma^2}
    \Bigr|_{\sigma=\sigma^{(g)}}
    =\frac{2}{\sigma^{(g)}}
    (\lambda-\mu)\binom{\lambda}{\mu}
    \int_{t=-\infty}^{t=\infty}
    t\frac{1}{{(\sqrt{2\pi})}^2\sigma^{(g)}}
    e^{-\frac{2}{2}t^2}\\
    \times[\Phi\left(t\right)
    ]^{\lambda-\mu-1}[
    1-\Phi\left(t\right)
    ]^{\mu-1}\,\mathrm{d}t
  \end{multlined}\\
  &\begin{multlined}
    \phantom{\frac{\partial^2 f_{\text{feas}}}{\partial \sigma^2} }
    =\frac{2}{{\sigma^{(g)}}^2}
    \underbrace{\frac{\lambda-\mu}{{(\sqrt{2\pi})}^{1+1}}
    \binom{\lambda}{\mu}
    \int_{t=-\infty}^{t=\infty}
    t\,
    e^{-\frac{1+1}{2}t^2}
    [\Phi\left(t\right)
    ]^{\lambda-\mu-1}[
    1-\Phi\left(t\right)
    ]^{\mu-1}\,\mathrm{d}t}_{=e_{\mu,\lambda}^{1,1}}.
  \end{multlined}
  \label{chapter:analysis_repairbyprojection_multirecombinative:sec:theoreticalanalysis:eq:fsecondderivativefeas}
\end{align}
In
\Cref{chapter:analysis_repairbyprojection_multirecombinative:sec:theoreticalanalysis:eq:fsecondderivativefeas},
one of the so-called generalized progress coefficients
(see \Cref{chapter:analysis_repairbyprojection_multirecombinative:sec:theoreticalanalysis:eq:eabml}) appears. 
Therefore,
\begin{equation}
  \label{chapter:analysis_repairbyprojection_multirecombinative:sec:theoreticalanalysis:eq:fsecondderivativeinsertedfeas}
  \frac{\partial^2 f_{\text{feas}}}{\partial \sigma^2}
  \Bigr|_{\sigma=\sigma^{(g)}}=
  \frac{2}{{\sigma^{(g)}}^2}e_{\mu,\lambda}^{1,1}
\end{equation}
follows.
Insertion of
\Cref{chapter:analysis_repairbyprojection_multirecombinative:sec:theoreticalanalysis:eq:fsecondderivativeinsertedfeas}
into
\Cref{chapter:analysis_repairbyprojection_multirecombinative:sec:theoreticalanalysis:eq:psiapproxwithf}
yields the \gls{SAR} for the feasible case
\begin{align}
  \label{chapter:analysis_repairbyprojection_multirecombinative:sec:theoreticalanalysis:eq:psiapproxfeas}
  \psi_{\text{feas}}
  \approx
  0
  + \frac{\tau^2}{2}
  + \frac{\tau^2}{2}{\sigma^{(g)}}^2
  \frac{2}{{\sigma^{(g)}}^2}e_{\mu,\lambda}^{1,1}
  =\frac{\tau^2}{2}\left(1+2 e_{\mu,\lambda}^{1,1}\right)
  =\tau^2\left(\frac{1}{2}+e_{\mu,\lambda}^{1,1}\right).
\end{align}

\subsection{The Approximate \glsfmttext{SAR}
  in the case that the
  probability of feasible offspring tends to \texorpdfstring{$0$}{\$0\$}}
In order to treat
\Cref{chapter:analysis_repairbyprojection_multirecombinative:sec:theoreticalanalysis:eq:fsecondderivative}
further for the infeasible case,
$\frac{\partial}{\partial \sigma}{P_Q}_{\text{infeas}}
(y \,|\, x^{(g)}, r^{(g)}, \sigma)$
needs to be derived.
To this end,
\Cref{sec:theoreticalanalysis:eq:approximatedPQforprogress}
is simplified further using $\sqrt{1+x} \simeq 1 + \frac{x}{2} + O(x^2)$
\begin{equation}
  \bar{r}
  \simeq \sqrt{1+\frac{{\sigma^{(g)}}^2N}{{r^{(g)}}^2}}
  \simeq 1+\frac{{\sigma^{(g)}}^2N}{2{r^{(g)}}^2}
  = 1+\frac{{{\sigma^{(g)}}^*}^2}{2N}
\end{equation}
(which is a justified approximation if ${{\sigma^{(g)}}^*}^2 \ll N$)
and for $N \rightarrow \infty$
\begin{equation}
  \sigma^{(g)} \simeq \sigma_r
\end{equation}
(see \Cref{sec:theoreticalanalysis:eq:approximatedr})
yielding
\begin{equation}
  \label{chapter:analysis_repairbyprojection_multirecombinative:sec:theoreticalanalysis:eq:PQinfeasforsar}
  {P_Q}_{\text{infeas}}(q\,|\,x^{(g)}, r^{(g)}, \sigma) \approx
  \Phi\left(\frac{\left(1+\frac{1}{\xi}\right)q
    -x^{(g)}-\frac{r^{(g)}}{\sqrt{\xi}}
    \left(1+\frac{\sigma^2N}{2{r^{(g)}}^2}\right)}
           {\sigma\sqrt{1 + \frac{1}{\xi}}}\right)
\end{equation}
and
\begin{align}
  {p_Q}_{\text{infeas}}&(q\,|\,x^{(g)}, r^{(g)}, \sigma) \\
  &=\frac{\mathrm{d}}{\mathrm{d}q}{P_Q}_{\text{infeas}}(q)\\
  &\approx
  \frac{1}{\sqrt{2\pi}}
  \frac{\sqrt{1+\frac{1}{\xi}}}
       {\sigma}
       \exp\left[
         -\frac{1}{2}
         \left(
         \frac{\left(1+\frac{1}{\xi}\right)q
           -x^{(g)}-\frac{r^{(g)}}{\sqrt{\xi}}
           \left(1+\frac{\sigma^2N}{2{r^{(g)}}^2}\right)}
           {\sigma\sqrt{1 + \frac{1}{\xi}}}
         \right)^2
         \right].
  \label{chapter:analysis_repairbyprojection_multirecombinative:sec:theoreticalanalysis:eq:pqinfeasforsar}
\end{align}
Insertion of
\Cref{chapter:analysis_repairbyprojection_multirecombinative:sec:theoreticalanalysis:eq:PQinfeasforsar}
into
\Cref{chapter:analysis_repairbyprojection_multirecombinative:sec:theoreticalanalysis:eq:fsecondderivative}
and subsequent application of the chain rule results in
\begin{align}
  &\begin{multlined}
    \frac{\partial}{\partial \sigma}
    {P_Q}_{\text{infeas}}
    (y \,|\, x^{(g)}, r^{(g)}, \sigma)
    \approx
    \frac{\partial}{\partial \sigma}
    \Phi\left(\frac{\left(1+\frac{1}{\xi}\right)y
      -x^{(g)}-\frac{r^{(g)}}{\sqrt{\xi}}
      \left(1+\frac{\sigma^2N}{2{r^{(g)}}^2}\right)}
             {\sigma\sqrt{1 + \frac{1}{\xi}}}\right)
  \end{multlined}
  \label{chapter:analysis_repairbyprojection_multirecombinative:sec:theoreticalanalysis:eq:PQinfeasderivative1}\\
  &\begin{multlined}
    \phantom{\frac{\partial}{\partial \sigma}
    }
    =\frac{\partial}{\partial \sigma}
    \Phi\left(\frac{\left(1+\frac{1}{\xi}\right)y
    -x^{(g)}-\frac{r^{(g)}}{\sqrt{\xi}}}
           {\sigma\sqrt{1 + \frac{1}{\xi}}}
           -\frac{\sigma N}{2 r^{(g)}\sqrt{\xi}\sqrt{1 + \frac{1}{\xi}}}
           \right)
  \end{multlined}
  \label{chapter:analysis_repairbyprojection_multirecombinative:sec:theoreticalanalysis:eq:PQinfeasderivative2}\\
  &\begin{multlined}
    \phantom{\frac{\partial}{\partial \sigma}
    }
    =\phi\left(\frac{\left(1+\frac{1}{\xi}\right)y
      -x^{(g)}-\frac{r^{(g)}}{\sqrt{\xi}}}
           {\sigma\sqrt{1 + \frac{1}{\xi}}}
           -\frac{\sigma N}{2 r^{(g)}\sqrt{\xi}\sqrt{1 + \frac{1}{\xi}}}
           \right)\\
    \times\left(-\frac{\left(1+\frac{1}{\xi}\right)y
      -x^{(g)}-\frac{r^{(g)}}{\sqrt{\xi}}}
           {\sigma^2\sqrt{1 + \frac{1}{\xi}}}
           -\frac{N}{2 r^{(g)}\sqrt{\xi}\sqrt{1 + \frac{1}{\xi}}}
           \right)
  \end{multlined}
  \label{chapter:analysis_repairbyprojection_multirecombinative:sec:theoreticalanalysis:eq:PQinfeasderivative3}\\
  &\begin{multlined}
    \phantom{\frac{\partial}{\partial \sigma}
    }
    =\frac{1}{\sqrt{2\pi}}
      \exp\left[-\frac{1}{2}\left(\frac{\left(1+\frac{1}{\xi}\right)y
      -x^{(g)}-\frac{r^{(g)}}{\sqrt{\xi}}}
           {\sigma\sqrt{1 + \frac{1}{\xi}}}
           -\frac{\sigma N}{2 r^{(g)}\sqrt{\xi}\sqrt{1 + \frac{1}{\xi}}}
           \right)^2\right]\\
    \times\left(-\frac{\left(1+\frac{1}{\xi}\right)y
      -x^{(g)}-\frac{r^{(g)}}{\sqrt{\xi}}}
           {\sigma^2\sqrt{1 + \frac{1}{\xi}}}
           -\frac{N}{2 r^{(g)}\sqrt{\xi}\sqrt{1 + \frac{1}{\xi}}}
           \right).
  \end{multlined}
  \label{chapter:analysis_repairbyprojection_multirecombinative:sec:theoreticalanalysis:eq:PQinfeasderivative4}
\end{align}
Insertion of
\Cref{chapter:analysis_repairbyprojection_multirecombinative:sec:theoreticalanalysis:eq:PQinfeasderivative4,%
chapter:analysis_repairbyprojection_multirecombinative:sec:theoreticalanalysis:eq:pqinfeasforsar,%
chapter:analysis_repairbyprojection_multirecombinative:sec:theoreticalanalysis:eq:PQinfeasforsar}
into
\Cref{chapter:analysis_repairbyprojection_multirecombinative:sec:theoreticalanalysis:eq:fsecondderivative}
yields
\begin{align}
  &\begin{multlined}
    \frac{\partial^2 f}{\partial \sigma^2}\Bigr|_{\sigma=\sigma^{(g)}}
    =\frac{2}{\sigma^{(g)}}
    (\lambda-\mu)\binom{\lambda}{\mu}
    \int_{y=0}^{y=\infty}
    \frac{1}{\sqrt{2\pi}}
    \frac{\sqrt{1+\frac{1}{\xi}}}
       {{\sigma^{(g)}}}\\
       \times\exp\left[
         -\frac{1}{2}
         \left(
         \frac{\left(1+\frac{1}{\xi}\right)y
           -x^{(g)}-\frac{r^{(g)}}{\sqrt{\xi}}
           \left(1+\frac{{\sigma^{(g)}}^2N}{2{r^{(g)}}^2}\right)}
           {{\sigma^{(g)}}\sqrt{1 + \frac{1}{\xi}}}
         \right)^2
         \right]\\
    \times\left[1-
    \Phi\left(\frac{\left(1+\frac{1}{\xi}\right)y
    -x^{(g)}-\frac{r^{(g)}}{\sqrt{\xi}}
    \left(1+\frac{{\sigma^{(g)}}^2N}{2{r^{(g)}}^2}\right)}
           {{\sigma^{(g)}}\sqrt{1 + \frac{1}{\xi}}}\right)
    \right]^{\lambda-\mu-1}\\
    \times\left[
    \Phi\left(\frac{\left(1+\frac{1}{\xi}\right)y
    -x^{(g)}-\frac{r^{(g)}}{\sqrt{\xi}}
    \left(1+\frac{{\sigma^{(g)}}^2N}{2{r^{(g)}}^2}\right)}
           {{\sigma^{(g)}}\sqrt{1 + \frac{1}{\xi}}}\right)
    \right]^{\mu-1}\\
    \times\frac{1}{\sqrt{2\pi}}
      \underbrace{\exp\left[-\frac{1}{2}
      \left(\frac{\left(1+\frac{1}{\xi}\right)y
      -x^{(g)}-\frac{r^{(g)}}{\sqrt{\xi}}}
           {{\sigma^{(g)}}\sqrt{1 + \frac{1}{\xi}}}
           -\frac{{\sigma^{(g)}} N}{2 r^{(g)}\sqrt{\xi}\sqrt{1 + \frac{1}{\xi}}}
           \right)^2
           \right]}
           _{=\exp\left[-\frac{1}{2}\left(\frac{\left(1+\frac{1}{\xi}\right)y
               -x^{(g)}-\frac{r^{(g)}}{\sqrt{\xi}}
                   \left(1+\frac{{\sigma^{(g)}}^2N}{2{r^{(g)}}^2}\right)}
                   {{\sigma^{(g)}}\sqrt{1 + \frac{1}{\xi}}}\right)^2\right]}\\
           \times\left(-\frac{\left(1+\frac{1}{\xi}\right)y
           -x^{(g)}-\frac{r^{(g)}}{\sqrt{\xi}}}
           {{\sigma^{(g)}}^2\sqrt{1 + \frac{1}{\xi}}}
           -\frac{N}{2 r^{(g)}\sqrt{\xi}\sqrt{1 + \frac{1}{\xi}}}
           \right)
    \,\mathrm{d}y.
  \end{multlined}
\end{align}
For solving this integral,
\begin{equation}
  \frac{\left(1+\frac{1}{\xi}\right)y
    -x^{(g)}-\frac{r^{(g)}}{\sqrt{\xi}}
    \left(1+\frac{{\sigma^{(g)}}^2N}{2{r^{(g)}}^2}\right)}
           {{\sigma^{(g)}}\sqrt{1 + \frac{1}{\xi}}} := -t
\end{equation}
is substituted.
It further implies
\begin{align}
  \frac{\mathrm{d}t}{\mathrm{d}y}&=-\frac{1+1/\xi}{\sigma^{(g)}\sqrt{1+1/\xi}}\\
  \mathrm{d}y &= -\frac{\sigma^{(g)}}{\sqrt{1+1/\xi}}\mathrm{d}t.
\end{align}
Expressing $t$ with normalized
$\sigma^{(g)} = \frac{r^{(g)}{\sigma^{(g)}}^*}{N}$
one obtains
\begin{align}
  t &= -N\frac{\left(1+\frac{1}{\xi}\right)y
    -x^{(g)}-\frac{r^{(g)}}{\sqrt{\xi}}
    \left(1+\frac{{{\sigma^{(g)}}^*}^2{r^{(g)}}^2N}{2{r^{(g)}}^2N^2}\right)}
  {{\sigma^{(g)}}^*r^{(g)}\sqrt{1 + \frac{1}{\xi}}}\\
    &= -N\frac{\left(1+\frac{1}{\xi}\right)y
    -x^{(g)}-\frac{r^{(g)}}{\sqrt{\xi}}
    \left(1+\frac{{{\sigma^{(g)}}^*}^2}{2N}\right)}
  {{\sigma^{(g)}}^*r^{(g)}\sqrt{1 + \frac{1}{\xi}}}.
\end{align}
For $N \rightarrow \infty$, the integration bounds after substitution follow
as $t_l=\infty$ and $t_u=-\infty$.
The transformed integral reads
\begin{align}
  &\begin{multlined}
    \frac{\partial^2 f_{\text{infeas}}}{\partial \sigma^2}
    \Bigr|_{\sigma=\sigma^{(g)}}
    =-\frac{2}{\sigma^{(g)}}
    (\lambda-\mu)\binom{\lambda}{\mu}
    \int_{t=\infty}^{t=-\infty}
    \frac{1}{\sqrt{2\pi}}
       e^{-\frac{1}{2}t^2}
    \left[1-
    \Phi\left(-t\right)
    \right]^{\lambda-\mu-1}
    \left[
    \Phi\left(-t\right)
    \right]^{\mu-1}\\
    \times\frac{1}{\sqrt{2\pi}}
      e^{-\frac{1}{2}t^2}
      \left(
      \frac{t}{\sigma^{(g)}}
      -\frac{2 N}{2 r^{(g)}\sqrt{\xi}\sqrt{1 + \frac{1}{\xi}}}
      \right)
    \,\mathrm{d}t
  \end{multlined}
  \label{chapter:analysis_repairbyprojection_multirecombinative:sec:theoreticalanalysis:eq:fsecondderivativeinfeas1}\\
  &\begin{multlined}
    \phantom{\frac{\partial^2 f_{\text{infeas}}}{\partial \sigma^2} }
    =\frac{2}{{\sigma^{(g)}}^2}
    \underbrace{\frac{\lambda-\mu}{{(\sqrt{2\pi})}^{1+1}}
    \binom{\lambda}{\mu}
    \int_{t=-\infty}^{t=\infty}
    t\,e^{-\frac{1+1}{2}t^2}
    \left[
    \Phi\left(t\right)
    \right]^{\lambda-\mu-1}
    \left[1-
    \Phi\left(t\right)
    \right]^{\mu-1}
    \,\mathrm{d}t}_{=e_{\mu,\lambda}^{1,1}}\\
    -\frac{2}{{\sigma^{(g)}}}
    \frac{N}{r^{(g)}\sqrt{\xi}\sqrt{1 + \frac{1}{\xi}}}\\
    \times\underbrace{\frac{\lambda-\mu}{{(\sqrt{2\pi})}^{1+1}}
    \binom{\lambda}{\mu}
    \int_{t=-\infty}^{t=\infty}
    e^{-\frac{1+1}{2}t^2}
    \left[
    \Phi\left(t\right)
    \right]^{\lambda-\mu-1}
    \left[1-
    \Phi\left(t\right)
    \right]^{\mu-1}
    \,\mathrm{d}t}_{=c_{\mu/\mu,\lambda}}
  \end{multlined}
  \label{chapter:analysis_repairbyprojection_multirecombinative:sec:theoreticalanalysis:eq:fsecondderivativeinfeas2}\\
  &\begin{multlined}
    \phantom{\frac{\partial^2 f_{\text{infeas}}}{\partial \sigma^2} }
    =\frac{2}{{\sigma^{(g)}}^2}
      e_{\mu,\lambda}^{1,1}
     -\frac{2}{{\sigma^{(g)}}}
     \frac{N}{r^{(g)}\sqrt{\xi}\sqrt{1 + \frac{1}{\xi}}}
     c_{\mu/\mu,\lambda}.
  \end{multlined}
  \label{chapter:analysis_repairbyprojection_multirecombinative:sec:theoreticalanalysis:eq:fsecondderivativeinfeas3}
\end{align}
From
\Cref{chapter:analysis_repairbyprojection_multirecombinative:sec:theoreticalanalysis:eq:fsecondderivativeinfeas1}
to
\Cref{chapter:analysis_repairbyprojection_multirecombinative:sec:theoreticalanalysis:eq:fsecondderivativeinfeas2}
the fact that taking the negative of
an integral can be expressed by exchanging the lower and upper bounds,
and the identity $\Phi(t)=1-\Phi(-t)$ have been used.
Two of the generalized progress coefficients
(see
\Cref{chapter:analysis_repairbyprojection_multirecombinative:sec:theoreticalanalysis:eq:eabml})
appear and have been inserted into
\Cref{chapter:analysis_repairbyprojection_multirecombinative:sec:theoreticalanalysis:eq:fsecondderivativeinfeas3}.
Insertion of
\Cref{chapter:analysis_repairbyprojection_multirecombinative:sec:theoreticalanalysis:eq:fsecondderivativeinfeas3}
into
\Cref{chapter:analysis_repairbyprojection_multirecombinative:sec:theoreticalanalysis:eq:psiapproxwithf}
yields the \gls{SAR} for the infeasible case
\begin{align}
  \psi_{\text{infeas}}
  &\approx
  0
  + \frac{\tau^2}{2}
  + \frac{\tau^2}{2}{\sigma^{(g)}}^2
  \left(
  \frac{2}{{\sigma^{(g)}}^2}
  e_{\mu,\lambda}^{1,1}
  -\frac{2}{{\sigma^{(g)}}}
  \frac{N}{r^{(g)}\sqrt{\xi}\sqrt{1 + \frac{1}{\xi}}}
  c_{\mu/\mu,\lambda}
  \right)
  \label{chapter:analysis_repairbyprojection_multirecombinative:sec:theoreticalanalysis:eq:psiapproxinfeas1}\\
  &=\frac{\tau^2}{2}
  \left(1
  +2 e_{\mu,\lambda}^{1,1}
  -\frac{2 N \sigma^{(g)} c_{\mu/\mu,\lambda}}{r^{(g)}\sqrt{1+\xi}}
  \right)
  \label{chapter:analysis_repairbyprojection_multirecombinative:sec:theoreticalanalysis:eq:psiapproxinfeas2}\\
  &=\tau^2
  \left(\frac{1}{2}
  +e_{\mu,\lambda}^{1,1}
  -\frac{{\sigma^{(g)}}^* c_{\mu/\mu,\lambda}}{\sqrt{1+\xi}}
  \right)
  \label{chapter:analysis_repairbyprojection_multirecombinative:sec:theoreticalanalysis:eq:psiapproxinfeas3}
\end{align}

\subsection{The Approximate \glsfmttext{SAR}
  Progress Rate - Combination Using the
  Single Offspring Feasibility Probability}
Both cases are combined into
\begin{equation}
  \label{chapter:analysis_repairbyprojection_multirecombinative:sec:theoreticalanalysis:eq:psicombined}
  \psi \approx
  P_{\text{feas}}(x^{(g)}, r^{(g)}, \sigma^{(g)})
  {\psi}_{\text{feas}}
  + [1 - P_{\text{feas}}(x^{(g)}, r^{(g)}, \sigma^{(g)})]
  {\psi}_{\text{infeas}}.
\end{equation}

\newpage

\chapter{Derivation of
  \texorpdfstring{$\mathrm{E}\left[{q_r^2}_{m;\lambda}\right]$}
                 {\$\textbackslash mathrm\{E\}%
\textbackslash left[\{q\_r\textasciicircum 2\}\_\{m;\textbackslash lambda\}%
\textbackslash right]\$} and
\texorpdfstring{$\mathrm{E}\left[q^2_{m;\lambda}\right]$}
                 {\$\textbackslash mathrm\{E\}%
\textbackslash left[q\textasciicircum 2\_\{m;\textbackslash lambda\}%
\textbackslash right]\$}}
\chaptermark{Derivation of
  \texorpdfstring{$\mathrm{E}\left[{q_r^2}_{m;\lambda}\right]$}
                 {\$\textbackslash mathrm\{E\}%
\textbackslash left[\{q\_r\textasciicircum 2\}\_\{m;\textbackslash lambda\}%
\textbackslash right]\$} and
\texorpdfstring{$\mathrm{E}\left[q^2_{m;\lambda}\right]$}
                 {\$\textbackslash mathrm\{E\}%
\textbackslash left[q\textasciicircum 2\_\{m;\textbackslash lambda\}%
\textbackslash right]\$}}
\label{sec:appendix:derivation_expectation_qrsquareml}

$\mathrm{E}\left[{q_r^2}_{m;\lambda}\right]$ can be derived similarly to
the $(1,\lambda)$ case as done for deriving
$\mathrm{E}\left[{q_r}_{1;\lambda}\right]$
(see~\citeapp[Sec. 3.1.2.2, pp. 49-56]{SpettelBeyer2018SigmaSaEsConeAPP}):
\begin{align}
  \mathrm{E}[{q_r^2}_{m;\lambda}\,|\,x^{(g)}, r^{(g)}, \sigma^{(g)}]
  := \mathrm{E}[{q_r^2}_{m;\lambda}]
  &= \int_{q_r=0}^{q_r=\infty}q_r^2\,p_{{q_r}_{m;\lambda}}(q_r)\,\mathrm{d}q_r.
  \label{sec:appendix:derivation_expectation_qrsquareml:eq:expectedqr2mlambda1}
\end{align}
The density
\begin{equation}
  p_{{q_r}_{m;\lambda}}(q_r) := p_{{q_r}_{m\lambda}}(q_r \,|\, x^{(g)}, r^{(g)}, \sigma^{(g)})
\end{equation}
indicates the probability density function of
the m-th best (offspring with m-th smallest $q$ value) offspring's $q_r$ value.
The derivation is analogous to the $(1,\lambda)$ case
with the additional consideration for the order statistics
(m-th best instead of best as done for the multi-recombinative $x$ progress rate
(\Cref{sec:theoreticalanalysis:eq:pqmlambda})).
It yields
\begin{equation}
  \label{sec:appendix:derivation_expectation_qrsquareml:eq:pqrmlambda}
  p_{{q_r}_{m;\lambda}}(q_r) = \frac{\lambda!}{(\lambda-m)!(m-1)!}
  \int_{q=0}^{q=\infty}
  p_{Q,Q_r}(q,q_r)[1-P_Q(q)]^{\lambda-m}[P_Q(q)]^{m-1}\,\mathrm{d}q.
\end{equation}
Insertion of
\Cref{sec:appendix:derivation_expectation_qrsquareml:eq:pqrmlambda}
into
\Cref{sec:appendix:derivation_expectation_qrsquareml:eq:expectedqr2mlambda1}
results in
\begin{equation}
  \begin{multlined}
    \mathrm{E}[{q_r^2}_{m;\lambda}]
    = \int_{q_r=0}^{q_r=\infty}q_r^2\,
    \frac{\lambda!}{(\lambda-m)!(m-1)!}
    \int_{q=0}^{q=\infty}
    p_{Q,Q_r}(q,q_r)\\
    \times[1-P_Q(q)]^{\lambda-m}
    [P_Q(q)]^{m-1}\,\mathrm{d}q\,\mathrm{d}q_r.
  \end{multlined}
  \label{sec:appendix:derivation_expectation_qrsquareml:eq:expectedqr2mlambda2}
\end{equation}
By changing the order of integration,
\Cref{sec:appendix:derivation_expectation_qrsquareml:eq:expectedqr2mlambda2}
can be rewritten to
\begin{equation}
  \begin{multlined}
    \mathrm{E}[{q_r^2}_{m;\lambda}]
    = \frac{\lambda!}{(\lambda-m)!(m-1)!}
    \int_{q=0}^{q=\infty}
    \underbrace{\left[\int_{q_r=0}^{q_r=\infty}
    q_r^2\,p_{Q,Q_r}(q,q_r)\,\mathrm{d}q_r\right]}_{=:I_2(q)}\\
    \times[1-P_Q(q)]^{\lambda-m}
    [P_Q(q)]^{m-1}\,\mathrm{d}q.
  \end{multlined}
  \label{sec:appendix:derivation_expectation_qrsquareml:eq:expectedqr2mlambda3}
\end{equation}
Note that $I_2(q)$ is very similar to $I(q)$ from
from~\citeapp[Sec. 3.1.2.2, pp. 49-56]{SpettelBeyer2018SigmaSaEsConeAPP}.
It only differs in that $I_2(q)$ contains $q_r^2$ and $I(q)$ contains $q_r$.
As already done for $I(q)$, $I_2(q)$ is expressed in terms
of the values before projection. Because $q_r^2$ represents
the squared distance from the cone's axis after projection, it
only takes values in the interval $[0,(x/\sqrt{\xi})^2]$ for
$x \in \mathbb{R},x \ge 0$. This is because all individuals that
happen to be generated outside the cone are projected onto the
cone boundary. The integral in $I_2(q)$ is therefore split into
two summands. The first summand represents the case of immediately feasible
offspring individuals. The second summand represents the case
of projected offspring individuals. All points that lie on the
projection line for a given value of $q$ are projected to $q/\sqrt{\xi}$.
This writes
\begin{align}
  &\begin{multlined}
     I_2(q)\,\mathrm{d}q
     =\int_{\tilde{r}=0}^{\tilde{r}=q/\sqrt{\xi}}\tilde{r}^2
     p_{1;1}(q,\tilde{r})\,\mathrm{d}\tilde{r}
     \,\mathrm{d}q\\
     \hspace{4cm}+\left(\frac{q}{\sqrt{\xi}}\right)^2
     \underbrace{\left(p_Q(q)\,\mathrm{d}q
     -\int_{\tilde{r}=0}^{\tilde{r}=q/\sqrt{\xi}}
     p_{1;1}(q,\tilde{r})
     \,\mathrm{d}\tilde{r}
     \,\mathrm{d}q\right)}_{=\,\mathrm{d}P_{\text{line}}}
  \end{multlined}
  \label{sec:appendix:derivation_expectation_qrsquareml:eq:iq2_1}
\end{align}
With the same arguments and assumptions leading
to~\citeapp[Eq. (3.206), page 56]{SpettelBeyer2018SigmaSaEsConeAPP},
the approximation
\begin{equation}
  \label{sec:appendix:derivation_expectation_qrsquareml:eq:eqrmlambdainfeas}
  \mathrm{E}[{{q_r^2}_{m;\lambda}}_{\text{infeas}}]
     \approx
     \frac{1}{\xi}\mathrm{E}[{{q}^2_{m;\lambda}}_{\text{infeas}}]
\end{equation}
can be derived for the case that the m-th best offspring is infeasible
with high probability.
For the case that the m-th best offspring is feasible
almost surely, the complete probability mass lies inside the cone.
Consequently, the second summand in
\Cref{sec:appendix:derivation_expectation_qrsquareml:eq:iq2_1}
vanishes. Additionally, the integral in the
first summand yields the second (non-central)
moment $\bar{r}^2 + \sigma_r^2$ because
the bounds indicate the integration over the whole feasible region
for the given area $\mathrm{d}q$.
In the feasible case, $I_2(q)$ therefore reads
\begin{equation}
  {I_2}_{\text{feas}}(q) = p_x(q)(\bar{r}^2 + \sigma_r^2).
\end{equation}
By insertion of ${I_2}_{\text{feas}}(q)$ as $I_2(q)$ into
\Cref{sec:appendix:derivation_expectation_qrsquareml:eq:expectedqr2mlambda3},
use of ${P_Q}_{\text{feas}}$ from
\Cref{sec:theoreticalanalysis:eq:approximatedPQforprogressfeasible},
and use of $p_x(x)=\frac{1}{\sqrt{2\pi}\sigma^{(g)}}\exp\left[
-\frac{1}{2}\left(\frac{x - x^{(g)}}{\sigma^{(g)}}\right)^2\right]$
one obtains
\begin{align}
  \mathrm{E}[{{q^2_r}_{m;\lambda}}_{\text{feas}}]
  &\approx (\bar{r}^2 + \sigma_r^2)
  \frac{\lambda!}{(\lambda-m)!(m-1)!}
  \int_{q=0}^{q=\infty}
  \frac{1}{\sqrt{2\pi}\sigma^{(g)}}e^{
    -\frac{1}{2}\left(\frac{q - x^{(g)}}{\sigma^{(g)}}\right)^2}\notag\\
  &\hspace{2cm}\times\left[1-\Phi\left(\frac{q-x^{(g)}}
             {\sigma^{(g)}}\right)\right]^{\lambda-m}
  \left[\Phi\left(\frac{q-x^{(g)}}
             {\sigma^{(g)}}\right)\right]^{m-1}\,\mathrm{d}q\\
  &=\bar{r}^2 + \sigma_r^2.
  \label{sec:appendix:derivation_expectation_qrsquareml:eq:squaredqrfeas}
\end{align}
In the last step the fact that the integral over the whole probability
density function of the m-th order statistic yields 1 has been used.
Both can be combined with the (approximate) feasibility probability
\begin{equation}
  \label{sec:appendix:derivation_expectation_qrsquareml:eq:squaredqr}
  \mathrm{E}[{q^2_r}_{m;\lambda}] \approx
  P_{\text{feas}}(x^{(g)}, r^{(g)}, \sigma^{(g)})
  \mathrm{E}[{{q^2_r}_{m;\lambda}}_{\text{feas}}]
  + [1 - P_{\text{feas}}(x^{(g)}, r^{(g)}, \sigma^{(g)})]
  \mathrm{E}[{{q^2_r}_{m;\lambda}}_{\text{infeas}}].
\end{equation}
Because
$\mathrm{E}[{{q_r^2}_{m;\lambda}}_{\text{infeas}}]
  \approx
    \frac{1}{\xi}\mathrm{E}[{{q}^2_{m;\lambda}}_{\text{infeas}}]$
(see \Cref{sec:appendix:derivation_expectation_qrsquareml:eq:eqrmlambdainfeas})
holds, $\mathrm{E}[{{q}^2_{m;\lambda}}_{\text{infeas}}]$ needs to
be determined next.
For
\Cref{sec:appendix:derivation_expectation_qrsquareml:eq:eqrmlambdainfeas},
$\mathrm{E}[{{q}^2_{m;\lambda}}_{\text{infeas}}]$ is derived
and finally combined with an approximation for the feasibility probability.
$\mathrm{E}[{q}^2_{m;\lambda}]$ writes\footnote{Note that here only
$\mathrm{E}[{{q}^2_{m;\lambda}}_{\text{infeas}}]$ is necessary. In order
to have an expression for the whole $\mathrm{E}[{q}^2_{m;\lambda}]$,
the derivations for the feasible and infeasible cases are presented
here for completeness.}
\begin{equation}
  \mathrm{E}[{q}^2_{m;\lambda}]=
  \frac{\lambda!}{(\lambda-m)!(m-1)!}
  \int_{q=0}^{q=\infty}
  q^2 p_Q(q)[1-P_Q(q)]^{\lambda-m}[P_Q(q)]^{m-1}\,\mathrm{d}q.
\end{equation}
Use of
\Cref{sec:theoreticalanalysis:eq:approximatedPQforprogressfeasible,%
sec:theoreticalanalysis:eq:approximatedpQforprogressfeasible}
yields for the feasible case
\begin{align}
  &\begin{multlined}
  \mathrm{E}[{{q}^2_{m;\lambda}}_{\text{feas}}]\approx
  \frac{\lambda!}{(\lambda-m)!(m-1)!}
  \int_{q=\bar{r}\sqrt{\xi}}^{q=\infty}
  q^2
  \frac{1}{\sqrt{2\pi}\sigma^{(g)}}
     e^{-\frac{1}{2}\left(\frac{q-x^{(g)}}{\sigma^{(g)}}\right)^2}\\
  \times\left[1-
  \Phi\left(\frac{q-x^{(g)}}{\sigma^{(g)}}\right)
  \right]^{\lambda-m}\left[
  \Phi\left(\frac{q-x^{(g)}}{\sigma^{(g)}}\right)
  \right]^{m-1}\,\mathrm{d}q.
  \end{multlined}
\end{align}
The substitution
\begin{equation}
  \frac{q-x^{(g)}}{\sigma^{(g)}} := -t
\end{equation}
is used.
It follows that
\begin{equation}
  q=-t\sigma^{(g)}+x^{(g)}
\end{equation}
and
\begin{equation}
  \mathrm{d}q=-\sigma^{(g)}\,\mathrm{d}t.
\end{equation}
Using normalized quantities, $t$ can be expressed as
\begin{equation}
  t = \frac{-N(q-x^{(g)})}{{\sigma^{(g)}}^* r^{(g)}}.
\end{equation}
Assuming $N \rightarrow \infty$ yields for the upper bound $t_u=-\infty$.
For the lower bound it follows that
\begin{align}
  t_l
  &= \frac{-N(\bar{r}\sqrt{\xi}-x^{(g)})}{{\sigma^{(g)}}^* r^{(g)}}\\
  &= -\frac{N}{{\sigma^{(g)}}^*}
  \left(\frac{\bar{r}\sqrt{\xi}}{r^{(g)}}-\frac{x^{(g)}}{r^{(g)}}\right).
\end{align}
Because $\bar{r} \simeq r^{(g)}\sqrt{1+\frac{{{\sigma^{(g)}}^*}^2}{N}}$
(\Cref{sec:theoreticalanalysis:eq:approximatedr}),
for $N \rightarrow \infty$ and ${{\sigma^{(g)}}^*}^2 \ll N$,
$\bar{r}\sqrt{\xi} \simeq r^{(g)}\sqrt{\xi} \le x^{(g)}$ follows.
The last inequality follows from the fact that the parental
individual is feasible (ensured by projection).
Hence, it follows that for $N \rightarrow \infty$
\begin{align}
  t_l
  &= \frac{N}{{\sigma^{(g)}}^*}
  \underbrace{
    \left(\frac{x^{(g)}}{r^{(g)}}-\frac{\bar{r}\sqrt{\xi}}{r^{(g)}}\right)}_
             {\ge 0}\\
  &\simeq \infty
\end{align}
holds.
Applying the substitution results in
\begin{align}
  &\begin{multlined}
  \mathrm{E}[{{q}^2_{m;\lambda}}_{\text{feas}}]\approx
  -\frac{\lambda!}{(\lambda-m)!(m-1)!}
  \frac{1}{\sqrt{2\pi}}
  \int_{t=\infty}^{t=-\infty}
  (-\sigma^{(g)}t + x^{(g)})^2
  e^{-\frac{1}{2}t^2}\\
  \times\left[1-
  \Phi(-t)
  \right]^{\lambda-m}\left[
  \Phi(-t)
  \right]^{m-1}\,\mathrm{d}t
  \end{multlined}\\
  &\begin{multlined}
  \phantom{\mathrm{E}[{{q}^2_{m;\lambda}}_{\text{feas}}]}=
  \frac{\lambda!}{(\lambda-m)!(m-1)!}
  \frac{1}{\sqrt{2\pi}}
  \int_{t=-\infty}^{t=\infty}
  \left({\sigma^{(g)}}^2t^2 - 2 \sigma^{(g)}t x^{(g)} + {x^{(g)}}^2\right)
  e^{-\frac{1}{2}t^2}\\
  \times\left[
  \Phi(t)
  \right]^{\lambda-m}\left[
  1-\Phi(t)
  \right]^{m-1}\,\mathrm{d}t
  \end{multlined}\\
  &\begin{multlined}
  \phantom{\mathrm{E}[{{q}^2_{m;\lambda}}_{\text{feas}}]}=
  {\sigma^{(g)}}^2 e^{0,2}_{(m-1),\lambda}
  - 2 \sigma^{(g)} x^{(g)} e^{0,1}_{(m-1),\lambda}
  + {x^{(g)}}^2
  \end{multlined}
  \label{sec:appendix:derivation_expectation_qrsquareml:eq:squaredqfeas}
\end{align}
where the generalized progress coefficients have been used.
Those are defined in~\citeapp[Eq. (5.112), p. 172]{Beyer2001APP}.
They are defined because
those integrals cannot be solved analytically for large $\lambda$ and large
$\mu$. The generalized progress coefficients are defined as
\begin{equation}
  \label{chapter:analysis_repairbyprojection_multirecombinative:sec:theoreticalanalysis:eq:eabml}
  e_{\mu,\lambda}^{\alpha,\beta} :=
  \frac{\lambda-\mu}{{(\sqrt{2\pi})}^{\alpha+1}}\binom{\lambda}{\mu}
  \int_{t=-\infty}^{t=\infty}
  t^\beta
  e^{-\frac{\alpha+1}{2}t^2}
  [\Phi(t)]^{\lambda-\mu-1}[1-\Phi(t)]^{\mu-\alpha}
  \,\mathrm{d}t.
\end{equation}
Use of
\Cref{sec:theoreticalanalysis:eq:approximatedPQforprogress,%
sec:theoreticalanalysis:eq:approximatedpQforprogress}
yields for the infeasible case
\begin{align}
  &\begin{multlined}
  \mathrm{E}[{{q}^2_{m;\lambda}}_{\text{infeas}}]\approx
  \frac{\lambda!}{(\lambda-m)!(m-1)!}
  \int_{q=0}^{q=\bar{r}\sqrt{\xi}}
  q^2
  \left(\frac{(1+1/\xi)}
  {\sqrt{{\sigma^{(g)}}^2 + \sigma_r^2/\xi}}\right)
  \frac{1}{\sqrt{2\pi}}\\\times\exp
  \left[-\frac{1}{2}\left(\frac{(1+1/\xi)q-x^{(g)}-\bar{r}/\sqrt{\xi}}
  {\sqrt{{\sigma^{(g)}}^2 + \sigma_r^2/\xi}}\right)^2\right]\\
  \times\left[1-
      \Phi\left(\frac{(1+1/\xi)q-x^{(g)}-\bar{r}/\sqrt{\xi}}
      {\sqrt{{\sigma^{(g)}}^2 + \sigma_r^2/\xi}}\right)
  \right]^{\lambda-m}\\\times\left[
      \Phi\left(\frac{(1+1/\xi)q-x^{(g)}-\bar{r}/\sqrt{\xi}}
      {\sqrt{{\sigma^{(g)}}^2 + \sigma_r^2/\xi}}\right)
  \right]^{m-1}\,\mathrm{d}q.
  \end{multlined}
\end{align}
The substitution
\begin{equation}
  \frac{(1+1/\xi)q-x^{(g)}-\bar{r}/\sqrt{\xi}}
       {\sqrt{{\sigma^{(g)}}^2 + \sigma_r^2/\xi}} := -t
\end{equation}
is used.
It follows that
\begin{equation}
  q=\frac{1}{(1+1/\xi)}
  \left(-\sqrt{{\sigma^{(g)}}^2+\sigma_r^2/\xi}\,t
  +x^{(g)}+\bar{r}/\sqrt{\xi}\right)
\end{equation}
and
\begin{equation}
  \mathrm{d}q=
  -\frac{\sqrt{{\sigma^{(g)}}^2+\sigma_r^2/\xi}}{(1+1/\xi)}\,\mathrm{d}t.
\end{equation}
Using the normalized ${\sigma^{(g)}}^*$, $\sigma_r \simeq \sigma^{(g)}$
for $N \rightarrow \infty$, and ${\sigma^{(g)}}^* \ll N$
(derived from \Cref{sec:theoreticalanalysis:eq:approximatedr}),
$t$ can be expressed as
\begin{align}
  t &= -\frac{(1+1/\xi)q-x^{(g)}-\bar{r}/\sqrt{\xi}}
  {\sqrt{\frac{{{\sigma^{(g)}}^*}^2{r^{(g)}}^2}{N^2}
      +\frac{{{\sigma^{(g)}}^*}^2{r^{(g)}}^2}{N^2\xi}}}\\
  &= -\frac{(1+1/\xi)q-x^{(g)}-\bar{r}/\sqrt{\xi}}
  {\frac{1}{N\sqrt{\xi}}\sqrt{\xi{{\sigma^{(g)}}^*}^2{r^{(g)}}^2
      +{{\sigma^{(g)}}^*}^2{r^{(g)}}^2}}\\
  &= -N\sqrt{\xi}\left[\frac{(1+1/\xi)q-x^{(g)}-\bar{r}/\sqrt{\xi}}
  {\sqrt{\xi{{\sigma^{(g)}}^*}^2{r^{(g)}}^2
      +{{\sigma^{(g)}}^*}^2{r^{(g)}}^2}}\right]\\
  &= -N\sqrt{\xi}\left[\frac{(1+1/\xi)q-x^{(g)}-\bar{r}/\sqrt{\xi}}
  {{{\sigma^{(g)}}^*}{r^{(g)}}\sqrt{\xi+1}}\right].
\end{align}
The lower bound in the transformed integral therefore follows
assuming $\xi \gg 1$, $N \rightarrow \infty$, using
$\infty > \bar{r} \simeq r^{(g)} \ge 0$,
and knowing that $0 \le x^{(g)} < \infty$ as
\begin{align}
  t_l &= -N\sqrt{\xi}\left[\frac{(1+1/\xi)0-x^{(g)}-\bar{r}/\sqrt{\xi}}
  {{{\sigma^{(g)}}^*}{r^{(g)}}\sqrt{\xi+1}}\right]\\
  &= -N\sqrt{\xi}\left[\frac{-x^{(g)}-\bar{r}/\sqrt{\xi}}
  {{{\sigma^{(g)}}^*}{r^{(g)}}\sqrt{\xi+1}}\right]\\
  &\simeq -N\underbrace{\left[\frac{-x^{(g)} - \bar{r}/\sqrt{\xi}}
  {{{\sigma^{(g)}}^*}{r^{(g)}}}\right]}_{\le 0}\\
  &\simeq \infty.
\end{align}
Similarly, the upper bound follows with the same assumptions
and using the fact that the case under consideration is
the infeasible case, i.e., $x^{(g)}\le\bar{r}\sqrt{\xi}$
\begin{align}
  t_u &= -N\sqrt{\xi}\left[\frac{(1+1/\xi)\bar{r}\sqrt{\xi}
  -x^{(g)}-\bar{r}/\sqrt{\xi}}
  {{{\sigma^{(g)}}^*}{r^{(g)}}\sqrt{\xi+1}}\right]\\
  &=-N\underbrace{\left[\frac{\bar{r}\sqrt{\xi}-x^{(g)}}
  {{{\sigma^{(g)}}^*}{r^{(g)}}}\right]}_{\ge 0}\\
  &\simeq -\infty.
\end{align}
Actually applying the substitution leads to
\begin{align}
  &\begin{multlined}
  \mathrm{E}[{{q}^2_{m;\lambda}}_{\text{infeas}}]\approx
  -\frac{\lambda!}{(\lambda-m)!(m-1)!}\\
  \times\int_{t=\infty}^{t=-\infty}
  \left[\frac{1}{(1+1/\xi)}
  \left(-\sqrt{{\sigma^{(g)}}^2+\sigma_r^2/\xi}\,t
  +x^{(g)}+\bar{r}/\sqrt{\xi}\right)\right]^2\\
  \times\frac{1}{\sqrt{2\pi}}e^{-\frac{1}{2}t^2}
  \left[1-
      \Phi\left(-t\right)
  \right]^{\lambda-m}\left[
      \Phi\left(-t\right)
  \right]^{m-1}\,\mathrm{d}t
  \end{multlined}\\
  &\begin{multlined}
  \phantom{\mathrm{E}[{{q}^2_{m;\lambda}}_{\text{infeas}}]}=
  \frac{\lambda!}{(\lambda-m)!(m-1)!}\\
  \times\int_{t=-\infty}^{t=\infty}
  \left[\frac{1}{(1+1/\xi)}
  \left(-\sqrt{{\sigma^{(g)}}^2+\sigma_r^2/\xi}\,t
  +x^{(g)}+\bar{r}/\sqrt{\xi}\right)\right]^2\\
  \times\frac{1}{\sqrt{2\pi}}e^{-\frac{1}{2}t^2}
  \left[
      \Phi\left(t\right)
  \right]^{\lambda-m}\left[1-
      \Phi\left(t\right)
  \right]^{m-1}\,\mathrm{d}t
  \end{multlined}\\
  &\begin{multlined}
  \phantom{\mathrm{E}[{{q}^2_{m;\lambda}}_{\text{infeas}}]}=
  \frac{({\sigma^{(g)}}^2+\sigma_r^2/\xi)}{(1+1/\xi)^2}e^{0,2}_{(m-1),\lambda}
  -\frac{\sqrt{{\sigma^{(g)}}^2+\sigma_r^2/\xi}}{(1+1/\xi)^2}2(x^{(g)}+\bar{r}/\sqrt{\xi})e^{0,1}_{(m-1),\lambda}\\
  +\frac{(x^{(g)}+\bar{r}/\sqrt{\xi})^2}{(1+1/\xi)^2}.
  \end{multlined}
  \label{sec:appendix:derivation_expectation_qrsquareml:eq:squaredqinfeas}
\end{align}
Both can be combined with the (approximate) feasibility probability
\begin{equation}
  \label{sec:appendix:derivation_expectation_qrsquareml:eq:squaredq}
  \mathrm{E}[{q}^2_{m;\lambda}] \approx
  P_{\text{feas}}(x^{(g)}, r^{(g)}, \sigma^{(g)})
  \mathrm{E}[{{q}^2_{m;\lambda}}_{\text{feas}}]
  + [1 - P_{\text{feas}}(x^{(g)}, r^{(g)}, \sigma^{(g)})]
  \mathrm{E}[{{q}^2_{m;\lambda}}_{\text{infeas}}].
\end{equation}

\newpage

\chapter{Derivation of
  \texorpdfstring{$\mathrm{E}\left[\langle q^2 \rangle\right]$}
                 {\$\textbackslash mathrm\{E\}%
\textbackslash left[\textbackslash%
langle q\textasciicircum 2 \textbackslash rangle%
\textbackslash right]\$}}
\chaptermark{}
\label{sec:appendix:derivation_expectation_qrsquarecentroid}

\begin{equation}
  \mathrm{E}\left[\langle q^2 \rangle\right]
  =
  \mathrm{E}\left[
    \frac{1}{\mu}\sum_{m=1}^{\mu}q^2_{m;\lambda}
  \right]
  =
  \frac{1}{\mu}
  \sum_{m=1}^{\mu}
  \mathrm{E}\left[
    q^2_{m;\lambda}
  \right]
\end{equation}
follows by expanding the notation for the centroid computation
and linearity of expectation.
An approximation for
$\mathrm{E}\left[q^2_{m;\lambda}\right]$ has been derived in
\Cref{sec:appendix:derivation_expectation_qrsquareml}
as
\Cref{sec:appendix:derivation_expectation_qrsquareml:eq:squaredq}
together with
\Cref{sec:appendix:derivation_expectation_qrsquareml:eq:squaredqfeas}
and
\Cref{sec:appendix:derivation_expectation_qrsquareml:eq:squaredqinfeas}.
Using
\Cref{sec:appendix:derivation_expectation_qrsquareml:eq:squaredqfeas,%
sec:appendix:derivation_expectation_qrsquareml:eq:squaredqinfeas,%
sec:appendix:derivation_expectation_qrsquareml:eq:squaredq},
\begin{align}
  &\begin{multlined}
  \frac{1}{\mu}\sum_{m=1}^{\mu}\mathrm{E}[{q}^2_{m;\lambda}]
  \approx
  \frac{1}{\mu}\sum_{m=1}^{\mu}\bigg[
  P_{\text{feas}}(x^{(g)}, r^{(g)}, \sigma^{(g)})
  \mathrm{E}[{{q}^2_{m;\lambda}}_{\text{feas}}]\\
  \hspace{3cm}+ [1 - P_{\text{feas}}(x^{(g)}, r^{(g)}, \sigma^{(g)})]
  \mathrm{E}[{{q}^2_{m;\lambda}}_{\text{infeas}}]
  \bigg]
  \end{multlined}\\
  &\begin{multlined}
  \phantom{\frac{1}{\mu}\sum_{m=1}^{\mu}\mathrm{E}[{q}^2_{m;\lambda}]}
  \approx
  P_{\text{feas}}(x^{(g)}, r^{(g)}, \sigma^{(g)})
  \left({\sigma^{(g)}}^2
  \left[\frac{1}{\mu}\sum_{m=1}^{\mu}e^{0,2}_{(m-1),\lambda}\right]\right.\\\left.
  \hspace{3cm}- 2 \sigma^{(g)} x^{(g)}
  \left[\frac{1}{\mu}\sum_{m=1}^{\mu}e^{0,1}_{(m-1),\lambda}\right]
  + {x^{(g)}}^2\right)\\
  + [1 - P_{\text{feas}}(x^{(g)}, r^{(g)}, \sigma^{(g)})]
  \bigg(\frac{({\sigma^{(g)}}^2+\sigma_r^2/\xi)}{(1+1/\xi)^2}
  \left[\frac{1}{\mu}\sum_{m=1}^{\mu}e^{0,2}_{(m-1),\lambda}\right]\\
  \hspace{1cm}-\frac{\sqrt{{\sigma^{(g)}}^2+\sigma_r^2/\xi}}{(1+1/\xi)^2}2(x^{(g)}+\bar{r}/\sqrt{\xi})
  \left[\frac{1}{\mu}\sum_{m=1}^{\mu}e^{0,1}_{(m-1),\lambda}\right]\\
  +\frac{(x^{(g)}+\bar{r}/\sqrt{\xi})^2}{(1+1/\xi)^2}\bigg)
  \end{multlined}
  \label{sec:appendix:derivation_expectation_qrsquarecentroid:eq:squarecentroid}
\end{align}
can be written.

An expression for
$\frac{1}{\mu}\sum_{m=1}^{\mu}e^{0,2}_{(m-1),\lambda}$
can be derived using the definition of the generalized progress
coefficients (\Cref{chapter:analysis_repairbyprojection_multirecombinative:sec:theoreticalanalysis:eq:eabml}).
By this definition,
\begin{equation}
  \begin{multlined}
    \frac{1}{\mu}\sum_{m=1}^{\mu}e^{0,2}_{(m-1),\lambda} =
    \frac{1}{\mu}\sum_{m=1}^{\mu}\bigg[
    \frac{\lambda-(m-1)}{\sqrt{2\pi}}\binom{\lambda}{m-1}\\
    \times
    \int_{t=-\infty}^{t=\infty}
    t^2
    e^{-\frac{1}{2}t^2}
    [\Phi(t)]^{\lambda-m}[1-\Phi(t)]^{m-1}\bigg]
    \,\mathrm{d}t
  \end{multlined}
\end{equation}
follows. Rewriting this results in
\begin{equation}
  \begin{multlined}
    \frac{1}{\mu}\sum_{m=1}^{\mu}e^{0,2}_{(m-1),\lambda} =
    \frac{1}{\mu}
    \frac{\lambda !}{\sqrt{2\pi}}
    \int_{t=-\infty}^{t=\infty}
    t^2
    e^{-\frac{1}{2}t^2}
    \sum_{m=1}^{\mu}
    \frac{[\Phi(t)]^{\lambda-m}[1-\Phi(t)]^{m-1}}
         {(\lambda-m)!(m-1)!}
    \,\mathrm{d}t.
  \end{multlined}
\end{equation}
Substituting $s:=-t$ and using the identity $\Phi(-s)=1-\Phi(s)$ yields
further
\begin{equation}
  \begin{multlined}
    \frac{1}{\mu}\sum_{m=1}^{\mu}e^{0,2}_{(m-1),\lambda} =
    \frac{1}{\mu}
    \frac{\lambda !}{\sqrt{2\pi}}
    \int_{s=-\infty}^{s=\infty}
    s^2
    e^{-\frac{1}{2}s^2}
    \sum_{m=1}^{\mu}
    \frac{[1-\Phi(s)]^{\lambda-m}[\Phi(s)]^{m-1}}
         {(\lambda-m)!(m-1)!}
    \,\mathrm{d}s.
  \end{multlined}
\end{equation}
Now, with the same arguments that were used to get from
\Cref{sec:theoreticalanalysis:eq:qcentroid3}
to
\Cref{sec:theoreticalanalysis:eq:qcentroid4},
\begin{equation}
  \begin{multlined}
    \frac{1}{\mu}\sum_{m=1}^{\mu}e^{0,2}_{(m-1),\lambda} =
    (\lambda-\mu)\binom{\lambda}{\mu}
    \int_{y=-\infty}^{y=\infty}
    \phi(y)
    [1-\Phi(y)]^{\lambda-\mu-1}
    [\Phi(y)]^{\mu-1}
    \int_{s=-\infty}^{s=y}s^2\,\phi(s)\,\mathrm{d}s\,\mathrm{d}y
  \end{multlined}
\end{equation}
can be derived. Use of the first identity
of~\citeapp[Eq. (A.17), p. 331)]{Beyer2001APP}
\begin{equation}
  \int_{-\infty}^{x}t^2\,e^{-\frac{1}{2}t^2}\,\mathrm{d}t
  =
  \sqrt{2\pi}\Phi(x)-x e^{-\frac{1}{2}x^2}
\end{equation}
results in $\Phi(y)-y \phi(y)$ for the inner integral.
Consequently,
\begin{align}
  &\begin{multlined}
    \frac{1}{\mu}\sum_{m=1}^{\mu}e^{0,2}_{(m-1),\lambda}
    =
    (\lambda-\mu)\binom{\lambda}{\mu}
    \int_{y=-\infty}^{y=\infty}
    \phi(y)
    [1-\Phi(y)]^{\lambda-\mu-1}
    [\Phi(y)]^{\mu-1}
    (\Phi(y)-y \phi(y))
    \,\mathrm{d}y
  \end{multlined}\\
  &\begin{multlined}
    \phantom{\frac{1}{\mu}\sum_{m=1}^{\mu}e^{0,2}_{(m-1),\lambda}}
    =
    (\lambda-\mu)\binom{\lambda}{\mu}
    \int_{y=-\infty}^{y=\infty}
    \phi(y)
    [1-\Phi(y)]^{\lambda-\mu-1}
    [\Phi(y)]^{\mu}
    \,\mathrm{d}y\\
    \hspace{5cm}-
    \frac{(\lambda-\mu)}{2\pi}
    \binom{\lambda}{\mu}
    \int_{y=-\infty}^{y=\infty}
    y
    e^{-y^2}
    [1-\Phi(y)]^{\lambda-\mu-1}
    [\Phi(y)]^{\mu-1}
    \,\mathrm{d}y
  \end{multlined}
\end{align}
follows. For the first summand, the same arguments leading to
\Cref{sec:theoreticalanalysis:eq:Eqcentroidfeasaddend}
reveal that it is the integration over
the density of the $(\mu+1)$-th order statistic
of independently standard normally distributed variables.
Hence, it integrates to $1$. By substituting $s:=-y$,
the second summand is recognized as
one of the generalized progress coefficients
(\Cref{chapter:analysis_repairbyprojection_multirecombinative:sec:theoreticalanalysis:eq:eabml}).
Thus, one gets
\begin{equation}
  \frac{1}{\mu}\sum_{m=1}^{\mu}e^{0,2}_{(m-1),\lambda}
  =
  1+e_{\mu,\lambda}^{1,1}.
\end{equation}
Analogously,
\begin{equation}
  \frac{1}{\mu}\sum_{m=1}^{\mu}e^{0,1}_{(m-1),\lambda}
  =
  e_{\mu,\lambda}^{1,0}=c_{\mu/\mu,\lambda}
\end{equation}
can be derived.
Reinsertion of these results into
\Cref{sec:appendix:derivation_expectation_qrsquarecentroid:eq:squarecentroid}
yields
\begin{align}
  &\begin{multlined}
  \frac{1}{\mu}\sum_{m=1}^{\mu}\mathrm{E}[{q}^2_{m;\lambda}]
  \approx
  P_{\text{feas}}(x^{(g)}, r^{(g)}, \sigma^{(g)})
  \left({\sigma^{(g)}}^2
  \left[1+e_{\mu,\lambda}^{1,1}\right]
  - 2 \sigma^{(g)} x^{(g)}
  c_{\mu/\mu,\lambda}
  + {x^{(g)}}^2\right)\\
  + [1 - P_{\text{feas}}(x^{(g)}, r^{(g)}, \sigma^{(g)})]
  \bigg(\frac{({\sigma^{(g)}}^2+\sigma_r^2/\xi)}{(1+1/\xi)^2}
  \left[1+e_{\mu,\lambda}^{1,1}\right]\\
  -\frac{\sqrt{{\sigma^{(g)}}^2+\sigma_r^2/\xi}}{(1+1/\xi)^2}2(x^{(g)}+\bar{r}/\sqrt{\xi})
  c_{\mu/\mu,\lambda}
  +\frac{(x^{(g)}+\bar{r}/\sqrt{\xi})^2}{(1+1/\xi)^2}\bigg)
  \end{multlined}
  \label{sec:appendix:derivation_expectation_qrsquarecentroid:eq:squarecentroidinserted}
\end{align}

\newpage

\section{Additional Plots Comparing Derived Approximations with
  Experiments}
\label{sec:algorithm:subsec:additionalplots}
\begin{figure}
  \centering
  \begin{tabular}{@{\hspace{-0.025\textwidth}}c@{\hspace{-0.025\textwidth}}c}
    \includegraphics[width=0.5\textwidth]{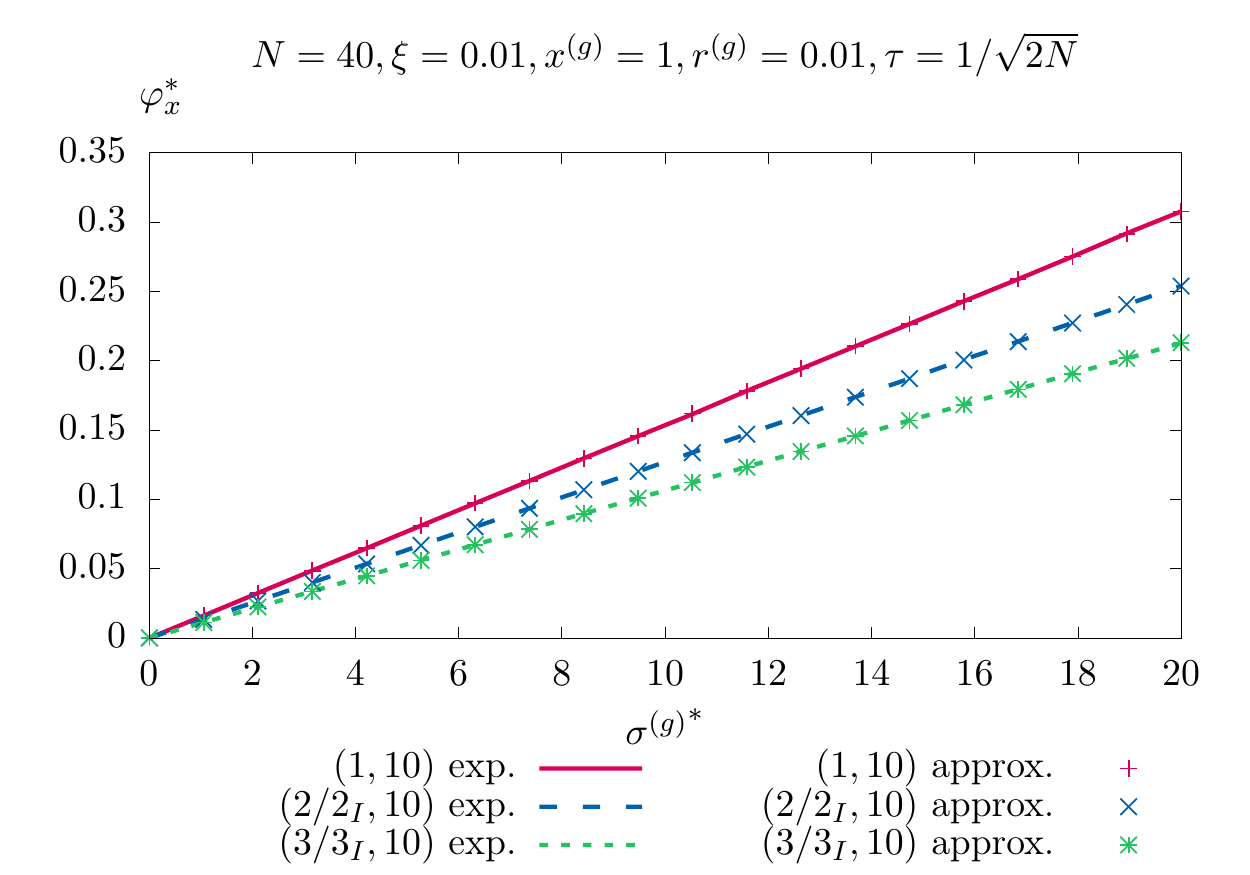}&
    \includegraphics[width=0.5\textwidth]{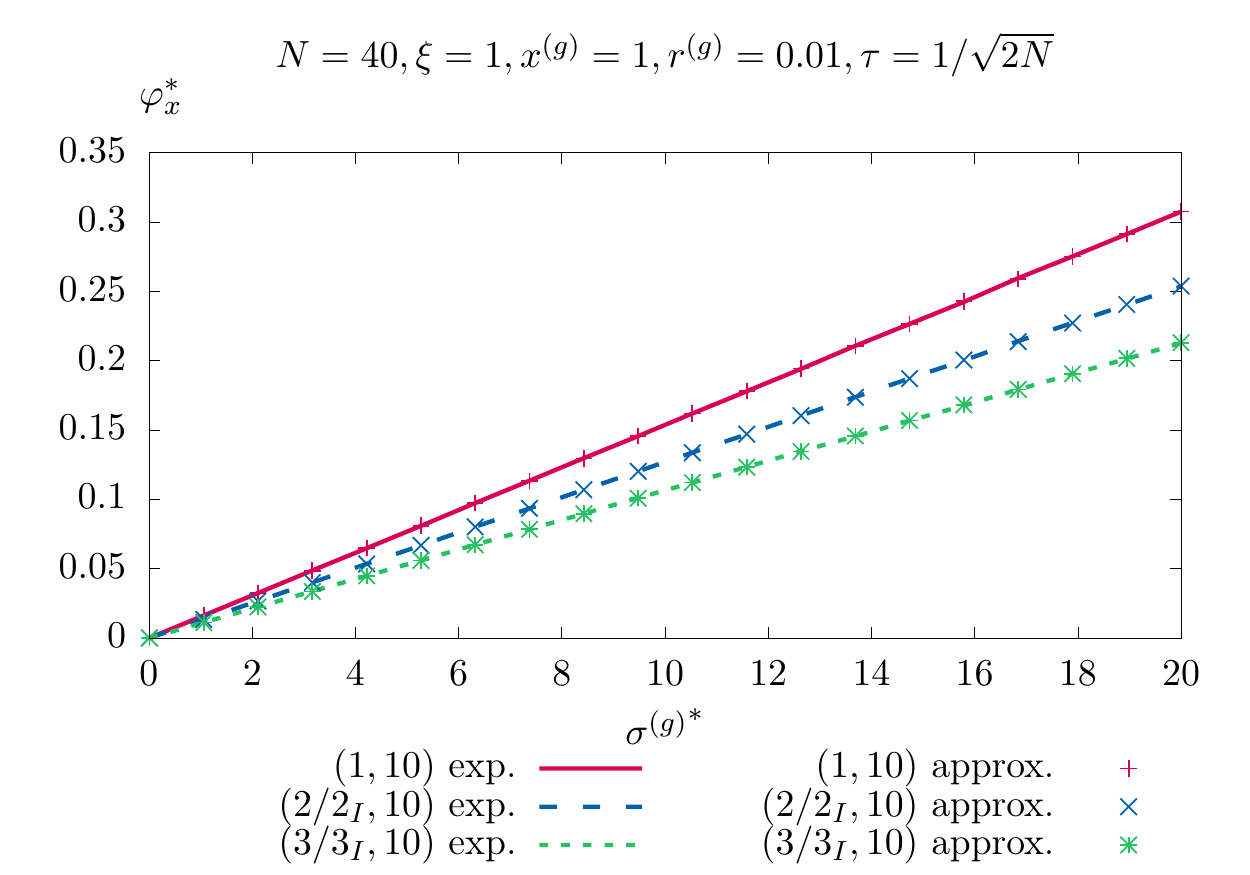}\\
    \includegraphics[width=0.5\textwidth]{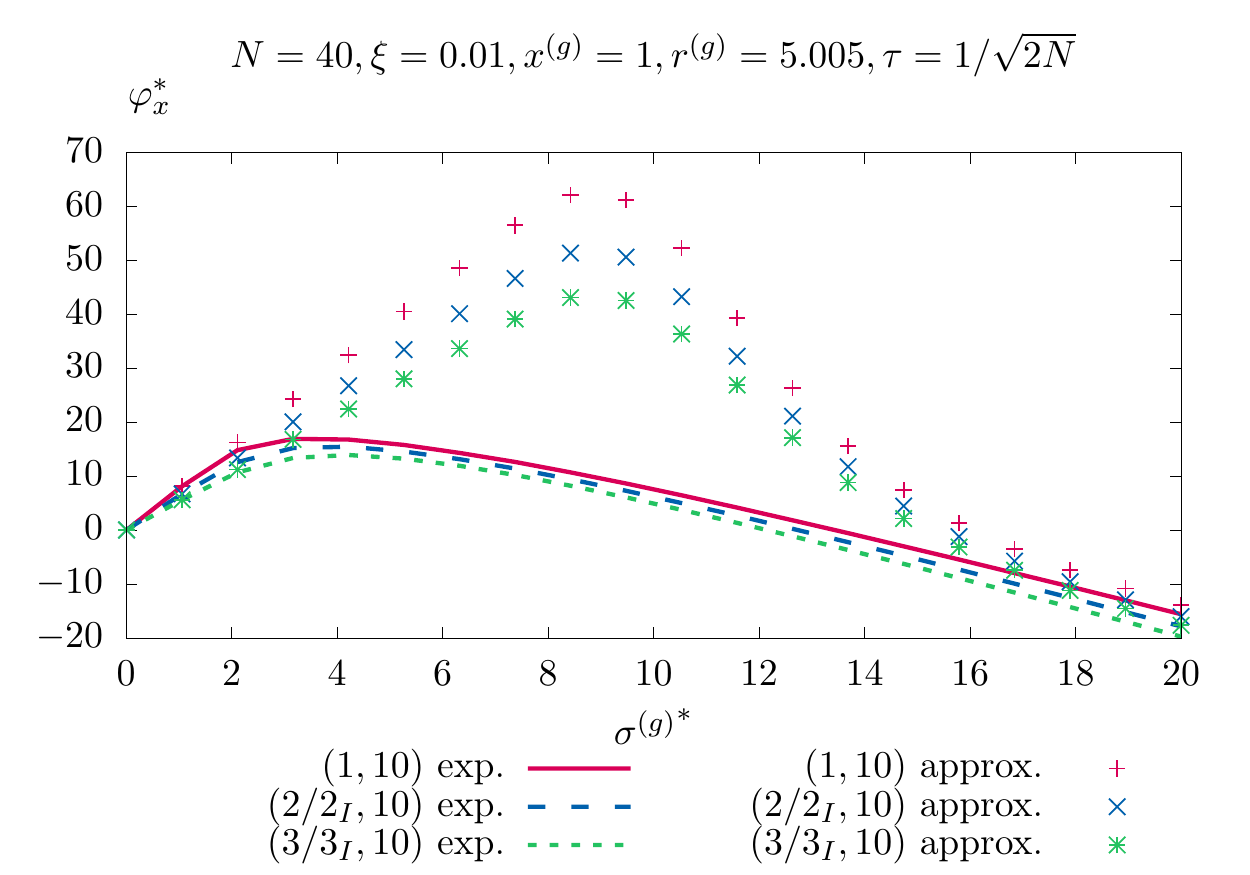}&
    \includegraphics[width=0.5\textwidth]{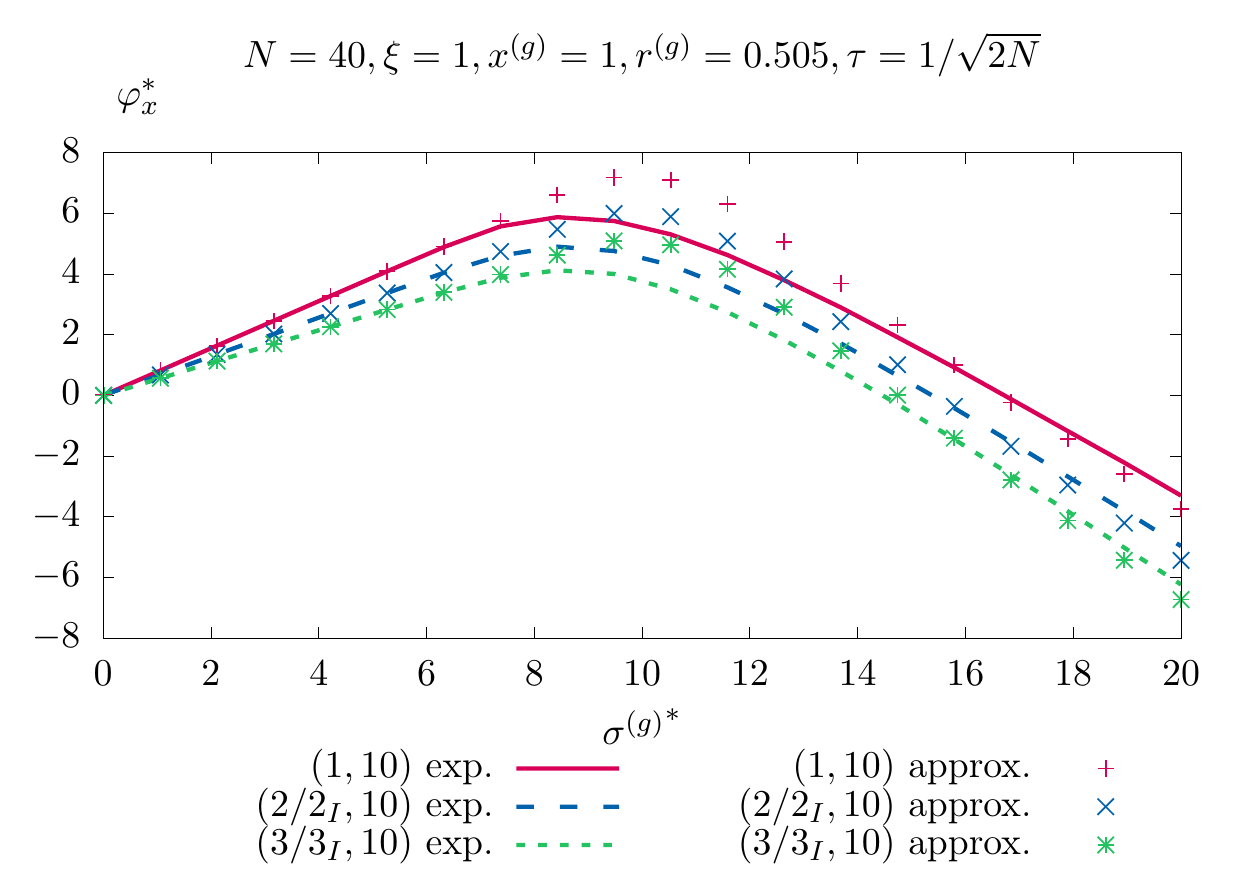}\\
    \includegraphics[width=0.5\textwidth]{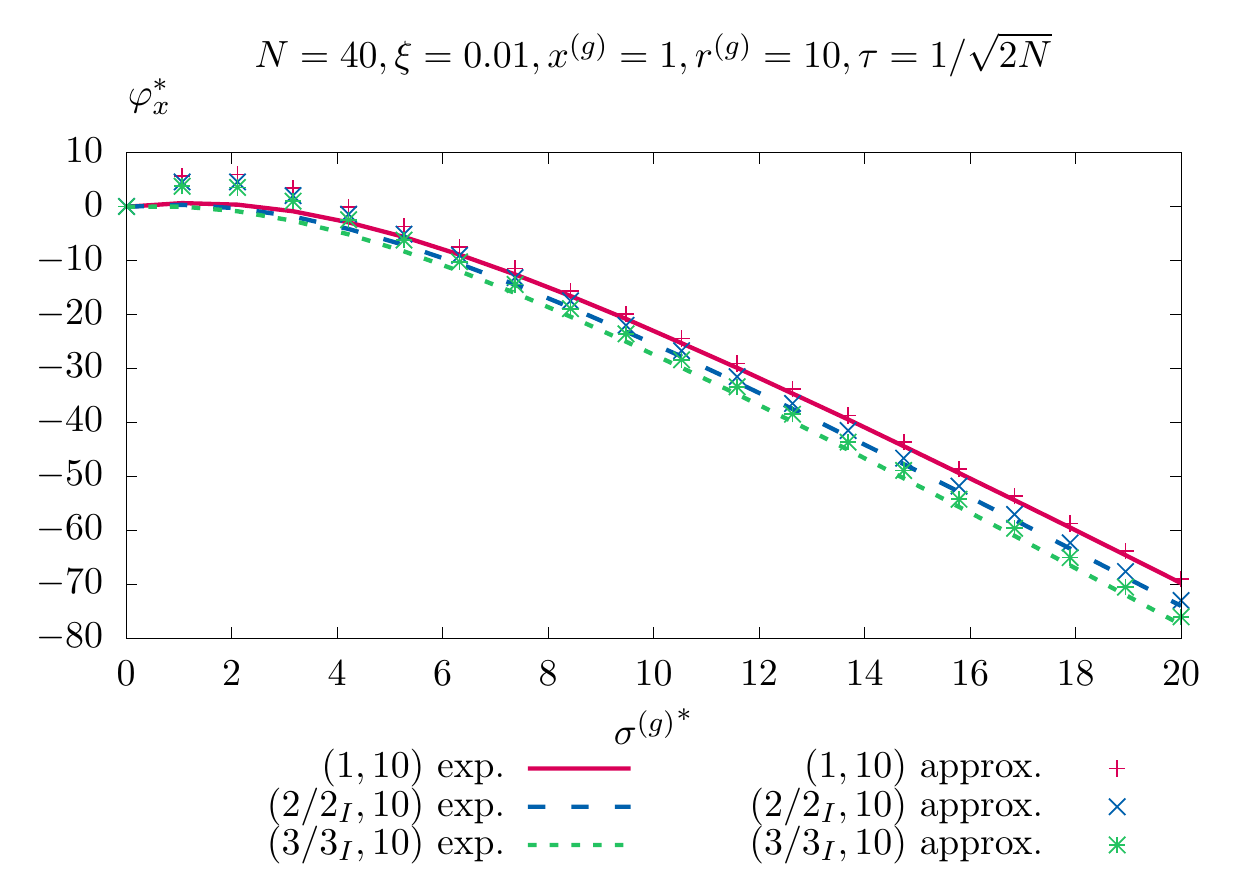}&
    \includegraphics[width=0.5\textwidth]{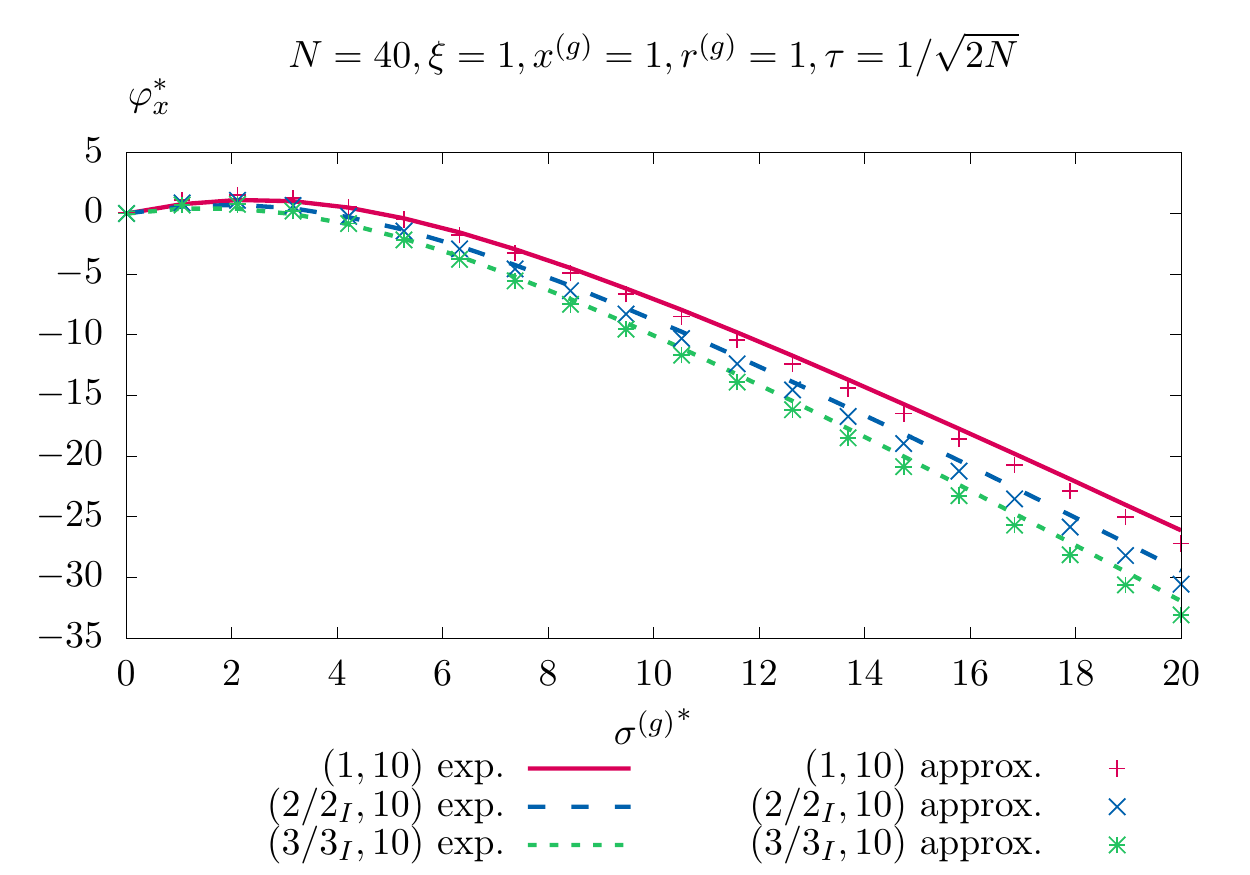}\\
  \end{tabular}
  \caption{Comparison of the $x$ progress rate approximation with simulations. (Part 1)}
  \label{sec:theoreticalanalysis:fig:xprogresscomparisonsdetail1}
\end{figure}

\begin{figure}
  \centering
  \begin{tabular}{@{\hspace{-0.025\textwidth}}c@{\hspace{-0.025\textwidth}}c}
    \includegraphics[width=0.5\textwidth]{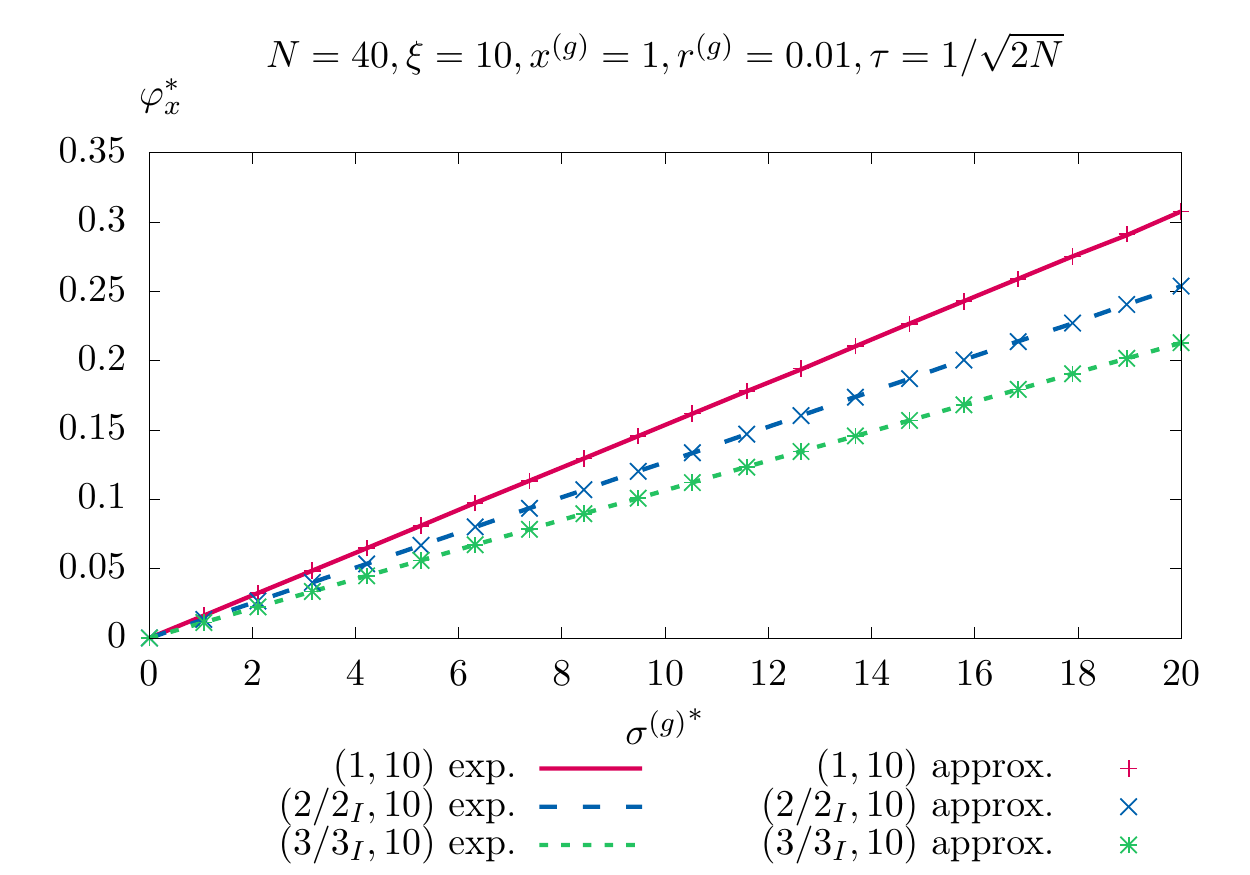}&
    \includegraphics[width=0.5\textwidth]{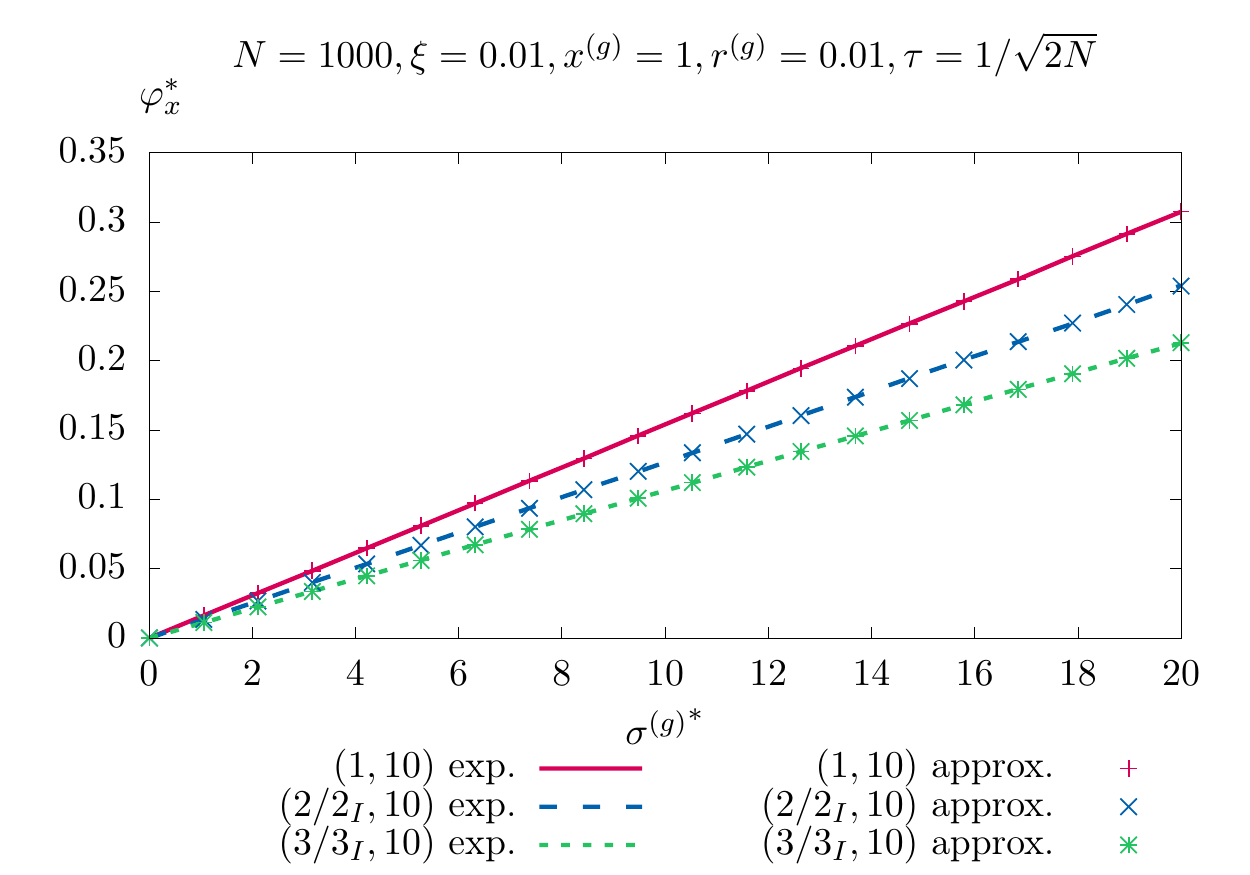}\\
    \includegraphics[width=0.5\textwidth]{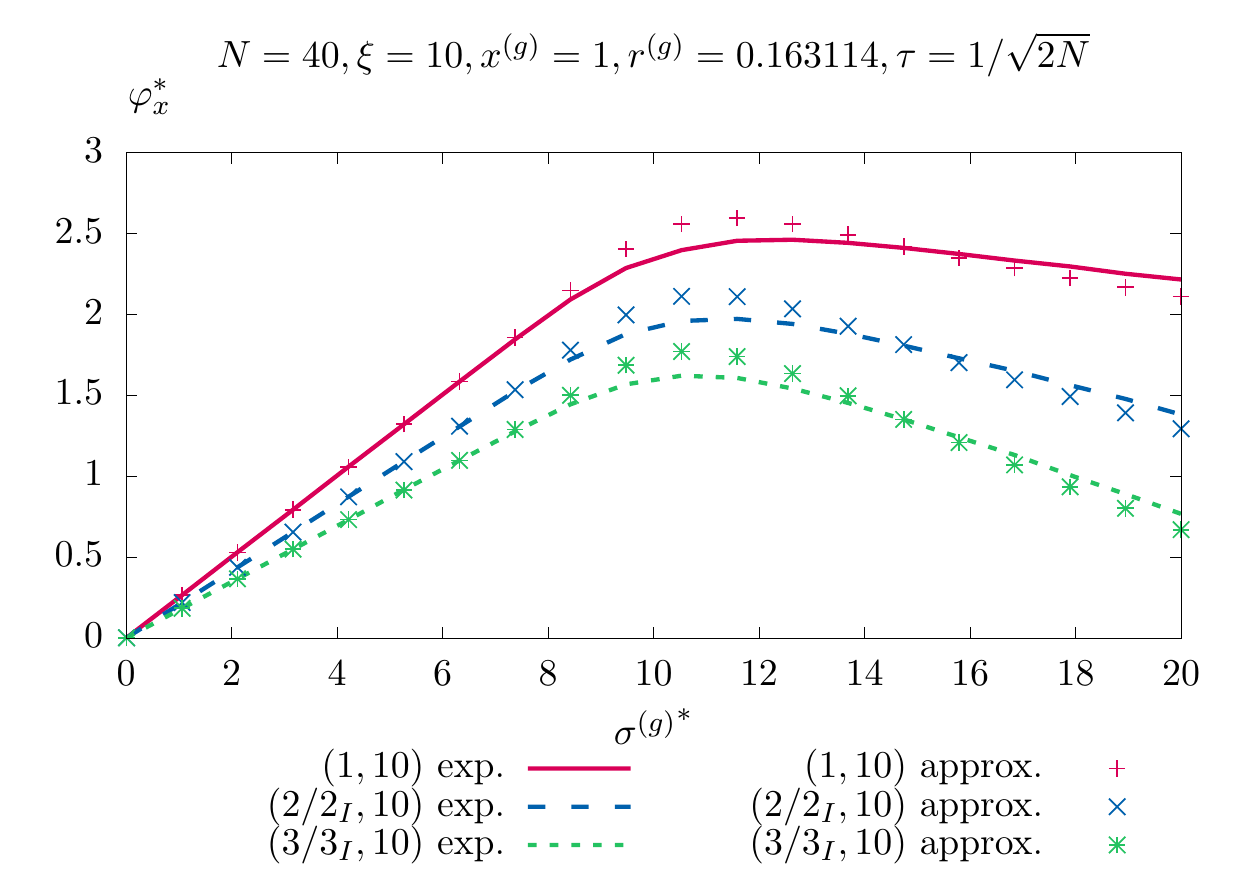}&
    \includegraphics[width=0.5\textwidth]{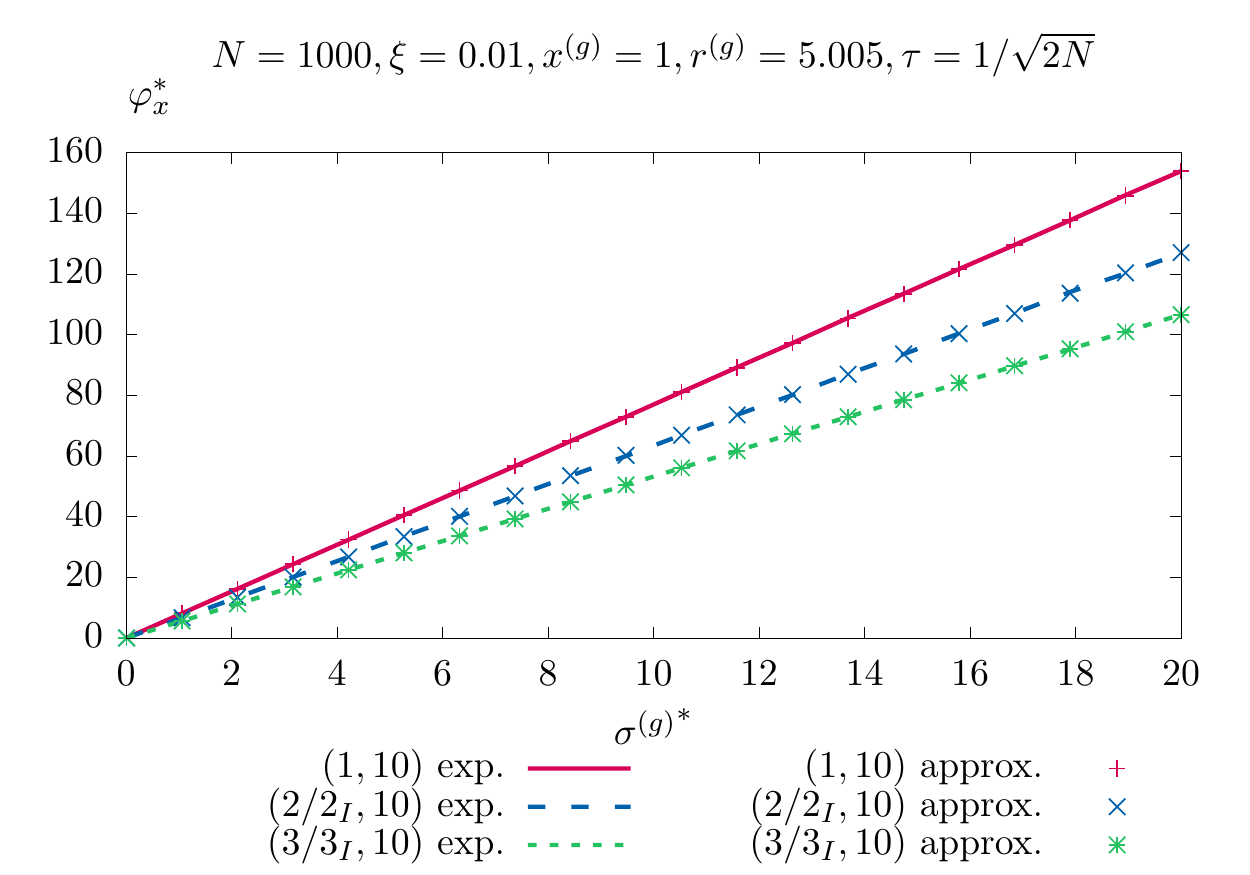}\\
    \includegraphics[width=0.5\textwidth]{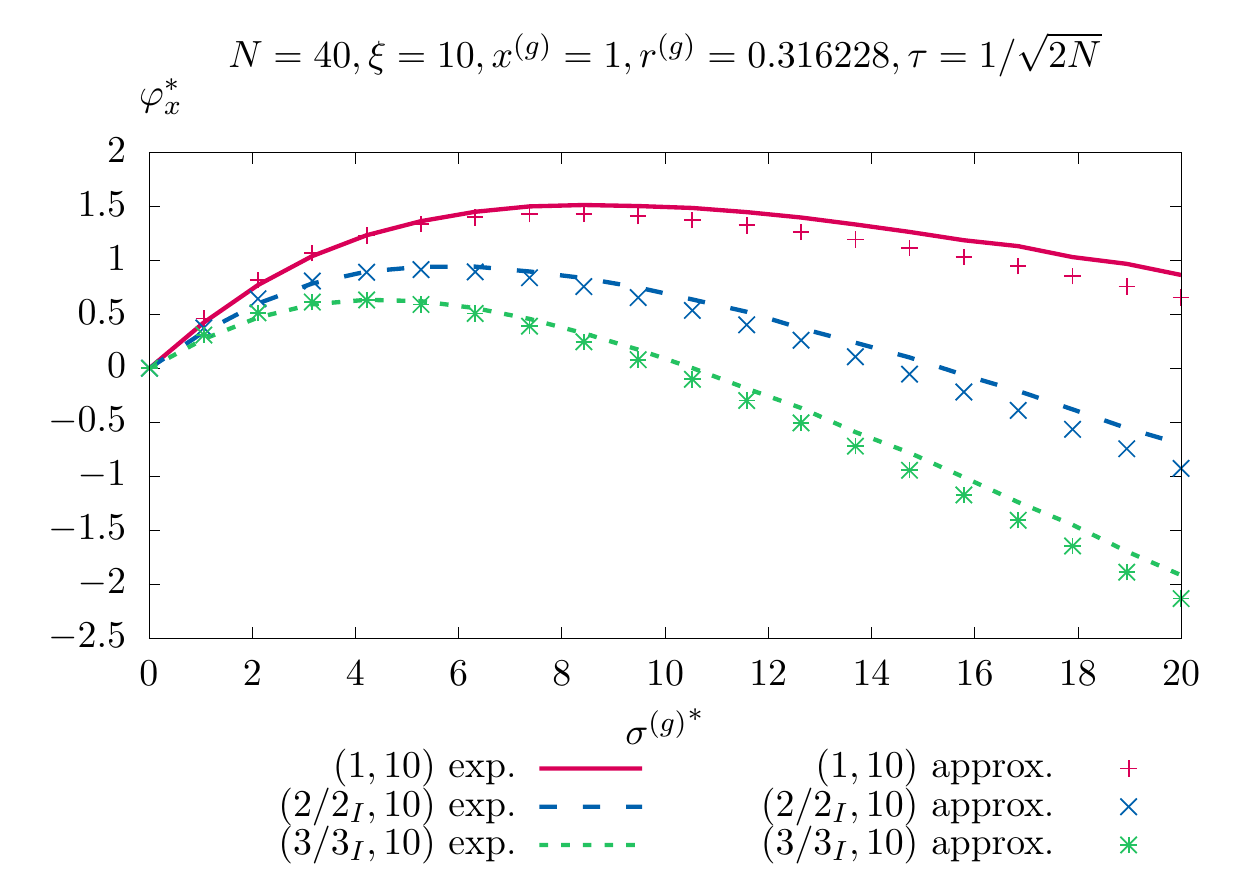}&
    \includegraphics[width=0.5\textwidth]{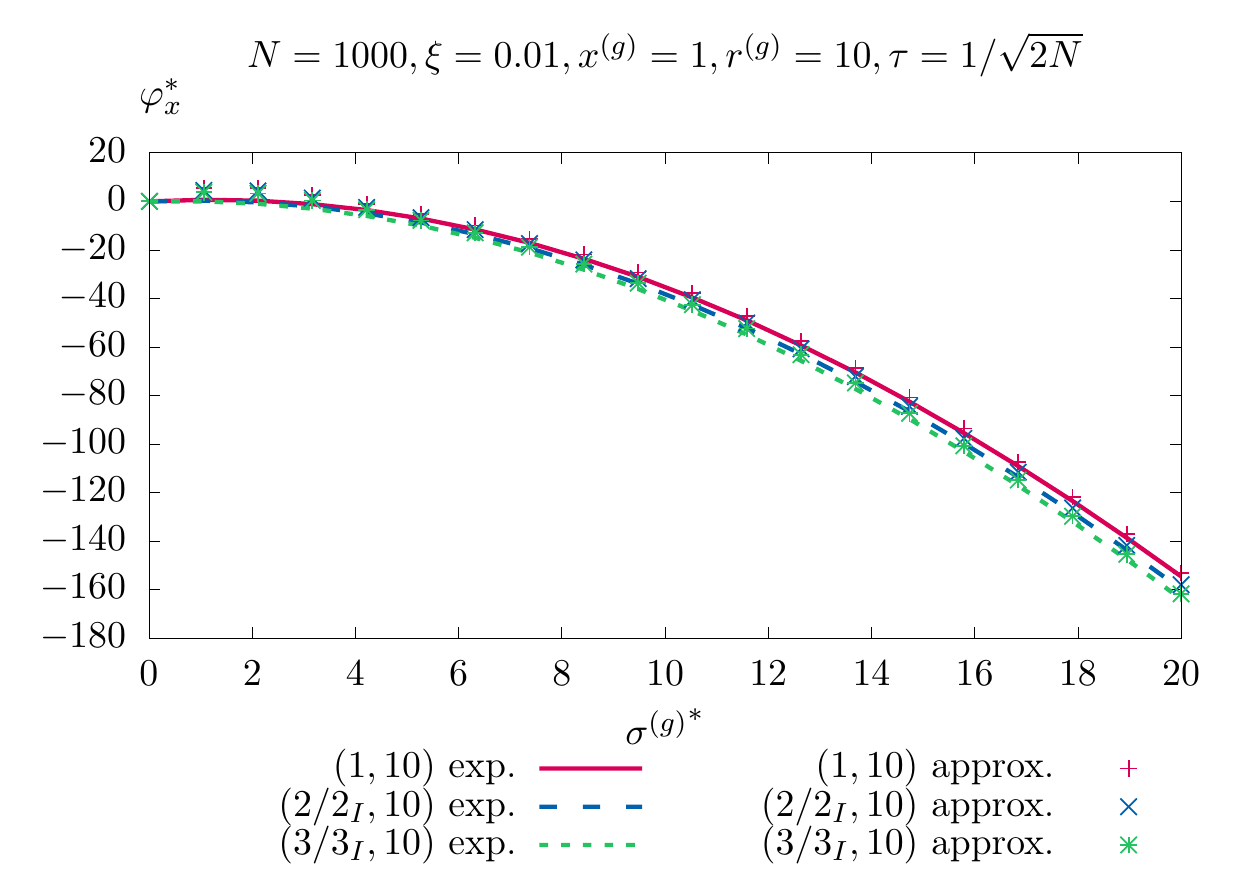}\\
  \end{tabular}
  \caption{Comparison of the $x$ progress rate approximation with simulations. (Part 2)}
  \label{sec:theoreticalanalysis:fig:xprogresscomparisonsdetail2}
\end{figure}

\begin{figure}
  \centering
  \begin{tabular}{@{\hspace{-0.025\textwidth}}c@{\hspace{-0.025\textwidth}}c}
    \includegraphics[width=0.5\textwidth]{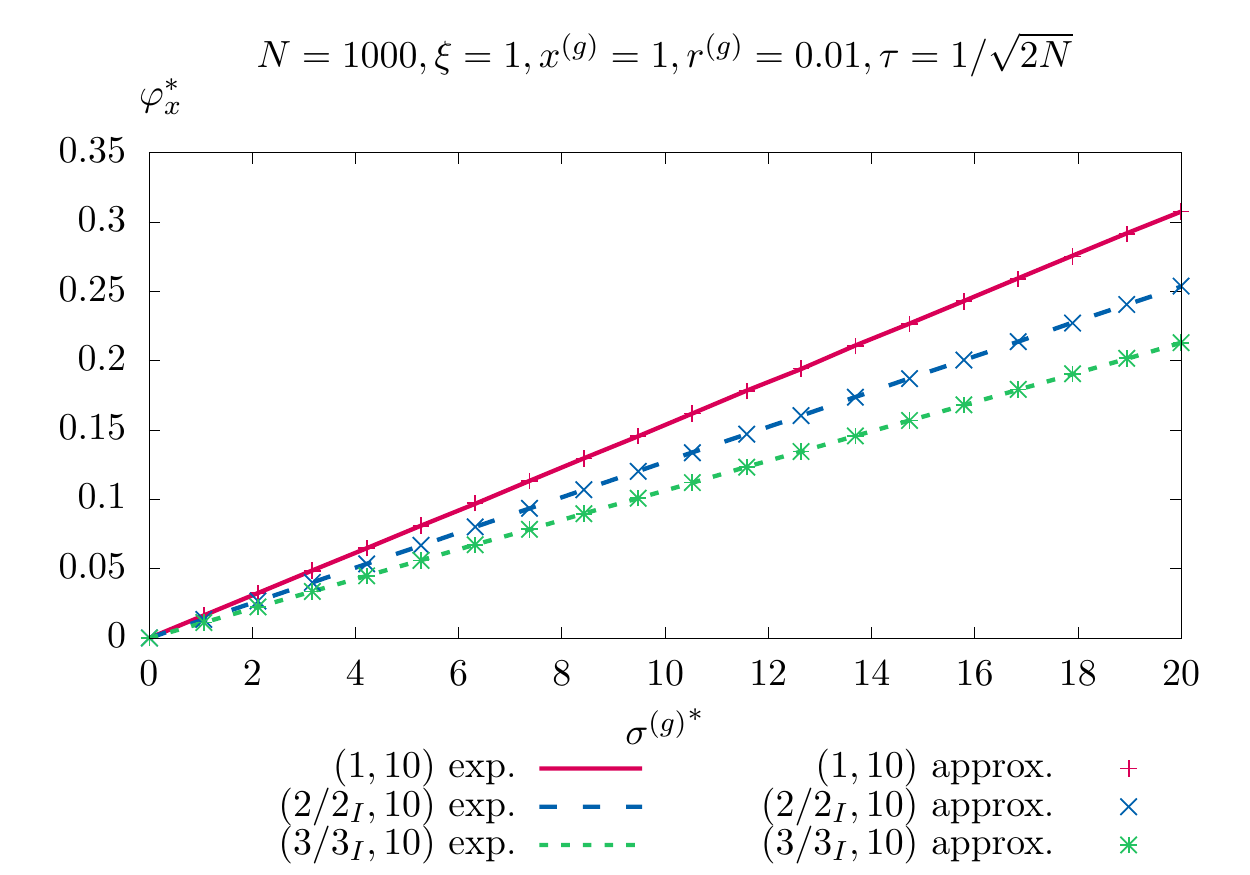}&
    \includegraphics[width=0.5\textwidth]{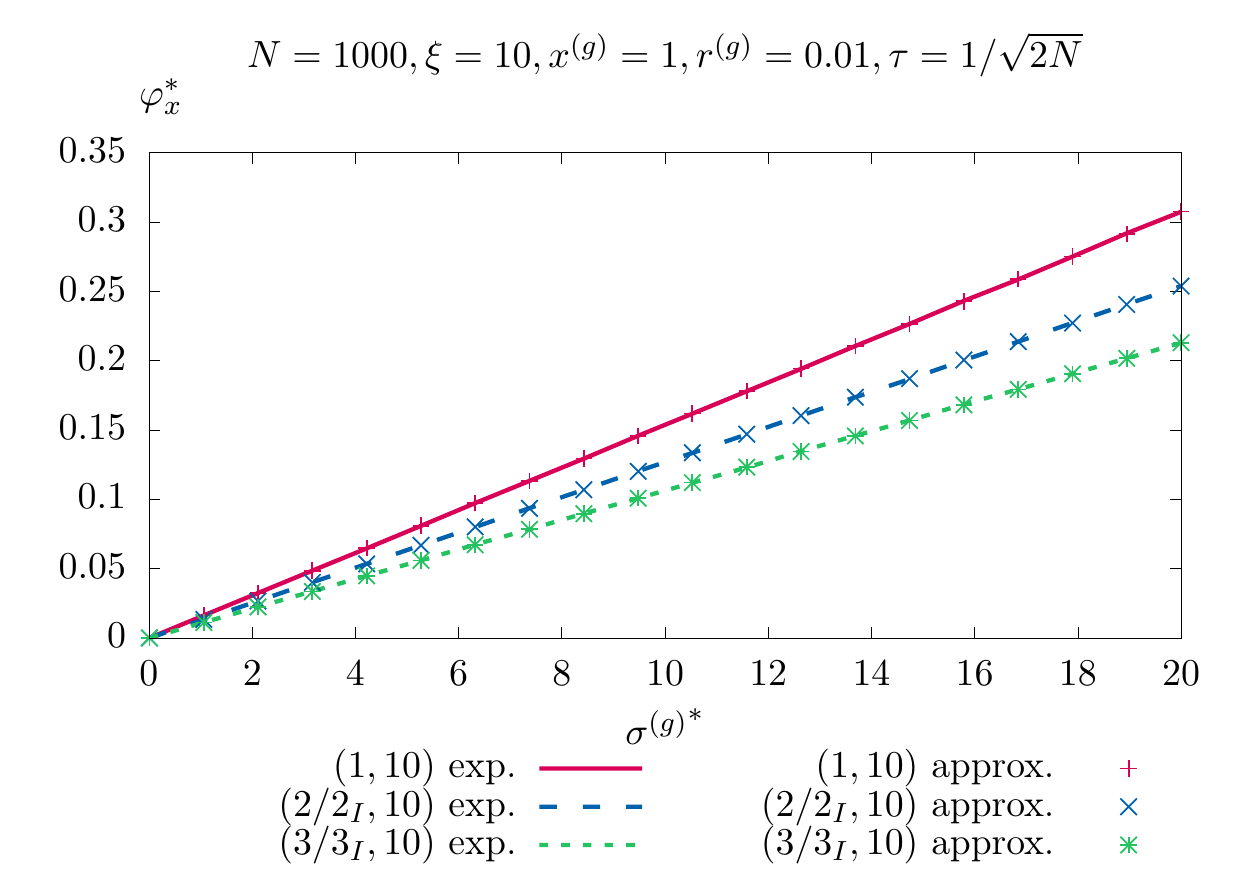}\\
    \includegraphics[width=0.5\textwidth]{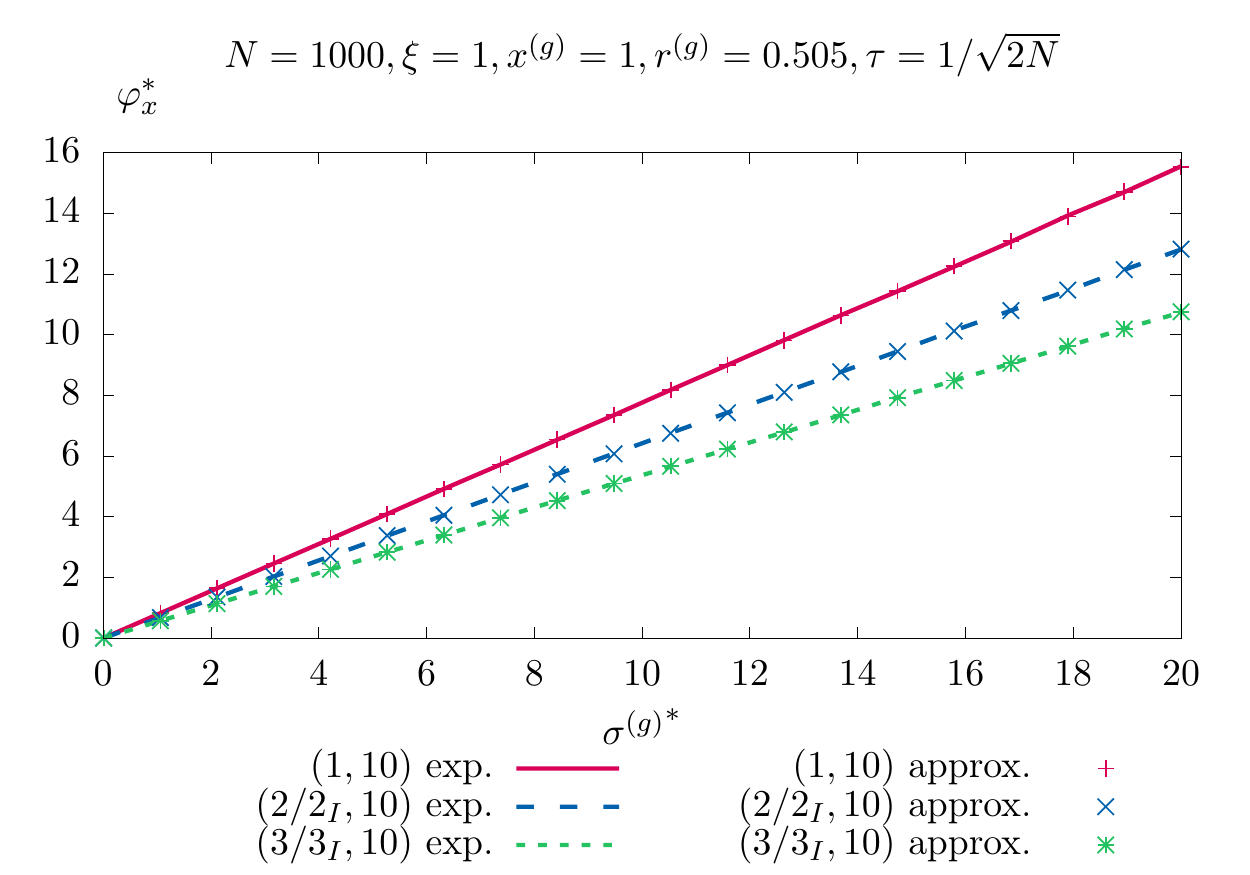}&
    \includegraphics[width=0.5\textwidth]{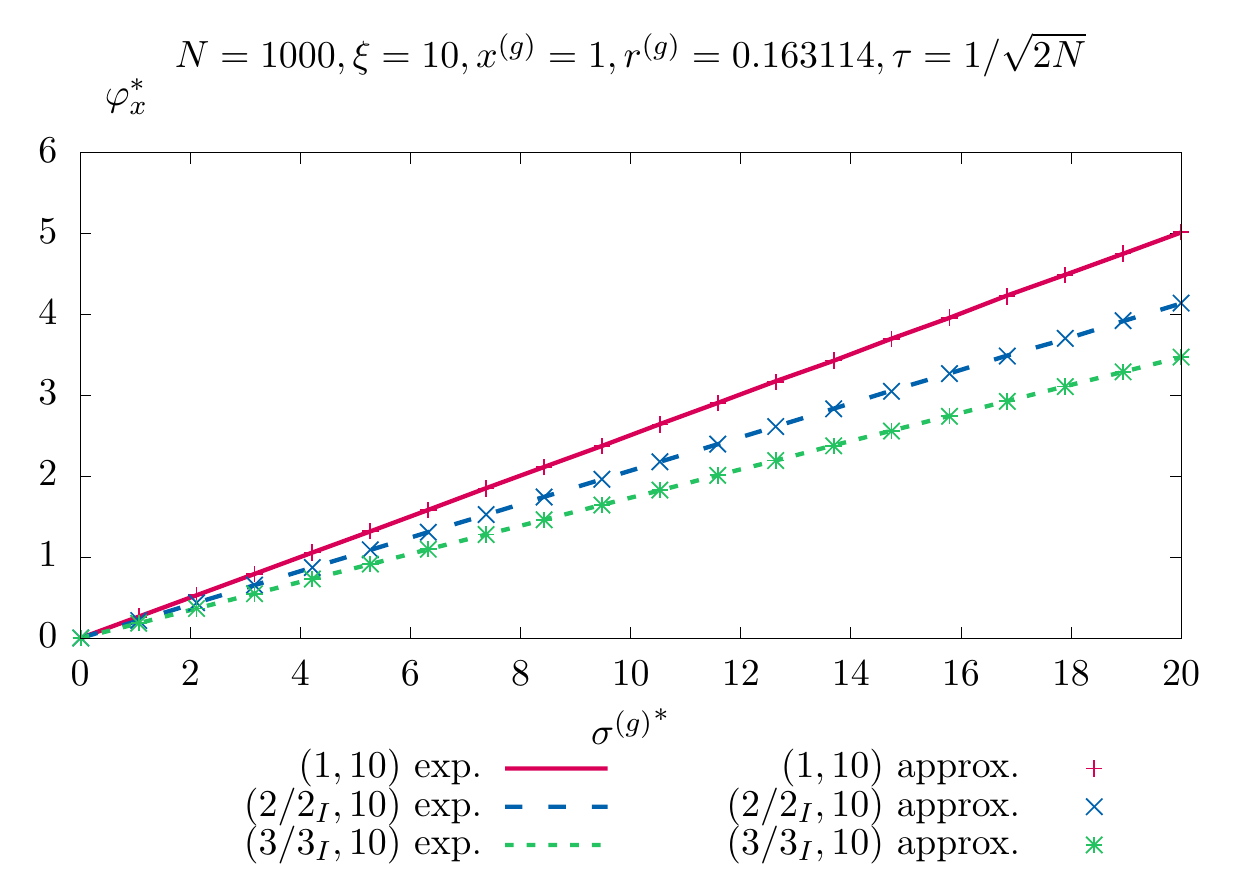}\\
    \includegraphics[width=0.5\textwidth]{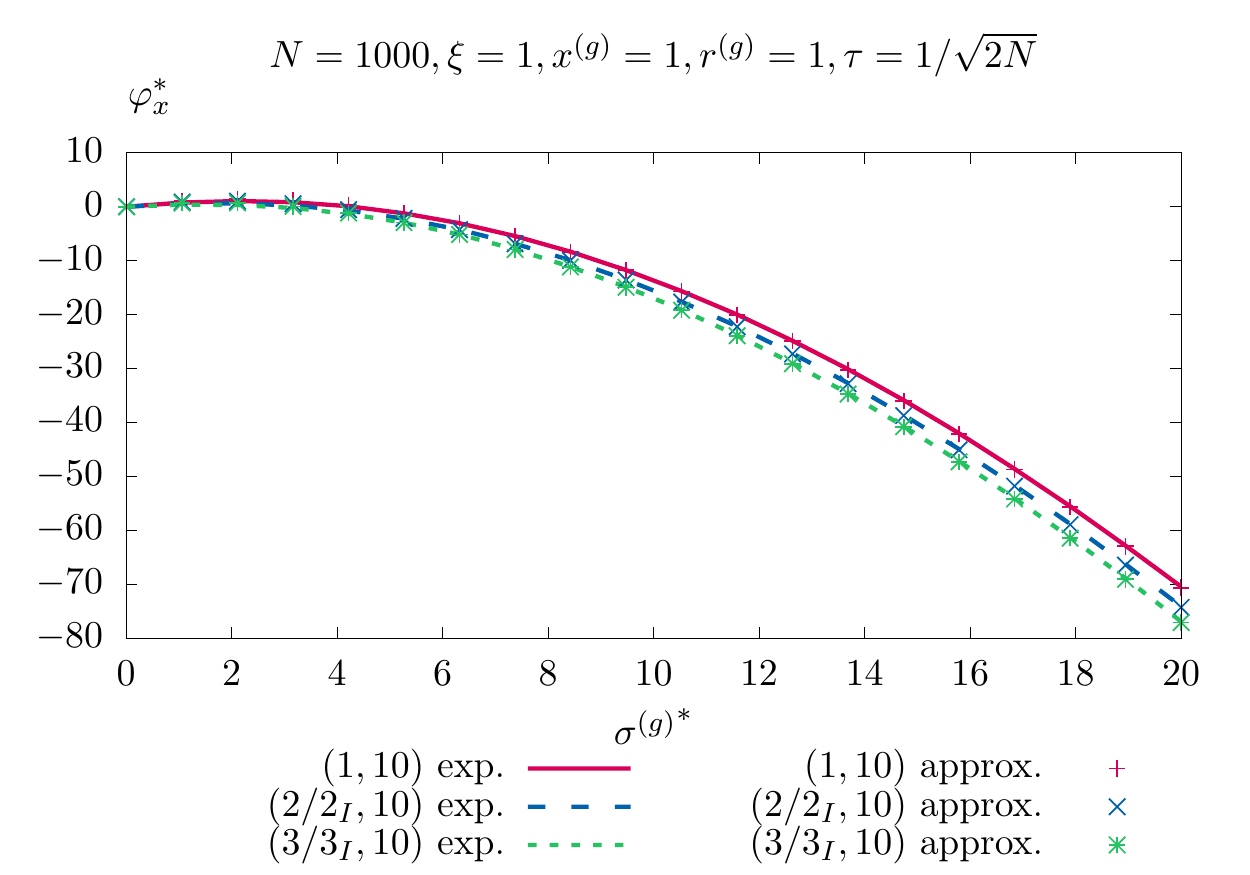}&
    \includegraphics[width=0.5\textwidth]{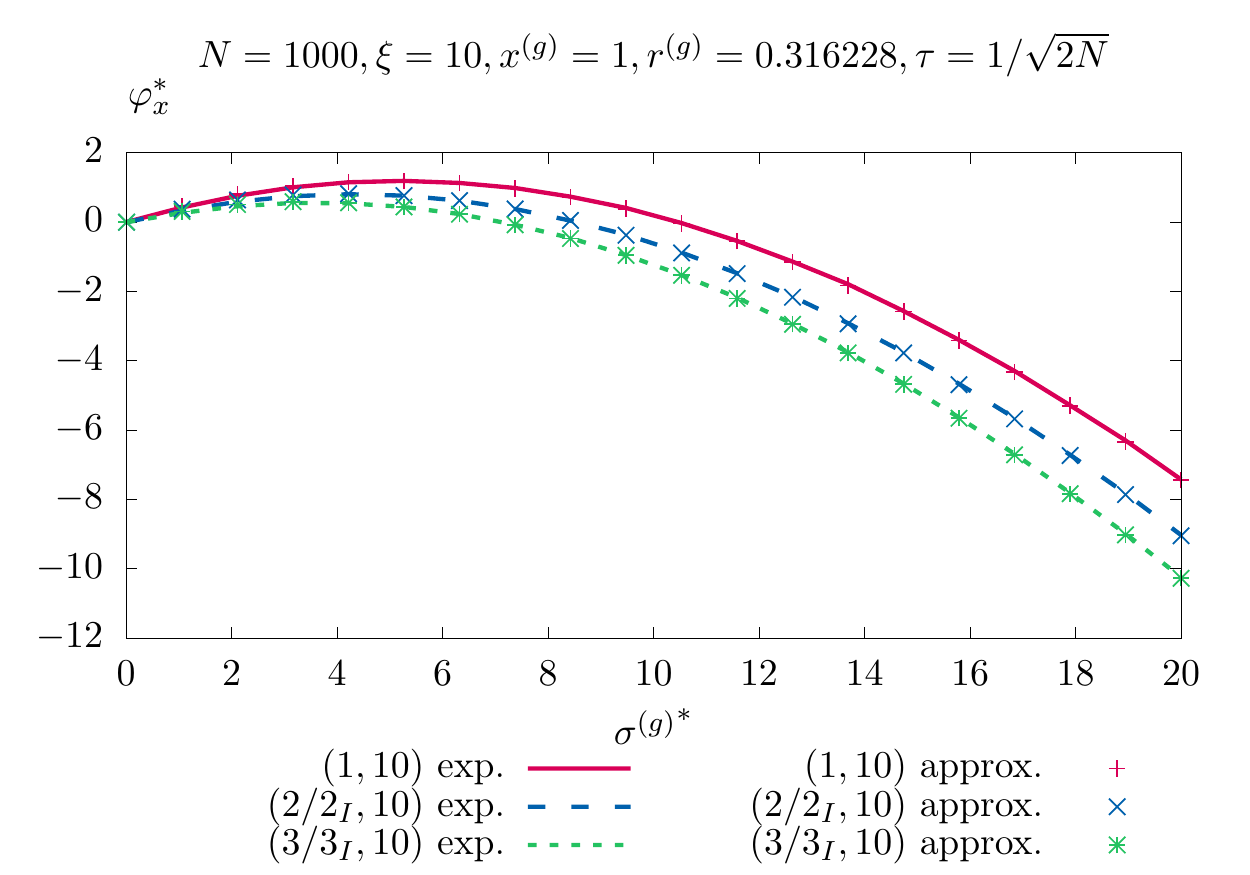}\\
  \end{tabular}
  \caption{Comparison of the $x$ progress rate approximation with simulations. (Part 3)}
  \label{sec:theoreticalanalysis:fig:xprogresscomparisonsdetail3}
\end{figure}

\begin{figure}
  \centering
  \begin{tabular}{@{\hspace{-0.025\textwidth}}c@{\hspace{-0.025\textwidth}}c}
    \includegraphics[width=0.5\textwidth]{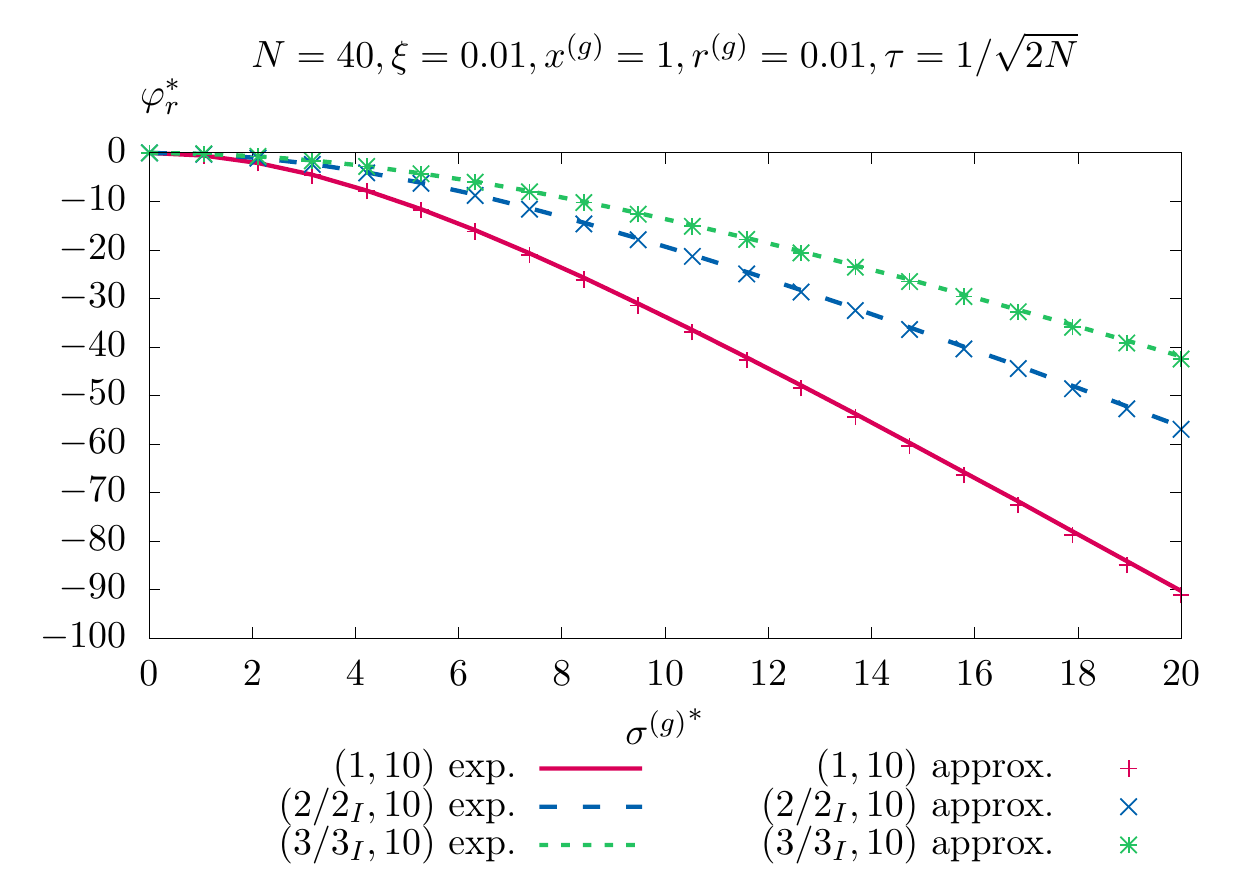}&
    \includegraphics[width=0.5\textwidth]{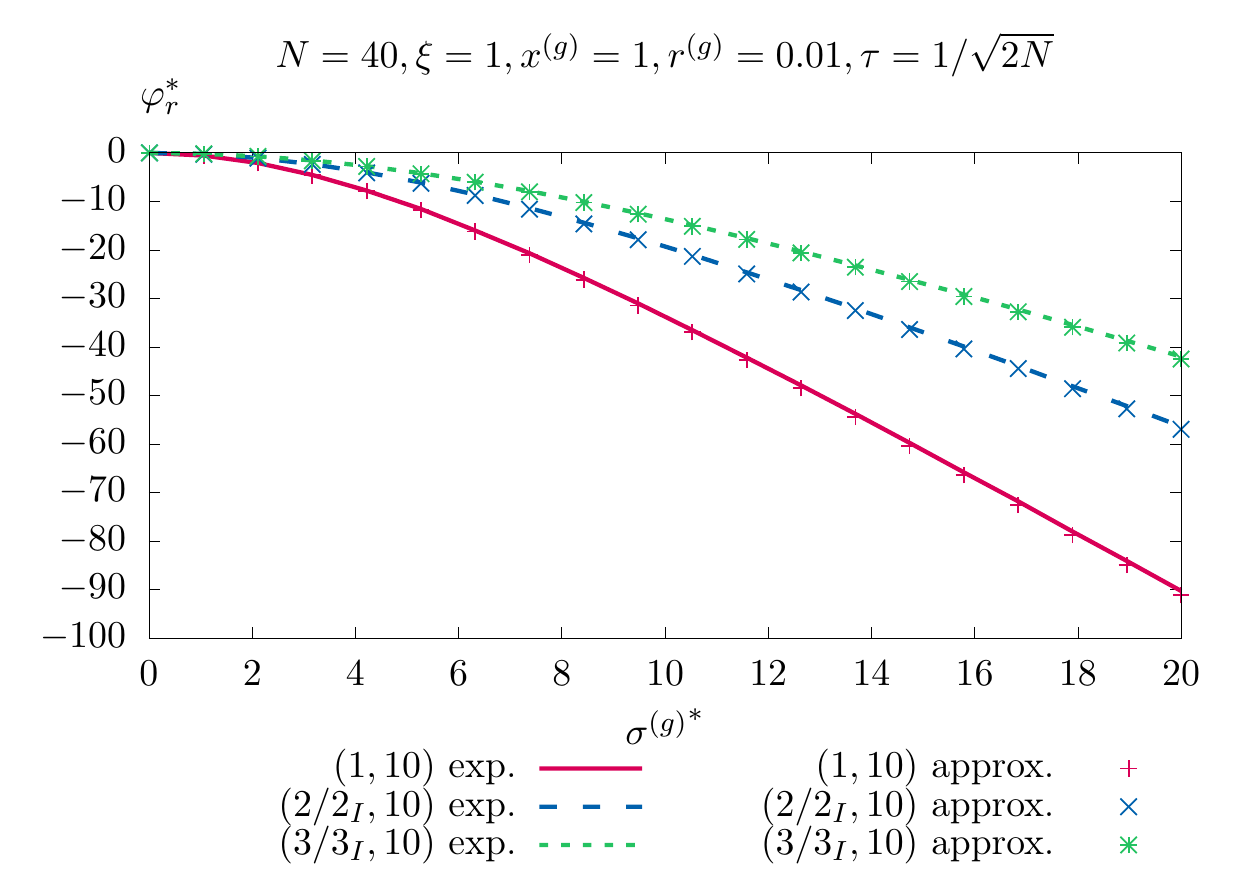}\\
    \includegraphics[width=0.5\textwidth]{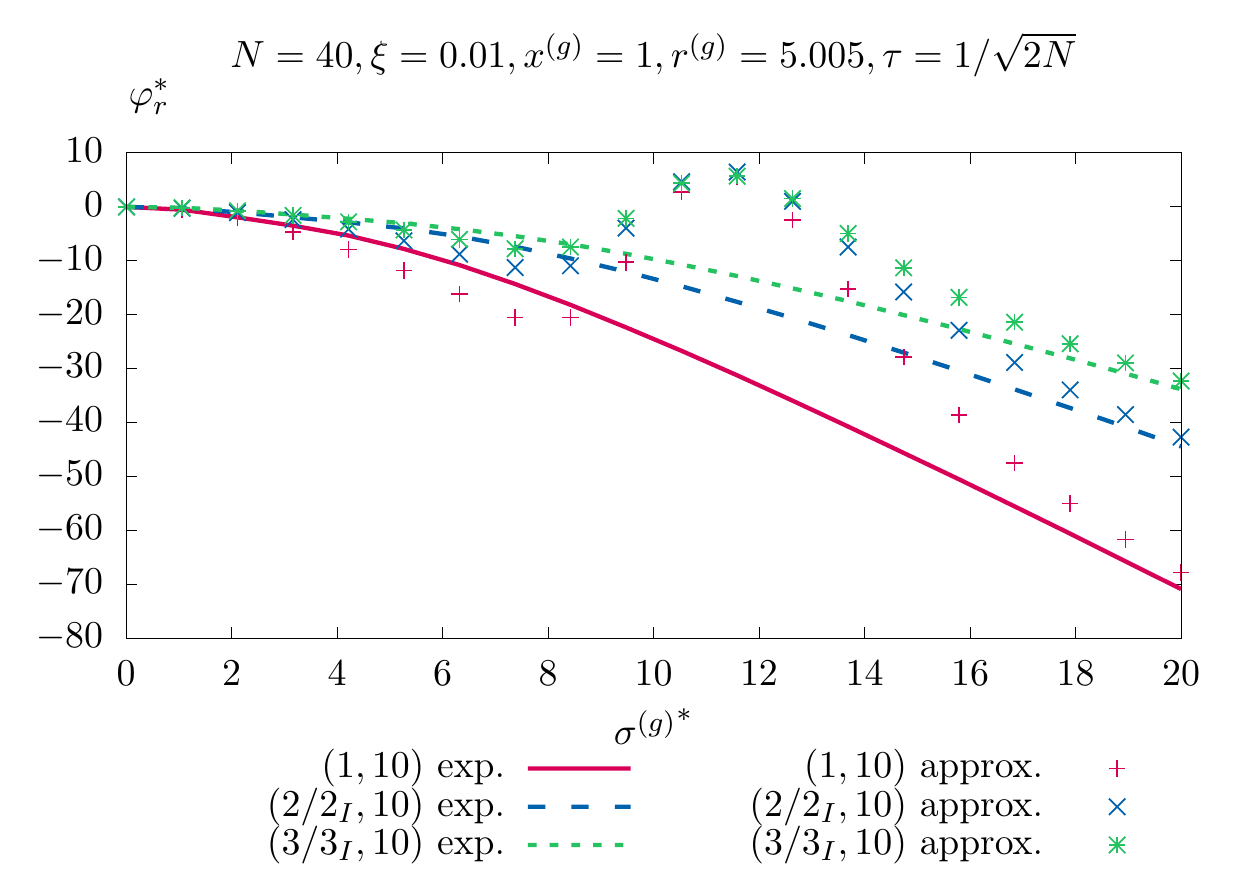}&
    \includegraphics[width=0.5\textwidth]{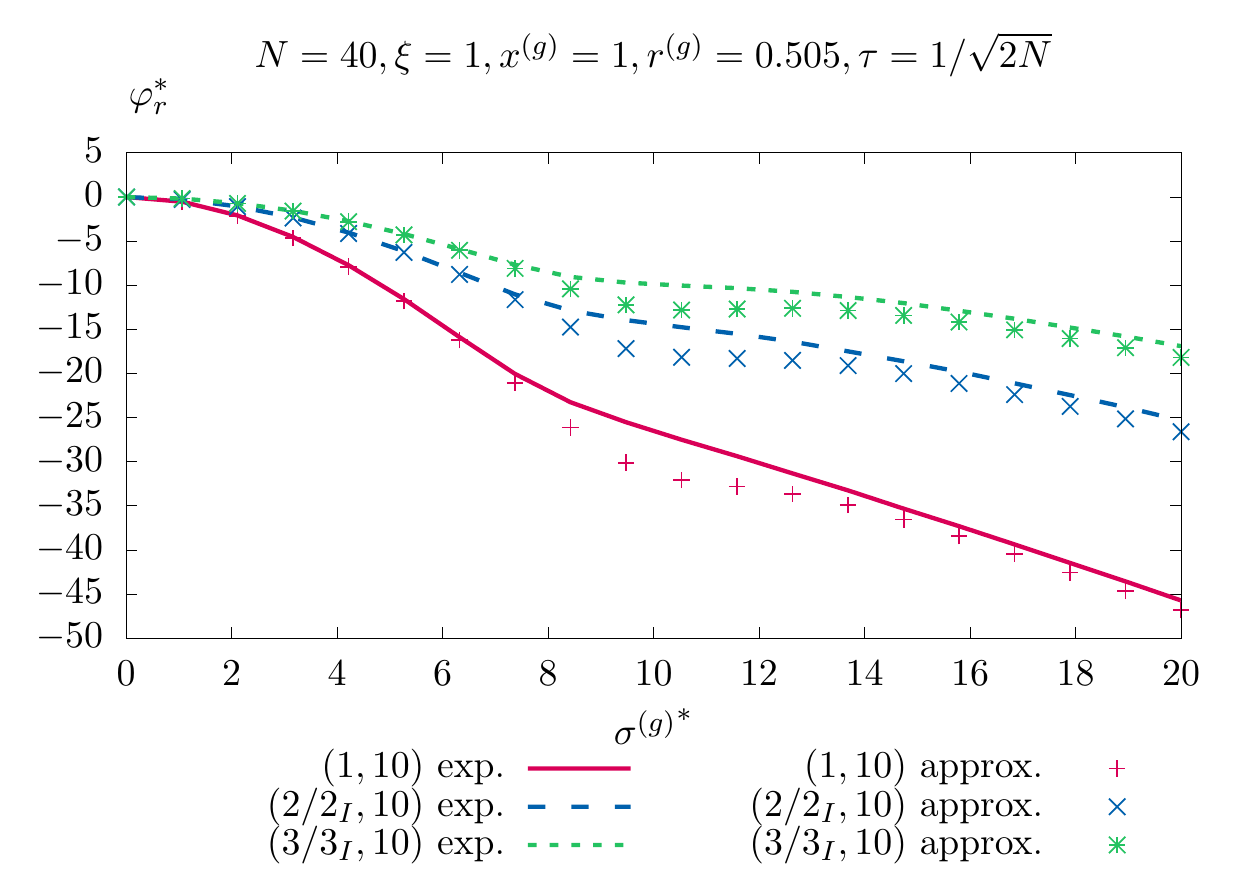}\\
    \includegraphics[width=0.5\textwidth]{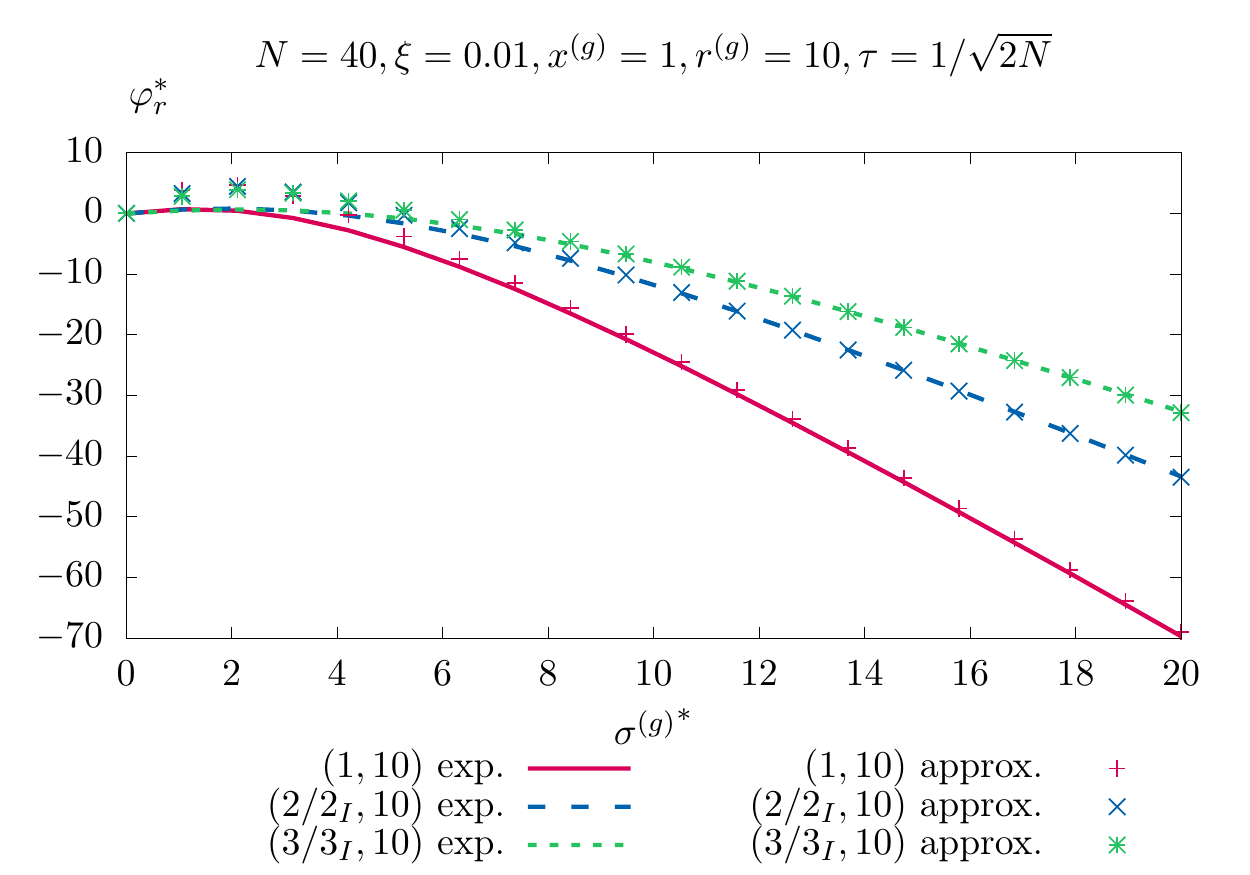}&
    \includegraphics[width=0.5\textwidth]{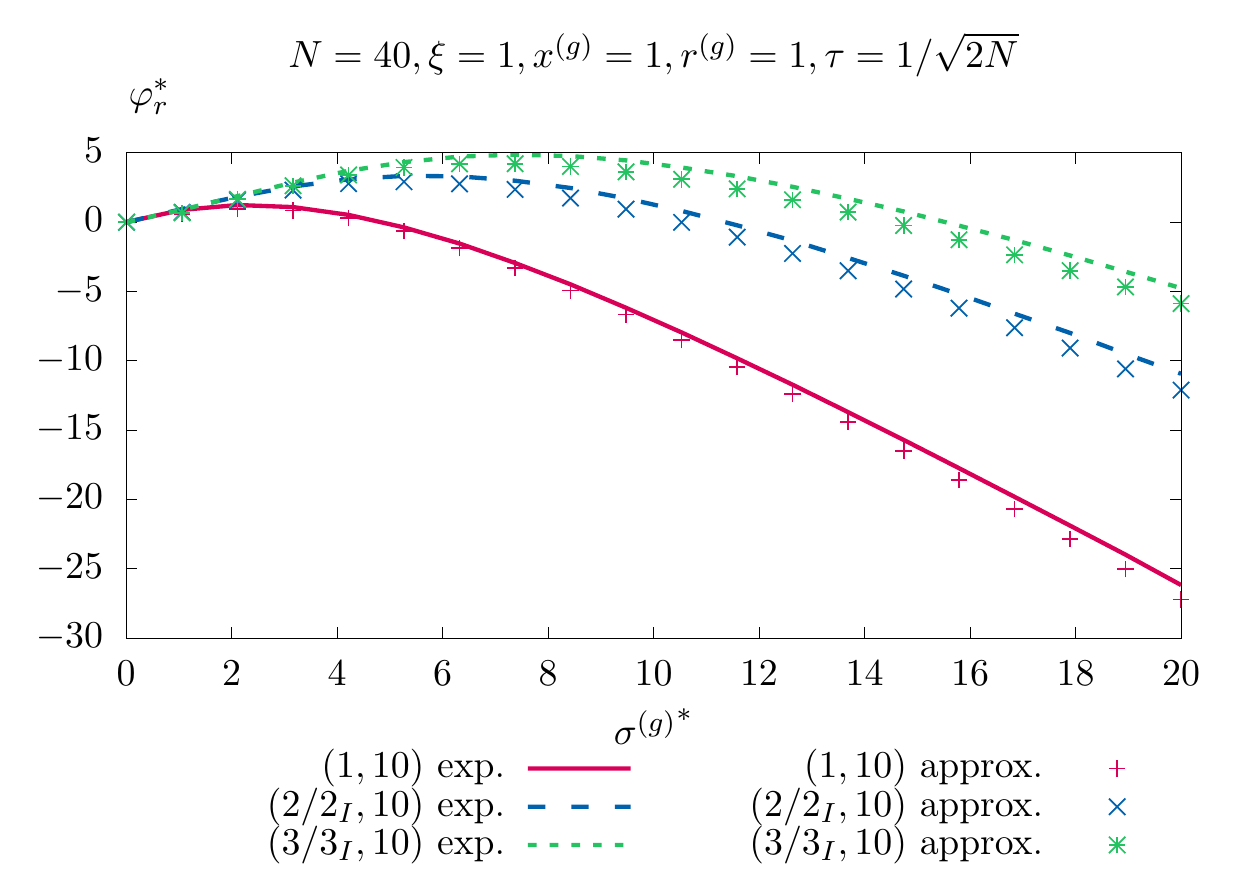}\\
  \end{tabular}
  \caption{Comparison of the $r$ progress rate approximation with simulations. (Part 1)}
  \label{sec:theoreticalanalysis:fig:rprogresscomparisonsdetail1}
\end{figure}

\begin{figure}
  \centering
  \begin{tabular}{@{\hspace{-0.025\textwidth}}c@{\hspace{-0.025\textwidth}}c}
    \includegraphics[width=0.5\textwidth]{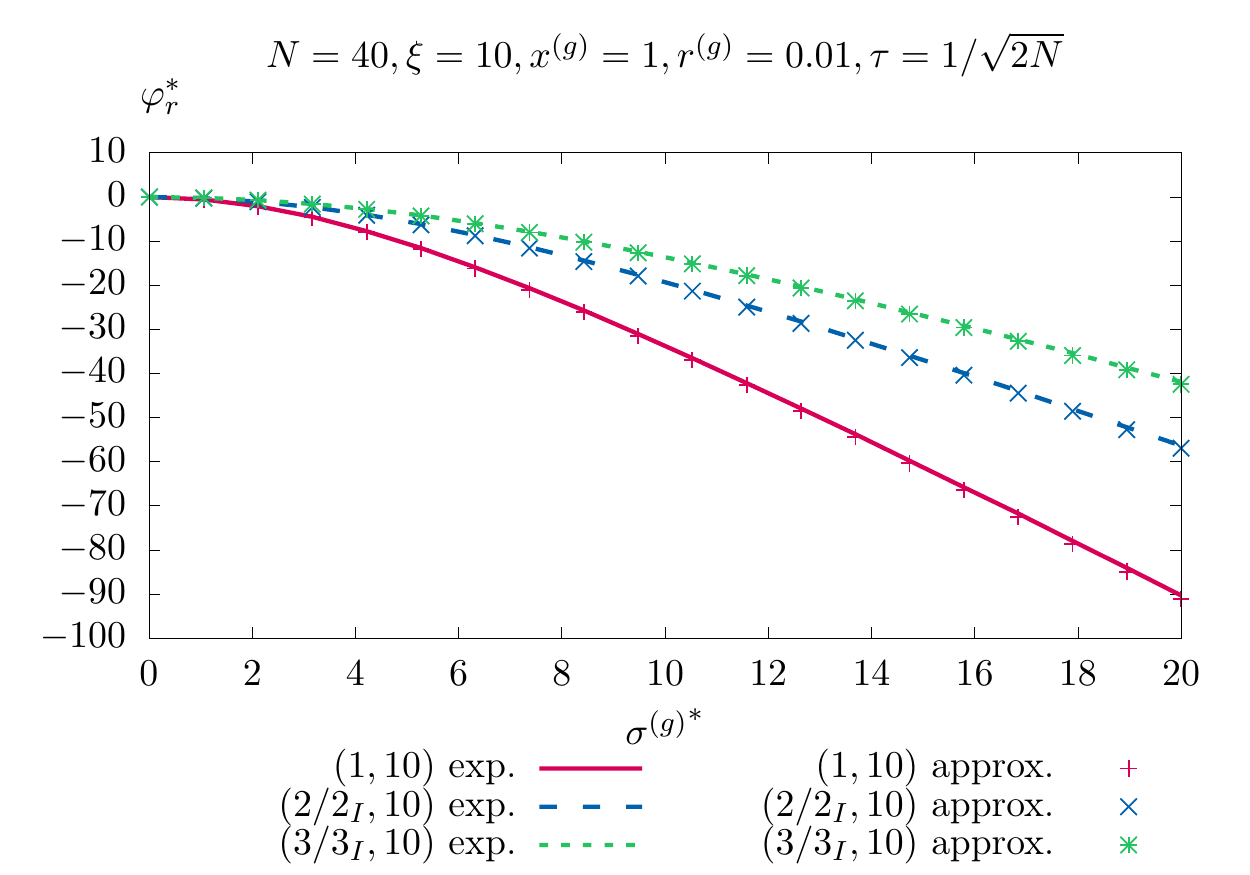}&
    \includegraphics[width=0.5\textwidth]{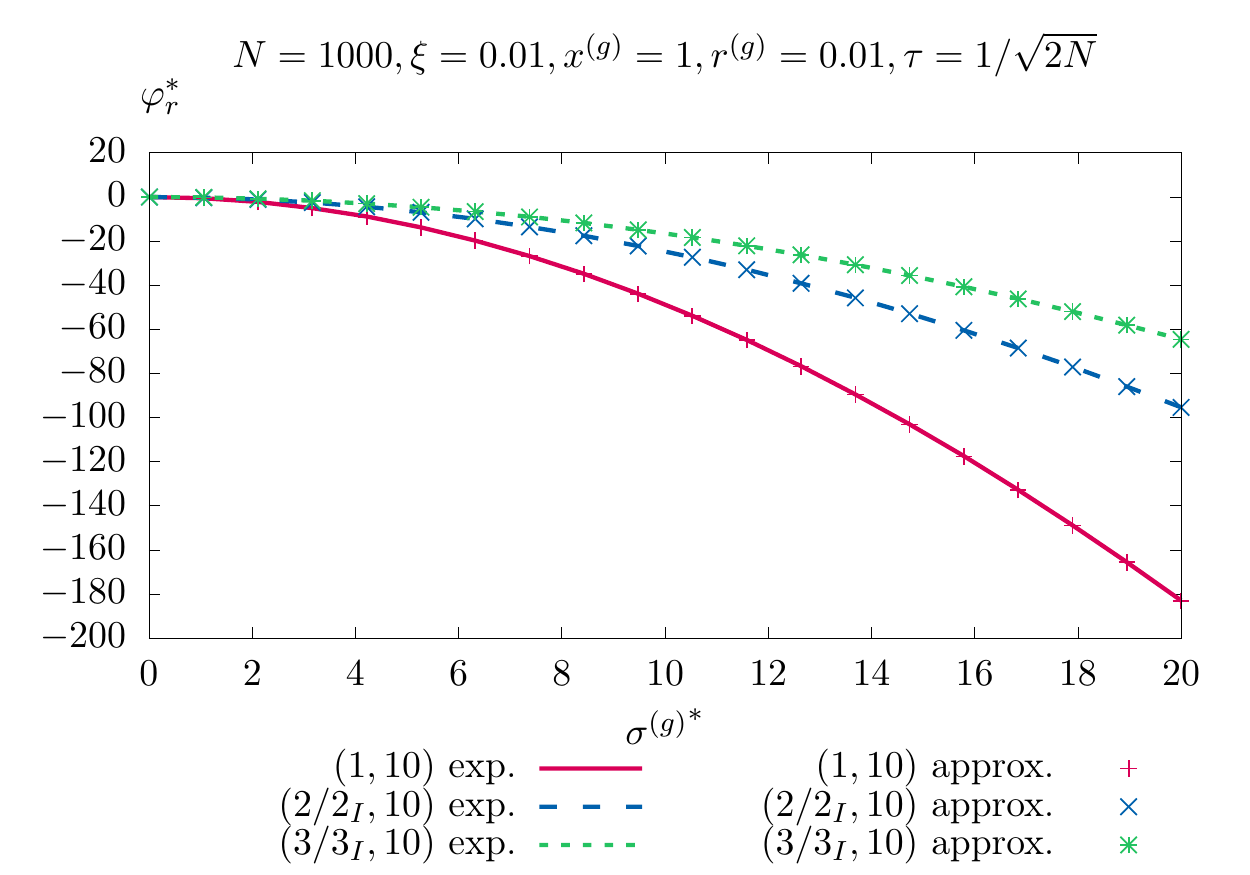}\\
    \includegraphics[width=0.5\textwidth]{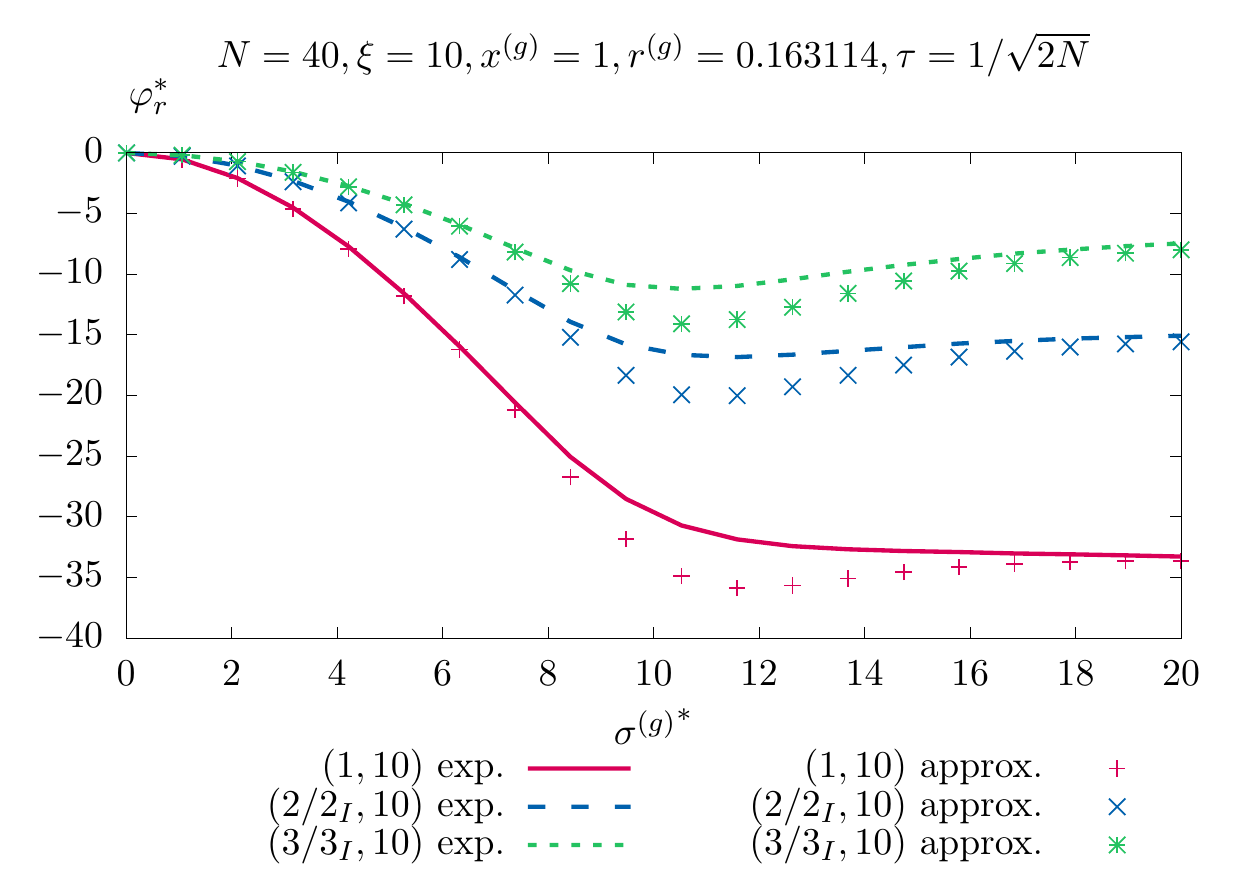}&
    \includegraphics[width=0.5\textwidth]{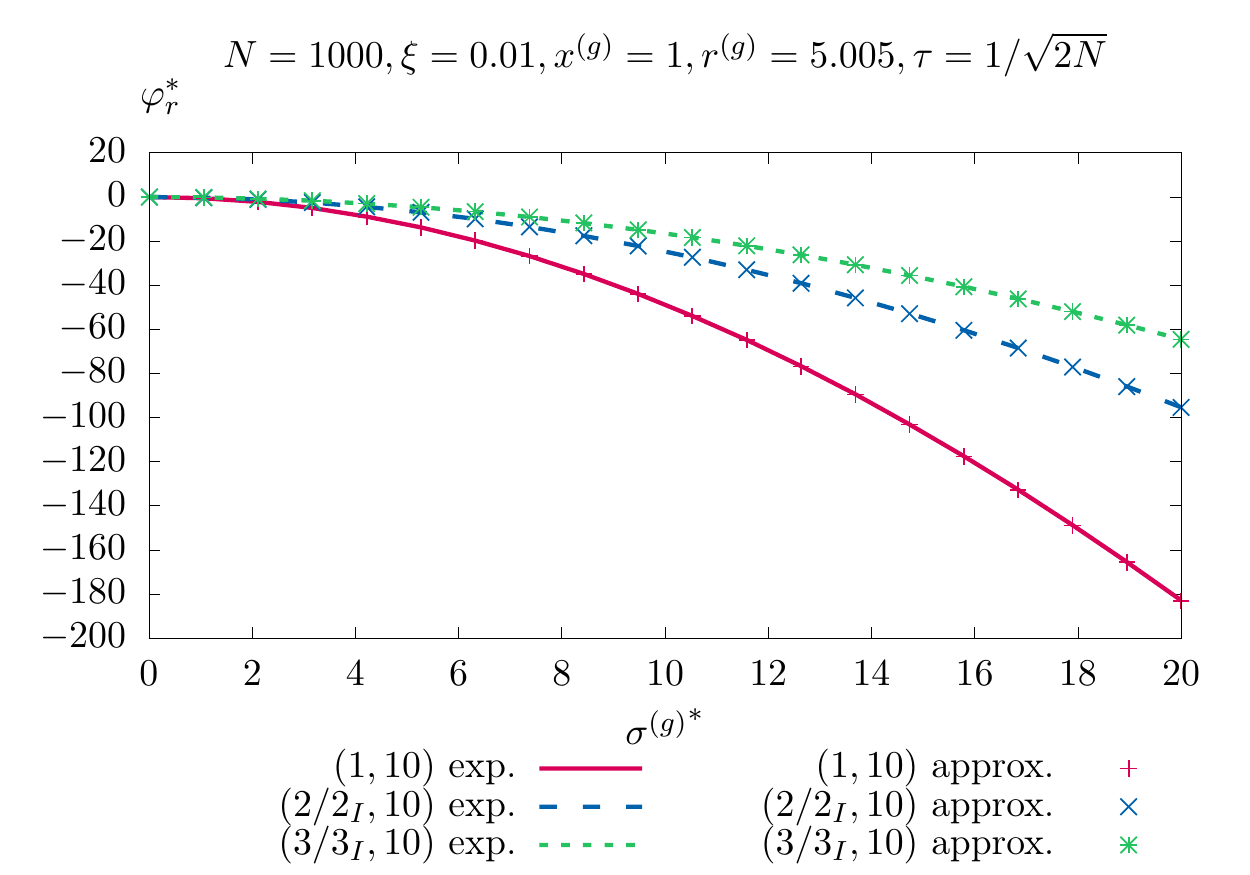}\\
    \includegraphics[width=0.5\textwidth]{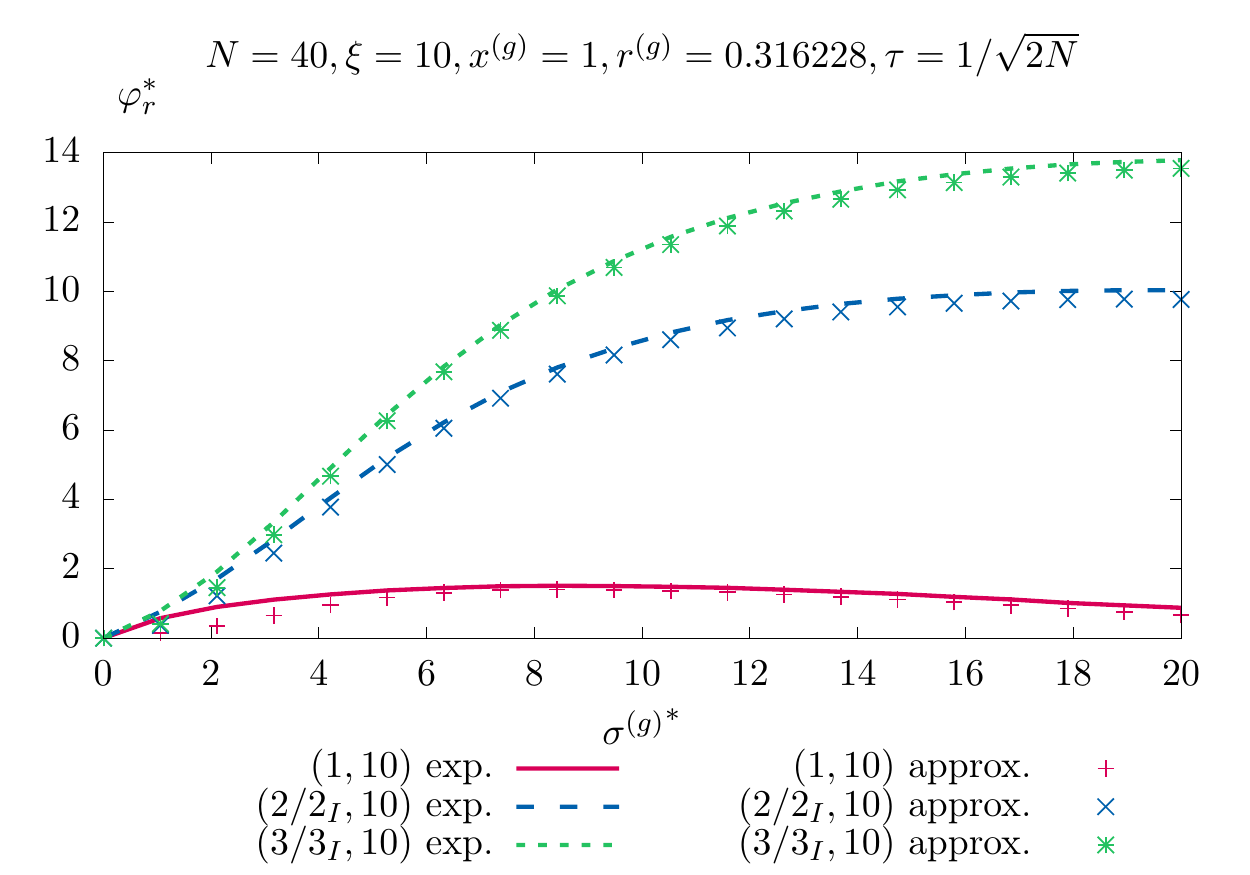}&
    \includegraphics[width=0.5\textwidth]{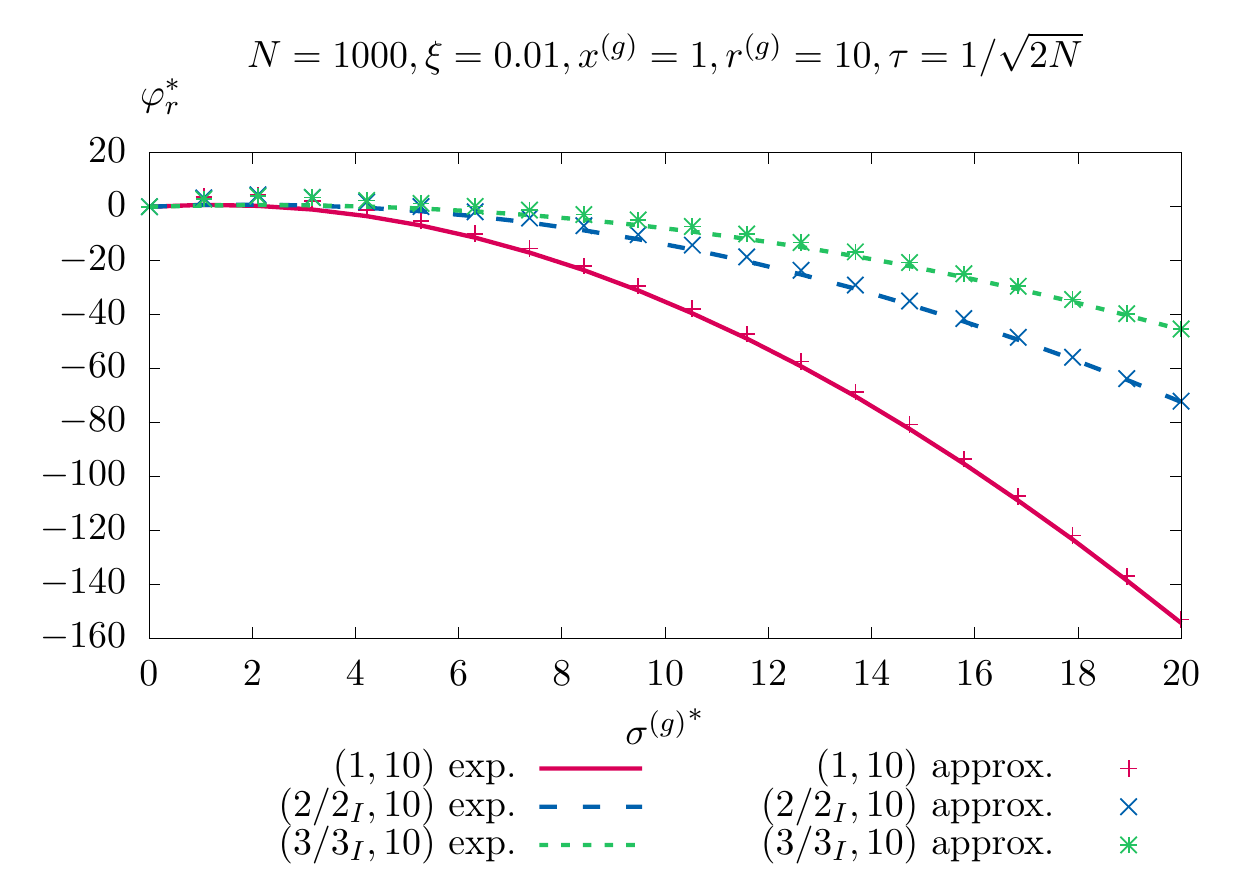}\\
  \end{tabular}
  \caption{Comparison of the $r$ progress rate approximation with simulations. (Part 2)}
  \label{sec:theoreticalanalysis:fig:rprogresscomparisonsdetail2}
\end{figure}

\begin{figure}
  \centering
  \begin{tabular}{@{\hspace{-0.025\textwidth}}c@{\hspace{-0.025\textwidth}}c}
    \includegraphics[width=0.5\textwidth]{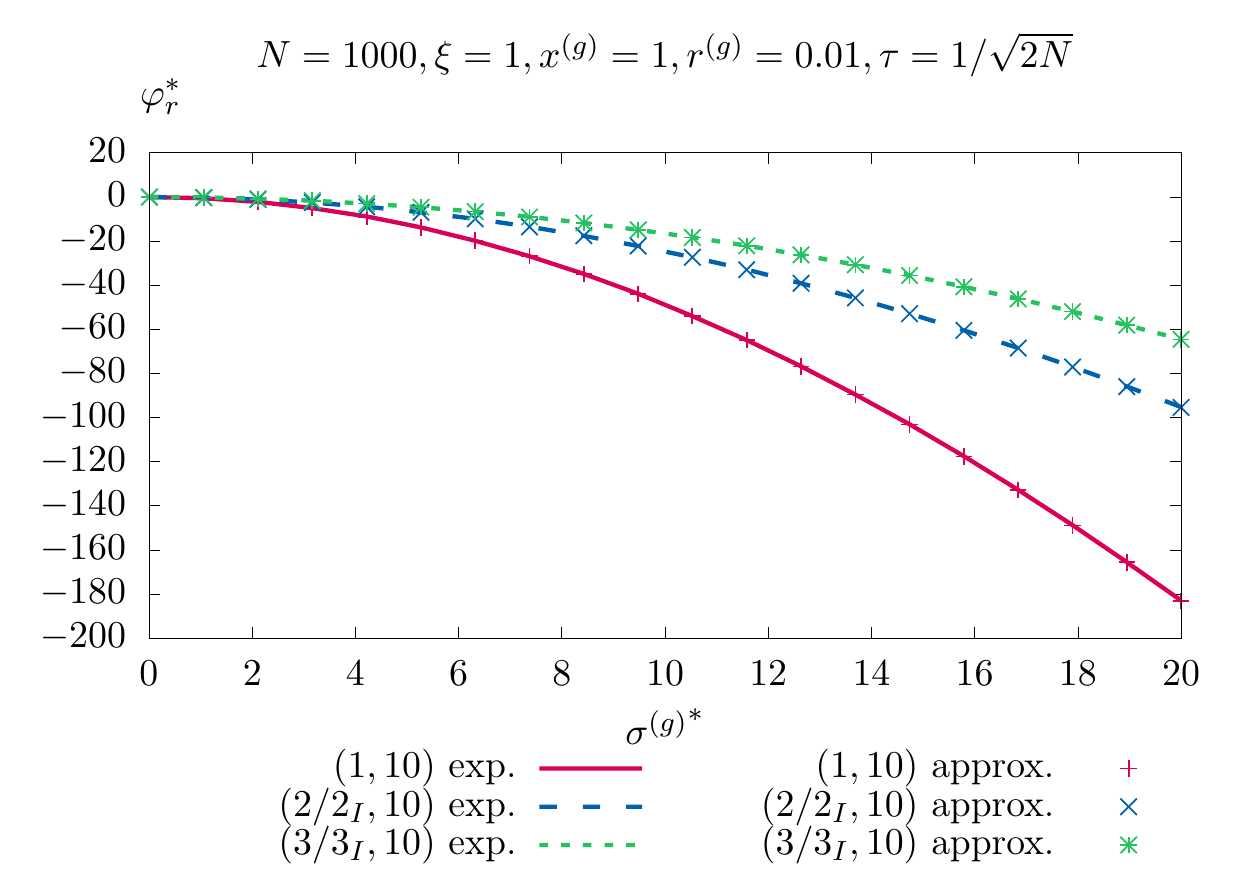}&
    \includegraphics[width=0.5\textwidth]{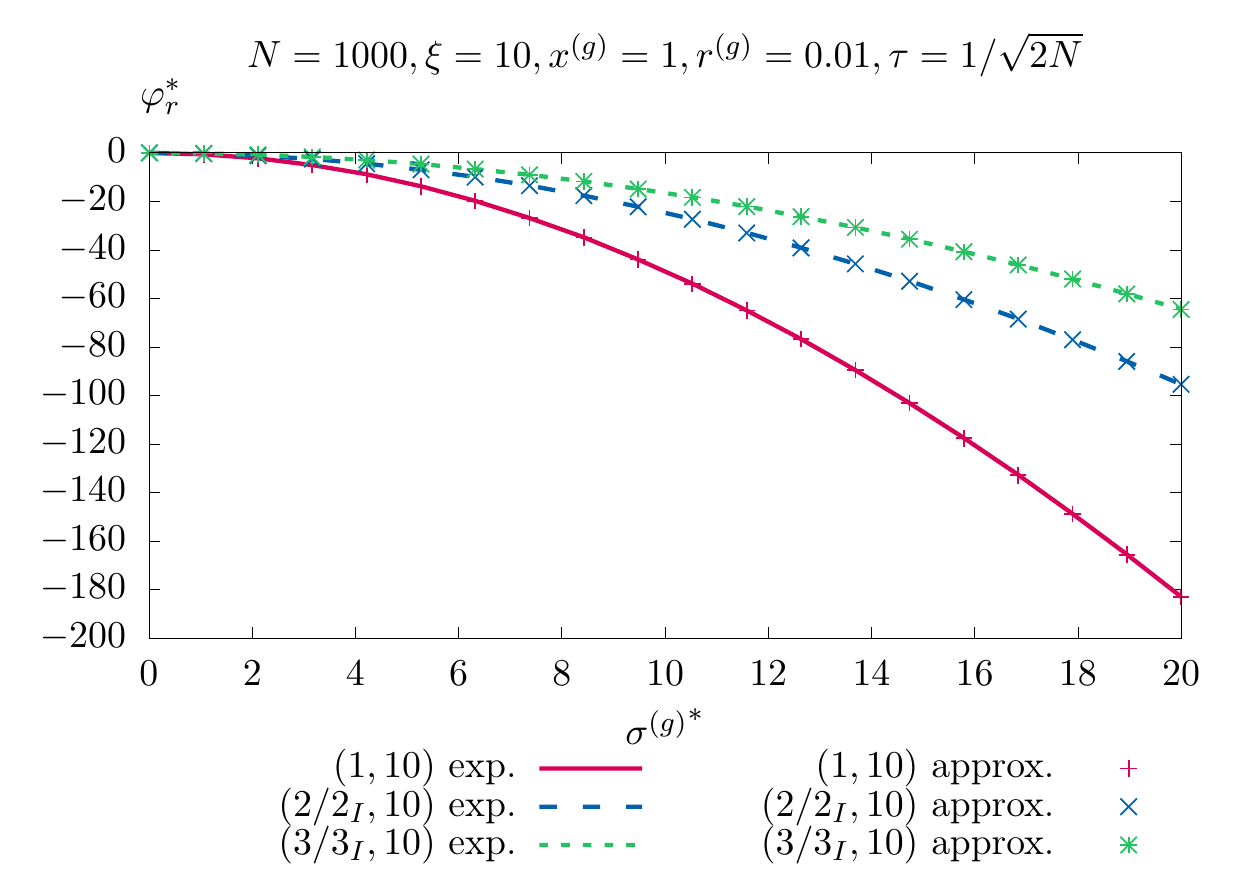}\\
    \includegraphics[width=0.5\textwidth]{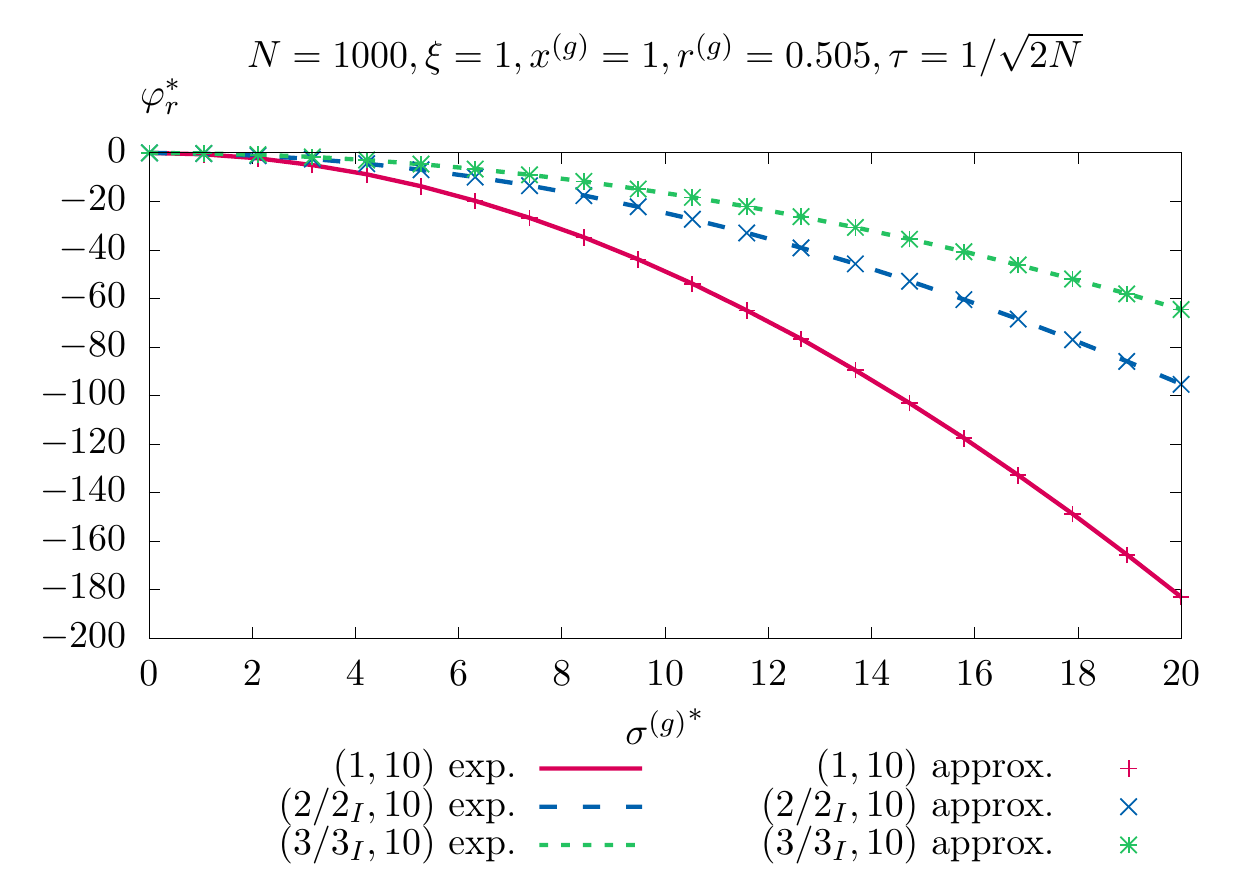}&
    \includegraphics[width=0.5\textwidth]{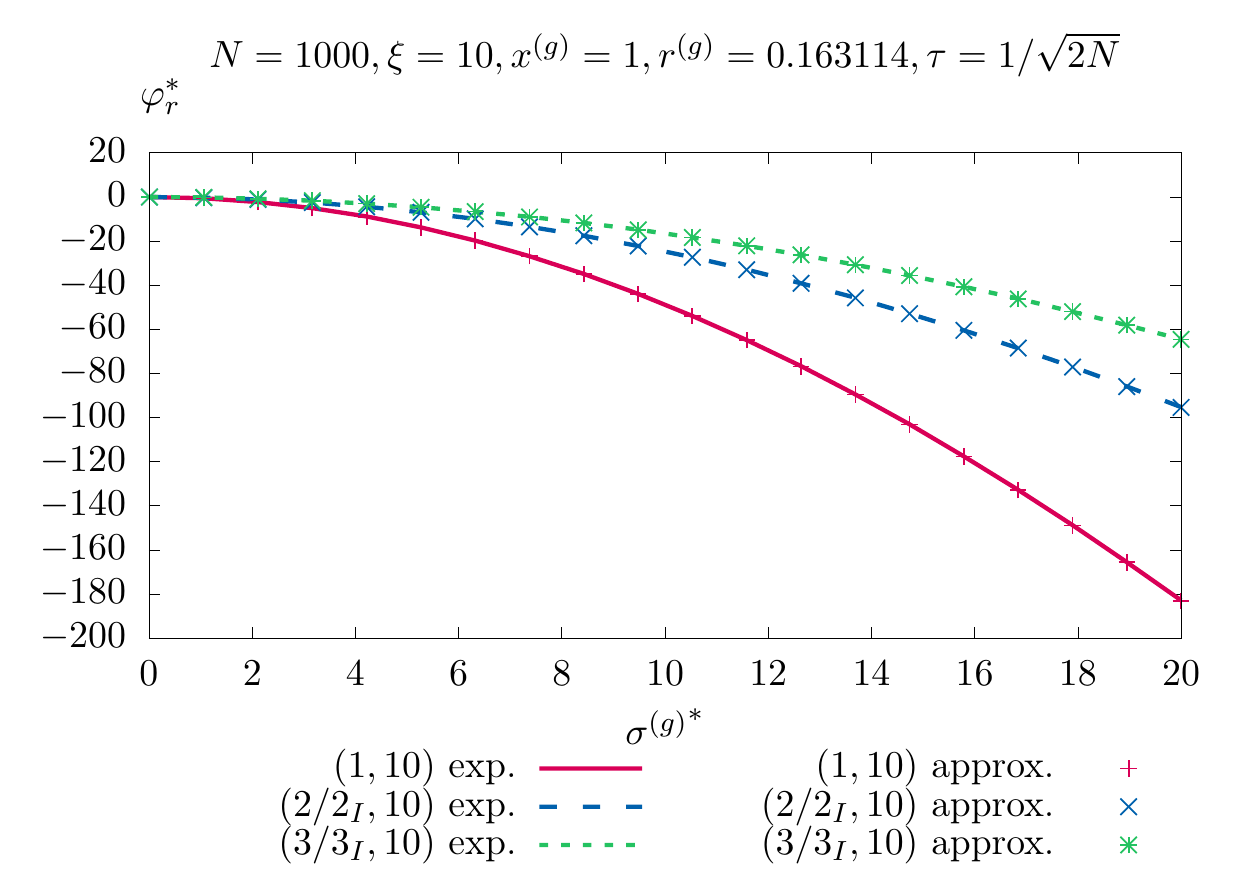}\\
    \includegraphics[width=0.5\textwidth]{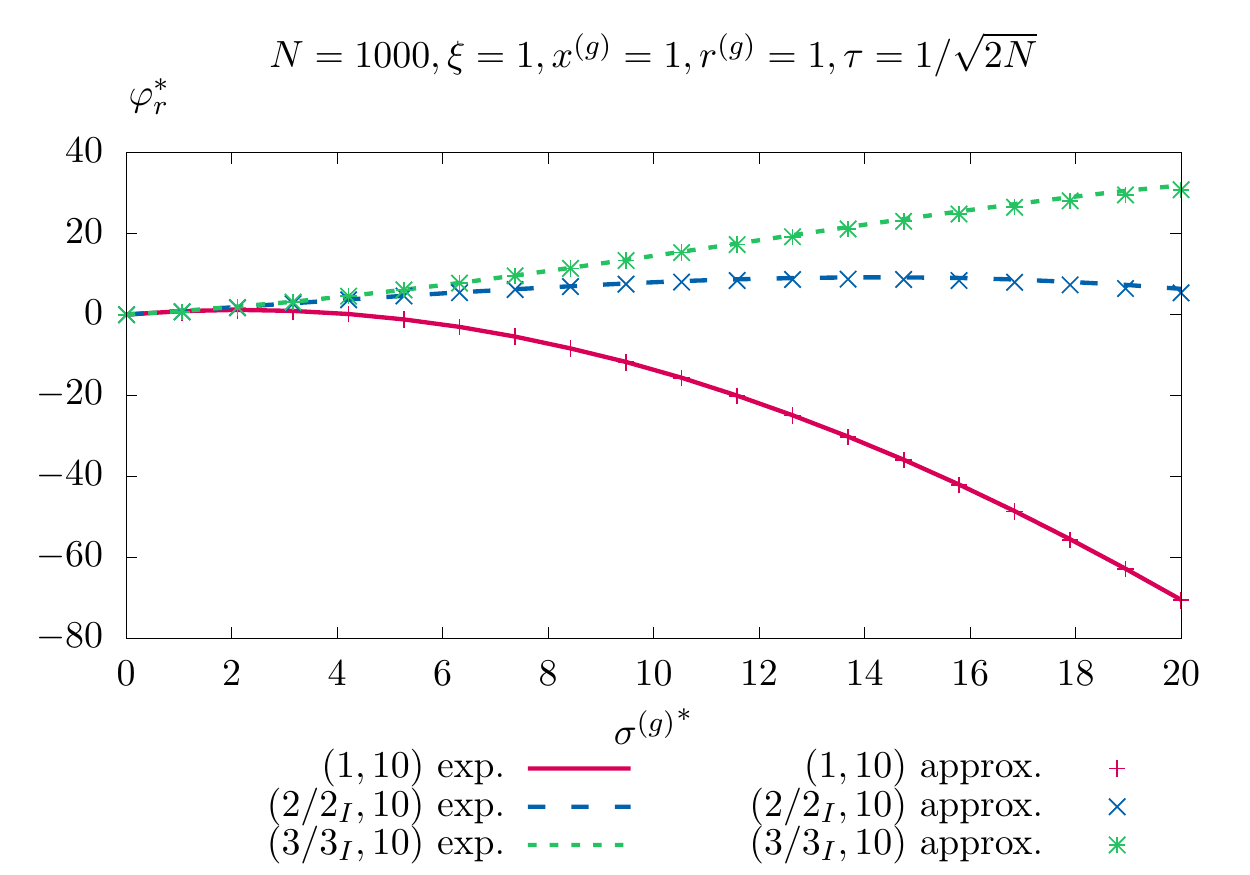}&
    \includegraphics[width=0.5\textwidth]{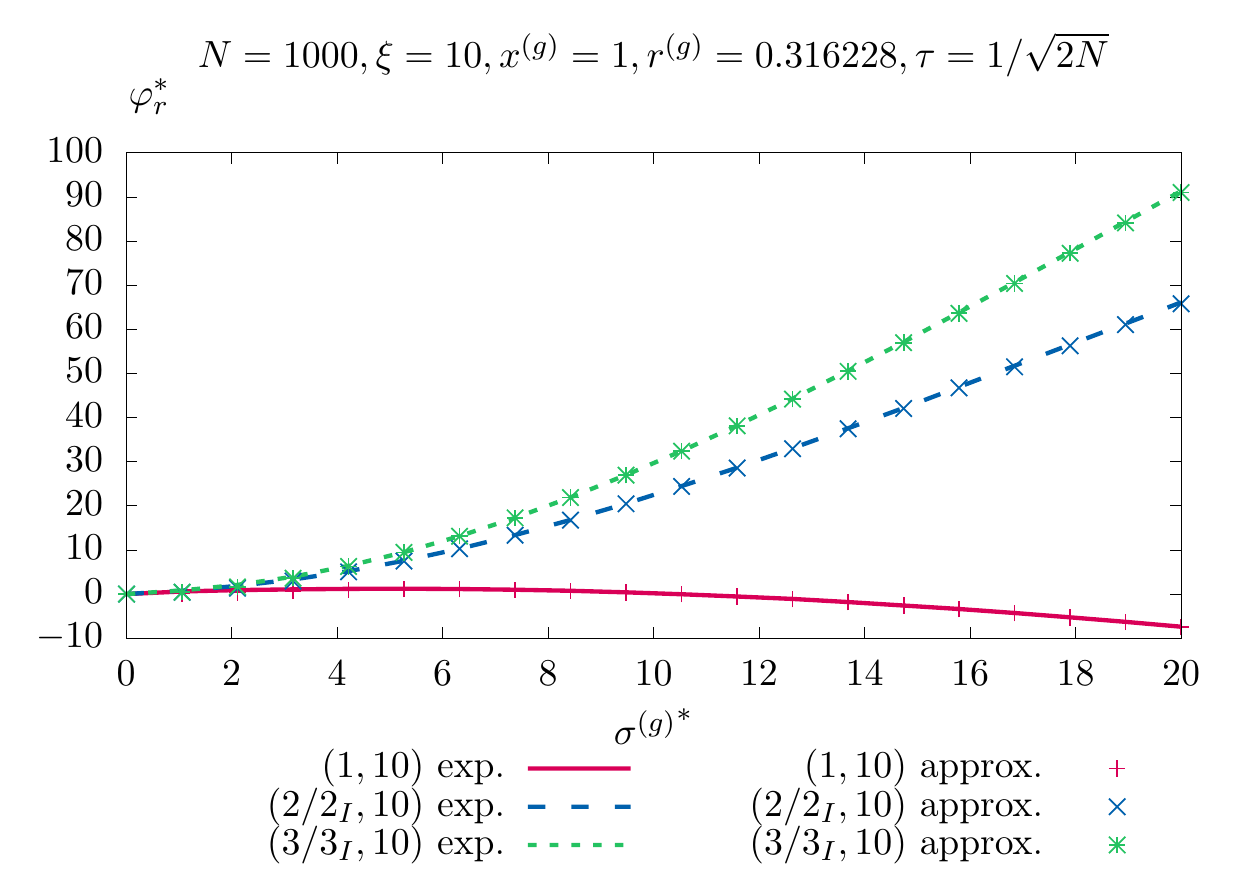}\\
  \end{tabular}
  \caption{Comparison of the $r$ progress rate approximation with simulations. (Part 3)}
  \label{sec:theoreticalanalysis:fig:rprogresscomparisonsdetail3}
\end{figure}

\begin{figure}
  \centering
  \begin{tabular}{@{\hspace{-0.025\textwidth}}c@{\hspace{-0.025\textwidth}}c}
    \includegraphics[width=0.5\textwidth]{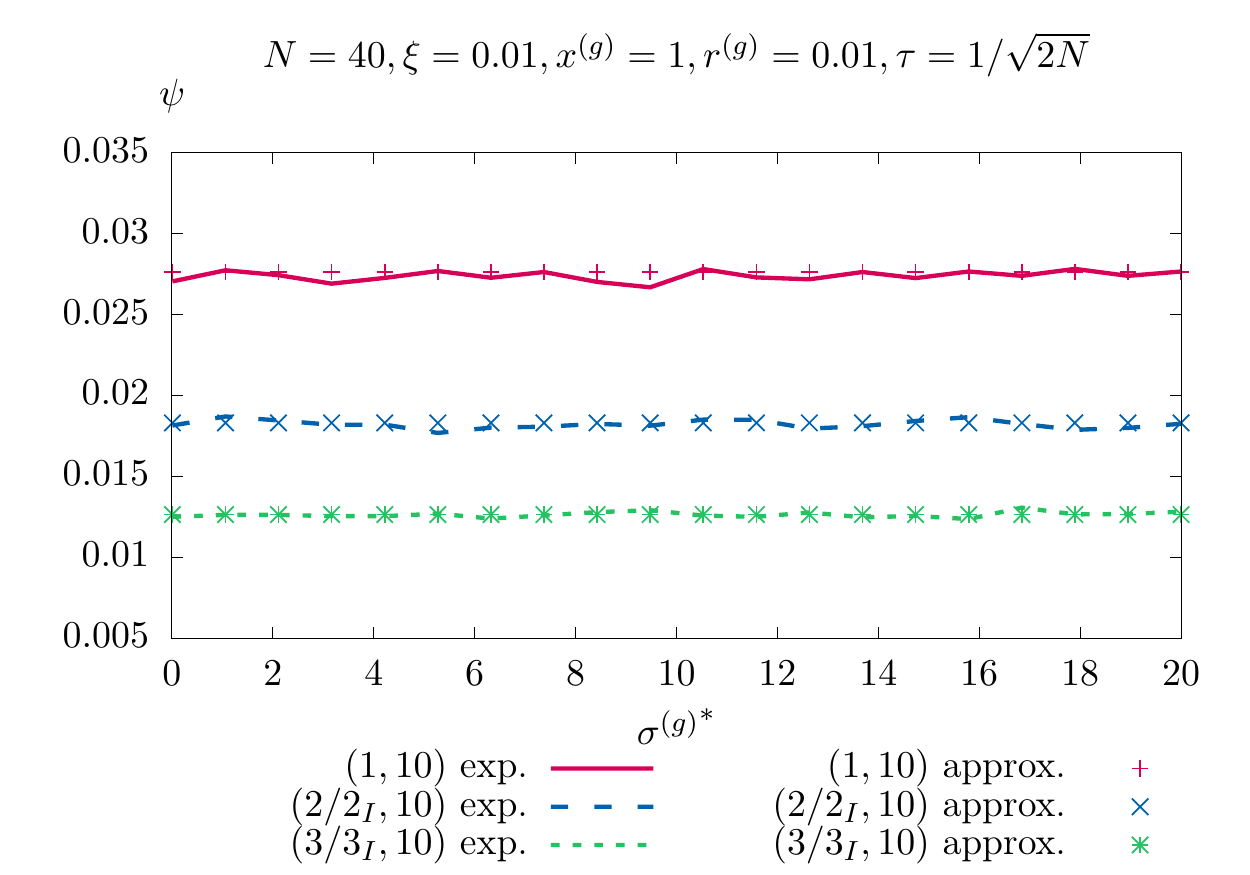}&
    \includegraphics[width=0.5\textwidth]{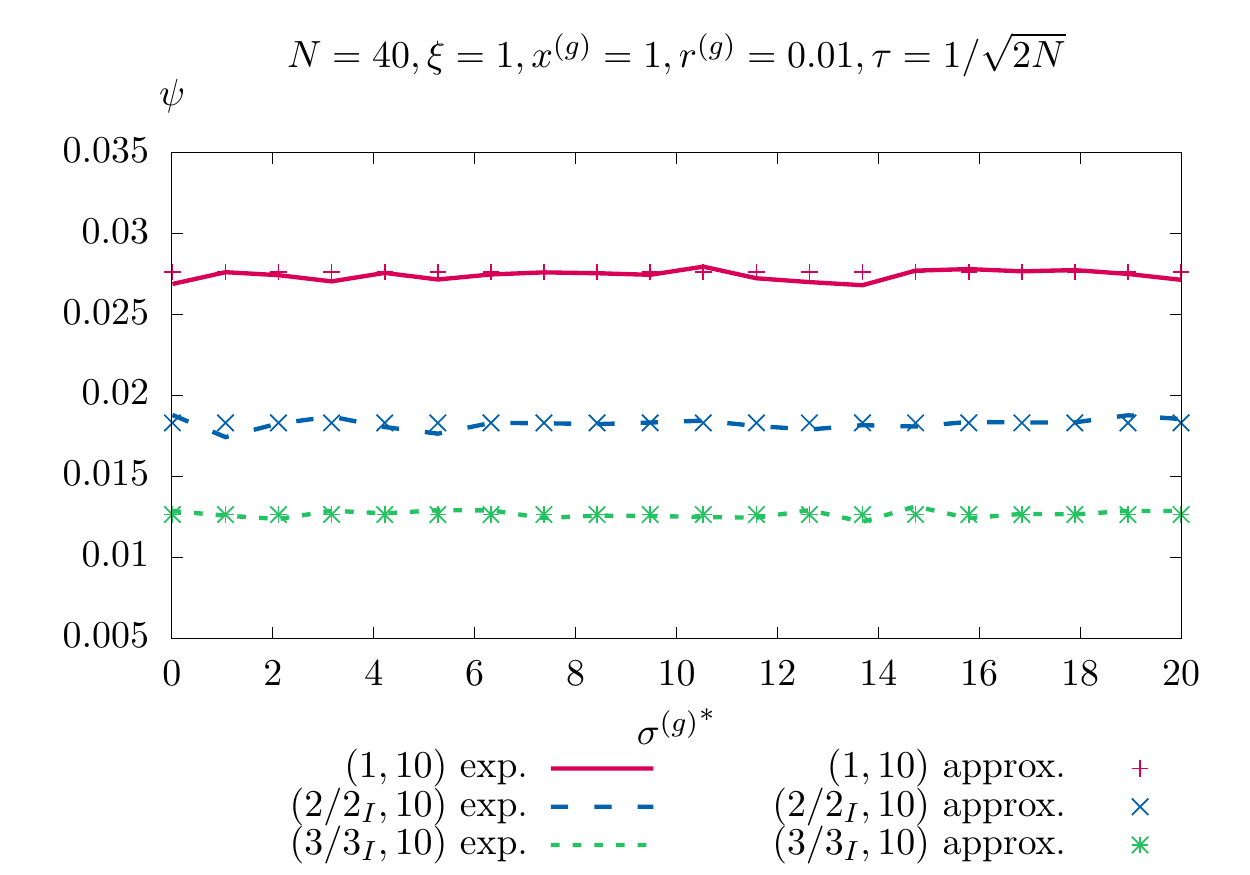}\\
    \includegraphics[width=0.5\textwidth]{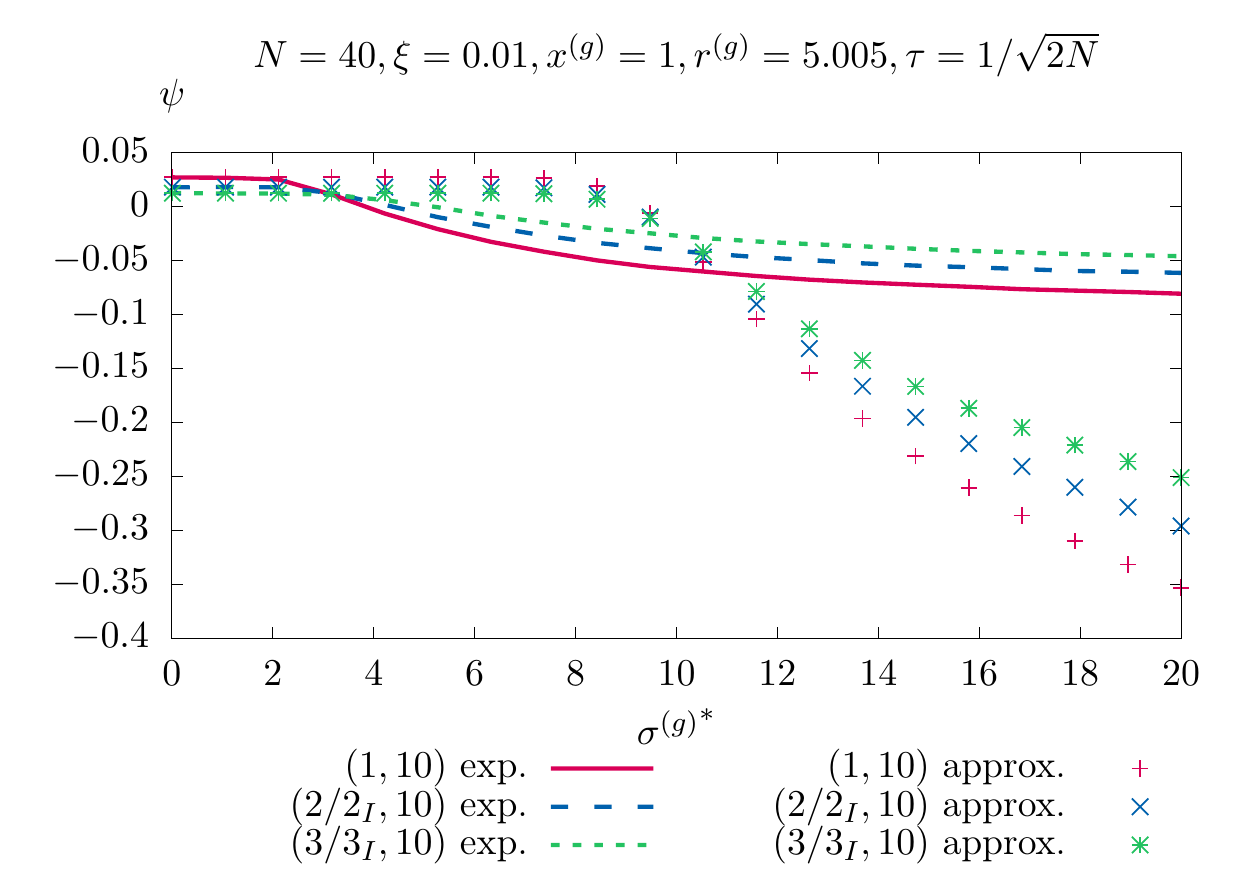}&
    \includegraphics[width=0.5\textwidth]{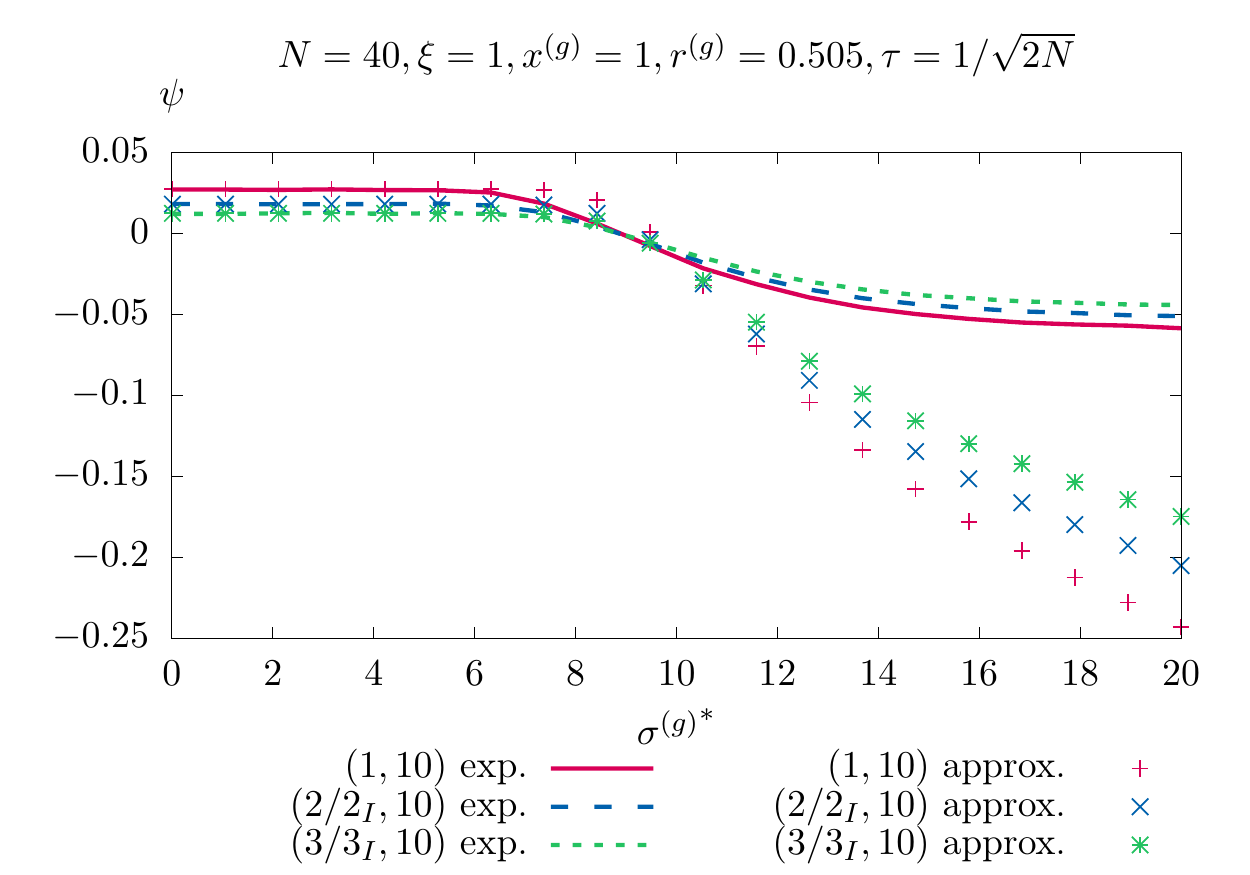}\\
    \includegraphics[width=0.5\textwidth]{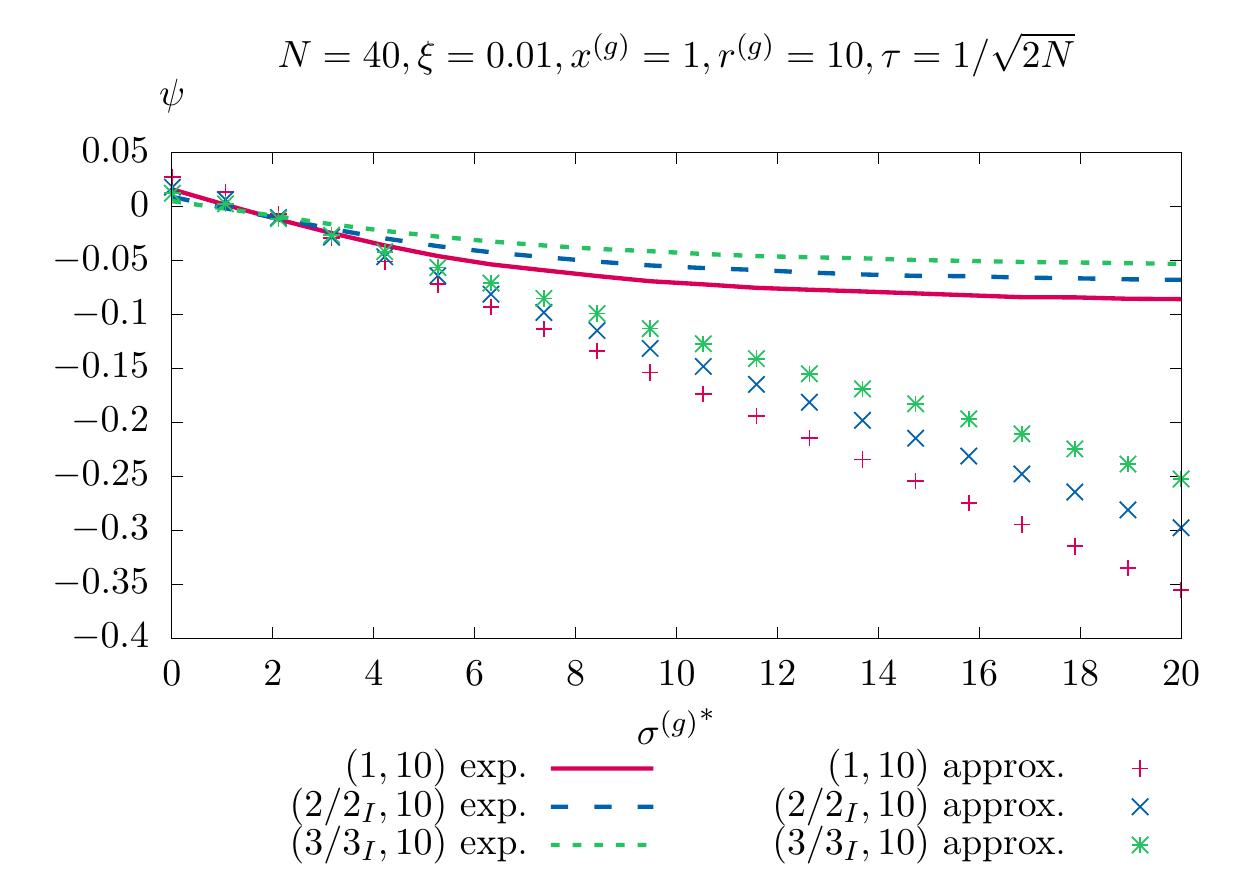}&
    \includegraphics[width=0.5\textwidth]{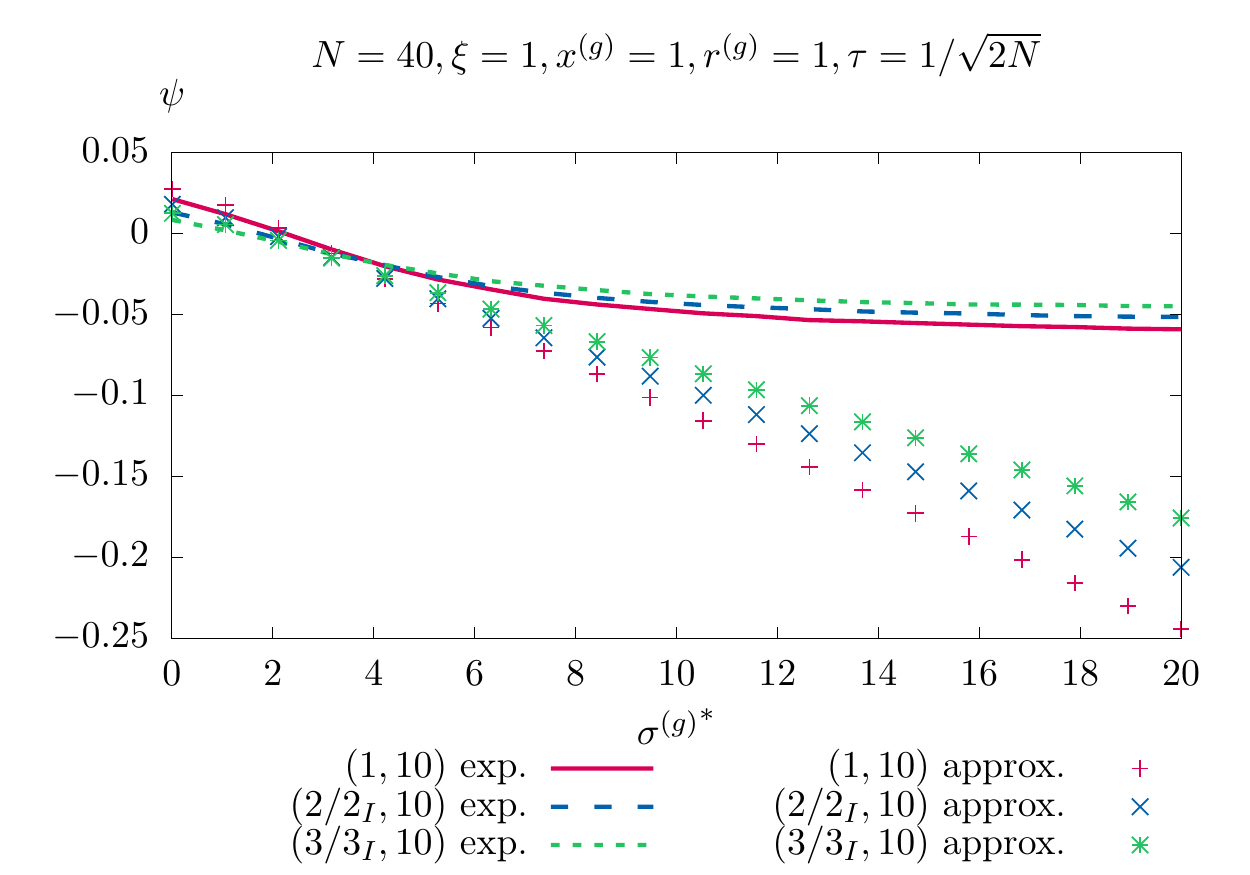}\\
  \end{tabular}
  \caption{Comparison of the SAR approximation with simulations. (Part 1)}
  \label{sec:theoreticalanalysis:fig:sarcomparisonsdetail1}
\end{figure}

\begin{figure}
  \centering
  \begin{tabular}{@{\hspace{-0.025\textwidth}}c@{\hspace{-0.025\textwidth}}c}
    \includegraphics[width=0.5\textwidth]{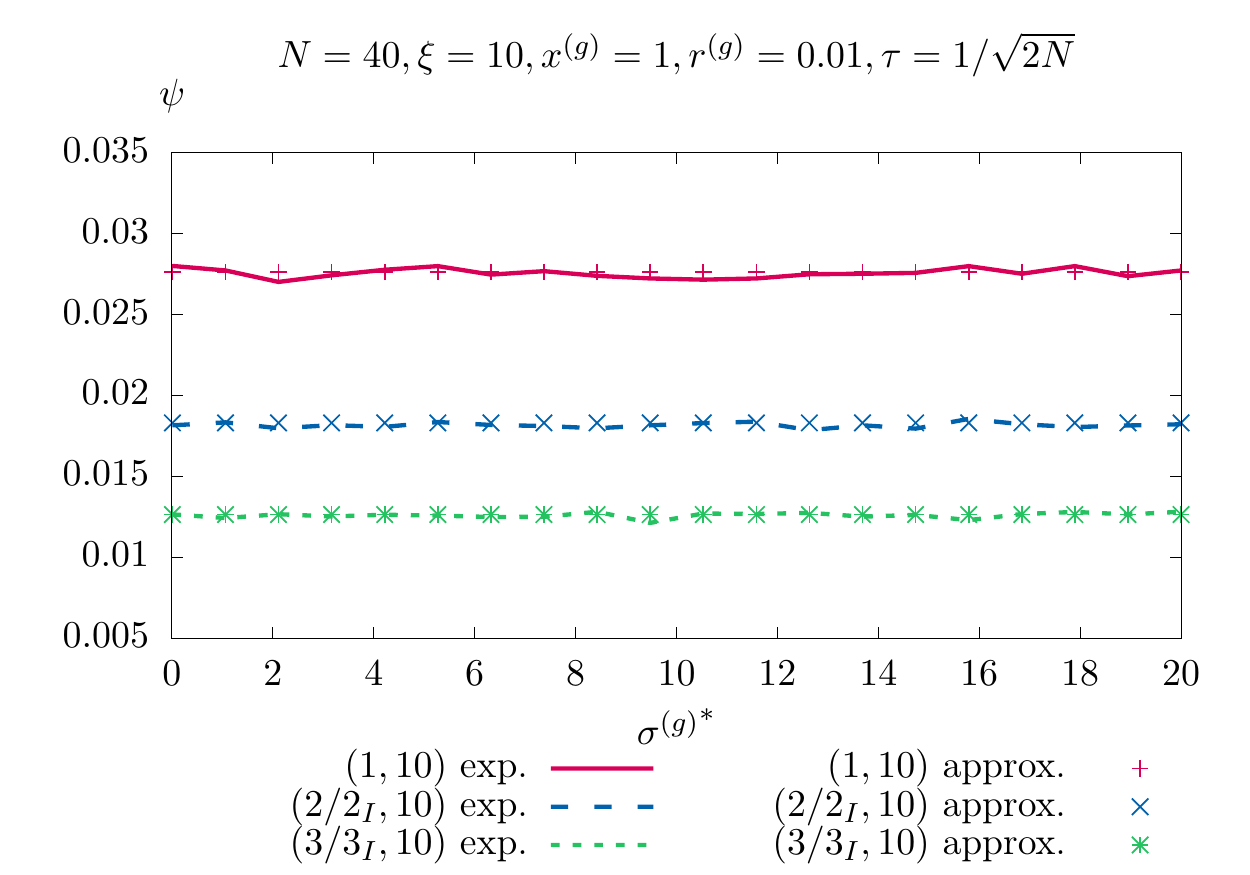}&
    \includegraphics[width=0.5\textwidth]{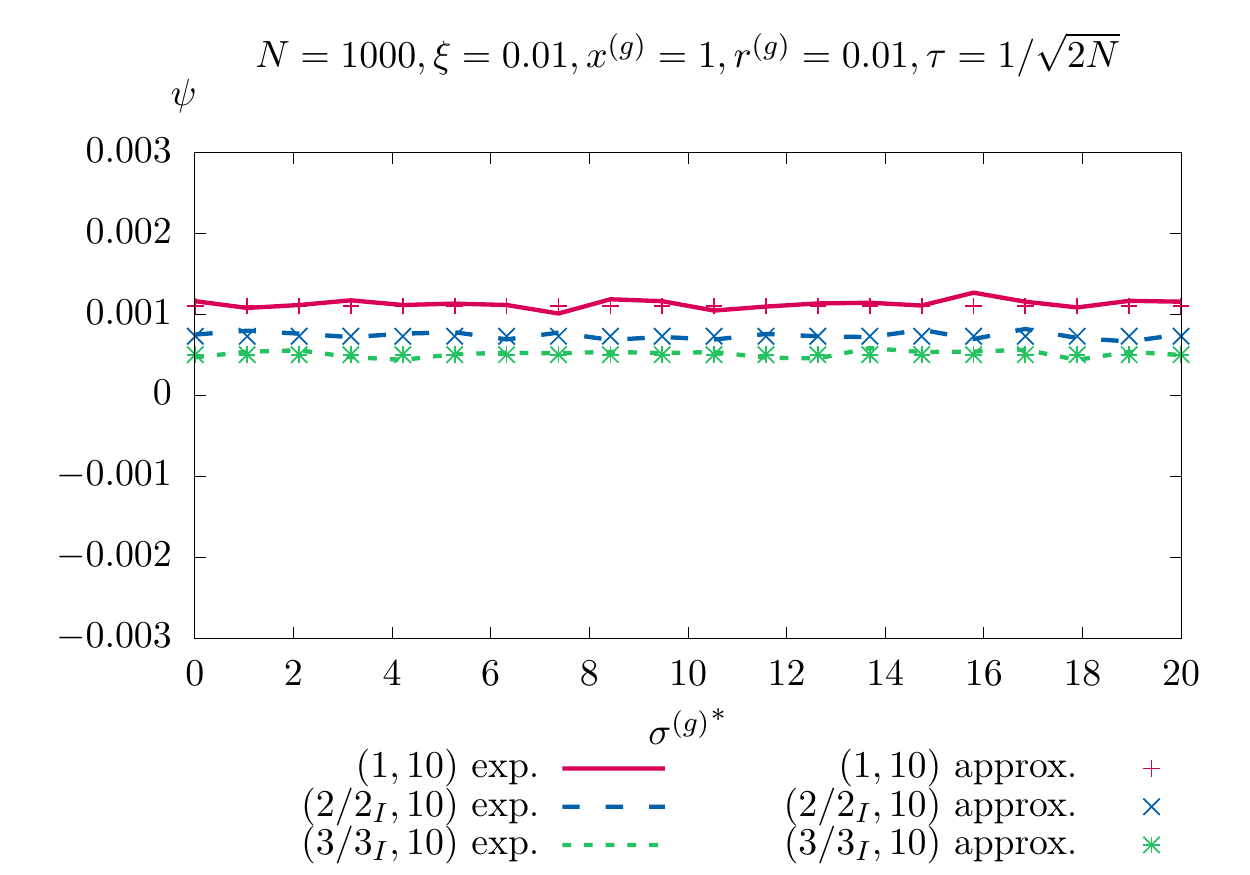}\\
    \includegraphics[width=0.5\textwidth]{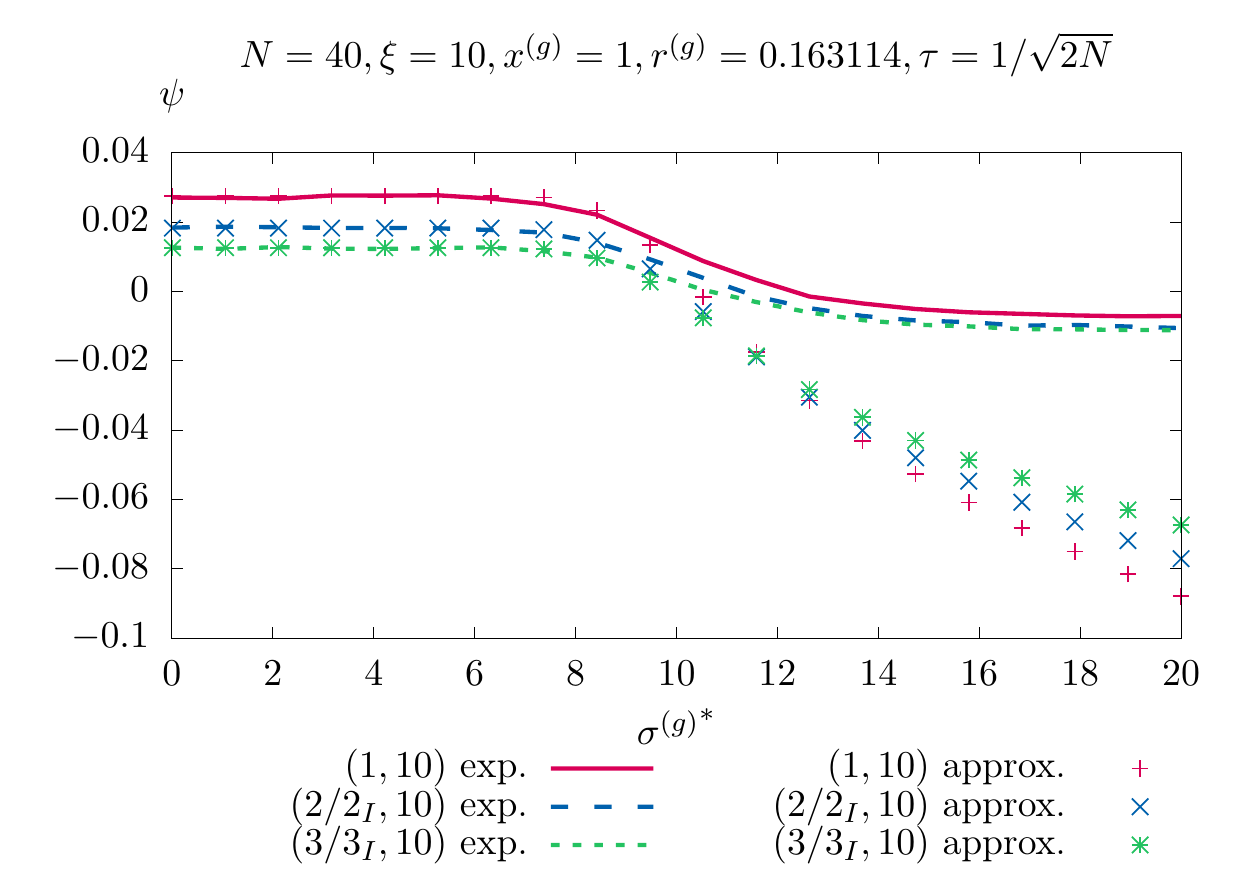}&
    \includegraphics[width=0.5\textwidth]{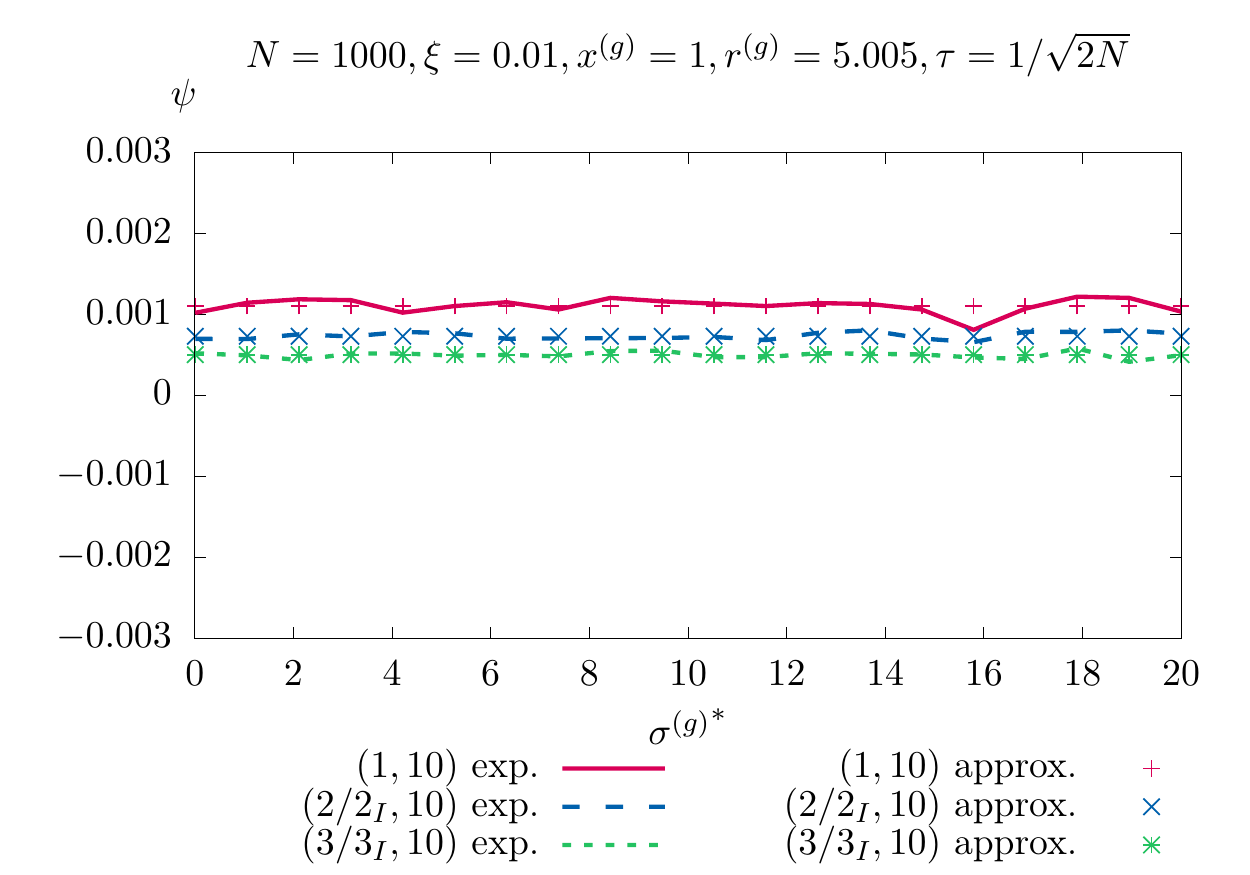}\\
    \includegraphics[width=0.5\textwidth]{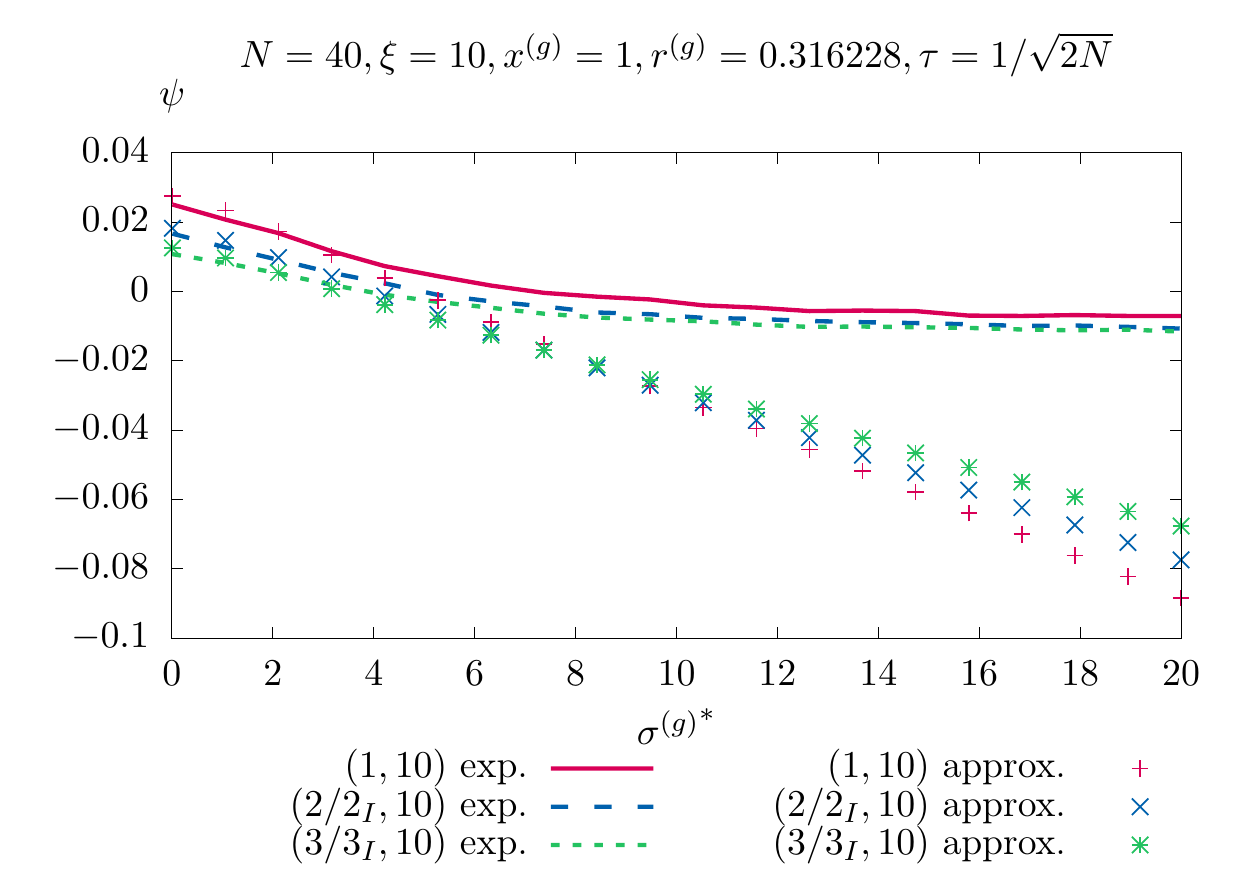}&
    \includegraphics[width=0.5\textwidth]{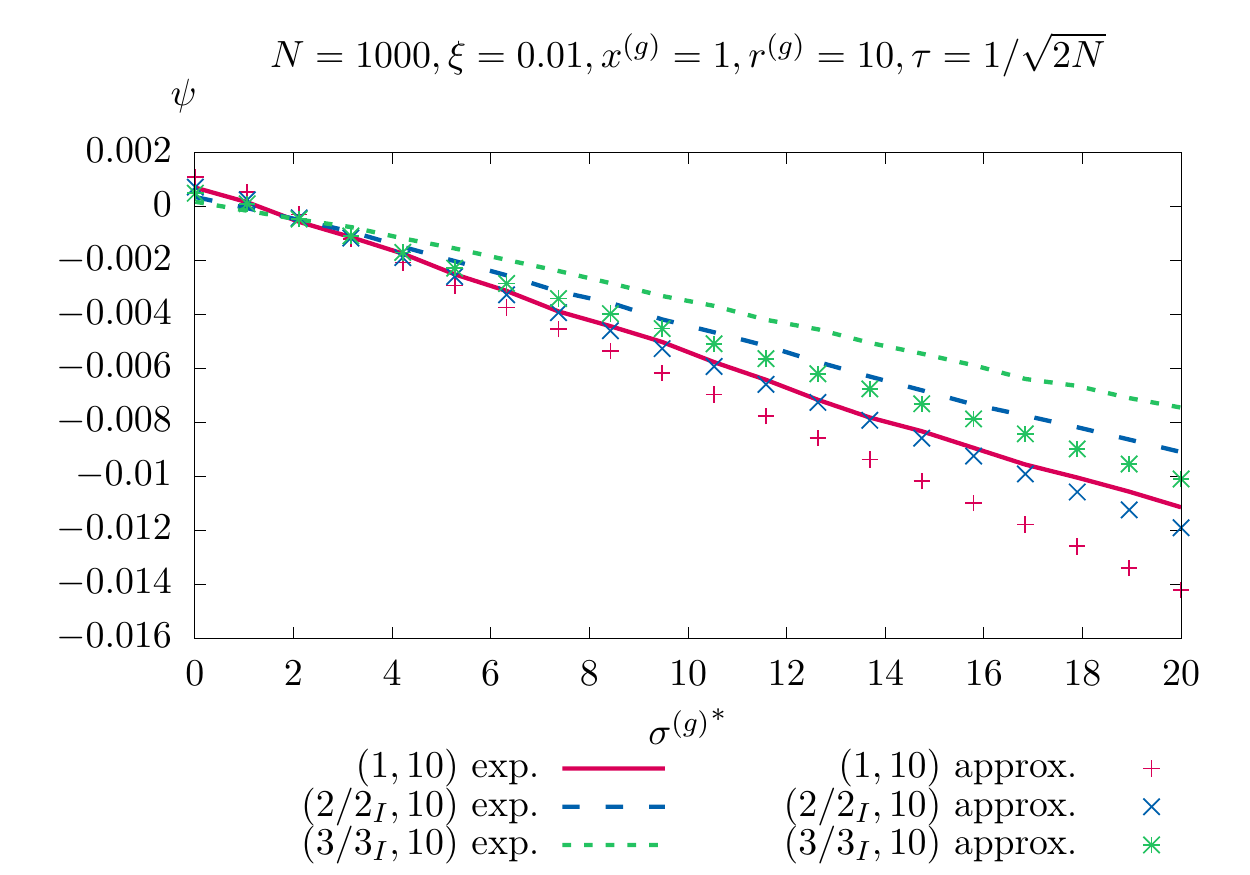}\\
  \end{tabular}
  \caption{Comparison of the SAR approximation with simulations. (Part 2)}
  \label{sec:theoreticalanalysis:fig:sarcomparisonsdetail2}
\end{figure}

\begin{figure}
  \centering
  \begin{tabular}{@{\hspace{-0.025\textwidth}}c@{\hspace{-0.025\textwidth}}c}
    \includegraphics[width=0.5\textwidth]{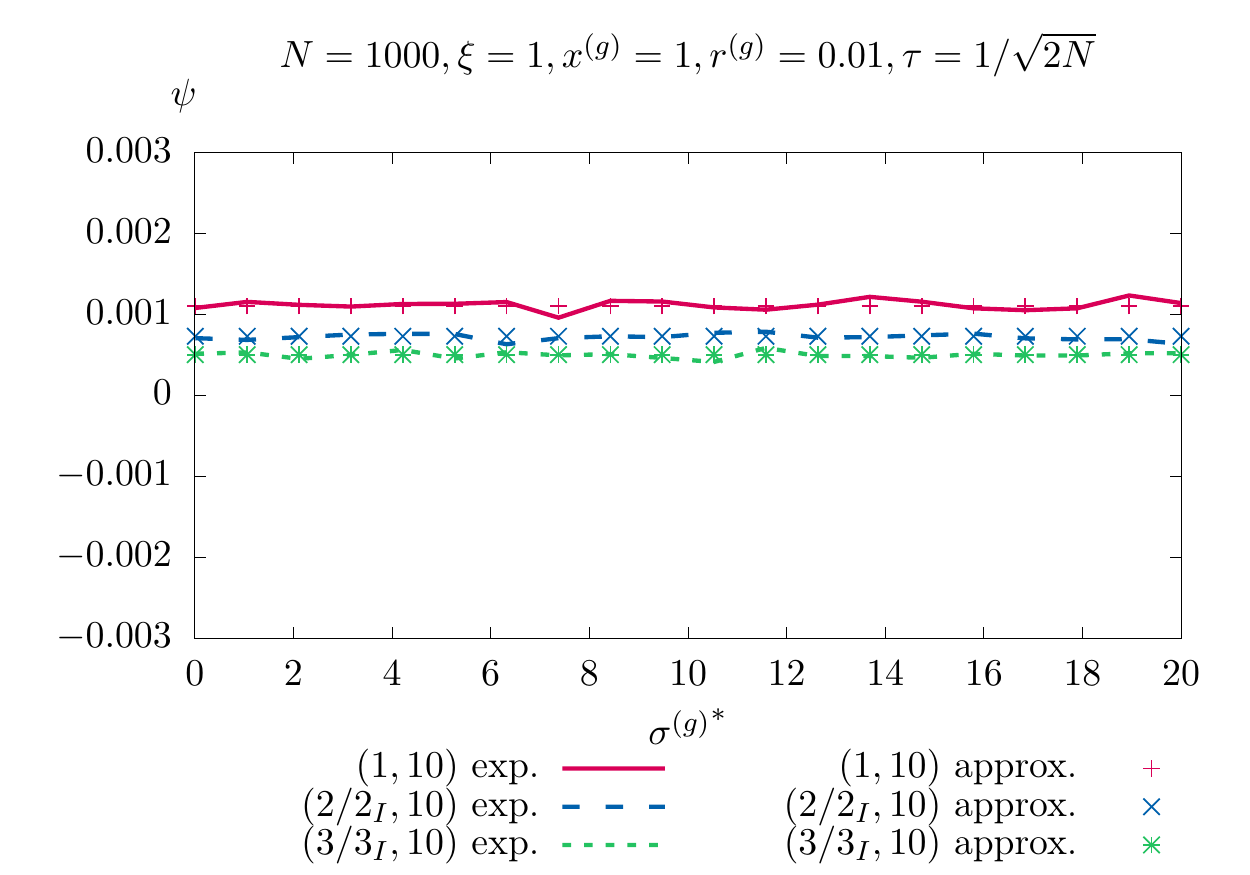}&
    \includegraphics[width=0.5\textwidth]{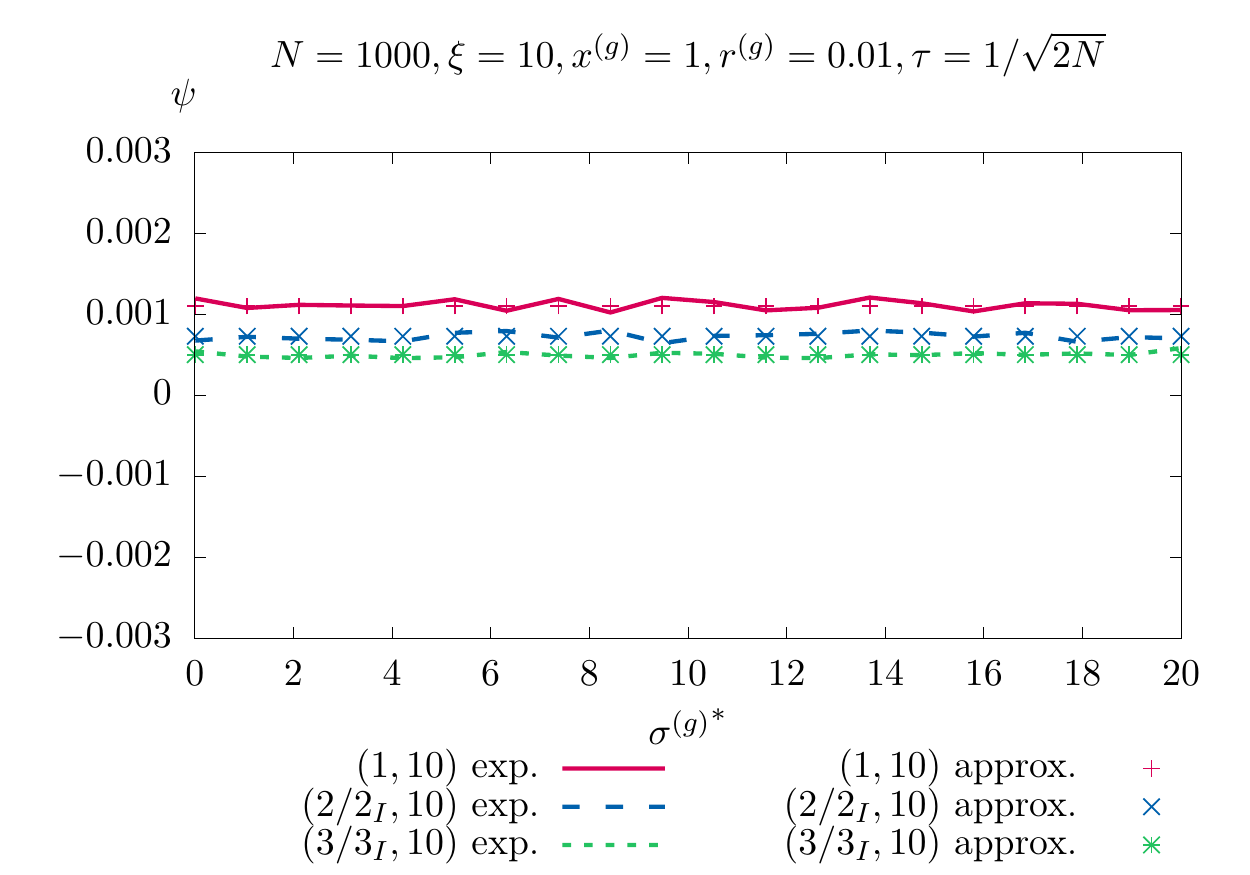}\\
    \includegraphics[width=0.5\textwidth]{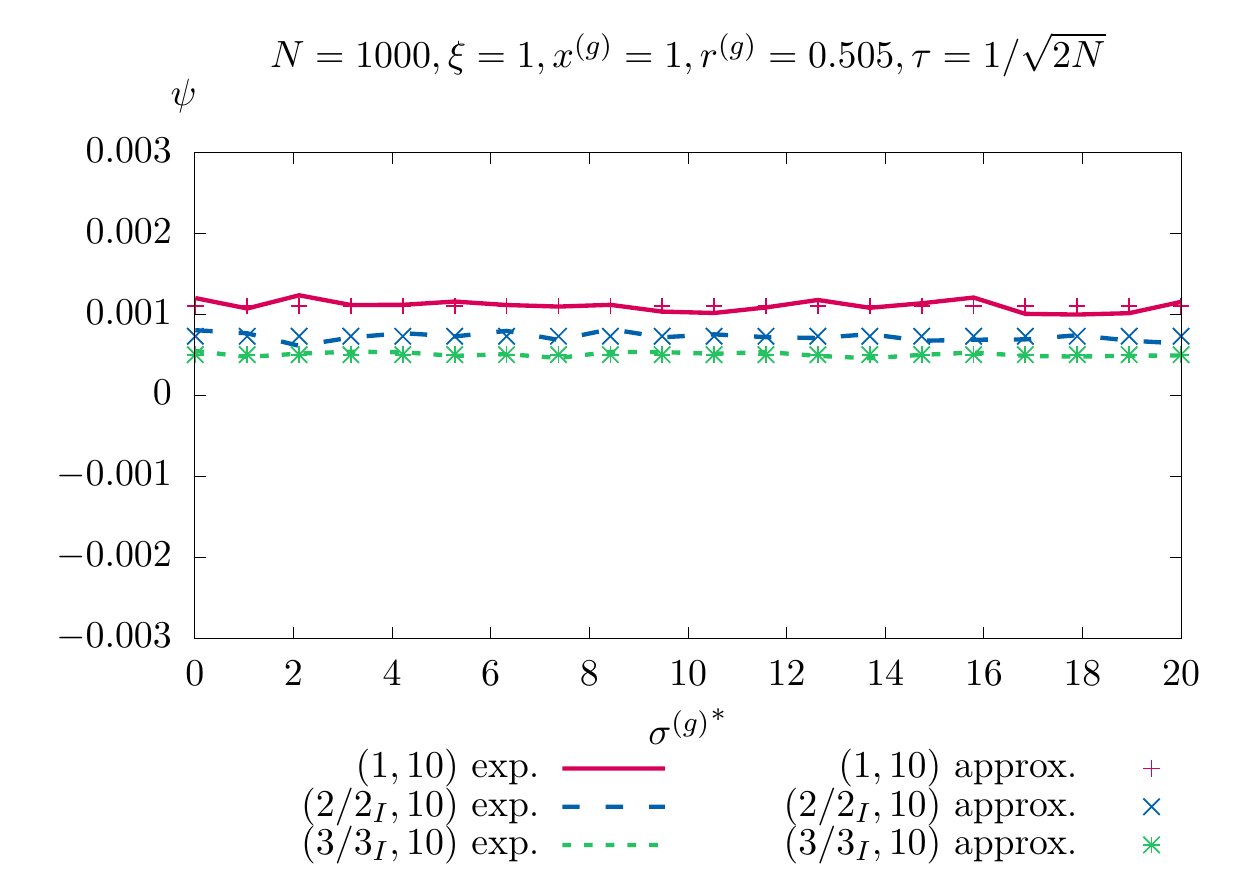}&
    \includegraphics[width=0.5\textwidth]{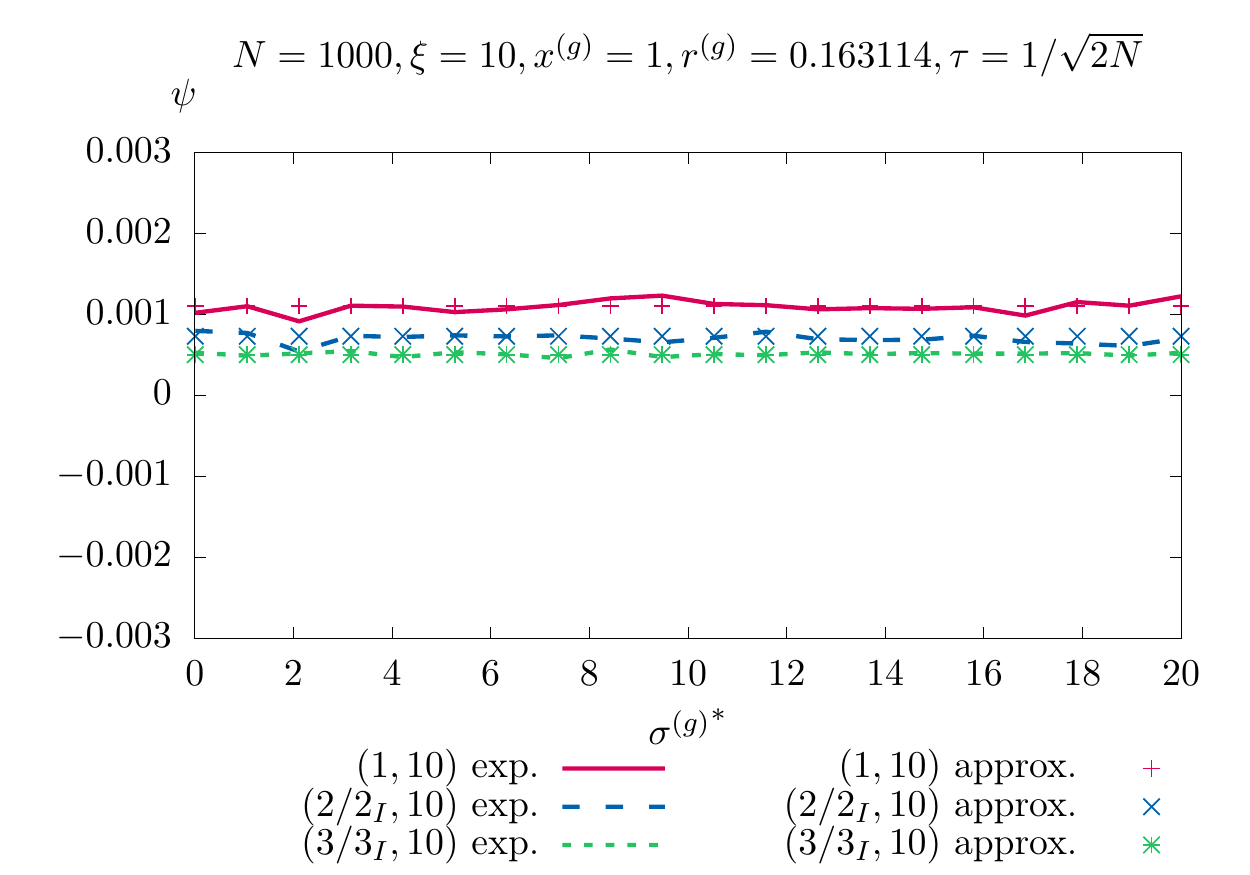}\\
    \includegraphics[width=0.5\textwidth]{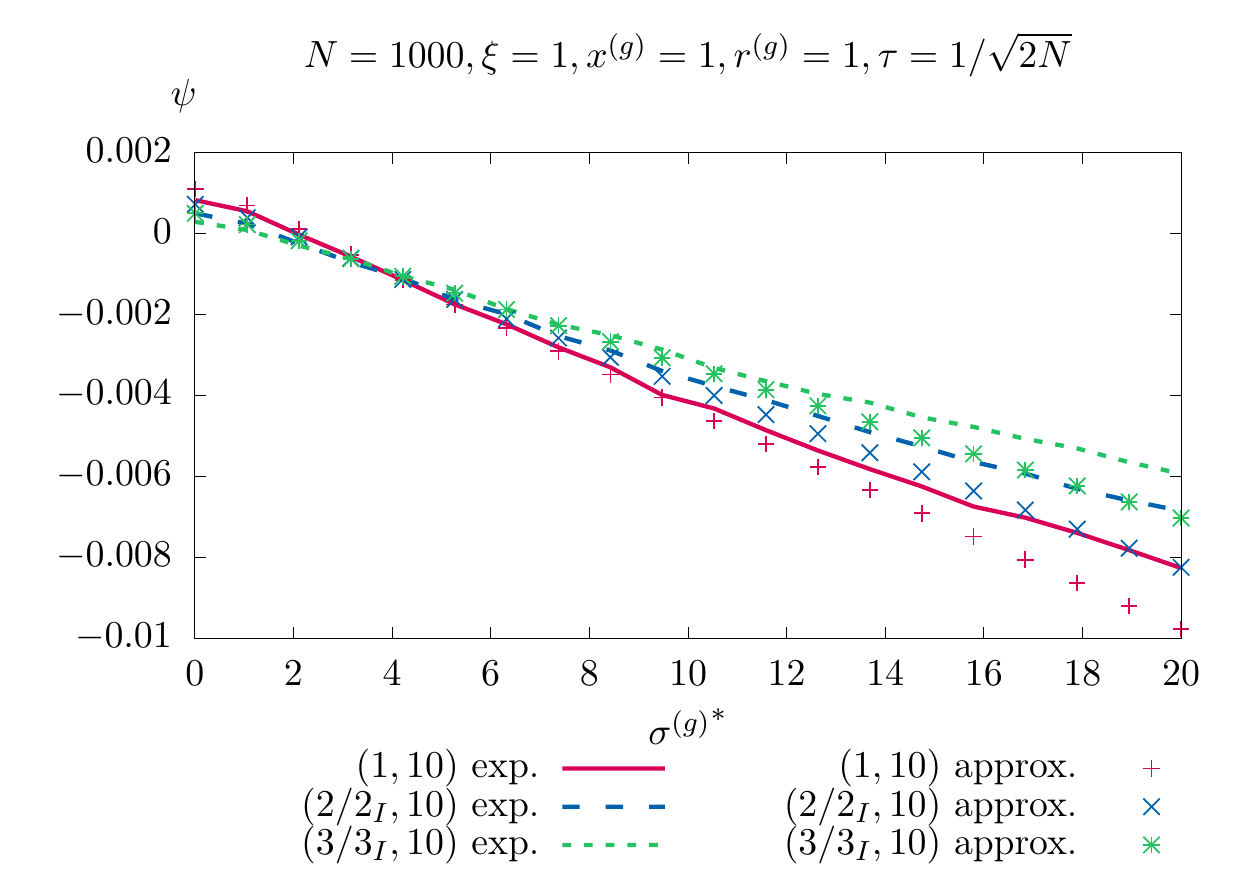}&
    \includegraphics[width=0.5\textwidth]{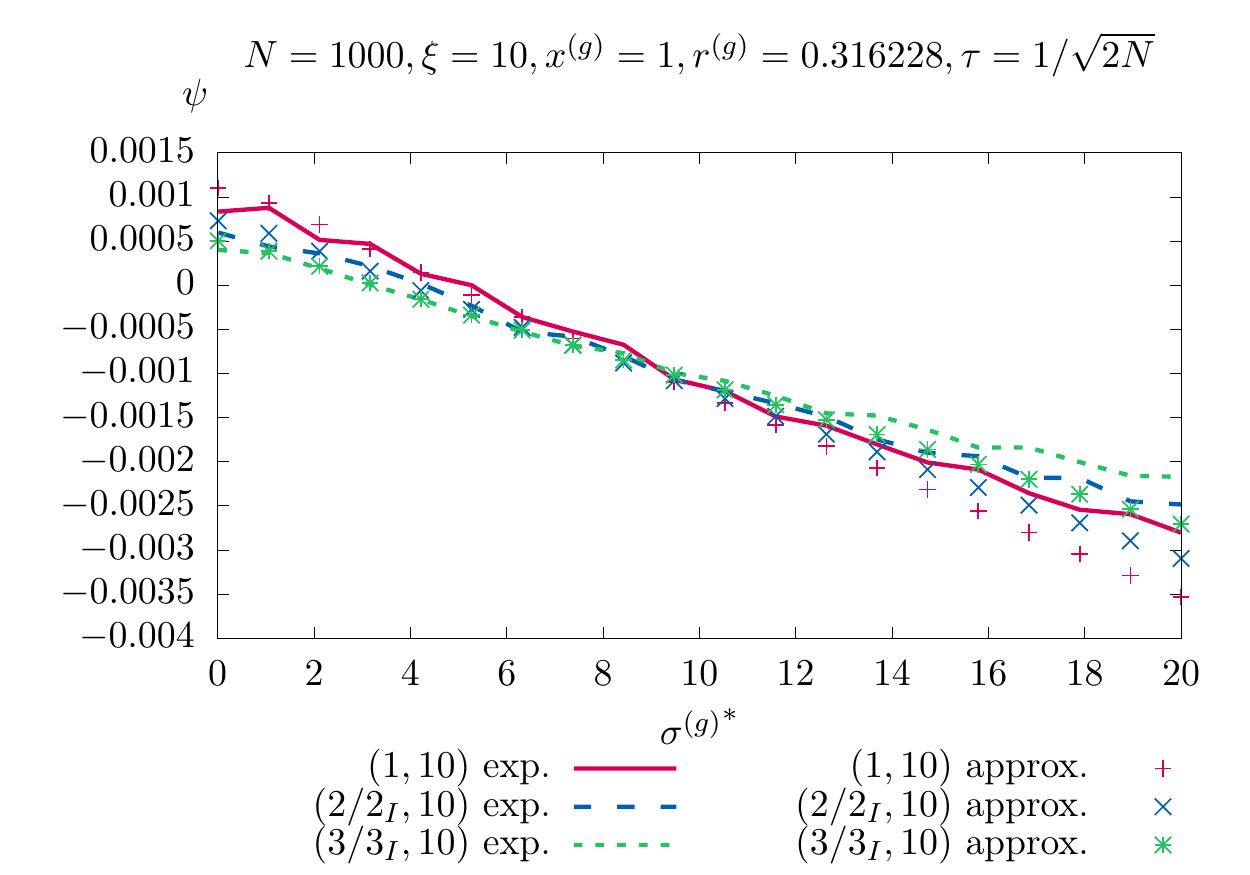}\\
  \end{tabular}
  \caption{Comparison of the SAR approximation with simulations. (Part 3)}
  \label{sec:theoreticalanalysis:fig:sarcomparisonsdetail3}
\end{figure}

\Crefrange{sec:theoreticalanalysis:fig:xprogresscomparisonsdetail1}
          {sec:theoreticalanalysis:fig:sarcomparisonsdetail3}
show plots comparing the derived closed-form
approximation for $\varphi_x^*$, $\varphi_r^*$, and $\psi$,
respectively, with one-generation experiments.
The pluses, crosses, and stars have been calculated by evaluating
\Cref{sec:theoreticalanalysis:eq:varphixnormalizedcombined}
with
\Cref{sec:theoreticalanalysis:eq:Pfeasapprox1,%
sec:theoreticalanalysis:eq:varphixnormalizedfeasible5,%
sec:theoreticalanalysis:eq:varphixnormalizedinfeasible5},
\Cref{sec:theoreticalanalysis:eq:varphirnormalizedcombinedmaintext}
with
\Cref{sec:theoreticalanalysis:eq:Pfeasapprox1,%
sec:theoreticalanalysis:eq:varphixnormalizedinfeasible5}, and
\Cref{sec:theoreticalanalysis:eq:psicombinedmaintext}
with
\Cref{sec:theoreticalanalysis:eq:Pfeasapprox1},
respectively.
The solid, dashed, and dotted lines have been
generated by one-generation experiments.
For this, the generational loop has been
run $10^5$ times for a fixed parental individual and constant parameters.
The experimentally determined values for
$\varphi_x^*$, $\varphi_r^*$, and $\psi$
from those $10^5$ runs have been averaged.
The top row in every
set of $3 \times 2$ plots shows the case that the parental individual
is near the cone axis (note that here the feasible case dominates).
The bottom row in every
set of $3 \times 2$ plots shows the case that the parental individual
is in the vicinity of the cone boundary (note that here the infeasible case
dominates). And the middle row shows a case in between.

\renewcommand{\sppapathtmp}{./figures/figure10/}
\renewcommand{\sppapathtmp}{./figures/figure11/}
\renewcommand{\sppapathtmp}{./figures/figure12/}

\renewcommand{\sppapathtmp}{./figures/figure13/}
\renewcommand{\sppapathtmp}{./figures/figure14/}
\renewcommand{\sppapathtmp}{./figures/figure15/}

\renewcommand{\sppapathtmp}{./figures/figure16/}
\renewcommand{\sppapathtmp}{./figures/figure17/}
\renewcommand{\sppapathtmp}{./figures/figure18/}

\Cref{sec:theoreticalanalysis:fig:dynamicsdetail1,%
sec:theoreticalanalysis:fig:dynamicsdetail2,%
sec:theoreticalanalysis:fig:dynamicsdetail3,%
sec:theoreticalanalysis:fig:dynamicsdetail4,%
sec:theoreticalanalysis:fig:dynamicsdetail5,%
sec:theoreticalanalysis:fig:dynamicsdetail6}
show the mean value dynamics of
the $(3/3_I,10)$-ES
applied to the conically constrained problem with different parameters
as indicated in the title of the subplots.
The plots are organized into
three rows and two columns. The first two rows show the
$x$ (first row, first column),
$r$ (first row, second column),
$\sigma$ (second row, first column),
and $\sigma^*$ (second row, second column) dynamics.
The third row shows $x$ and $r$ converted into each other by $\sqrt{\xi}$.
The third row shows that after some initial phase, the ES transitions into
a stationary state. In this steady state, the ES moves near the
cone boundary. This becomes clear in the plots because the equation for
the cone boundary is $r = x / \sqrt{\xi}$ or equivalently $x = r\sqrt{\xi}$.
Furthermore, in this stationary state, the normalized mutation
strength is constant on average. The lines for the real runs have been generated
by averaging $100$ real runs of the ES. The lines for the
iteration with one-generation experiments have been
determined by iterating the mean value iterative system
with one-generation experiments for
${\varphi^{(g)}_{x}}^*$, ${\varphi^{(g)}_{r}}^*$, and $\psi^{(g)}$.
The lines for the iteration by approximation have been
computed by iterating the mean value iterative system
with the derived approximations in
\Cref{sec:theoreticalanalysis:subsec:microscopic}
for ${\varphi^{(g)}_{x}}^*$, ${\varphi^{(g)}_{r}}^*$, and $\psi^{(g)}$.
Note that due to the approximations used
it is possible that in a generation $g$
the iteration of the mean value iterative system yields infeasible
$(x^{(g)},r^{(g)})^T$. In such cases, the particular $(x^{(g)},r^{(g)})^T$
values have been projected back and projected values used in the further
iterations.

\renewcommand{\sppapathtmp}{./figures/figure19/}
\begin{figure}[H]
  \centering
  \begin{tabular}{@{\hspace{-0.025\textwidth}}c@{\hspace{-0.025\textwidth}}c}
    \includegraphics[width=0.5\textwidth]{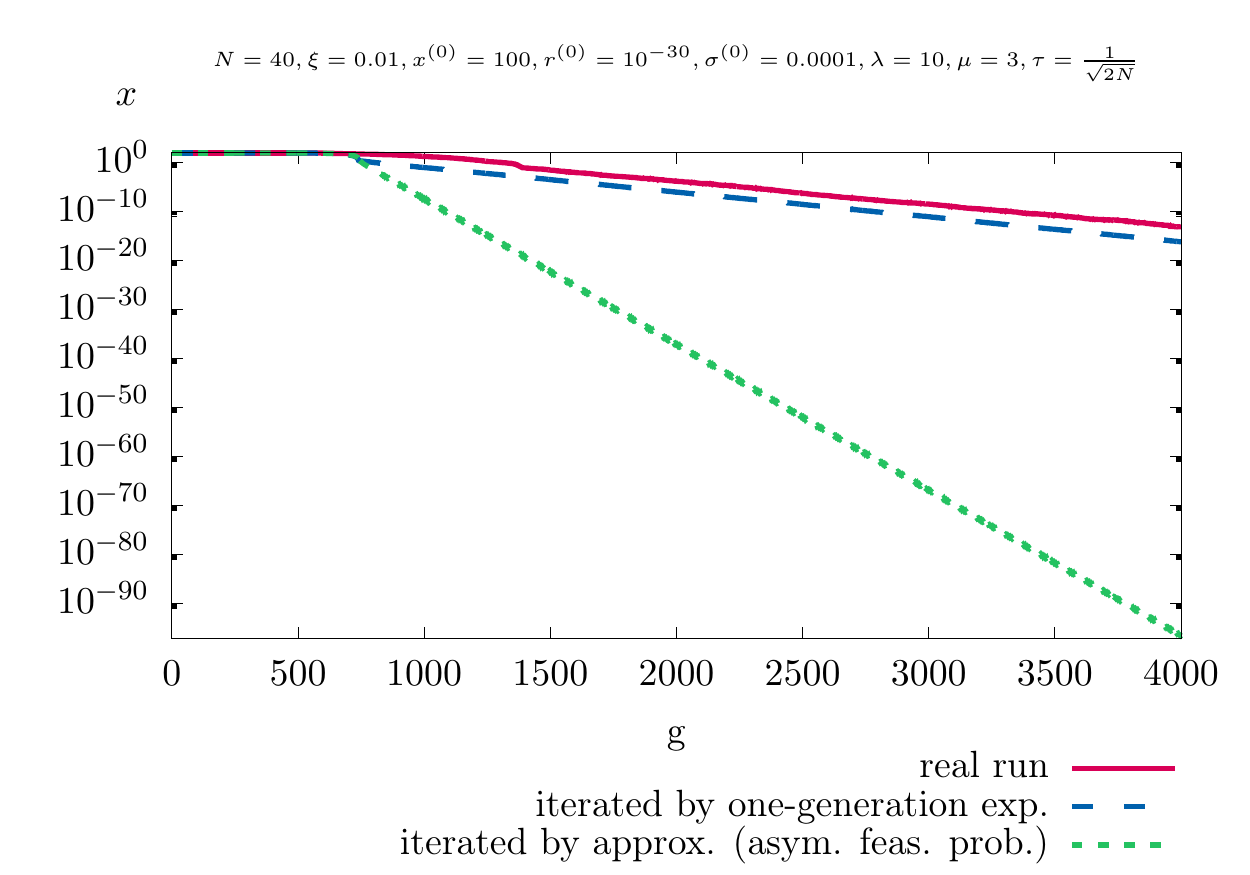}&
    \includegraphics[width=0.5\textwidth]{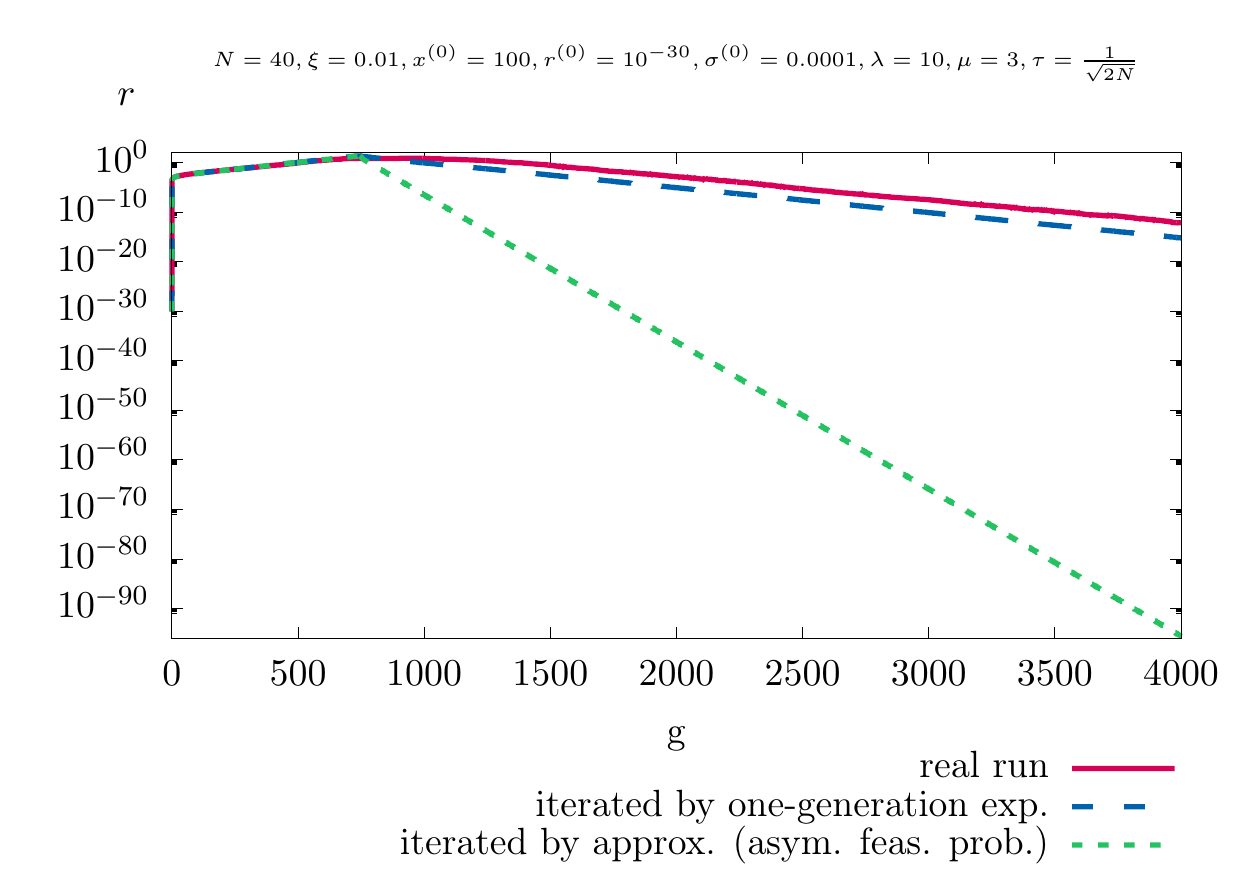}\\
    \includegraphics[width=0.5\textwidth]{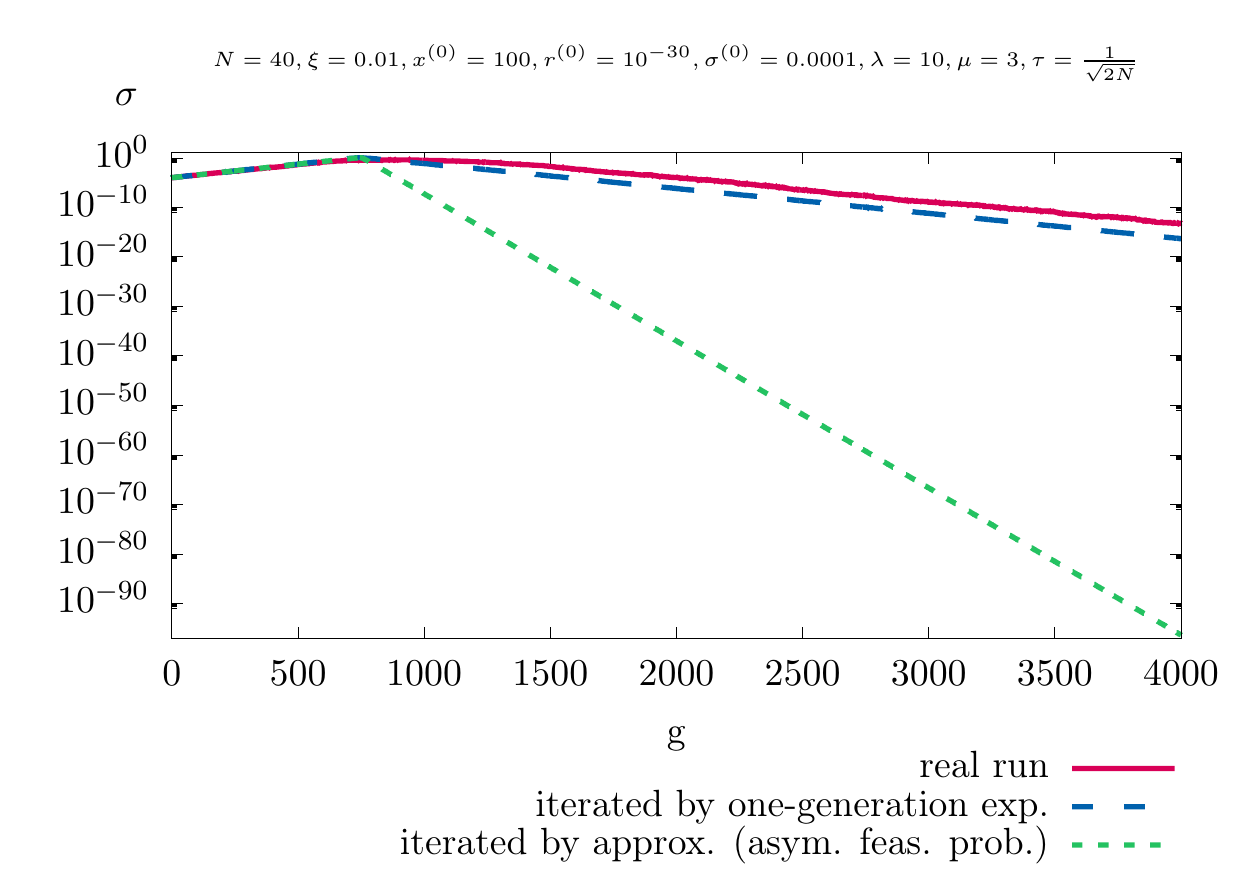}&
    \includegraphics[width=0.5\textwidth]{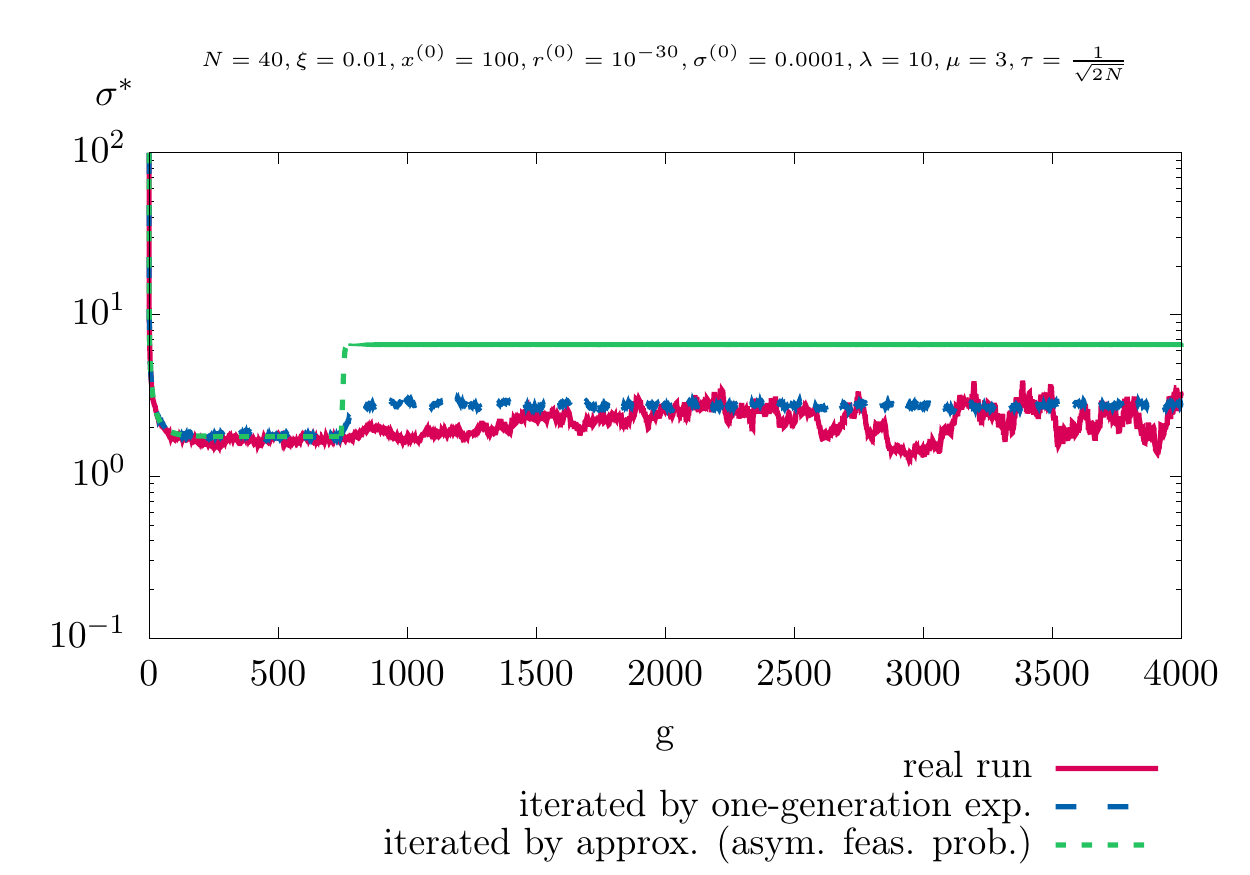}\\
    \includegraphics[width=0.5\textwidth]{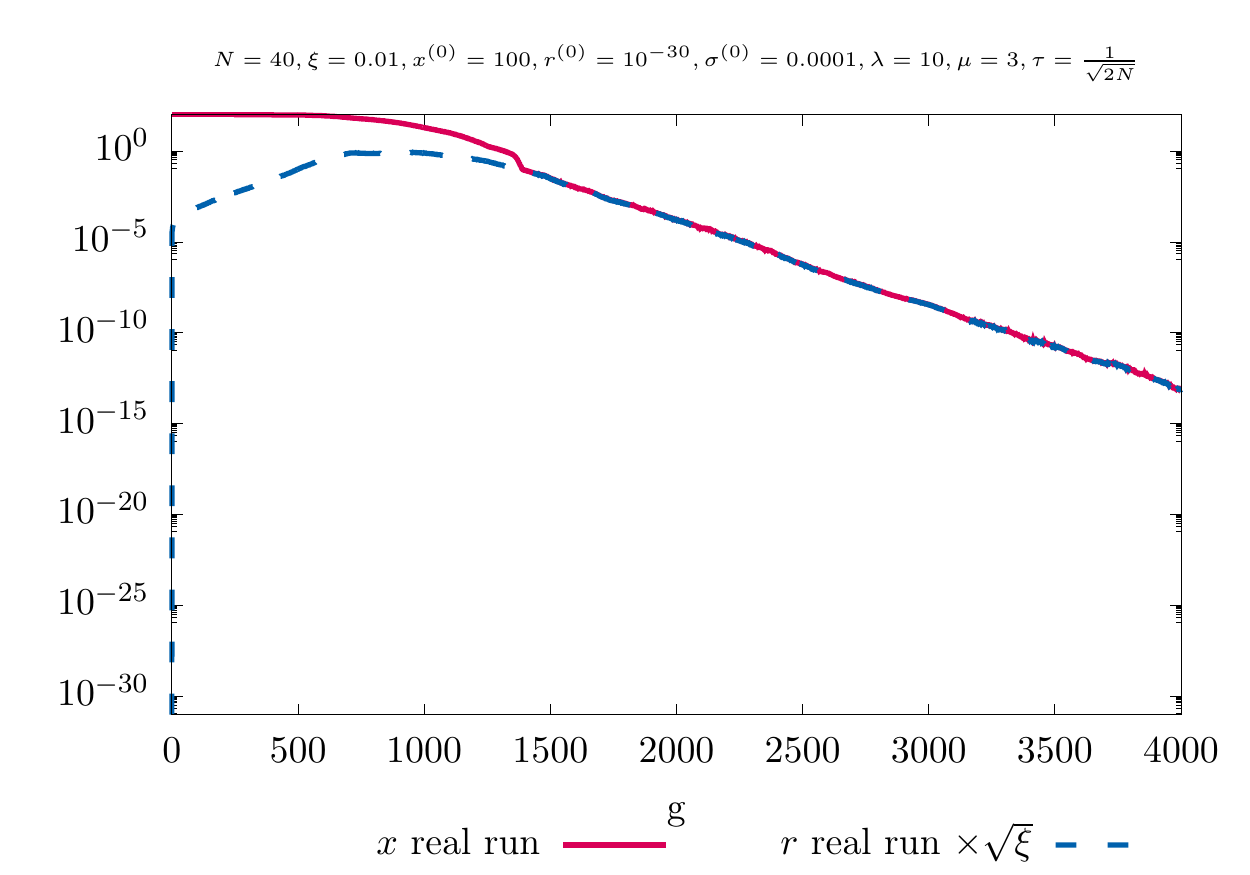}&
    \includegraphics[width=0.5\textwidth]{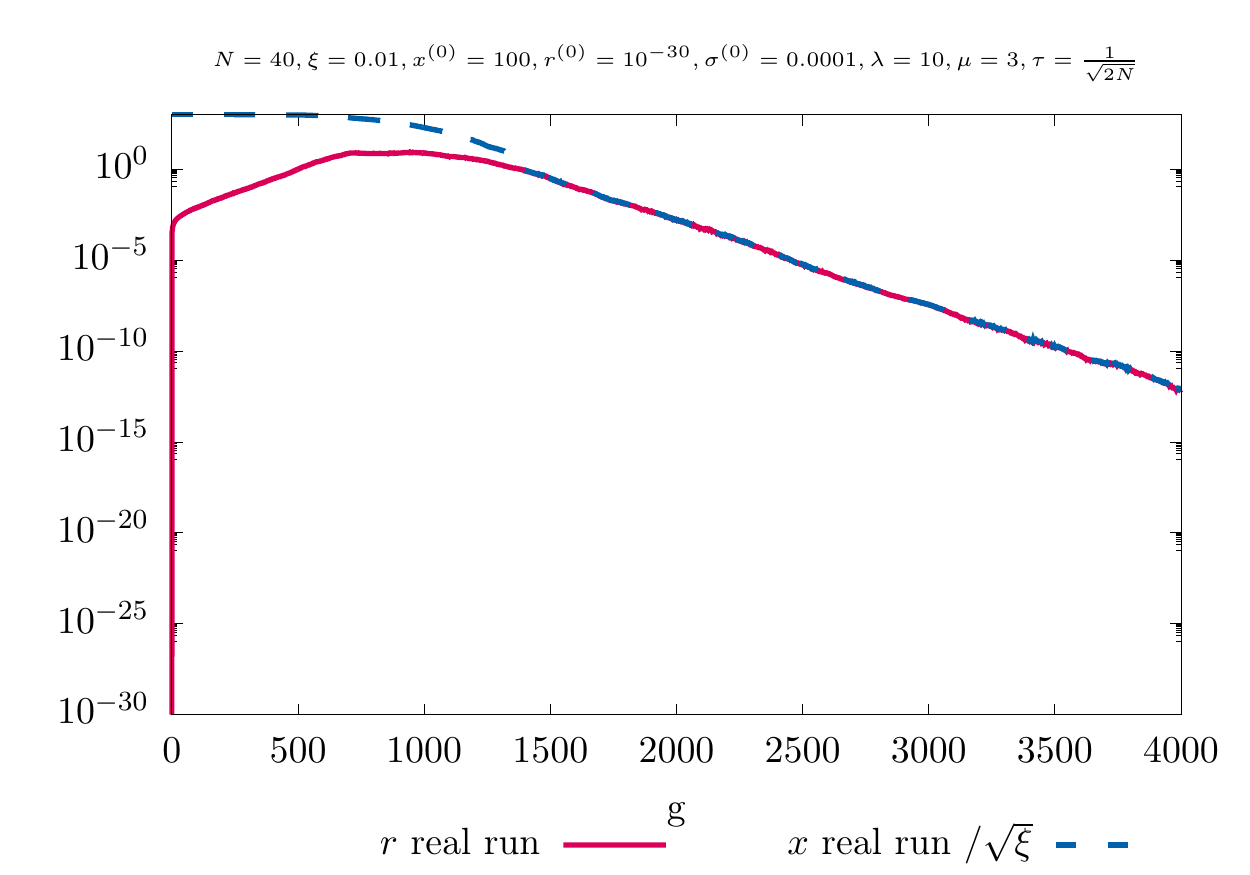}
  \end{tabular}
  \caption{Mean value dynamics closed-form approximation and real-run
    comparison of the
    $(3/3_I,10)$-ES
    with repair by projection
    applied to the conically constrained problem. (Part 1)}
  \label{sec:theoreticalanalysis:fig:dynamicsdetail1}
\end{figure}
\renewcommand{\sppapathtmp}{./figures/figure20/}
\begin{figure}[H]
  \centering
  \begin{tabular}{@{\hspace{-0.025\textwidth}}c@{\hspace{-0.025\textwidth}}c}
    \includegraphics[width=0.5\textwidth]{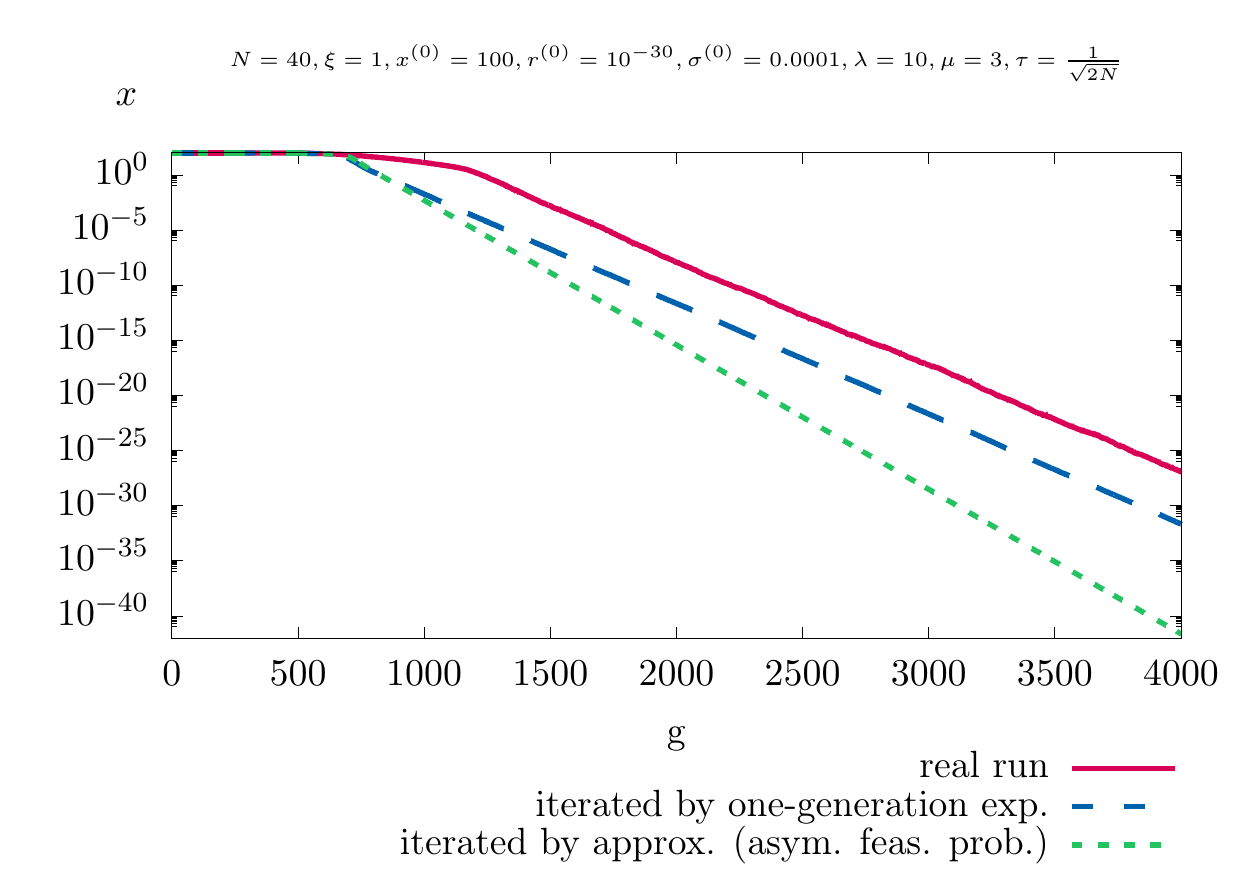}&
    \includegraphics[width=0.5\textwidth]{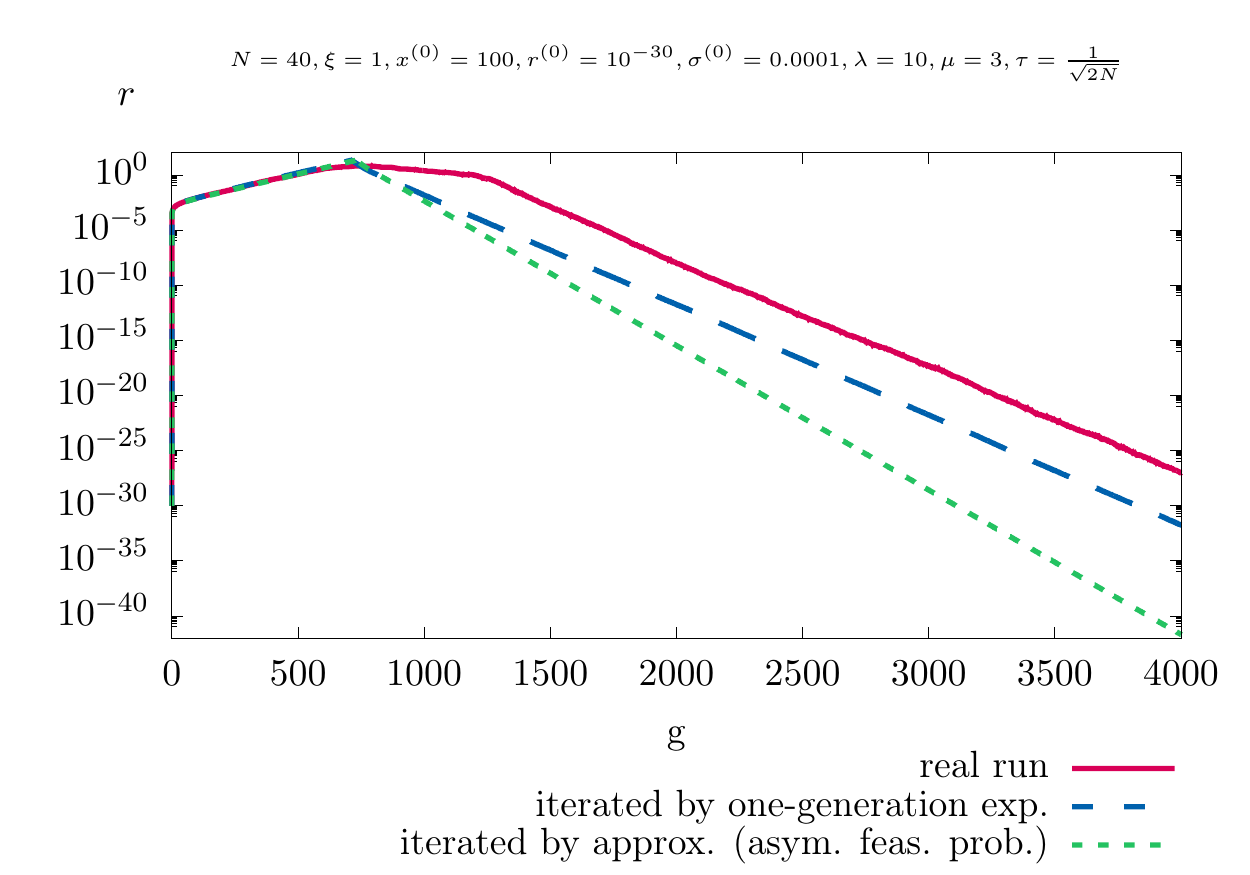}\\
    \includegraphics[width=0.5\textwidth]{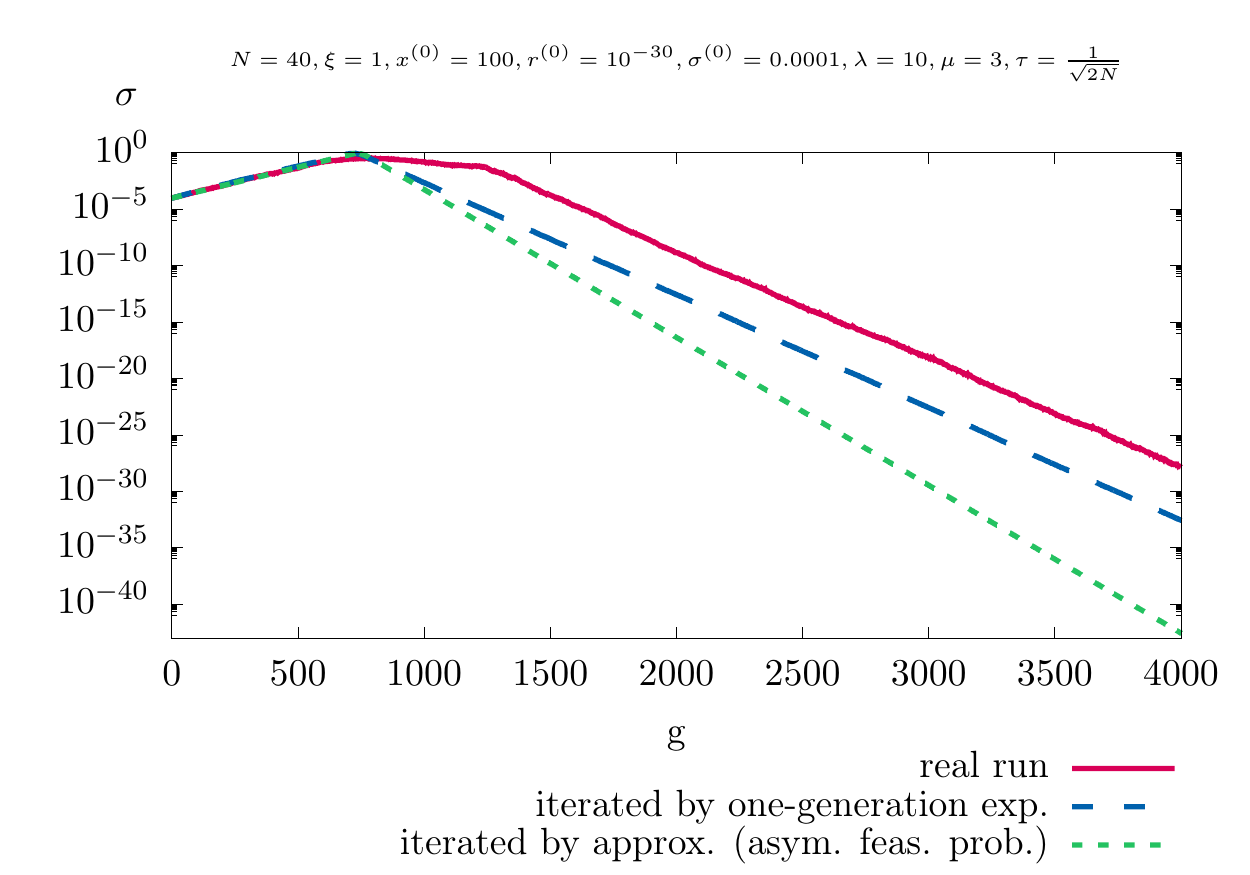}&
    \includegraphics[width=0.5\textwidth]{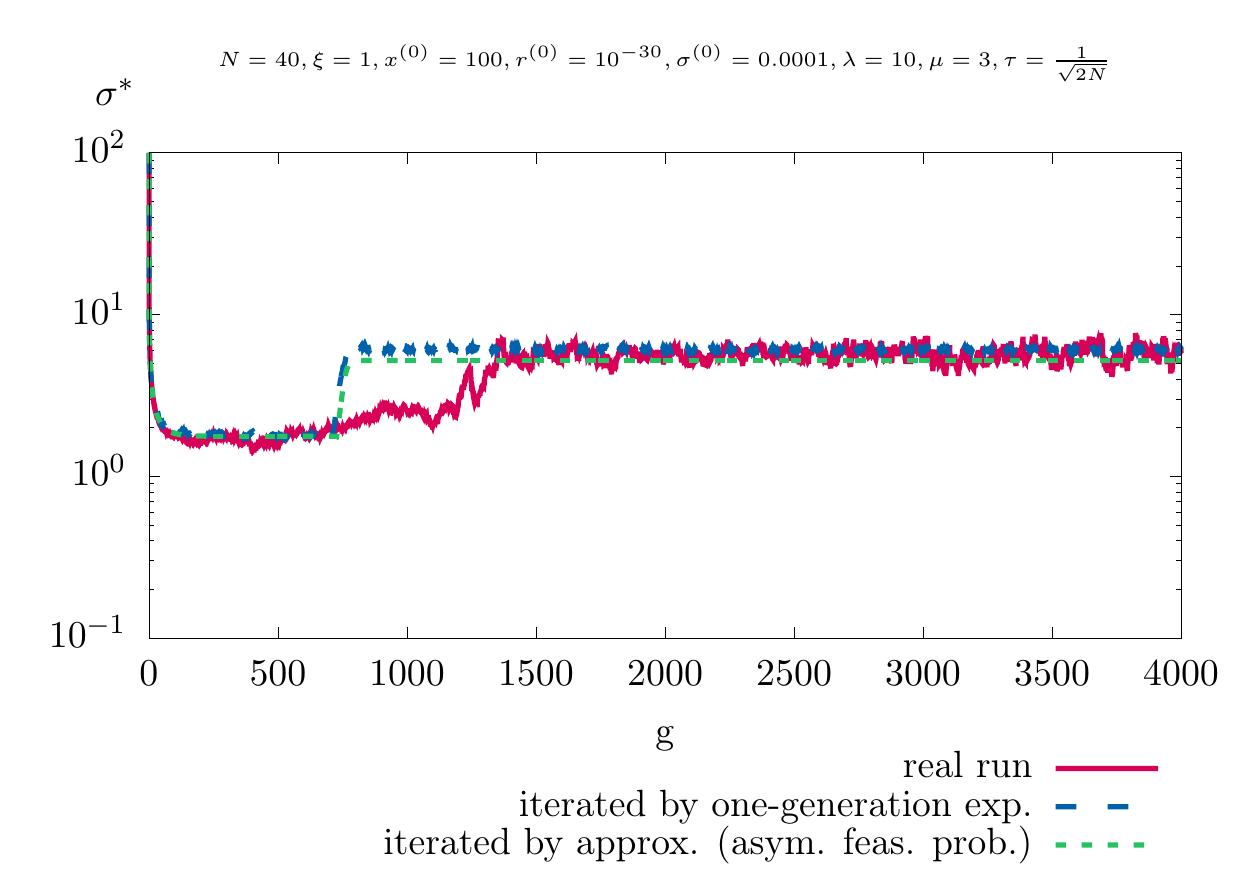}\\
    \includegraphics[width=0.5\textwidth]{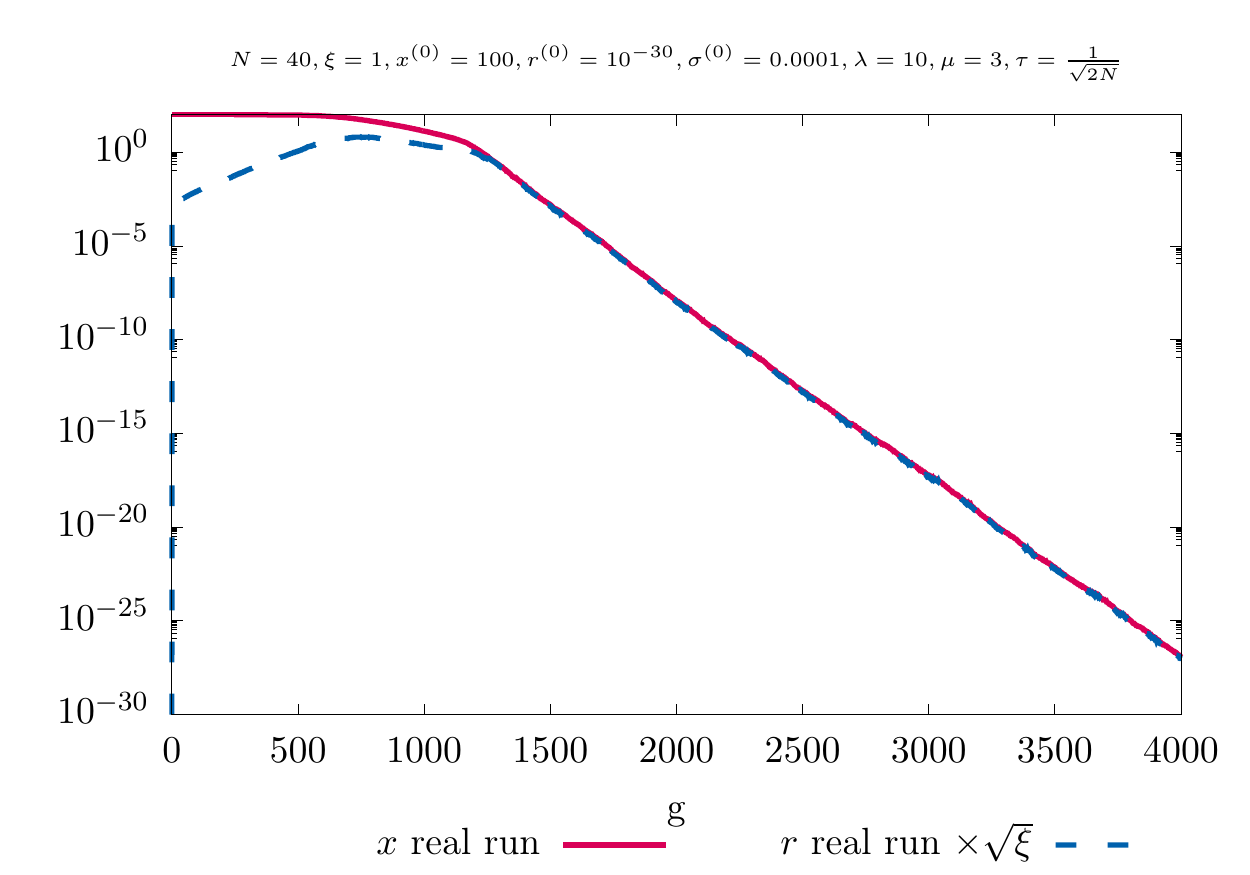}&
    \includegraphics[width=0.5\textwidth]{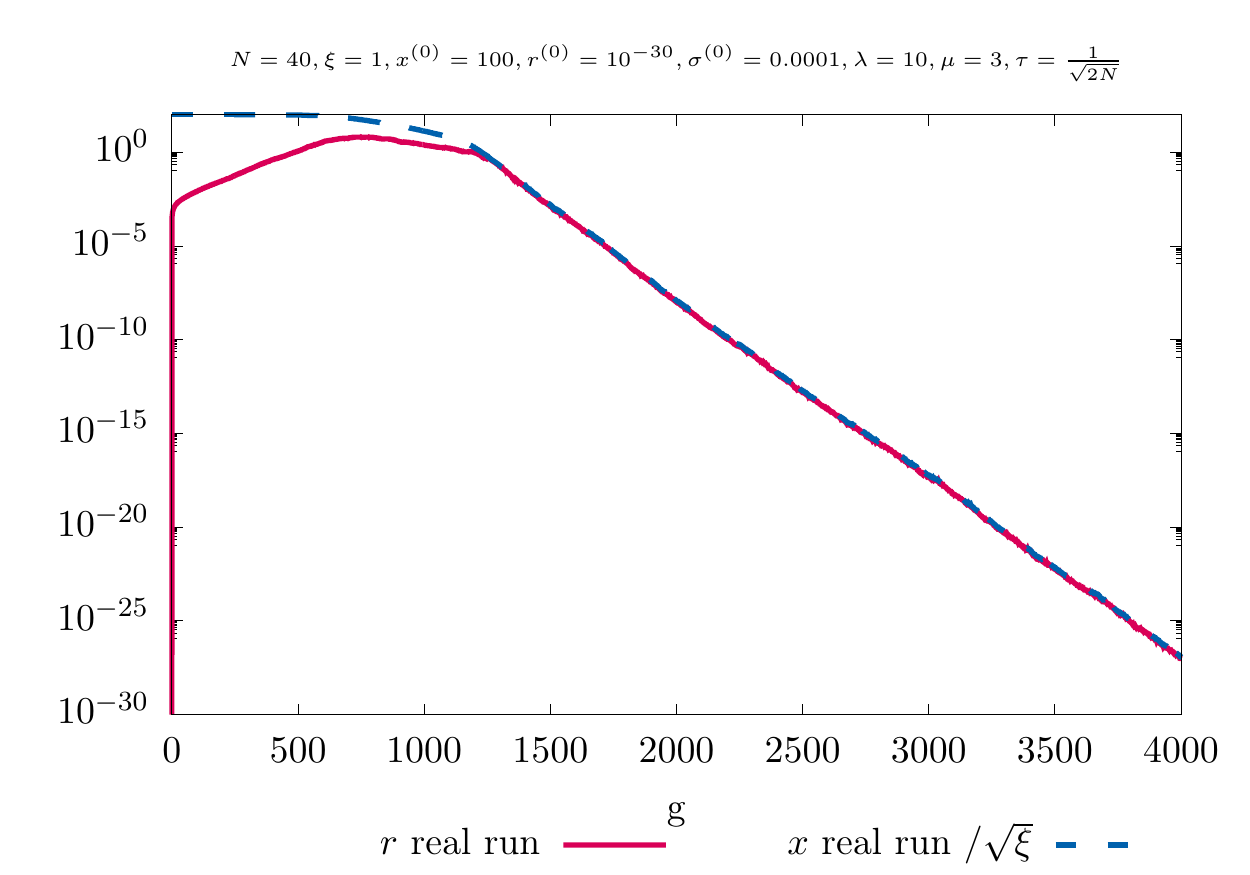}
  \end{tabular}
  \caption{Mean value dynamics closed-form approximation and real-run
    comparison of the
    $(3/3_I,10)$-ES
    with repair by projection
    applied to the conically constrained problem. (Part 2)}
  \label{sec:theoreticalanalysis:fig:dynamicsdetail2}
\end{figure}
\renewcommand{\sppapathtmp}{./figures/figure21/}
\begin{figure}[H]
  \centering
  \begin{tabular}{@{\hspace{-0.025\textwidth}}c@{\hspace{-0.025\textwidth}}c}
    \includegraphics[width=0.5\textwidth]{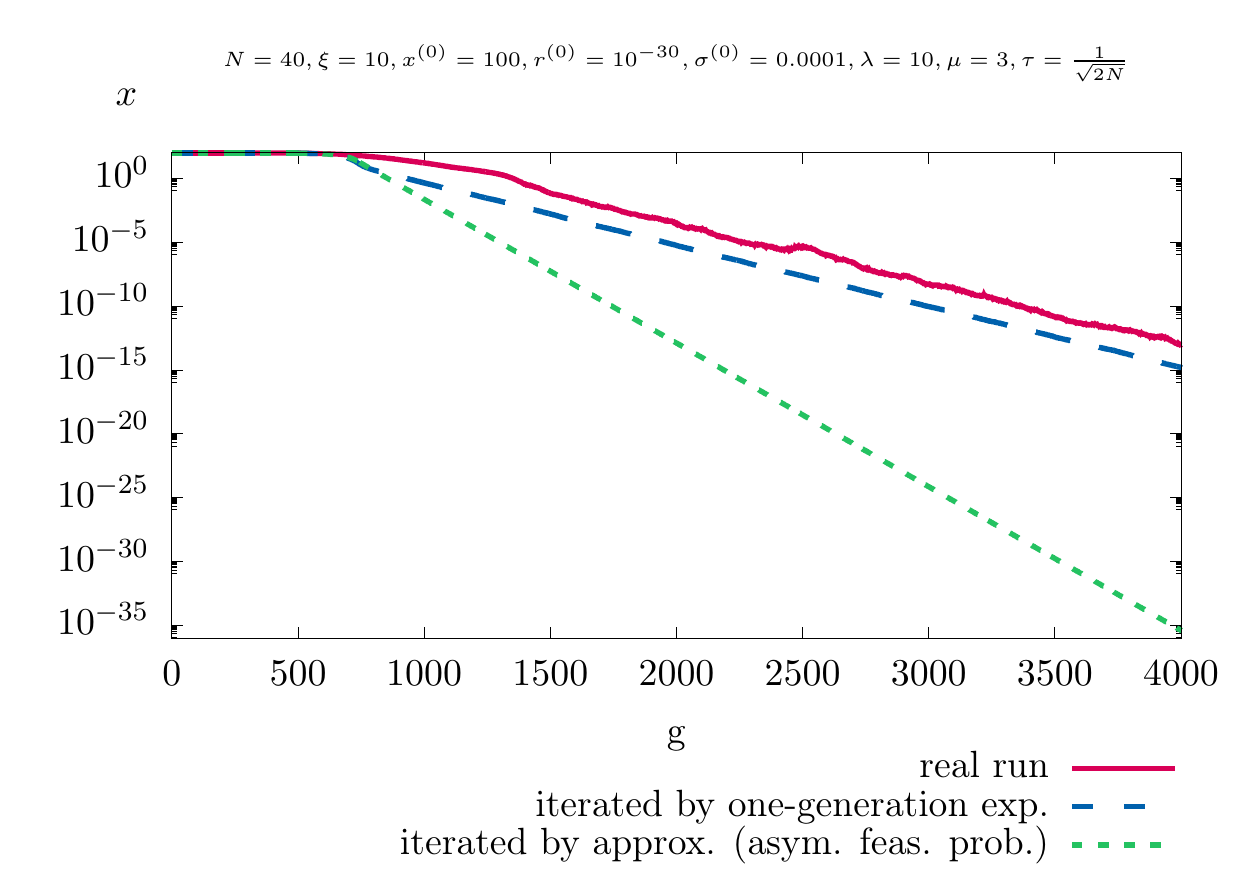}&
    \includegraphics[width=0.5\textwidth]{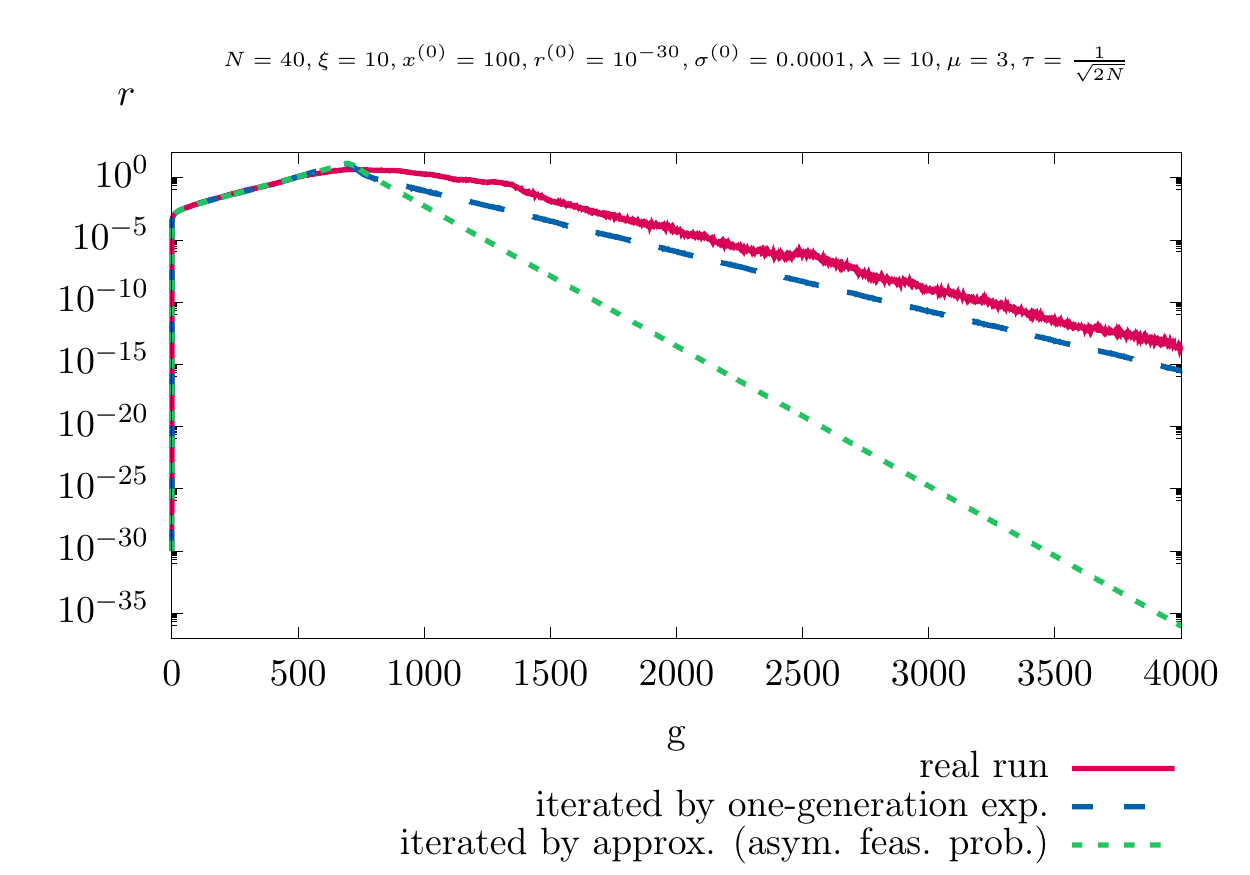}\\
    \includegraphics[width=0.5\textwidth]{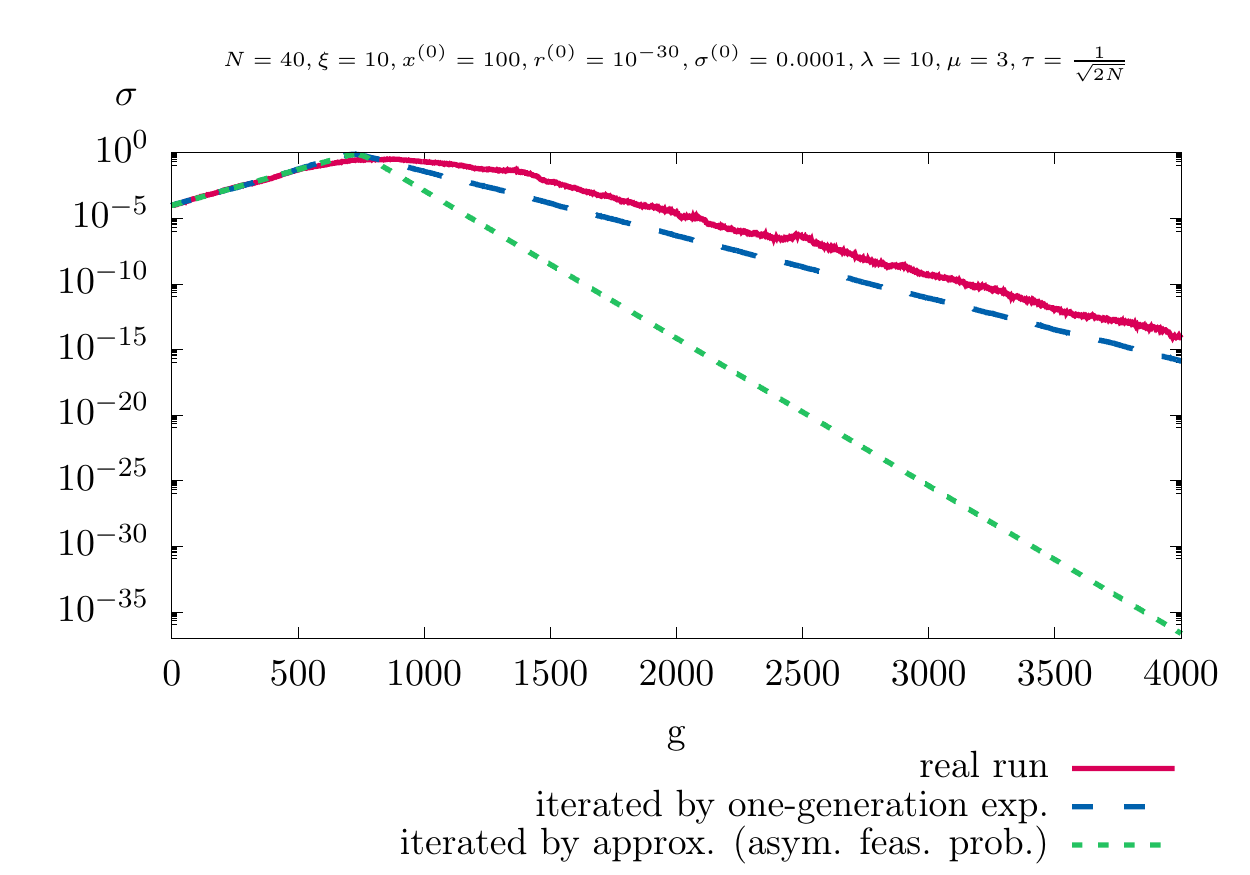}&
    \includegraphics[width=0.5\textwidth]{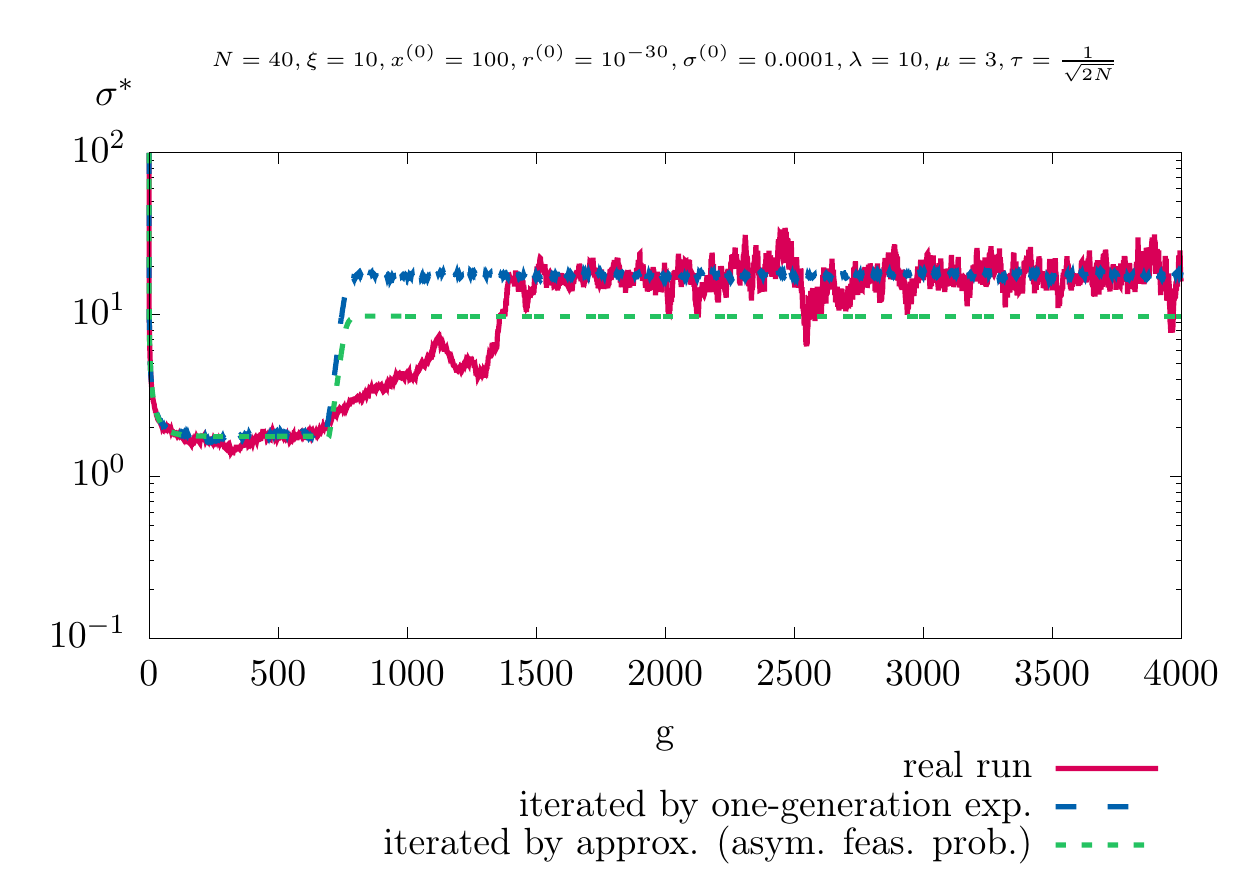}\\
    \includegraphics[width=0.5\textwidth]{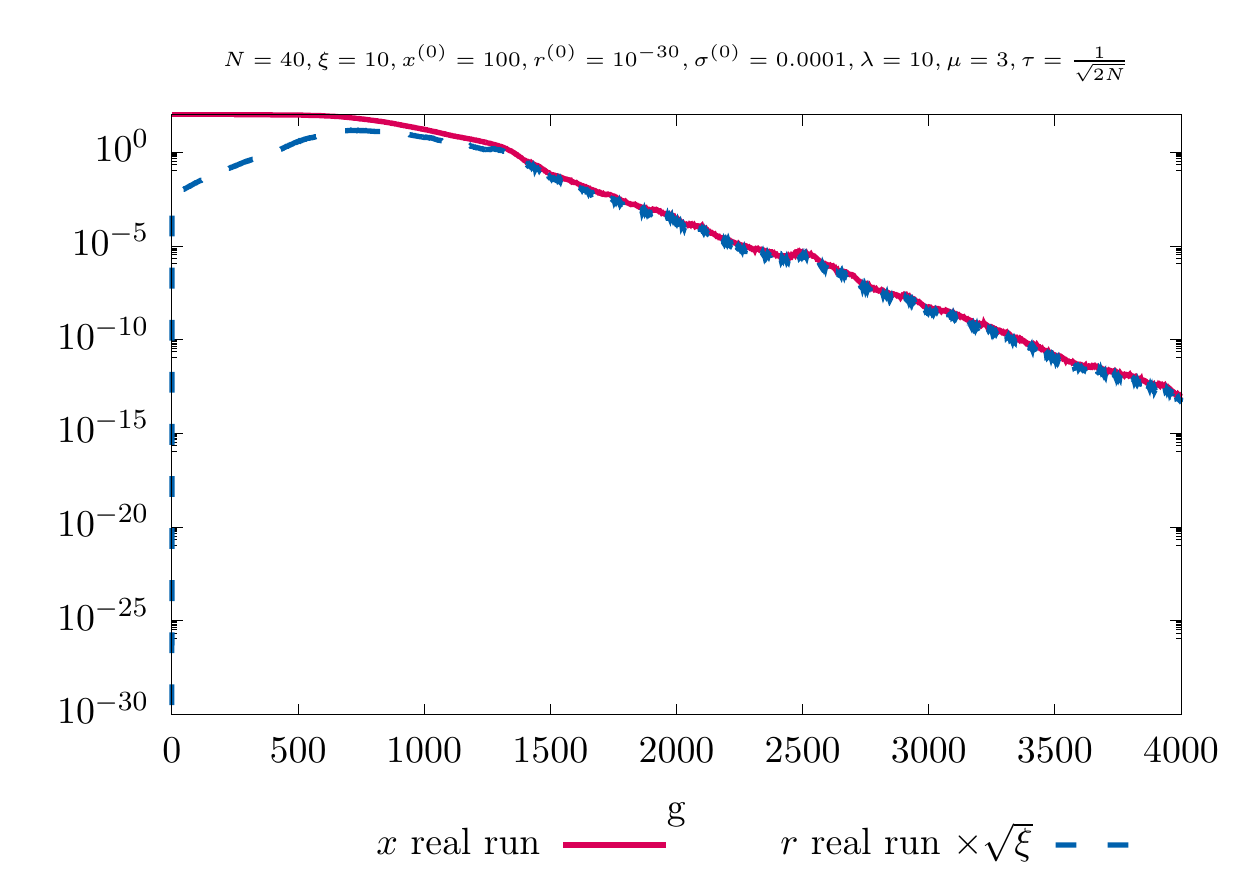}&
    \includegraphics[width=0.5\textwidth]{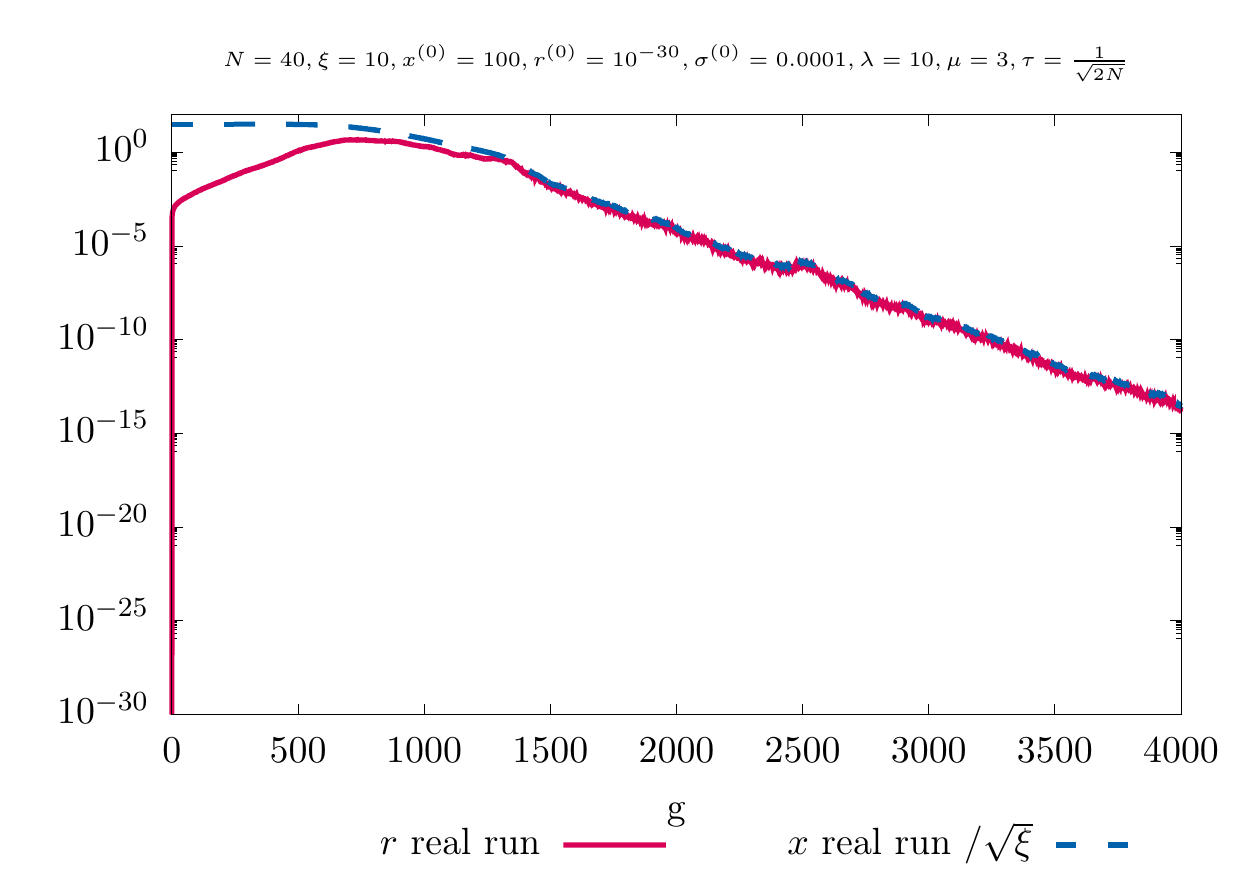}
  \end{tabular}
  \caption{Mean value dynamics closed-form approximation and real-run
    comparison of the
    $(3/3_I,10)$-ES
    with repair by projection
    applied to the conically constrained problem. (Part 3)}
  \label{sec:theoreticalanalysis:fig:dynamicsdetail3}
\end{figure}
\renewcommand{\sppapathtmp}{./figures/figure22/}
\begin{figure}[H]
  \centering
  \begin{tabular}{@{\hspace{-0.025\textwidth}}c@{\hspace{-0.025\textwidth}}c}
    \includegraphics[width=0.5\textwidth]{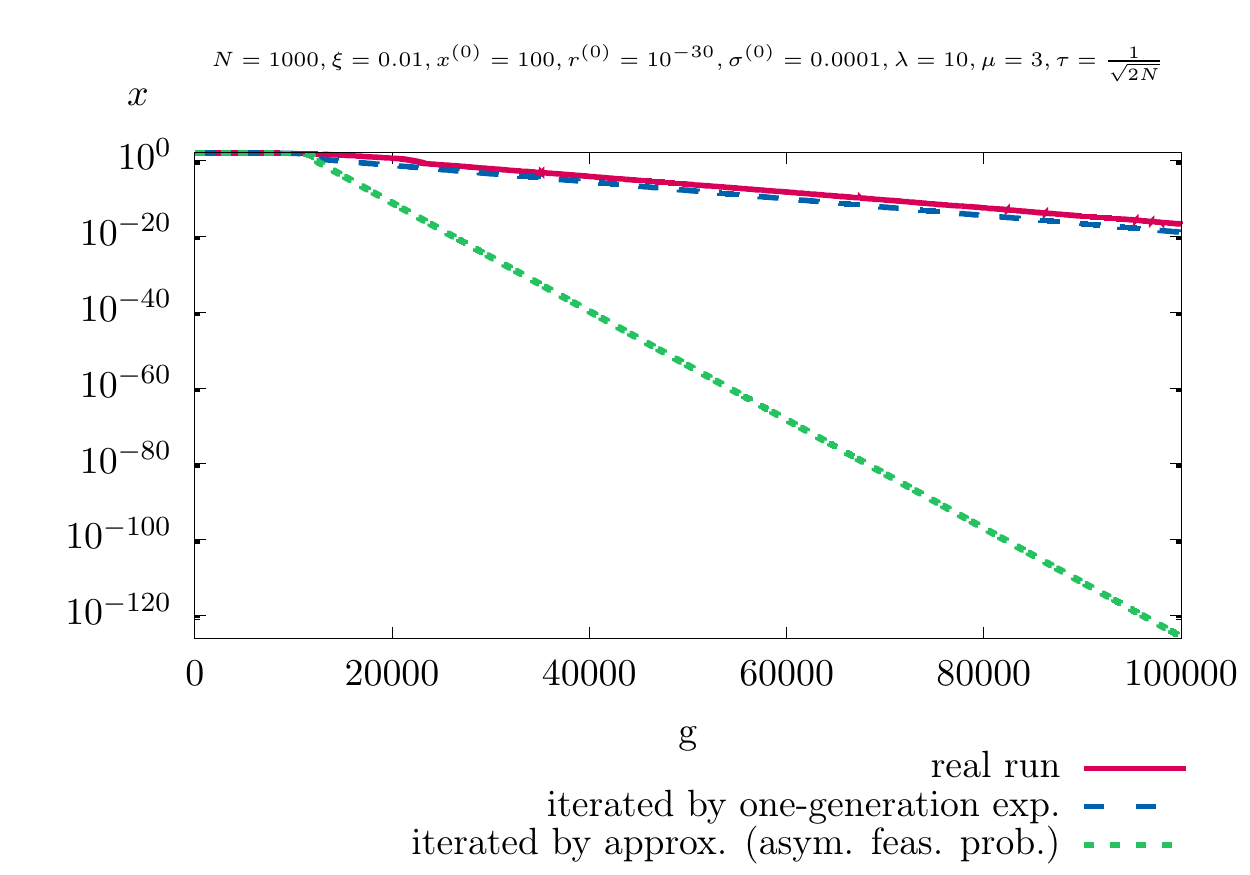}&
    \includegraphics[width=0.5\textwidth]{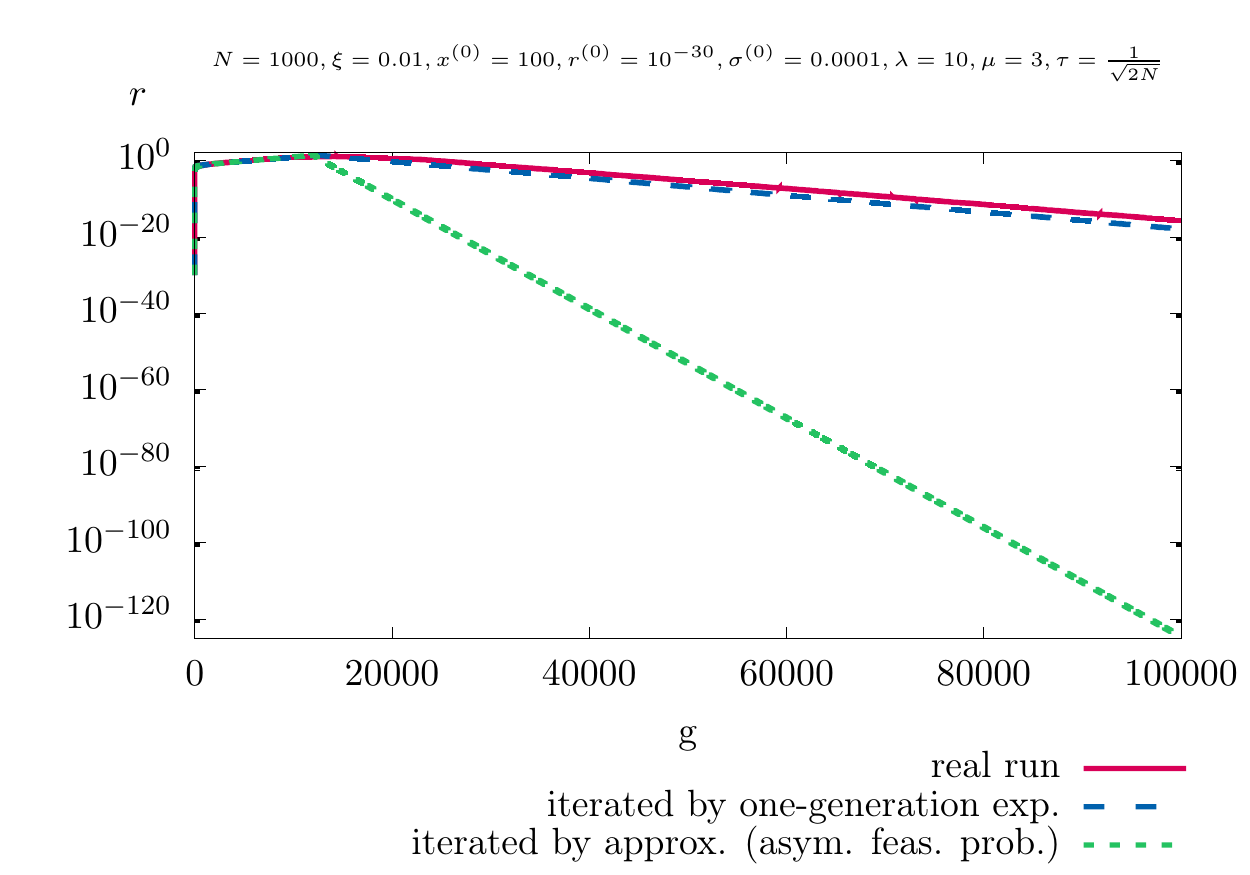}\\
    \includegraphics[width=0.5\textwidth]{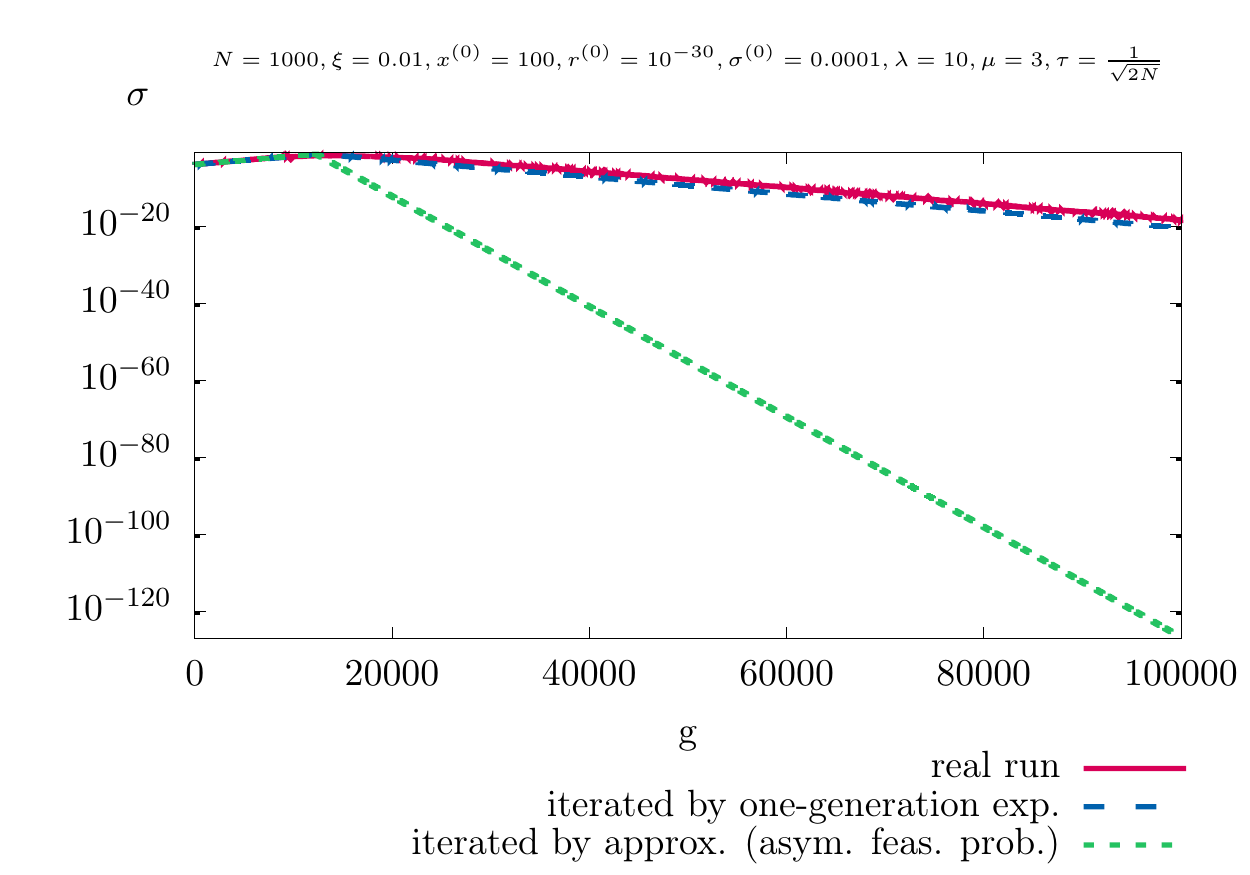}&
    \includegraphics[width=0.5\textwidth]{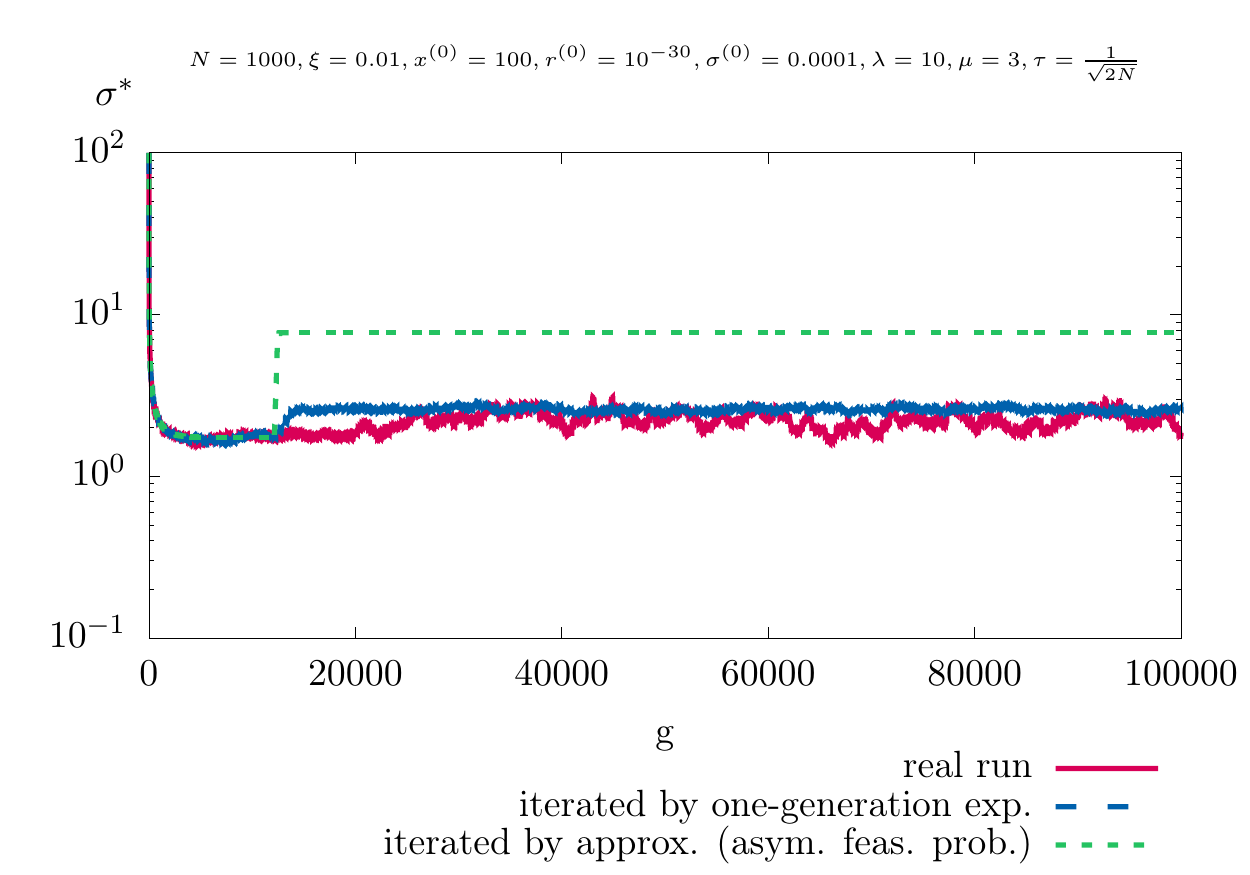}\\
    \includegraphics[width=0.5\textwidth]{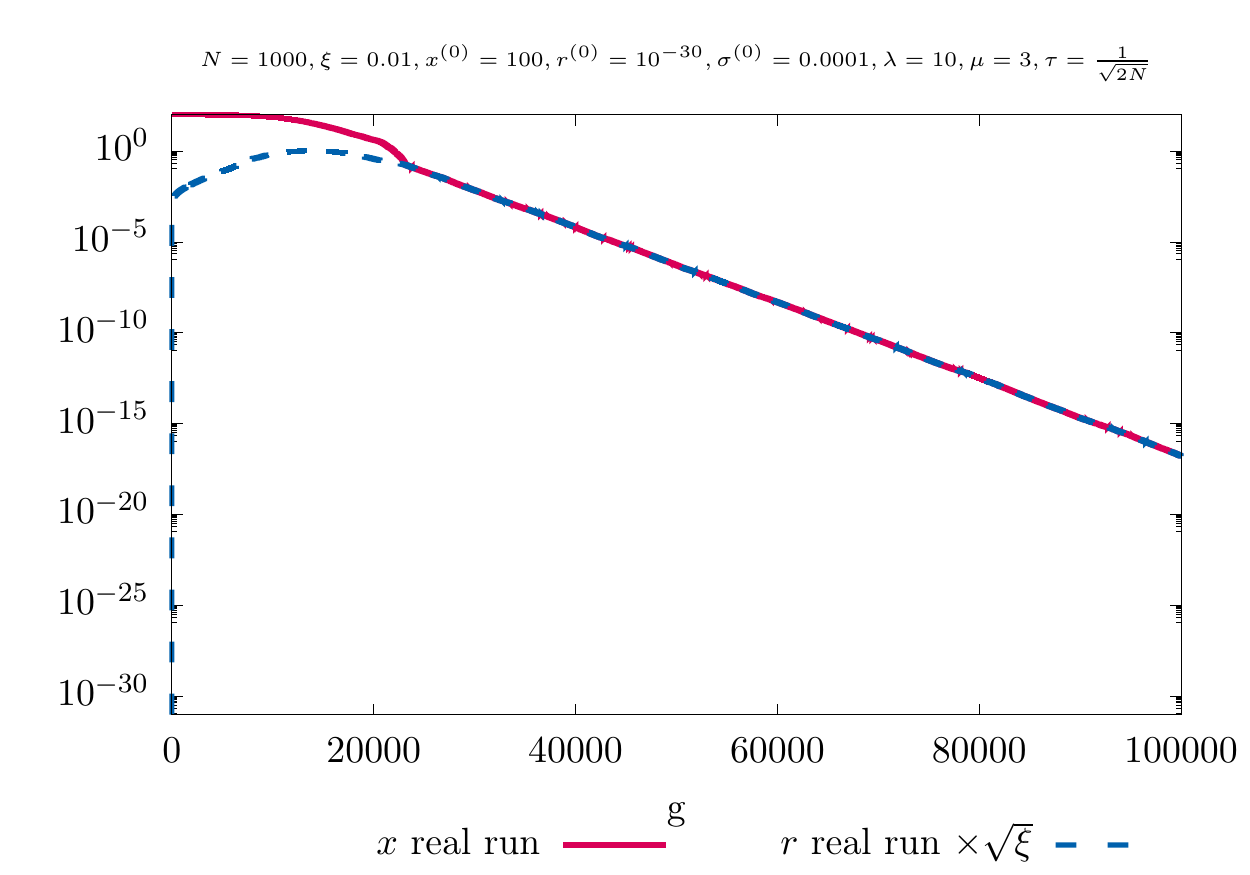}&
    \includegraphics[width=0.5\textwidth]{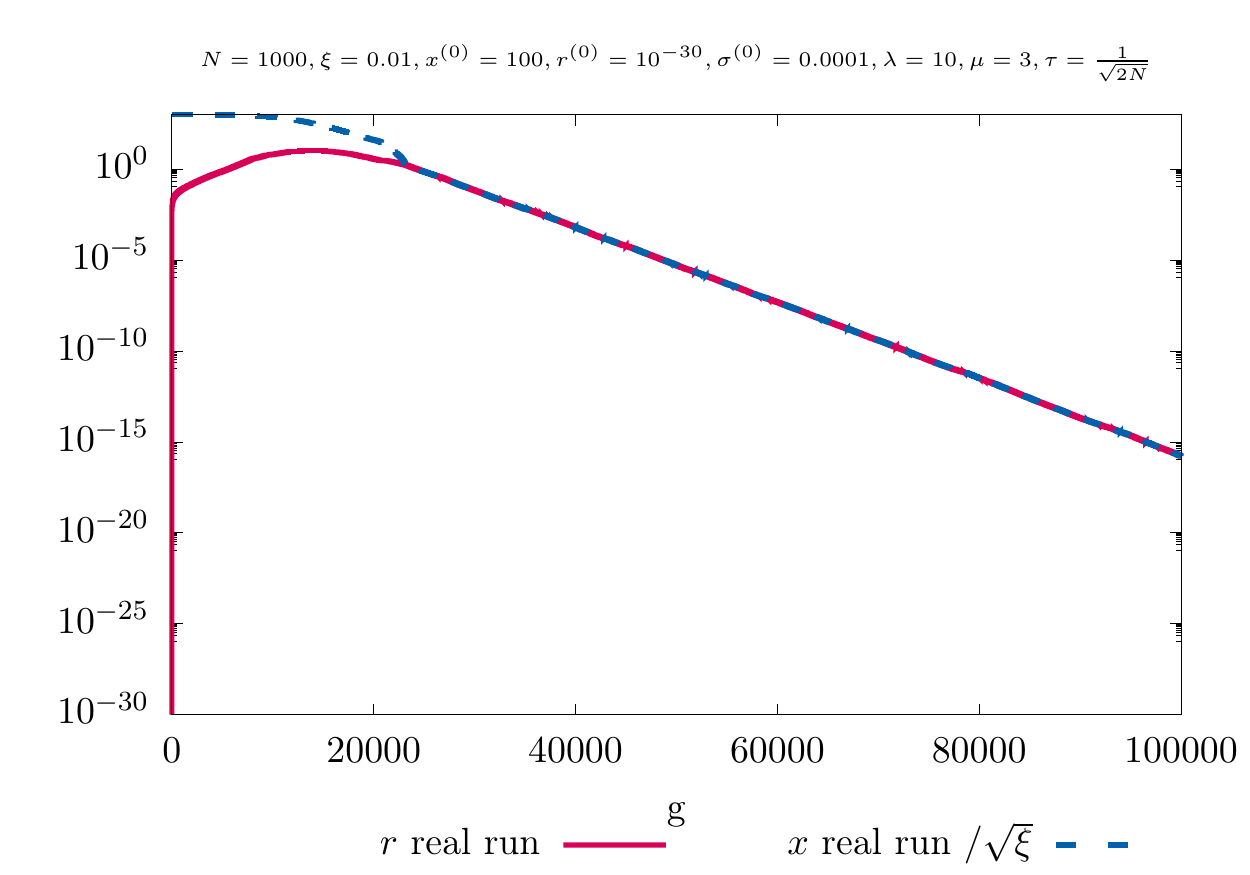}
  \end{tabular}
  \caption{Mean value dynamics closed-form approximation and real-run
    comparison of the
    $(3/3_I,10)$-ES
    with repair by projection
    applied to the conically constrained problem. (Part 4)}
  \label{sec:theoreticalanalysis:fig:dynamicsdetail4}
\end{figure}
\renewcommand{\sppapathtmp}{./figures/figure23/}
\begin{figure}[H]
  \centering
  \begin{tabular}{@{\hspace{-0.025\textwidth}}c@{\hspace{-0.025\textwidth}}c}
    \includegraphics[width=0.5\textwidth]{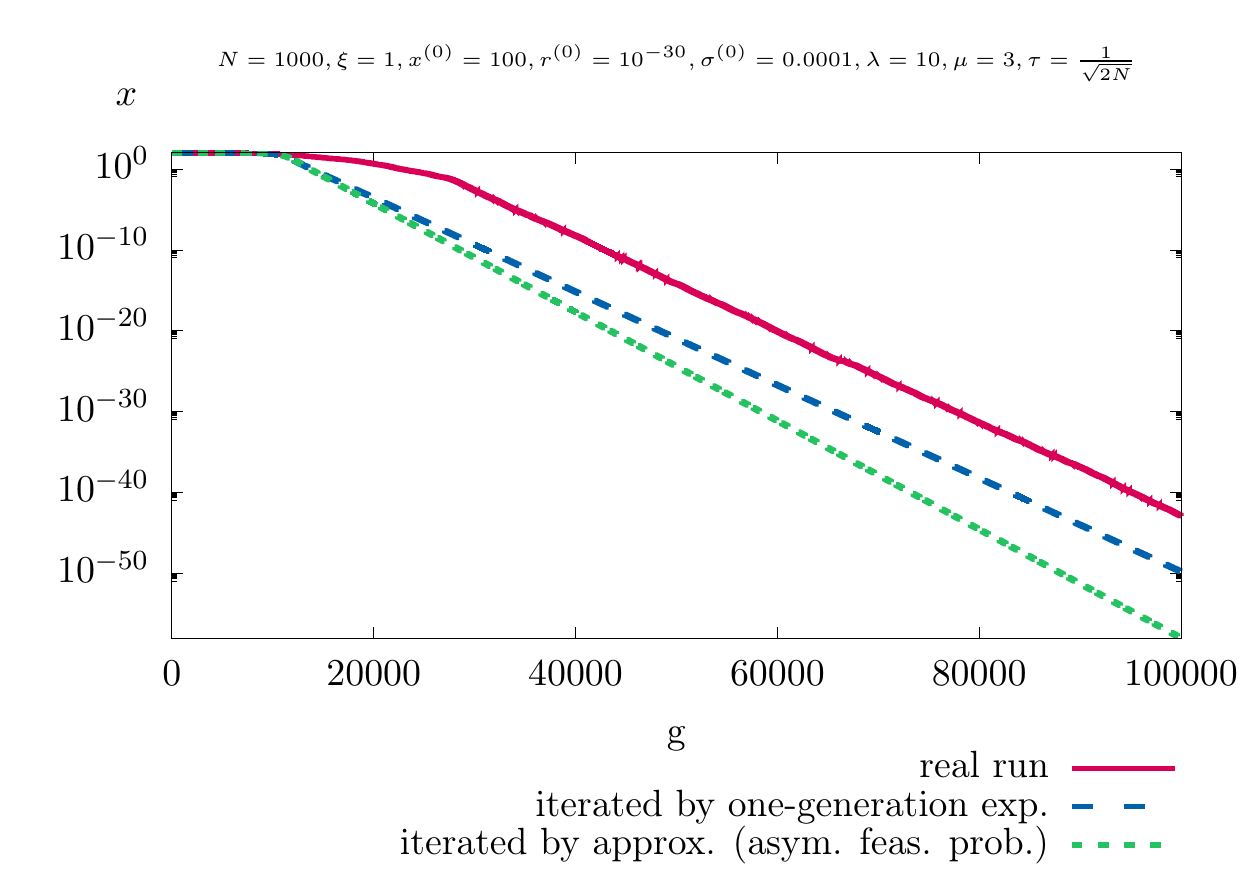}&
    \includegraphics[width=0.5\textwidth]{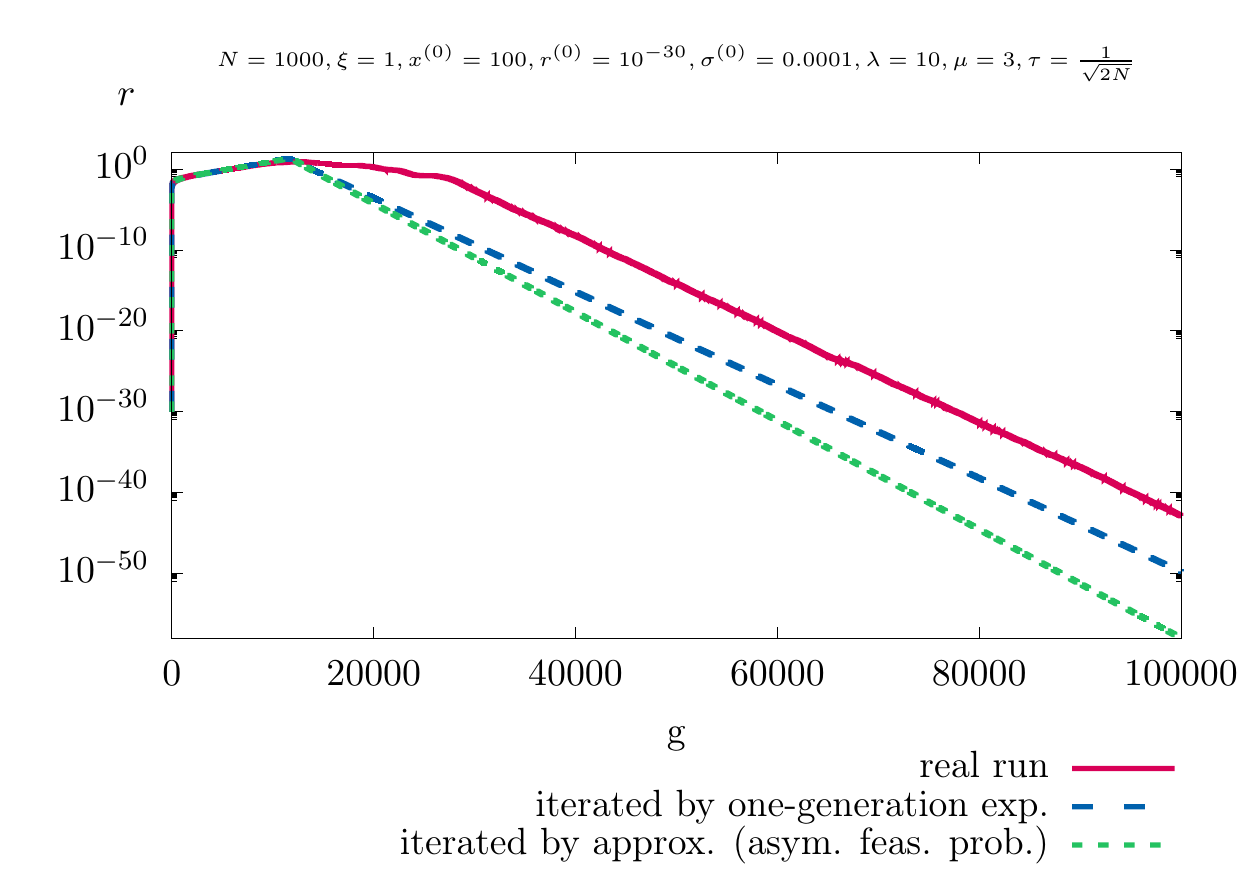}\\
    \includegraphics[width=0.5\textwidth]{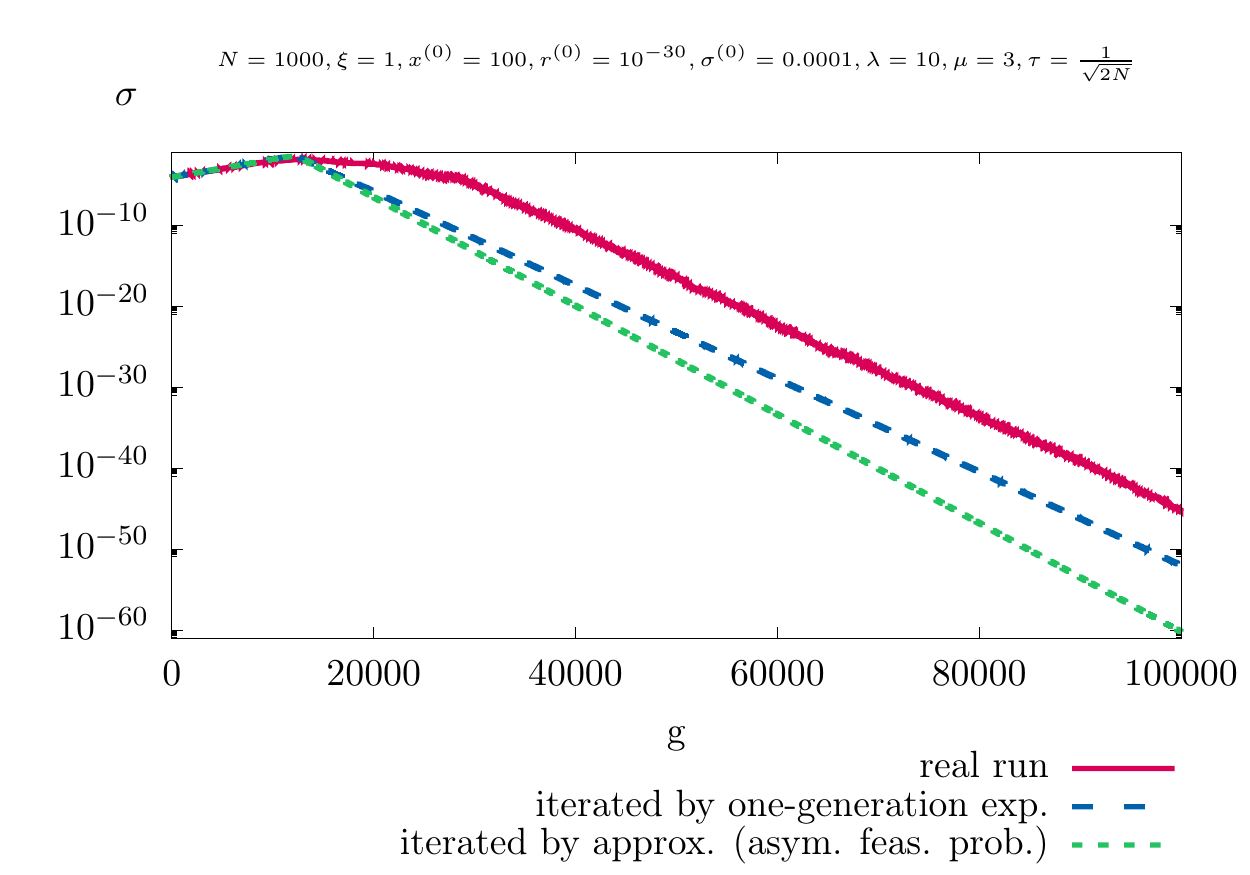}&
    \includegraphics[width=0.5\textwidth]{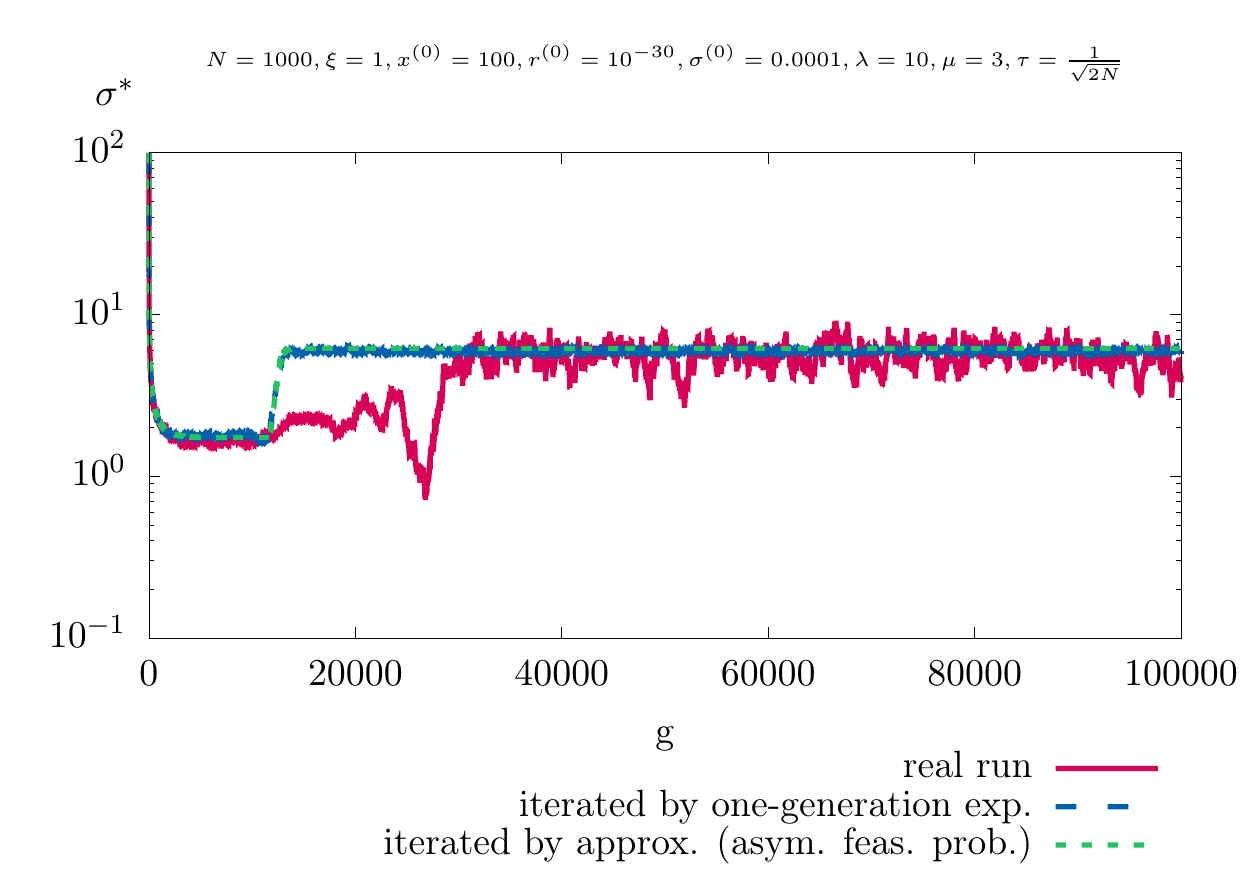}\\
    \includegraphics[width=0.5\textwidth]{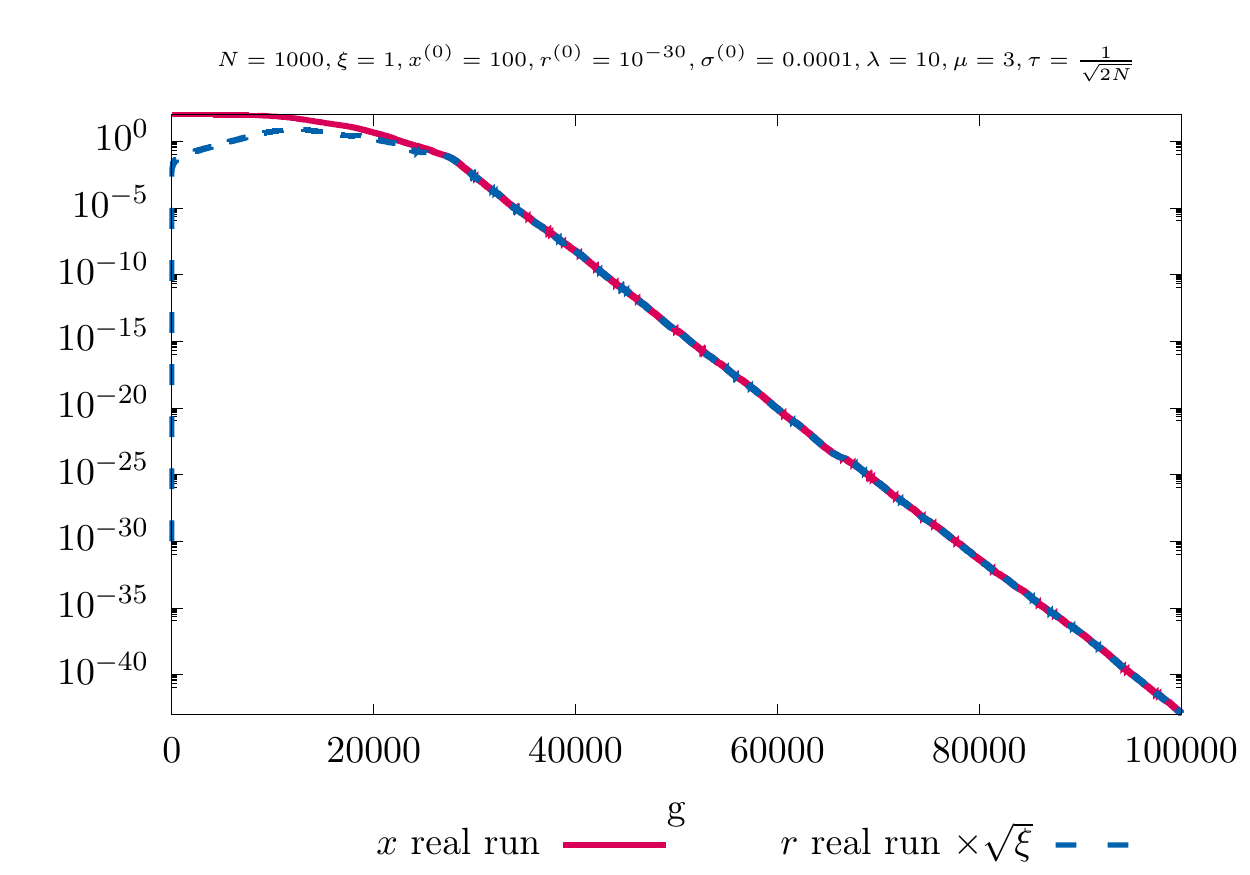}&
    \includegraphics[width=0.5\textwidth]{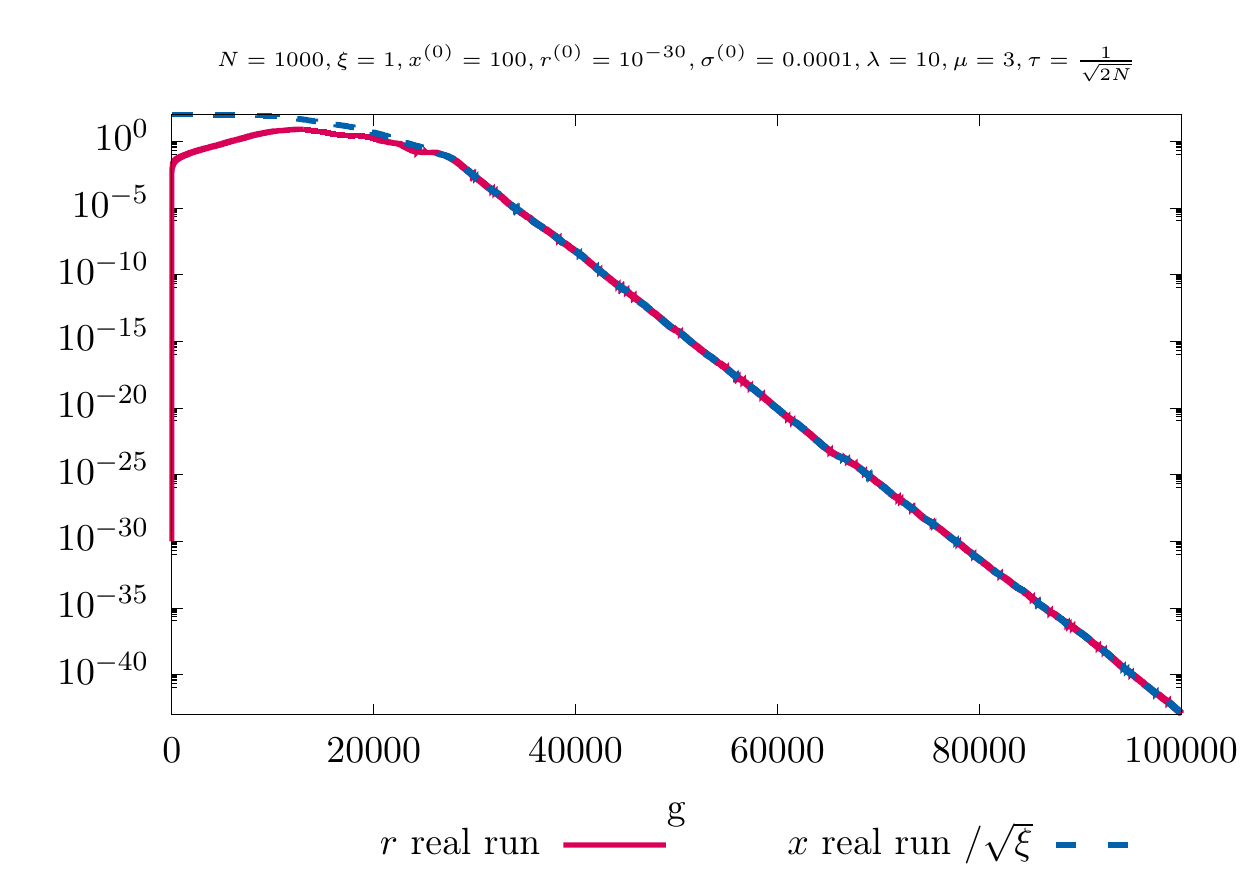}
  \end{tabular}
  \caption{Mean value dynamics closed-form approximation and real-run
    comparison of the
    $(3/3_I,10)$-ES
    with repair by projection
    applied to the conically constrained problem. (Part 5)}
  \label{sec:theoreticalanalysis:fig:dynamicsdetail5}
\end{figure}
\renewcommand{\sppapathtmp}{./figures/figure24/}
\begin{figure}[H]
  \centering
  \begin{tabular}{@{\hspace{-0.025\textwidth}}c@{\hspace{-0.025\textwidth}}c}
    \includegraphics[width=0.5\textwidth]{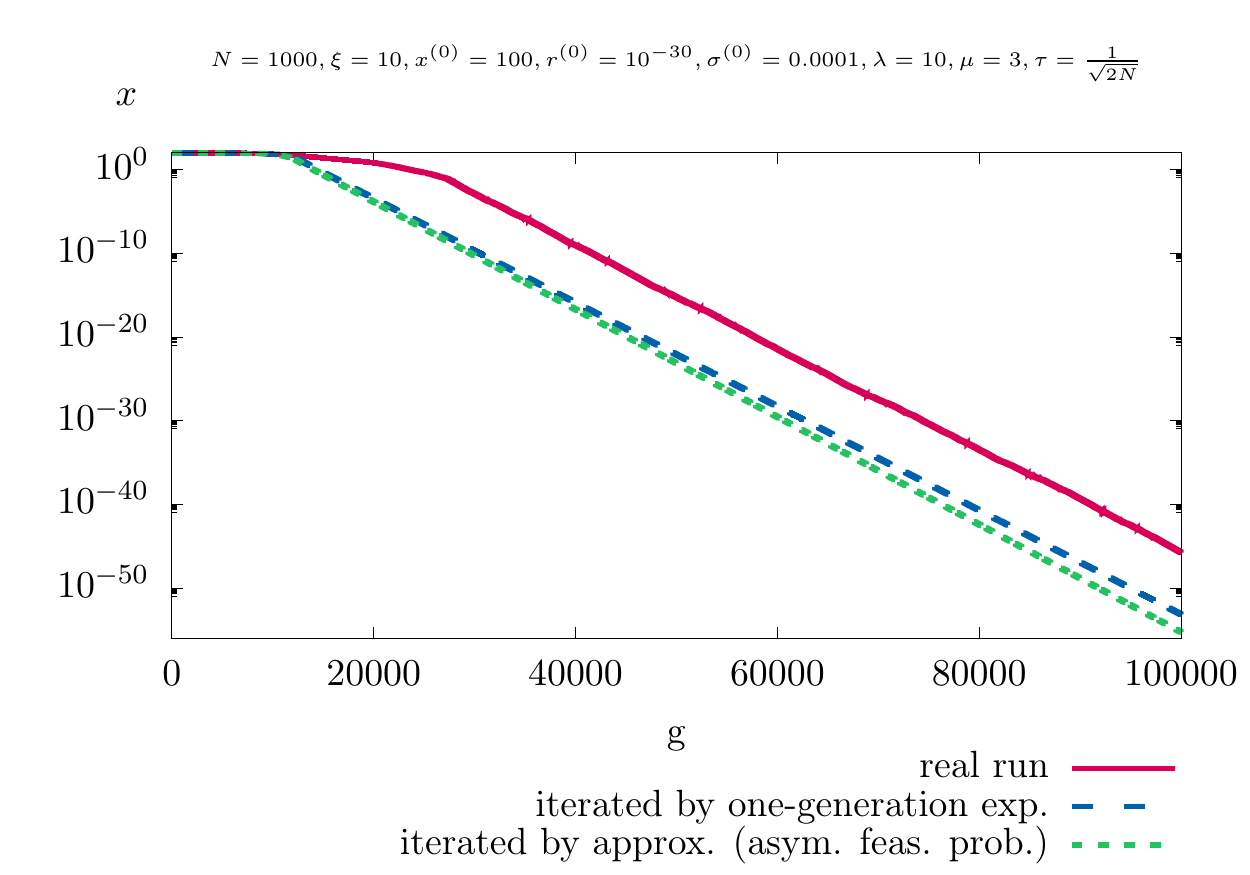}&
    \includegraphics[width=0.5\textwidth]{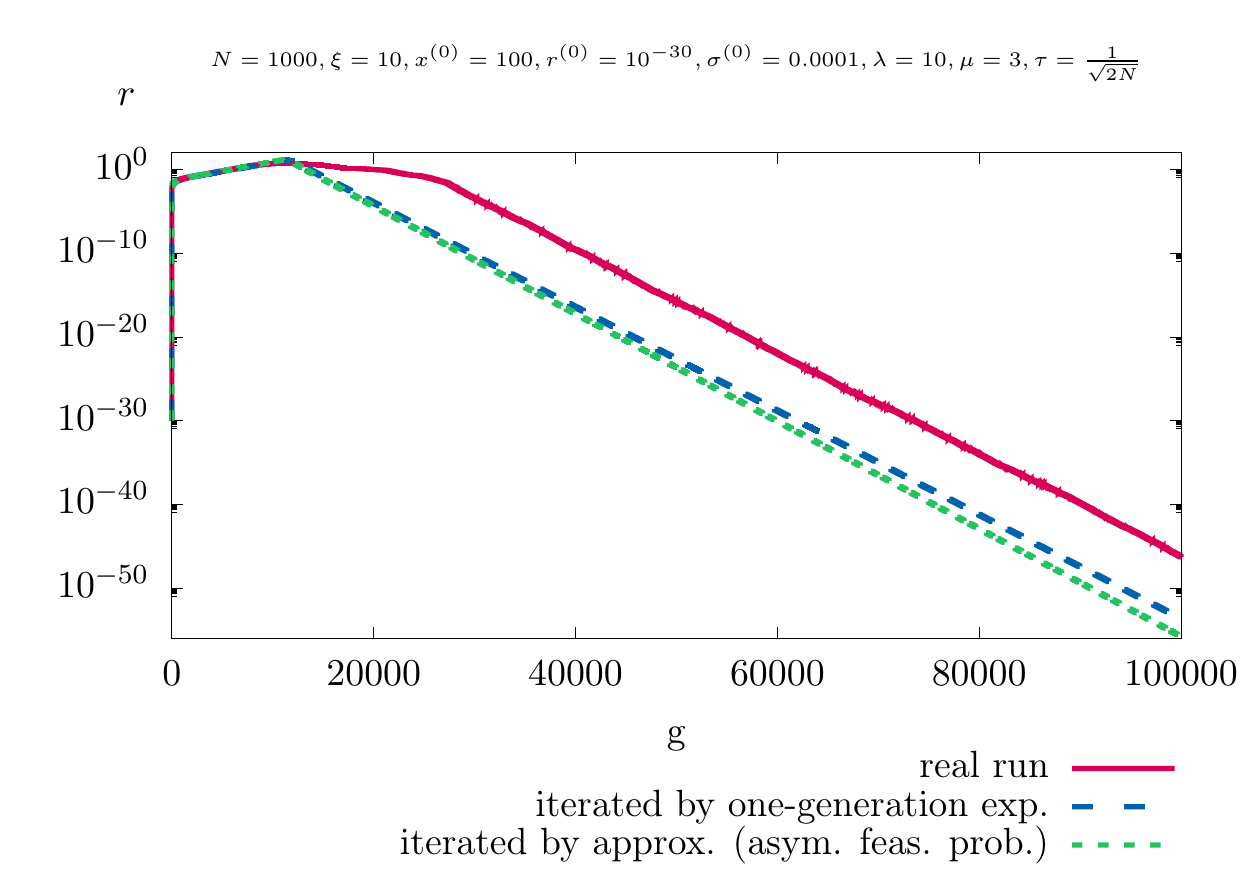}\\
    \includegraphics[width=0.5\textwidth]{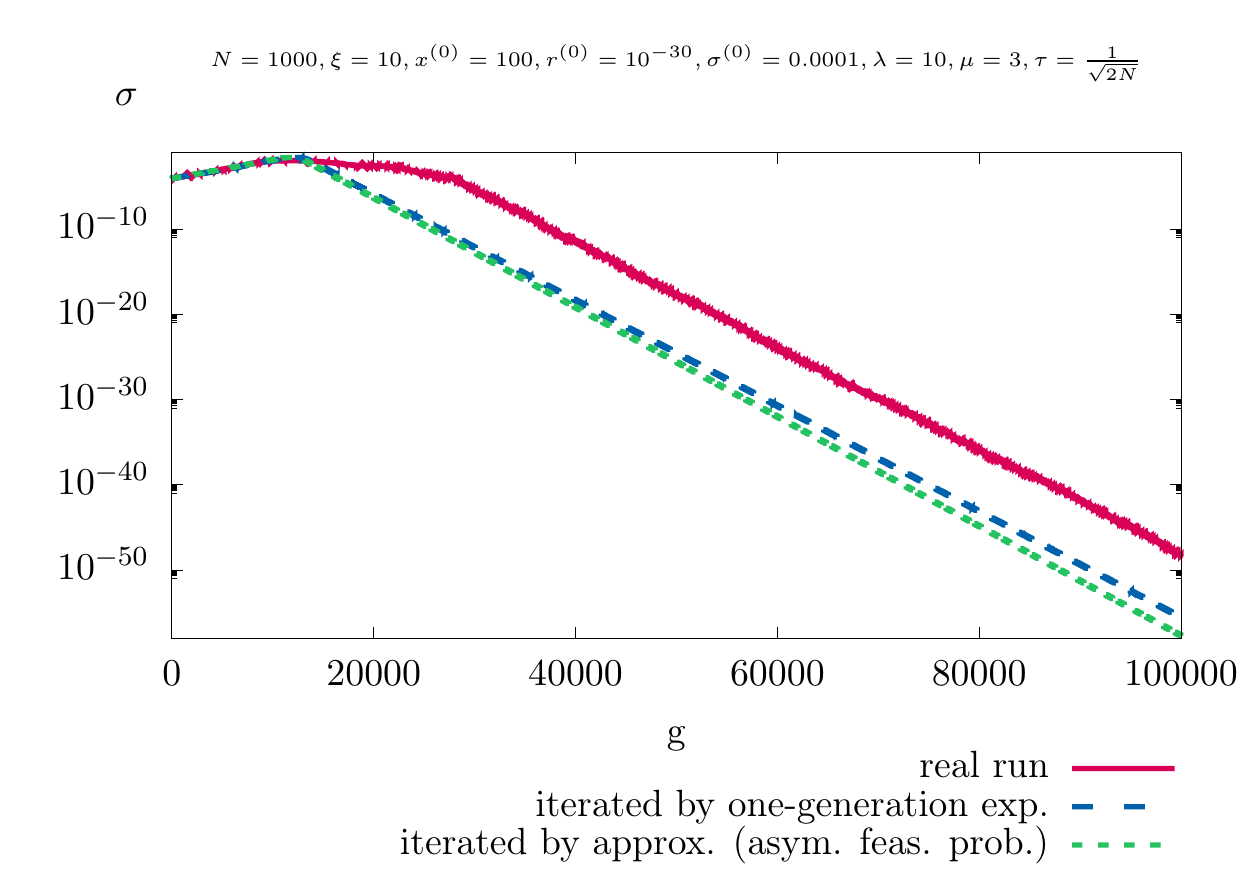}&
    \includegraphics[width=0.5\textwidth]{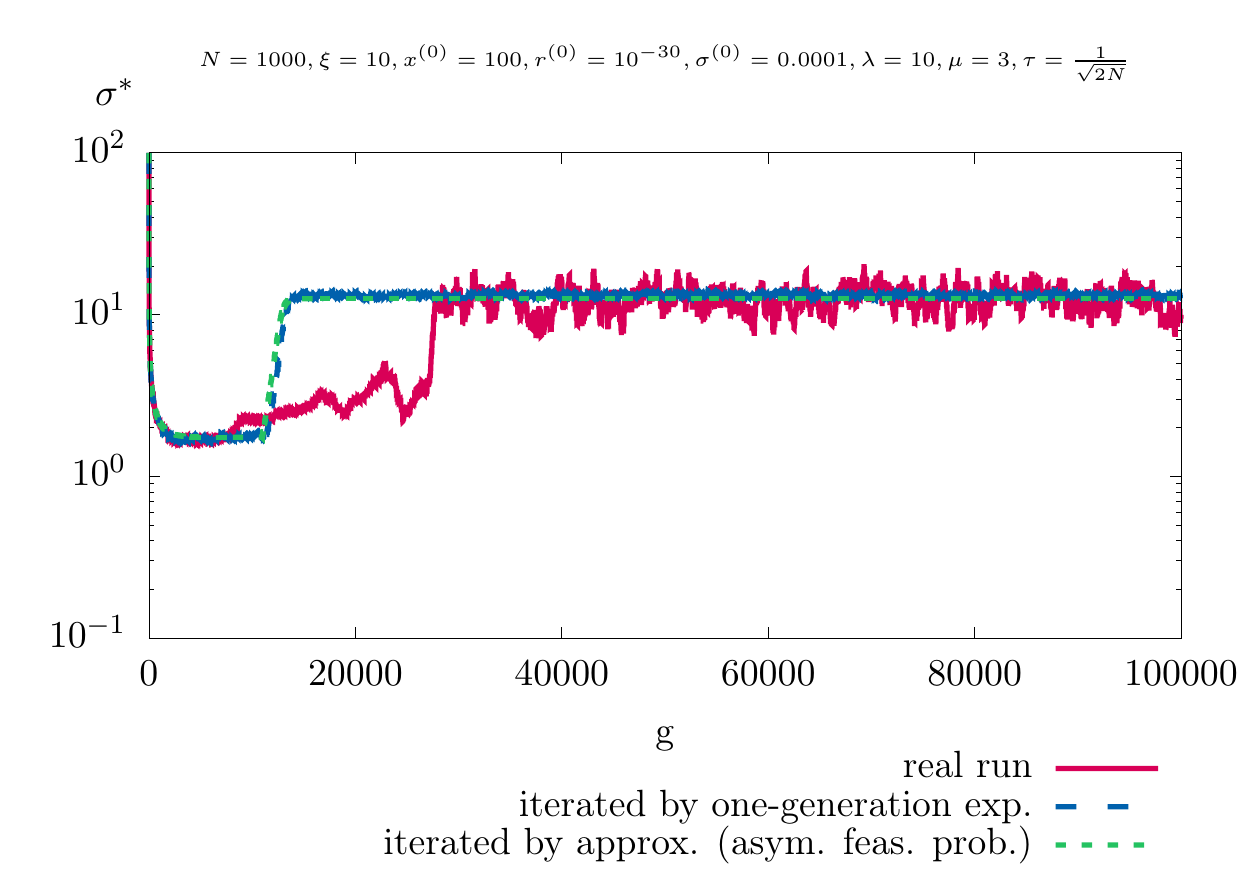}\\
    \includegraphics[width=0.5\textwidth]{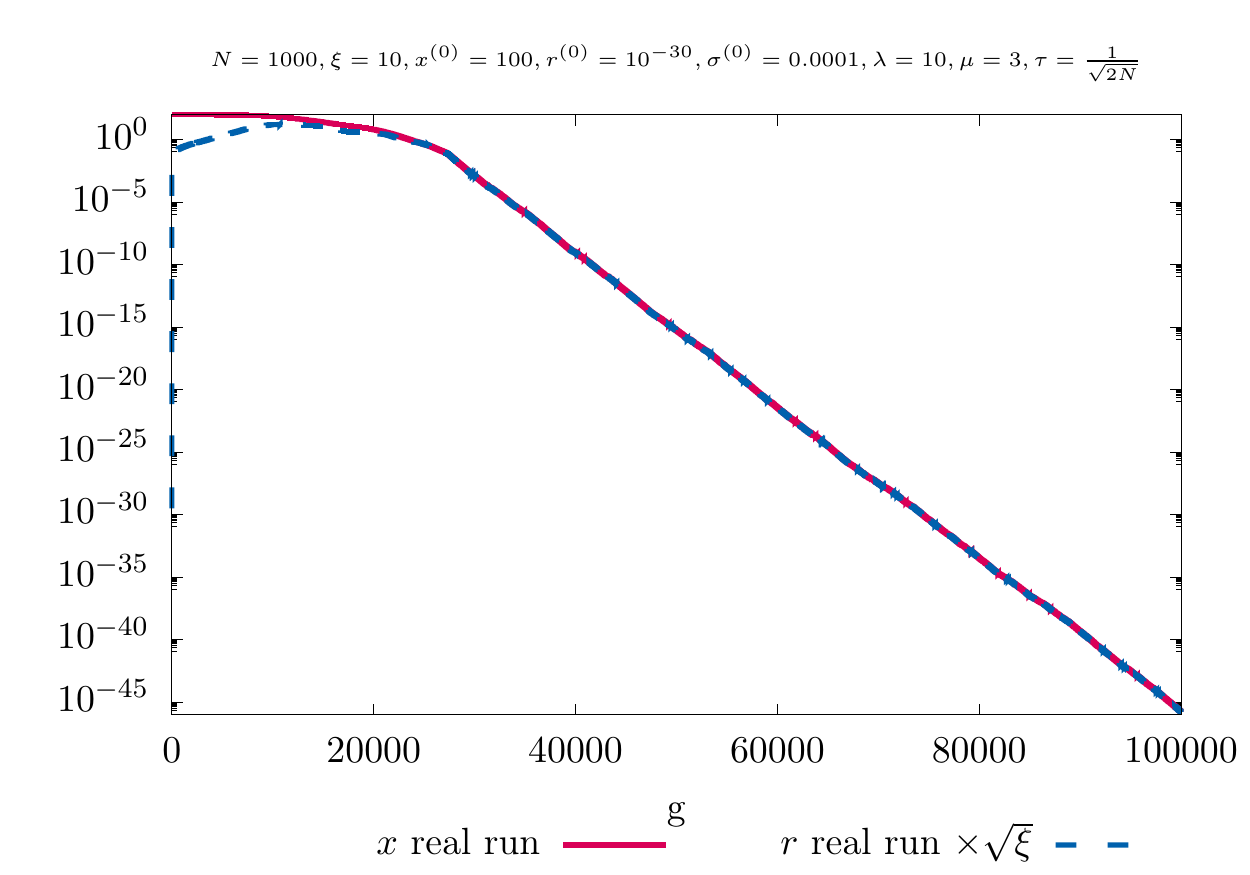}&
    \includegraphics[width=0.5\textwidth]{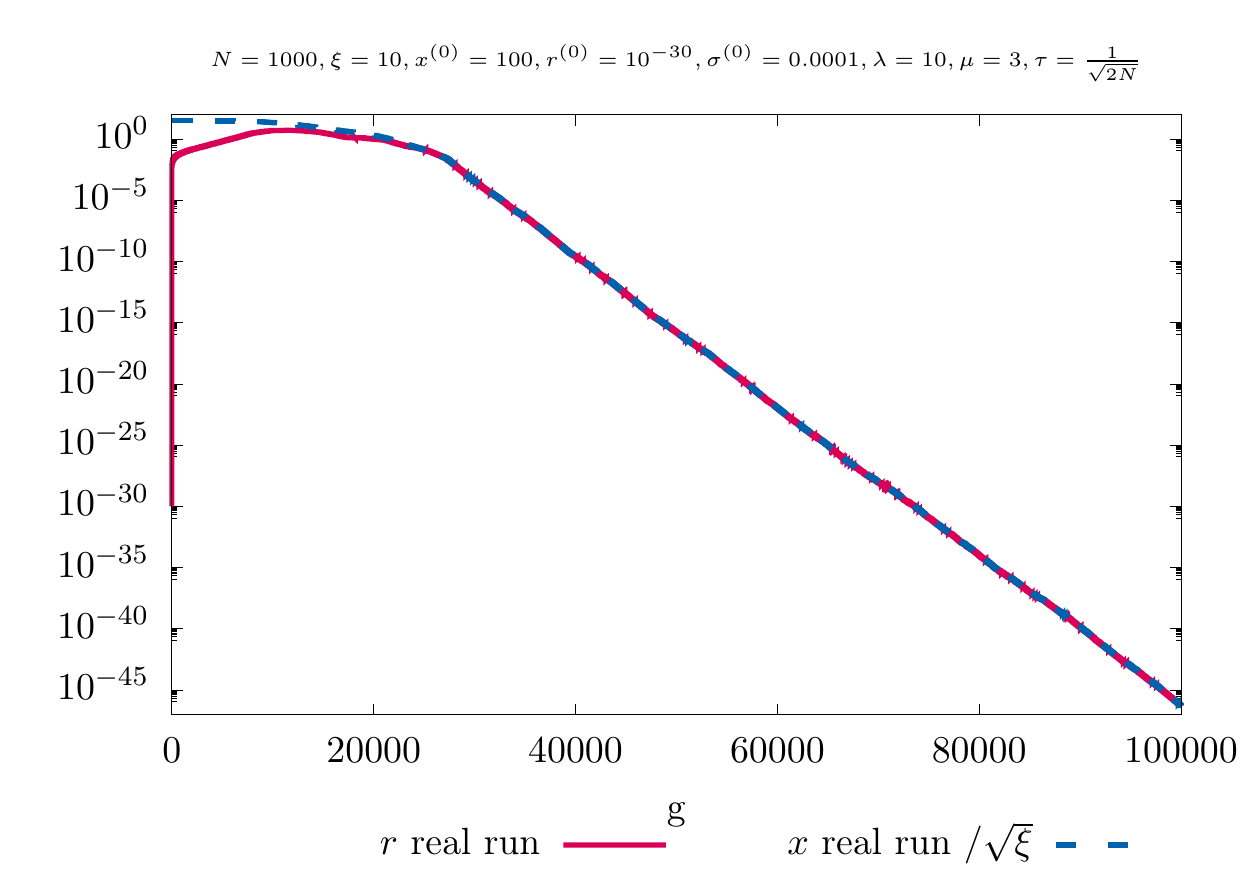}
  \end{tabular}
  \caption{Mean value dynamics closed-form approximation and real-run
    comparison of the
    $(3/3_I,10)$-ES
    with repair by projection
    applied to the conically constrained problem. (Part 6)}
  \label{sec:theoreticalanalysis:fig:dynamicsdetail6}
\end{figure}

\end{appendices}

\bibliographystyleapp{IEEEtran}
\bibliographyapp{app}

\end{document}